\documentclass{article}

    \usepackage[final]{neurips_2022}

\usepackage[utf8]{inputenc} 
\usepackage[T1]{fontenc}    
\usepackage{hyperref}       
\usepackage{url}            
\usepackage{booktabs}       
\usepackage{amsfonts}       
\usepackage{nicefrac}       
\usepackage{microtype}      

\usepackage{graphicx} 
\usepackage{subcaption} 
\usepackage{multirow}
\usepackage{mathtools}
\usepackage{relsize}
\usepackage{url}
\usepackage{nicefrac}
\usepackage{xspace}
\usepackage{algorithm,algorithmic}
\usepackage{multicol}
\usepackage{float}
\usepackage{bbm}
\usepackage{booktabs} 

\usepackage[utf8]{inputenc} 
\usepackage[T1]{fontenc}    
\usepackage{hyperref}       
\usepackage{url}            

\usepackage{nicefrac}       
\usepackage{microtype}      
 
\usepackage{layout}
\usepackage{multirow}

\usepackage{amsfonts}
\usepackage{amsmath}
\usepackage{amsthm}
\usepackage{amssymb} 

\usepackage{mdframed} 
\usepackage{thmtools}

\usepackage[font=normalsize, labelfont=bf]{caption}

\usepackage{makecell}
\usepackage{colortbl}
\definecolor{bgcolor2}{rgb}{0.65, 0.85, 0.99}
\definecolor{bgcolor}{rgb}{1, 0.96, 0.80}

\usepackage[dvipsnames]{xcolor}
\usepackage{pifont}

\newcommand{\makecellnew}[1]{{\renewcommand{\arraystretch}{0.8}\begin{tabular}{c} #1 \end{tabular}}}

\usepackage[colorinlistoftodos,bordercolor=orange,backgroundcolor=orange!20,linecolor=orange,textsize=scriptsize]{todonotes}

\graphicspath{{../figures/}}

\usepackage{amsmath,amsfonts,bm,mathtools}

\def\<{\left\langle}
\def\>{\right\rangle}
\def\[{\left[}
\def\]{\right]}
\def\({\left(}
\def\){\right)}

\newtheorem{theorem}{Theorem}  

\theoremstyle{definition}
\newtheorem{assumption}{Assumption} 
\newtheorem{definition}{Definition}

\theoremstyle{remark}

\newcommand{\mL}{\mathbf{L}}
\DeclareMathAlphabet\mathbfcal{OMS}{cmsy}{b}{n}

\def \E {\mathbb{E}}

\newcommand{\eqdef}{\stackrel{\mathrm{def}}{=}}

\newcommand{\R}{\mathbb R}

\newcommand{\bB}{\mathbb B}

\newcommand{\cL}{\mathcal L}
\newcommand{\mA}{\mathbf A}
\newcommand{\cO}{\mathcal O}
\newcommand{\cQ}{\mathcal Q}

\newcommand{\fC}{\mathcal{C}}
\newcommand{\mC}{\mathbf{C}}

\usepackage{lipsum}

\newcommand{\mI}{\mathbf{I}}

\newcommand{\Norm}[1]{\left\|#1\right\|}

\DeclareMathOperator{\prox}{prox}
\DeclareMathOperator{\sign}{sign}

\DeclareMathOperator{\diag}{\bf Diag}

\DeclareMathOperator{\range}{range}

\newcommand{\overbar}[1]{\mkern 1.5mu\overline{\mkern-1.5mu#1\mkern-1.5mu}\mkern 1.5mu}
\newcommand{\half}{{\nicefrac{1}{2}}}
\newcommand{\phalf}{{\dagger\nicefrac{1}{2}}}

\newcommand{\ML}{{\bf L}}
\newcommand{\MC}{{\bf C}}
\newcommand{\MI}{{\bf I}}

\newcommand{\MP}{{\bf P}}
\newcommand{\cC}{{\cal C}}

\newcommand{\squeeze}{\textstyle}

\title{Theoretically Better and Numerically Faster \\ Distributed Optimization with \\ Smoothness-Aware Quantization Techniques}

\author{%
  Bokun Wang \\
  Texas A\&M University, United States\thanks{Work done when the author was a research intern at KAUST, Saudi Arabia.} \\
  \texttt{bokunw.wang@gmail.com} \\
  \And
  Mher Safaryan \\
  KAUST, Saudi Arabia \\
  \texttt{mher.safaryan.1@kaust.edu.sa} \\
  \AND
  Peter Richt\'arik \\
  KAUST, Saudi Arabia \\
  \texttt{peter.richtarik@kaust.edu.sa}
}

\begin{document}

\maketitle

\begin{abstract}
To address the high communication costs of distributed machine learning, a large body of work has been devoted in recent years to designing various compression strategies, such as sparsification and quantization, and optimization algorithms capable of using them. Recently, \citet{safaryan2021smoothness} pioneered a dramatically different compression design approach: they first use the local training data to form local {\em smoothness matrices} and then propose to design a compressor capable of exploiting the smoothness information contained therein. While this novel approach leads to substantial savings in communication, it is limited to sparsification as it crucially depends on the linearity of the compression operator. In this work, we generalize their smoothness-aware compression strategy to {\em arbitrary unbiased compression} operators, which also include sparsification. Specializing our results to stochastic quantization, we guarantee significant savings in communication complexity compared to standard quantization. In particular, we prove that block quantization with $n$ blocks theoretically outperforms single block quantization, leading to a reduction in communication complexity by an $\mathcal{O}(n)$ factor, where $n$ is the number of nodes in the distributed system. Finally, we provide extensive numerical evidence with convex optimization problems that our smoothness-aware quantization strategies outperform existing quantization schemes as well as the aforementioned smoothness-aware sparsification strategies with respect to three evaluation metrics: the number of iterations, the total amount of bits communicated, and wall-clock time.
\end{abstract}

\section{Introduction}

Training modern machine learning models is typically cast in terms of (regularized) empirical risk minimization problem and requires increasingly more training data to make empirical risk closer to the true risk \citep{Sch,Vaswani2019-overparam}. This natural requirement makes it harder (and in some scenarios impossible) to collect all data in one place and carry out the training using a single data source. As a result, we reconciled with a flock of datasets disseminated across various compute nodes holding the actual training data \citep{bekkerman2011scaling,vogels}. However, such divide-and-conquer approach of handling vast amount of data means that local updates need to be  communicated  among the nodes (or through some central server orchestrating the process), which often forms the main bottleneck in modern distributed systems \citep{ZLKALZ, LHMWD}.
This issue is further exacerbated by the fact that modern highly performing models are typically overparameterized \citep{Brown2020fewshot,Narayanan2021megatron}.

{\bf 1.1. Distributed training.}
In this paper, we consider distributed training formalized as the following optimization problem
\begin{equation}\label{main-opt-problem-dist}
\squeeze
  \min\limits_{x\in\R^d} f(x) + R(x), \quad \text{where} \quad  f(x)\eqdef \frac{1}{n}\sum \limits_{i=1}^n f_i(x),
\end{equation}
and where $d$ is the number of parameters of model $x\in\R^d$ to be trained,  $n$ is the number of machines/nodes participating in the training, $f_i(x)$ is the loss/risk associated with the data stored on machine $i\in[n] \eqdef \{1,2,\dots,n\}$, $f(x)$ is the empirical loss/risk, and $R(x)$ is a regularizer.

Because of the communication constraints, large body of work has been devoted in recent years to the design of various compression strategies, such as sparsification \citep{RDME,tonko,alistarh2018convergence}, quantization \citep{Goodall1951:randdithering,Roberts1962:randdithering,alistarh2017qsgd}, low-rank approximation \citep{vogels}, three point compressor \citep{richtarik3PC}, and optimization algorithms capable of using them, such as Distributed Compressed Gradient Descent (DCGD) \citep{KFJ}, QSGD \citep{alistarh2017qsgd,Faghri2020adaptive}, NUQSGD \citep{Ramezani2021NUQSGD}, DIANA \citep{MGTR,horvath2019stochastic}, PowerSGD \citep{vogels}, signSGD \citep{bernstein2018signsgd, sign_descent_2019}, intSGD \citep{Mishchenko2021intSGD}, ADIANA \citep{ADIANA}, MARINA \citep{Gorbunov2021MARINA}.

{\bf 1.2. From scalar smoothness to matrix smoothness.}
Typically, distributed optimization algorithms in the literature that employ compressed communication, including all methods from the aforementioned works, use only shallow smoothness information of the loss function such as scalar $L$-smoothness \citep{Nest-ILCO}.
\begin{definition}[Scalar Smoothness] \label{def:scalar-smooth} Differentiable function $\phi:\R^d\to \R$ is called $L$-smooth if there exists a non-negative scalar value $L\ge0$ such that
\begin{equation}\label{eq:scalar-smooth}
\squeeze
\phi(x) \le \phi(y) + \<\nabla \phi(y), x-y\> + \frac{L}{2}\|x-y\|^2, \quad \forall x,y\in \R^d.
\end{equation}
\end{definition}
As pointed out by \cite{safaryan2021smoothness}, smoothness constant $L$ reflects small part of the rich smoothness information often easily available through the training data. In their recent work, \cite{safaryan2021smoothness} pioneered a dramatically different compression design approach. First, they propose to use the local training data to form local {\em smoothness matrices}, which they claim contain much more useful information than standard smoothness constants.
\begin{definition}[Matrix Smoothness] \label{def:matrix-smooth} Differentiable function $\phi:\R^d\to \R$ is called $\ML$-smooth if there exists a symmetric positive semidefinite matrix  $\ML\succeq\bm{0}$ such that
\begin{equation}\label{eq:matrix-smooth}
\squeeze
\phi(x) \le \phi(y) + \<\nabla \phi(y), x-y\> + \frac{1}{2}\|x-y\|^2_{\ML}, \quad \forall x,y\in \R^d.
\end{equation}
\end{definition}

Intuitively, the usefulness of $\mL$-smoothness over the standard $L$-smoothness is the tighter upper bound for functional growth. In other words, if for a function $\phi$ we have the tightest scalar smoothness $L$ and the tightest matrix smoothness $\mL$ parameters, then $L=\lambda_{\max}(\mL)$ and, hence, upper bound \eqref{eq:matrix-smooth} is better than \eqref{eq:scalar-smooth}. To understand the relationship deeper, consider the functional growth of $\phi$ along the direction $e\in\R^d$ (without loss of generality assume $\|e\|=1$). Let $x = y + te$, where $t>0$ is a positive scaling parameter, and consider the quadratic terms $\frac{t^2}{2}L$ of \eqref{eq:scalar-smooth} and $\frac{t^2}{2}\langle e, \mL e \rangle$ of \eqref{eq:matrix-smooth} bounded the functional growth. Obviously, depending on the direction $e$, the quadratic form $\langle e, \mL e \rangle$ can be much smaller than $L=\lambda_{\max}(\mL) = \sup\{\langle e, \mL e \rangle \colon \|e\| = 1\}$.

Non-uniform functional growths over different directions hints to design optimization algorithms that are aware of such properties of the objective. Several works on randomized coordinate descent have successfully exploited this approach \citep{Nsync,QuRich16-1,QuRich16-2,GJS-HR,ACD-HanzRich}. For example, the `NSync algorithm of \citet{Nsync} uses the smoothness matrix to estimate smaller, so-called {\em ESO (Expected Separable Overapproximation)} parameters for each coordinate, leading to larger stepsizes for the update rule and improved complexity for the algorithm. Note that randomized coordinate descent can be viewed as compressed gradient descent with random sparsification ($n=1$ number of workers).

In the context of distributed optimization, using smoothness matrices $\ML_i$ of all local loss functions $f_i(x),\;i\in[n]$, \cite{safaryan2021smoothness} design a compressor capable of exploiting the smoothness information contained within the smoothness matrices. In particular, under certain heterogeneity conditions on the smoothness matrices $\ML_i$, their new compressor reduces total communication cost by a factor of $\cO(\min(n,d))$.

{
    \begin{table*}[t]
    \scriptsize
        \centering
        \caption{Summary of main theoretical results of this work. Below constants and $\log\frac{1}{\varepsilon}$ factors are hidden, $n$ is the number of nodes, $d$ is the model size, $L_{\max} = \max_i L_i,\; L_i = \lambda_{\max}(\ML_i)$, the expected smoothness constant $\cL_{\max}$ is defined in \eqref{def:exp-smooth-L}, the variance of generic compression operator is denoted by $\omega$, parameters $\nu$ and $\nu_1$ are defined in (\ref{def:nu}). See Table \ref{tbl:notation} in the Appendix for the notations. We discuss some limitations of the proposed algorithms in Section~\ref{sec:limits}.}
        \label{tbl:summary}
        \renewcommand{\arraystretch}{1.7}
        \begin{tabular}{|c|c|c|}
        \hline
        \makecellnew{Regime}
            & \makecellnew{$\nabla f_i(x^*) \equiv 0$}
            & \makecellnew{arbitrary $\nabla f_i(x^*)$} \\
        \hline
            \hline
            \makecellnew{\bf Original Methods}
            & \makecellnew{\bf DCGD \citep{KFJ}}
            & \makecellnew{\bf DIANA \citep{MGTR}} \\
\hline
\makecellnew{Iteration Complexity}
&
$\frac{L}{\mu} + \frac{\omega L_{\max}}{n\mu}$
&
$\omega + \frac{L_{\max}}{\mu} + \frac{\omega L_{\max}}{n\mu}$
\\
\hline
\makecellnew{Communication Complexity \\ Standard Quantization ($\omega=\cO(n)$)}
&
$d \frac{L_{\max}}{\mu}$
&
$nd + d \frac{L_{\max}}{\mu}$
\\
\hline
            \hline
            \rowcolor{bgcolor}
            \makecellnew{\bf Redesigned Methods}
            & \makecellnew{\bf DCGD+ (Algorithm \ref{alg:DCGD+}) \\ {\bf with general compression}}
            & \makecellnew{\bf DIANA+ (Algorithm \ref{alg:DIANA+}) \\ {\bf with general compression}} \\
\hline
            \rowcolor{bgcolor}
\makecellnew{Iteration Complexity}
&
$\frac{L}{\mu} + \frac{\cL_{\max}}{n\mu}$
&
$\omega_{\max} + \frac{L}{\mu} + \frac{\cL_{\max}}{n\mu}$
\\
\hline
         \rowcolor{bgcolor}
\makecellnew{Communication Complexity \\ Block Quantization ($n = \cO(\sqrt{d})$)}
&
\makecellnew{
    \\
    $\frac{d}{{\color{red} n}}\frac{L_{\max}}{\mu}$ \\
    \\[-4pt]
    $\begin{smallmatrix} (\text{if}\; \nu,\;\nu_1 \;\text{are}\; \cO(1)) \end{smallmatrix}$
}
&
\makecellnew{
    \\
    $nd + \frac{d}{{\color{red} \sqrt{nd}}}\frac{L_{\max}}{\mu}$ \\
    \\[-4pt]
    $\begin{smallmatrix} (\text{if}\; \nu,\;\nu_1 \;\text{are}\; \cO(1)) \end{smallmatrix}$
}
\\
\hline
         \rowcolor{bgcolor}
\makecellnew{Communication Complexity \\ Quantization with varying steps}
&
\makecellnew{
    \\
    $\frac{d}{{\color{red} n}}\frac{L_{\max}}{\mu} + \frac{d}{{\color{red} d}}\frac{L_{\max}}{\mu}$ \\
    \\[-4pt]
    $\begin{smallmatrix} (\text{if}\; \nu,\;\nu_1 \;\text{are}\; \cO(1)) \end{smallmatrix}$
}
&
\makecellnew{
    \\
    $nd + \frac{d}{{\color{red} n}}\frac{L_{\max}}{\mu} + \frac{d}{{\color{red} d}}\frac{L_{\max}}{\mu}$ \\
    \\[-4pt]
    $\begin{smallmatrix} (\text{if}\; \nu,\;\nu_1 \;\text{are}\; \cO(1)) \end{smallmatrix}$
}
\\
\hline
         \rowcolor{bgcolor}
\makecellnew{Theorems}
&
{\small \ref{thm:dcgd+_arbitrary},\; \ref{thm-prop:block-dcgd+},\; \ref{thm-prop:quant-dcgd+}}
&
{\small \ref{thm:diana+_arbitrary},\; \ref{thm-prop:block-diana+},\; \ref{thm-prop:quant-diana+}} \\
\hline
\hline
         \rowcolor{bgcolor}
\makecellnew{Speedup factor (up to)}
&
${\color{red} \min(n,d)}$
&
${\color{red} \min(n,d)}$
\\
        \hline
        \end{tabular}   
    \end{table*}
}

{\em While this novel approach leads to substantial savings in communication, it is limited to random sparsification as it crucially depends on the linearity of the compression operator. It is not clear whether this approach can be useful in the design of other smoothness-aware compression techniques.}

\section{Summary of Contributions}

Motivated by the above mentioned development, in this work, we made the following contributions.

{\bf 2.1. Extending matrix-smoothness-aware sparsification to general compression schemes.}
First, we generalize the smoothness-aware sparsification strategy \citep{safaryan2021smoothness} to arbitrary unbiased compressors. Instead of sparsification operator, we consider the generic class $\bB^d(\omega)$ of (possibly randomized) unbiased compression operators $\cC\colon\R^d\to\R^d$ with bounded variance $\omega\ge0$, i.e.,
\begin{equation*}
  \E\[\cC(x)\] = x, \quad \E\[\|\fC(x) - x\|^2\] \le \omega\|x\|^2, \quad \forall x\in \R^d.
\end{equation*}
This class is quite broad including random sparsification and various quantization schemes. To benefit from the matrix smoothness information with general compressor $\cC$, we propose the following modification in the communication protocol. If $x\in\R^d$ is the vector to be communicated, instead of applying compressor $\fC$ directly to $x$ and sending $\fC(x)$, we compress it by $\fC(\ML^\phalf x)$ and decompress it by multiplying $\ML^\half$. Overall, the receiver estimates the original $x$ by $\ML^\half\fC(\ML^\phalf x)$.

{\bf 2.2. Distributed compressed methods with improved communication complexity.}
To highlight the appropriateness of our generalization, we redesign two distributed compressed methods---DCGD \citep{KFJ} and DIANA \citep{MGTR}---to effectively utilize both matrix smoothness information and general compression operators leading to new methods, which we call DCGD+ (Algorithm \ref{alg:DCGD+}) and DIANA+ (Algorithm \ref{alg:DIANA+}).
The key notion we introduce that enables the technical analysis is the following quantity describing interaction between compression operator $\fC\in\bB^d(\omega)$ and smoothness matrix $\ML\succeq\bm{0}$:
\begin{equation*}
\squeeze
\cL(\fC, \mL) \eqdef \inf\left\{\cL\ge0 \colon \E{\|\fC(x) - x\|^2_{\mL}} \le \cL\|x\|^2\right\} \le \omega \lambda_{\max}( \mL).
\end{equation*}
This quantity generalizes the one defined in \cite{safaryan2021smoothness} for sparsification, and provides means for tighter theoretical guarantees (Theorems \ref{thm:dcgd+_arbitrary} and \ref{thm:diana+_arbitrary}) and better compression design.

{\bf 2.3. Block quantization.}
As we are no longer constrained to sparsification to exploit matrix smoothness, we consider more aggressive quantization schemes to further reduce the communication cost. Our first extension of standard quantization \citep{alistarh2017qsgd} is {\em block quantization}, where each block is allowed to have a separate quantization parameter.
Notably, we show theoretically that our block quantization with $n$ blocks outperforms single block quantization and saves in communication by a factor of $\mathcal{O}(n)$ for both DCGD+ (Theorem \ref{thm-prop:block-dcgd+}) and DIANA+ (Theorem \ref{thm-prop:block-diana+}) when $n=\cO(\sqrt{d})$.

{\bf 2.3. Quantization with varying steps.}
In our second extension of standard quantization, we go even further and allow all coordinates to have their own quantization steps. This extension turns out to be more efficient in practice than block quantization and provides savings in communication cost by a factor of $\mathcal{O}(\min(n,d))$ for both DCGD+ (Theorem \ref{thm-prop:quant-dcgd+}) and DIANA+ (Theorem \ref{thm-prop:quant-diana+}).

{\bf 2.4. Experiments.}
Finally, we perform extensive numerical experiments using LibSVM data \citep{chang2011libsvm} and provide clear numerical evidence that the proposed smoothness-aware quantization strategies outperform existing quantization schemes as well the aforementioned smoothness-aware sparsification strategies with respect to three evaluation metrics: the number of iterations, the total amount of bits communicated, and wall-clock time (see Section \ref{sec:exps} and the Appendix).

\section{Smoothness-Aware Distributed Methods with General \mbox{Compressors}}\label{sec:ext-arb-comp}

In this section we extend methods DCGD+ and DIANA+ of \cite{safaryan2021smoothness} to handle arbitrary unbiased compression operators. We consider the problem \eqref{main-opt-problem-dist} with matrix smoothness assumption for all local losses $f_i(x)$ and with strong convexity of loss function $f(x)$.

\begin{assumption}[Matrix smoothness]\label{asm:Li-smooth-convex}
The  functions $f_i\colon\R^d\to\R$ are differentiable, convex, lower bounded and $\ML_i$-smooth. Besides, $f$ is $\ML$-smooth with the scalar smoothness constant $L \eqdef \lambda_{\max}(\ML)$.\end{assumption}

First, note that lower boundedness of $f_i(x)$ is not needed once $\ML_i\succ0$ is invertible. This part of the assumption is not a restriction in applications as all loss function are lower bounded. Regarding the relation between $\mL$ and $\mL_i$, notice that \eqref{main-opt-problem-dist} implies $\mathbf{L} \preceq \frac{1}{n}\sum_{i=1}^n \mathbf{L}_i$. This means that while $\frac{1}{n}\sum_{i=1}^n \mathbf{L}_i$ can serve as a smoothness matrix for $f$, there might be a tighter estimate, which we denote by $\ML$. Clearly, matrix smoothness provides much more information about the loss function than scalars smoothness. However, estimating dense smoothness matrix $\mL$ could be expensive for problems beyond generalized linear models because of the $d^2$ number of entries and lack of closed-form expression. On the other hand, estimating sparse, such as diagonal, smoothness matrix $\diag(L^1, L^2,\dots,L^d)$ should be feasible. Lastly, if $\mL$ is a smoothness matrix (could be dense, diagonal or any structure) of $f$, then any matrix $\widetilde{\mL}\preceq\mL$ is also a smoothness matrix for $f$. This implies that our theory would still work if the smoothness matrix is over-approximated.

\begin{assumption}[$\mu$-convexity]\label{asm:mu-convex}
The function $f\colon\R^d\to\R$ is $\mu$-convex for some $\mu>0$, i.e.,
$$
\squeeze
f(x) \ge f(y) + \<\nabla f(y), x-y\> + \frac{\mu}{2}\|x-y\|^2, \quad \forall x,y\in\R^d.
$$
\end{assumption}

This assumption is rather standard in the literature, sometimes referred to as strong convexity.

{\bf 3.1. DCGD+ with arbitrary unbiased compression.}
In our version of DCGD+, each node $i\in[n]$ is allowed to control its own compression operator $\fC_i\in\bB^d(\omega)$ independent of other nodes. Denote
\begin{equation}\label{def:exp-smooth-L}
\squeeze 
  \cL_{\max} \eqdef \max_{1\le i \le n} \cL_i, \quad \text{where} \quad
  \cL_i \eqdef \cL(\fC_i, \mL_i).
\end{equation}

Furthermore, as the compressor $\fC_i$ can be random, denote by $\fC_i^k$ a copy of $\fC_i$ generated at iteration $k$.

\begin{algorithm}[H]
\begin{algorithmic}[1]
\STATE \textbf{Input:} Initial point $x^0\in\R^d$, step size $\gamma>0$, compression operators $\{\fC^k_1,\dots,\fC^k_n\}$
\STATE \textbf{on} server
\STATE \quad send $x^k$ to all nodes
\STATE \quad get compressed updates $\fC_i^k(\ML_i^{\phalf} \nabla f_i(x^k))$ from all nodes $i\in[n]$
\STATE \quad update the model to $x^{k+1} = \prox_{\gamma R}(x^k - \gamma g^k)$, where $g^k = \frac{1}{n}\sum_{i=1}^n \ML_i^{\half}\fC_i^k(\ML_i^{\phalf} \nabla f_i(x^k))$
\end{algorithmic}
\caption{\sc DCGD+ with arbitrary unbiased compression}
\label{alg:DCGD+}
\end{algorithm}

Similar to the standard DCGD method, convergence of DCGD+ is linear up to some oscillation neighborhood. However, for the interpolation regime this neighborhood vanishes and the method converges linearly to the exact solution.

\begin{theorem}\label{thm:dcgd+_arbitrary} 
Let Assumptions  \ref{asm:Li-smooth-convex} and \ref{asm:mu-convex} hold and assume that each node $i\in[n]$ generates its own copy of compression operator $\fC_i^k\in\bB^d(\omega_i)$ independently from others. Then, for the step-size 
$
0<\gamma
\le \frac{1}{L+\frac{2}{n} \cL_{\max}},
$
the iterates $\{x^k\}$ of DCGD+ (Algorithm \ref{alg:DCGD+}) satisfy
\begin{equation}\label{rate-dskgd}
\squeeze
\E\[\|x^k - x^*\|^2\] \le \(1-\gamma\mu\)^k\|x^{0}  - x^*\|^2 + \frac{2\gamma\sigma_+^*}{\mu n},
\end{equation}
where $\sigma_+^* \eqdef \frac{1}{n}\sum_{i=1}^n \cL_i \|\nabla f_i(x^*)\|^2_{\ML_i^{\dagger}}.$ In particular, for the interpolation regime (i.e., $\nabla f_i(x^*) = 0$ for all $i\in[n]$), then DCGD+ converges linearly with iteration complexity
\begin{equation}\label{DCGD+-complexity}
\squeeze
\cO\big( ( \frac{L}{\mu} + \frac{\cL_{\max}}{n \mu} ) \log\frac{1}{\varepsilon} \big).
\end{equation}
\end{theorem}

We show later that the iteration complexity \eqref{DCGD+-complexity} of DCGD+ can be much better than one of DCGD. However, the size of the neighborhood of DCGD+ might be bigger than of DCGD. In case of standard (scalar) smoothness (i.e. $\mL_i = L_i\mI$) the size of the neighborhood would be $\sigma^* \eqdef \frac{1}{n}\sum_{i=1}^n \omega_i \|\nabla f_i(x^*)\|^2$, which might be smaller than $\sigma_+^*$. Even though we have $\cL_i \le \omega_i\lambda_{\max}(\mL_i)$ from the definition of $\cL_i$, it does not imply $\cL_i\mL_i^\dagger \preceq \omega_i\mI$. Thus, with matrix-smoothness-aware compression we ensure faster linear convergence at the cost of a possibly larger oscillation radius. This is not an issue for the interpolation regime, which can interpolate the whole training data with zero loss. Moreover, next we present an algorithmic solution to remove the neighborhood using the DIANA method.

{\bf 3.2. DIANA+ with arbitrary unbiased compression.}
The mechanism allowing to remove the neighborhood in DIANA+ is based on the DIANA method, which was initially introduced for ternary quantization by \cite{MGTR}, and then extended to arbitrary unbiased compression operators by \cite{horvath2019stochastic}. The high level idea is to learn the local optimal gradients $\nabla f_i(x^*)$ by estimates $u_i^k$ for all nodes $i\in[n]$ in a communication efficient manner. Nodes use these estimates $u_i^k$ to progressively construct better local gradient estimates $g_i^k$ reducing the variance induced from the compression.

\begin{algorithm}[H]
\begin{algorithmic}[1]
\STATE \textbf{Input:} Initial point $x^0\in\R^d$, initial shifts $u_i^0\in\range(\ML_i)$ and $u^0 \eqdef \frac{1}{n}\sum_{i=1}^n u_i^0$, step size parameters $\gamma>0$ and $\alpha>0$, compression operators $\{\fC^k_1,\dots,\fC^k_n\}$
\FOR{each node $i = 1, \dots, n$ in parallel} 
\STATE get $x^k$ from the server and compute local gradient $\nabla f_i(x^k)$
\STATE send compressed update $\Delta_i^k = \fC_i^k(\ML_i^\phalf (\nabla f_i(x^k) - u_i^k))$ to the server
\STATE update local gradient and shift $\overbar{\Delta}_i^k = \ML_i^\half\Delta_i^k,\; g_i^k = u_i^k + \overbar{\Delta}_i^k, u_i^{k+1} = u_i^k + \alpha\overbar{\Delta}_i^k$
\ENDFOR
\STATE \textbf{on} server
\STATE \quad get all sparse updates $\Delta_i^k,\, i\in[n]$ and $\overbar{\Delta}^k = \frac{1}{n}\sum_{i=1}^n \overbar{\Delta}_i^k = \frac{1}{n}\sum_{i=1}^n \ML_i^\half\Delta_i^k,\; g^k = \overbar{\Delta}^k + u^k$
\STATE \quad update the global model to $x^{k+1} = \prox_{\gamma R}(x^k - \gamma g^k)$ and global shift to $u^{k+1} = u^k + \alpha\overbar{\Delta}^k$
\end{algorithmic}
\caption{\sc DIANA+ with arbitrary unbiased compression}
\label{alg:DIANA+}
\end{algorithm}

We prove in the Appendix that both iterates $x^k$ and all local gradient estimates $u_i^k$ converge linearly to the exact solution $x^*$ and $\nabla f_i(x^*)$ respectively.

\begin{theorem}\label{thm:diana+_arbitrary}
Let Assumptions  \ref{asm:Li-smooth-convex} and \ref{asm:mu-convex} hold and assume that each node $i\in[n]$ generates its own copy of compression operator $\fC_i^k\in\bB^d(\omega_i)$ independently from others. Then, if $\omega_{\max} = \max_{1\le i\le n}\omega_i$ and the step-size 
$
\gamma = \frac{1}{L+\frac{6}{n} \cL_{\max}},
$
DIANA+ (Algorithm \ref{alg:DIANA+}) converges linearly with iteration complexity
\begin{equation}\label{DIANA+-complexity}
\squeeze
\cO\big( ( \omega_{\max} + \frac{L}{\mu} + \frac{\cL_{\max}}{n\mu} ) \log\frac{1}{\varepsilon} \big).
\end{equation}
\end{theorem}

Notice that the cost of removing the neighborhood is the extra $\cO(\omega_{\max}\log\frac{1}{\varepsilon})$ iterations, which is negligible in the overall complexity \eqref{DIANA+-complexity} above.
Another interesting observation is the second order flavor of the gradient learning technique employed by DIANA+. Let, for concreteness, matrices $\mL_i$ be invertible and $\fC_i^k(-x) = -\fC_i^k(x)$ for all $x\in\R^d$ (both random sparsification and quantization satisfy this). Typically, the learning procedure of the original DIANA method, $u_i^{k+1} = u_i^k - \alpha\fC_i^k(u_i^k - \nabla f_i(x^k))$, can be interpreted as a single step of CGD applied to the problem of minimizing the convex quadratic function
$
\squeeze
\varphi^k_i(u)\eqdef \frac{1}{2}\left\|u - \nabla f_i(x^k)\right\|^2,
$
which changes in each iteration because the gradient changes. In contrast, we observe  that the learning mechanism of DIANA+ can be interpreted as a single step of  a (damped) Newton's method with compressed gradients and with the true Hessian. Indeed, fix the iteration counter $k$ and denote 
$
\squeeze
\varphi^k_i(u) \eqdef \frac{1}{2}\left\|u-\nabla f_i(x^k)\right\|^2_{\mL_i^{-\half}}.
$
Then, the update rule of shifts $u_i^k$ in DIANA+ can be rewritten as
$
\squeeze
u_i^{k+1}
= u_i^k - \alpha\ML_i^\half\fC(\ML_i^{-\half}(u_i^k - \nabla f_i(x^k)))
= u_i^k - \alpha\[\nabla^2\varphi_i(u_i^k)\]^{-1}\fC_i^k(\nabla\varphi_i(u_i^k)).
$
This might serve as an extra explanation on why incorporating smoothness matrices properly can improve the performance of first order methods with communication compression.

{\bf 3.3. Baselines for the original methods.}
To make the theoretical comparison against DCGD and DIANA more transparent, we fix the following baselines using the standard quantization scheme.

$\bullet$ {\bf Baseline for DIANA.}
The iteration complexity of DIANA is $T = \widetilde{\cO}(\omega + \frac{L_{\max}}{\mu} + \frac{\omega L_{\max}}{n\mu})$. When applying standard quantization from \citep{alistarh2017qsgd}, the amount of bits each node communicates is $b = \cO(s^2 + s\sqrt{d}) = \max(s^2,s\sqrt{d}) = \cO(\frac{d}{\omega})$ since $\omega = \min(\frac{d}{s^2},\frac{\sqrt{d}}{s}) = \frac{d}{\max(s^2,s\sqrt{d})}$ (Lemma 3.1., \citet{alistarh2017qsgd}). Thus, the total communication complexity of DIANA is $n \cdot T \cdot b = \widetilde{\cO}(nd + \frac{n d L_{\max}}{\omega\mu} + \frac{d L_{\max}}{\mu})$. Thus, the optimal total communication complexity of DIANA is $\widetilde{\cO}(nd + \frac{d L_{\max}}{\mu})$, which is attained when $\omega = \cO(n)$.

$\bullet$ {\bf Baseline for DCGD.}
Based on the iteration complexity\footnote{this can be shown by specializing Theorem \ref{thm:dcgd+_arbitrary} or Theorem 2 of \cite{safaryan2021smoothness} to scalar smoothness setup and interpolation regime, namely $\mL_i = L_i\mI$ and $\|\nabla f_i(x^*)\| = 0$ for all $i\in[n]$.} $\widetilde{\cO}(\frac{L}{\mu} + \frac{\omega L_{\max}}{n\mu})$ of DCGD (in case $\nabla f_i(x^*)=0$ for all $i\in[n]$), we fix the same level of compression $\omega=\cO(n)$, which results in $\widetilde{\cO}(\frac{L_{\max}}{\mu})$ iterations complexity. From the estimate of quantization variance $\omega = \min\(\frac{d}{s^2},\frac{\sqrt{d}}{s}\)$ we conclude that $s=\cO(\frac{\sqrt{d}}{n})$ should be used. Finally, with this choice of $s$, each node communicates $\cO(s^2+s\sqrt{d}) = \cO(\frac{d}{n})$ amount of bits. Thus, total communication complexity (i.e. how many bits flows through the central server) of DCGD is $\widetilde{\cO}(\frac{d L_{\max}}{\mu})$.

To compare the proposed methods with these baselines and highlight improvement factors, define parameters $\nu$ and $\nu_{1}$ describing local smoothness matrices $\ML_i$ as follows
\begin{equation}\label{def:nu}
\squeeze
\nu \eqdef \frac{\sum_{i=1}^n L_i}{\max_{i\in[n]} L_i}, \quad
\nu_1 \eqdef \max_{i\in[n]}\frac{\sum_{j=1}^d \ML_{i;j}}{\max_{j\in[d]}\ML_{i;j}},
\end{equation}
where $L_i = \lambda_{\max}(\ML_i)$, $L_{\max} \eqdef \max_{1\le i\le n} L_i$ and $\mL_{i;j}$ is the $j$th diagonal element of matrix $\mL_i$. Parameters $\nu\in[1,n]$ and $\nu_1\in[1,d]$ describe the level of heterogeneity over the nodes and coordinates respectively. If $\ML_i$ matrices coincide, then $\nu=n$ and $\nu_1=d$. On the other extreme, when the values of $\mL_i$ are extremely non-uniform, we have $\nu\ll n$ and $\nu_1\ll d$.

Notice that the quantity $\frac{\cL_{\max}}{\mu n}$ in \eqref{DCGD+-complexity} and the quantity $\omega_{\max} + \frac{\cL_{\max}}{\mu n}$ in \eqref{DIANA+-complexity} depend on compression operators $\cC^k_i$ applied by the nodes. For the rest of the paper we are going to minimize these quantities with respect to the choice of $\cC^k_i$ in such a way to minimize total communication complexity of the proposed distributed methods. We specialize compressors $\cC_i$ to two different extensions of standard quantization and optimize with respect to compression parameters.

\section{Block Quantization}\label{sec:block-quant}

We now present our first extension to standard quantization in order to properly capture the matrix smoothness information. Instead of having a single quantization parameter (e.g. number of levels) for all coordinates, here we divide the space $\R^d$ into $B\in\{1,2,\dots,d\}$ blocks as $\R^d = \R^{d_1}\times\R^{d_2}\times\dots\times\R^{d_B}$ and for each subspace $\R^{d_l},\; l\in[B]$ we apply standard quantization independently from other blocks with different number of levels $s_l$. Thus, for any $l\in[B]$ we allocate one parameter $s_l$ for $l^{th}$ block of $x\in\R^d$. Hence quantization is applied block-wise: for each block we send the norm $\|x^l\|$ of the block $x^l\in\R^{d_l}$ and all entries within this block are quantized with levels $\{0, \frac{1}{s_l},\frac{2}{s_l},\dots,1\}$. In the special case of $B=1$, we get the standard quantization of \cite{alistarh2017qsgd}.

To get rid of the constraints on $s_l$ to be integers, instead of working with the number of levels $s_l$, we introduce the size of the quantization step $h_l = \frac{1}{s_l}$ and allow them to take any positive values (even bigger than $1$). Thus, for each block $l\in[B]$ we quantize with respect to levels $\{0,h_l,2h_l,\dots\}$.

\begin{definition}[Block Quantization]\label{def:block-quant}
For a given number of blocks $B\in[d]$ and fixed quantization steps $h=(h_1,\dots,h_B)$, define block-wise quantization operator $\cQ^B_h\colon\R^d\to\R^d$ as follows:
\begin{eqnarray*}
\squeeze
\[\cQ^B_h(x)\]_{t} &\eqdef& \|x^l\| \cdot \sign(x_{t}) \cdot \xi_l\(\frac{|x_{t}|}{\|x^l\|}\),
\end{eqnarray*}
where $t = (l-1)B + j,\; x\in\R^d,\; j\in[d_l],\; l\in[B]$ and $\xi_l(v)$ for $v\ge0$ is defined via the quantization levels $\{0,h_l,2h_l,\dots\}$ as follows: if $k h_l \le v < (k+1)h_l$ for some $k\in\{0,1,2,\dots\}$, then
\begin{equation}\label{quant-xi}
\squeeze
\xi_l(v) \eqdef
\left\{
\begin{smallmatrix}
  k h_l \hfill && \text{with probability}\quad k + 1 - \frac{v}{h_l}, \hfill \\
  (k+1)h_l \hfill && \text{with probability}\quad \frac{v}{h_l} - k. \hfill 
\end{smallmatrix}\right.
\end{equation}
\end{definition}
Note that $\cQ^B_h$ is an unbiased compression operator as $\E\[\xi_j(v)\] = v$ for any $v\ge0$.
To communicate a vector of the form $\cQ^B_h(x)$, we encode each block $\[\cQ^B_h(x)\]^l\in\R^{d_l}$ using Elias $\omega$-coding as in the standard quantization scheme \citep{alistarh2017qsgd}. Hence, for each block $l\in[B]$ we need to send $\widetilde{\cO}(\frac{1}{h_l^2} + \frac{\sqrt{d_l}}{h_l})$ bits and one floating point number for $\|x^l\|$. Overall, the number of encoding bits for $\cQ^B_h(x)$ (up to constant and $\log$ factors) can be given by $\sum_{l=1}^B(\frac{1}{h_l^2} + \frac{\sqrt{d_l}}{h_l}) + B$.
As for the compression noise, we prove in the Appendix the following upper bound for $\cL(\cQ_h^B, \mL)$:
\begin{equation}\label{var-bound:block-q}
\squeeze
\cL(\cQ_h^B, \mL) \le \max_{1\le l\le B} h_l\|\diag(\mL^{ll})\|,
\end{equation}
where $\mL^{ll}$ is the $l^{th}$ diagonal block matrix of $\mL$ with sizes $d_l\times d_l$. Next, we are going to minimize communication complexity of DCGD+ and DIANA+ by optimizing parameters of block quantization.

{\bf 4.1. DCGD+ with block quantization.}
We fix the number of blocks $B\in[d]$ for all nodes $i\in[n]$ and allow each node to apply different block quantization operator $\cQ^B_{h_i}$ with quantization steps $h_i = (h_{i,1},\dots,h_{i,B})$. To minimize communication complexity of DCGD+, we need to minimize $\cL_{\max}$ subject to the communication constraint mentioned above. Since $\cL_{\max} = \max_{i\in[n]}\cL(\cC_i,\mL_i)$,  each node $i\in[n]$ can minimize the impact of its own compression by minimizing $\cL(\cC_i,\mL_i)$ based on local smoothness matrix $\mL_i$. This leads to the following optimization problem for finding optimal values of $h_i$ for each node $i\in[n]$:
\begin{equation}\label{meta-opt-3}
\squeeze
\min\limits_{h\in\R^B} \max_{1\le l\le B} h_l\|\diag(\mL_i^{ll})\|, \quad
s.t. \; \sum_{l=1}^B(\frac{1}{h_l^2} + \frac{\sqrt{d_l}}{h_l}) + B = \beta,\; h_l>0,\; l\in[B],
\end{equation}
where $\beta$ is the ``budget'' of communication: Larger $\beta$ leads to finer quantization levels. Note that the constraint in \eqref{meta-opt-3} depends monotonically from each $h_l$. Therefore, the optimum is attained when $h_l\|\diag(\mL_i^{ll})\|$ is uniform over $l\in[B]$. Thus, the solution to this problem is given by
$
h_{i,l} = \frac{\delta_{i,B}}{\|\diag(\mL_i^{ll})\|},
$
where $\delta_{i,B}\ge0$ is uniquely determined by the constraint equality of \eqref{meta-opt-3} as the only positive solution of
$
\squeeze
\delta_{i,B}^2 - \delta_{i,B} \frac{d T_{i,B}}{\beta-B} - \frac{d T_{i,1}^2}{\beta-B} = 0,
$
which implies
$
\delta_{i,B} = \tfrac{d T_{i,B}}{2(\beta-B)} + \sqrt{\tfrac{d^2 T_{i,B}^2}{4(\beta-B)^2} + \tfrac{d T_{i,1}^2}{\beta-B}}
\le \frac{d}{\beta-B}T_{i,B} + \sqrt{\frac{d}{\beta-B}}T_{i,1},
$
where $T_{i,B} \eqdef \frac{1}{d}\sum_{l=1}^B \sqrt{d_l}\|\diag(\mL_i^{ll})\|$.
If this solution of quantization steps $h_i$ is used by all nodes $i\in[n]$, then we show reduction in communication complexity by a factor of $\cO(n)$.

\begin{theorem}\label{thm-prop:block-dcgd+}
Assume $n=\cO(\sqrt{d})$ and both $\nu,\nu_1$ are $\cO(1)$. Then DCGD+ using block quantization with $B=n$ blocks, $d_l = \cO(\nicefrac{d}{n})$ block sizes for all $l\in[n]$ and quantization steps $h_{i,l} = \nicefrac{\delta_{i,B}}{\|\diag(\mL_i^{ll})\|}$ with $\beta = \cO(\nicefrac{d}{n})$ reduces overall communication complexity by a factor of $\cO(n)$ compared to DCGD using $B=1$ single block quantization. Formally, to guarantee $\varepsilon>0$ accuracy, the communication complexity of DCGD+ is $\squeeze \cO\left(\frac{d}{n}\frac{L_{\max}}{\mu}\log\frac{1}{\varepsilon} \right),$ which is $\cO(n)$ times smaller over DCGD.
\end{theorem}

{\bf 4.2. DIANA+ with block quantization.}
For the rate \eqref{DIANA+-complexity} of DIANA+, we need to optimize $\omega_{\max} + \frac{\cL_{\max}}{n\mu}$ part of the complexity under the same communication constraint used in \eqref{meta-opt-3}. Since
\begin{equation}\label{eq:011}
\squeeze
\max_{i\in[n]}\(\omega_{i} + \frac{\cL_{i}}{n\mu}\)
\le
\omega_{\max} + \frac{\cL_{\max}}{n\mu}
\le
2\max_{i\in[n]}\(\omega_{i} + \frac{\cL_{i}}{n\mu}\),
\end{equation}
we can decompose the problem into subproblems for each node $i$ to optimize $\omega_{i} + \frac{\cL_{i}}{n\mu}$ with respect to its own quantization parameters $h_{i}$. Analogously, this leads to the following optimization problem for finding optimal values of $h_i$ for each node $i\in[n]$:
\begin{equation}\label{meta-opt-4}
\squeeze
\min\limits_{h\in\R^B} \max_{1\le l\le B} h_l\(\sqrt{d_l} + \frac{1}{\mu n}\|\diag(\mL_i^{ll})\|\), \;
s.t. \; \sum_{l=1}^B(\frac{1}{h_l^2} + \frac{\sqrt{d_l}}{h_l}) + B = \beta,\; h_l>0,
\end{equation}
which can be solved with a similar argument as done for \eqref{meta-opt-3}. Details are deferred to the Appendix.

\begin{theorem}\label{thm-prop:block-diana+}
Assume $n=\cO(\sqrt{d})$ and both $\nu,\nu_1$ are $\cO(1)$. Then DIANA+ using block quantization with $B=n$ blocks, $d_l = \cO(\nicefrac{d}{n})$ block sizes for all $l\in[n]$ and $h_{i,l}$ quantization steps (solution to \eqref{meta-opt-4}) with $\beta = \cO(\nicefrac{d}{n})$ reduces overall communication complexity by a factor of $\cO(n)$ compared to DIANA using $B=1$ single block quantization. Formally, to guarantee $\varepsilon>0$ accuracy, the communication complexity of DIANA+ is $\squeeze\cO\big(\big(nd + \sqrt{\frac{d}{n}}\frac{L_{\max}}{\mu}\big)\log\frac{1}{\varepsilon} \big),$ which (ignoring $n$ summand in the complexity) is $\cO(n)$ times smaller over DIANA.
\end{theorem}

\section{Quantization with Varying Steps}\label{sec:quant}

Our second extension of standard quantization scheme is to allow different quantization steps for all coordinates $\{1,2,\dots,d\}$. In other words, for each coordinate $j\in[d]$ we quantize with respect to levels $\{0,h_j,2h_j,\dots\}$. The standard quantization \citep{alistarh2017qsgd} is the special case when $h_j = \frac{1}{s}$ for all $j\in[d]$, where $s$ is the number of quantization levels.

\begin{definition}[Quantization with varying steps]\label{def:quant}
For fixed quantization steps $h = (h_1,\dots,h_d)^\top\in\R^d$, define quantization operator $\cQ_h\colon\R^d\to\R^d$ as follows:
\begin{eqnarray*}
\squeeze
\[\cQ_h(x)\]_j = \|x\| \cdot \sign(x_j) \cdot \xi_j\(\frac{|x_j|}{\|x\|}\), \quad x\in\R^d,\; j=1,2,\dots,d,
\end{eqnarray*}
where $\xi_j$ is defined via the quantization levels $\{0,h_j,2h_j,\dots\}$ as in \eqref{quant-xi}.
\end{definition}

Note that compression operator $\cQ_h$ is unbiased as $\E\[\xi_j(v)\] = v$ for any $v\ge0$ and is not a special case of block quantization defined earlier. To understand how the number of encoding bits of $\cQ_h(x)$ depends on $h$ exactly seems challenging, since it depends on the actual encoding scheme (i.e. binary representation of compressed information). Besides, even if we fix binary mapping, the closed form expression of total amount of bits is too complicated to be utilized in the further analysis.
We provide theoretical arguments and clear numerical evidence that $\|h^{-1}\| = \sqrt{\sum_{j=1}^d h_j^{-2}}$ is a reasonable proxy for the number of encoding bits for compressor $\cQ_h$.

\begin{assumption}\label{asm:quant-bits}
For any input vector $x\in\R^d$ and quantization steps $h\in\R^d$, compressed vector $\cQ_h(x)$ can be encoded with $\cO(\|h^{-1}\|)$ number of bits.
\end{assumption}

First, consider the special case when all quantization steps are the same, i.e. $h_j=\frac{1}{s}$. Then $\|h^{-1}\|=s\sqrt{d}$ recovers the dominant part (provided $s=\cO(\sqrt{d})$) in $\widetilde{\cO}(s^2+s\sqrt{d})$ showing total amount of bits for standard quantization scheme.
Second, in the Appendix we present an encoding scheme which (up to constant and $\log d$ factors) requires $\E\[\psi(\|\hat{x}\|_0)\] + \|h^{-1}\|$ number of bits in expectation to communicate $\hat{x} = \cQ_h(x)$, where $\psi(\tau)\eqdef d H_2(\nicefrac{\tau}{d}) + \tau \le d\log3$, if $\tau\in[0,d]$ and $H_2$ is the binary entropy function. Note that, based on the definition \eqref{quant-xi}, increasing quantization steps $h_j$ forces more sparsity in $\hat{x}$ and hence reduces $\|\hat{x}\|_0$. Thus, $\|\hat{x}\|_0$ and hence $\psi(\|\hat{x}\|_0)$ (notice that $\psi(0)=0$) are proportional to $\|h^{-1}\|$. Furthermore, we present a numerical experiment which shows that the number of encoding bits of $\cQ_h(x)$ and $\|h^{-1}\|$ are positively correlated.


Hence, in the further analysis, we fix the number of encoding bits of $\cQ_h(x)$ by the constraint $\|h^{-1}\|=\beta$ for some parameter $\beta>0$.
As for the variance induced by the compression operator $\cQ_h$, we prove the following upper bound for $\cL(\cQ_h,\mL)$:
\begin{equation}\label{eq:002}
\squeeze
\cL(\cQ_h,\mL) \le \|\diag(\mL)h\|.
\end{equation}

{\bf 5.1. DCGD+ with varying quantization steps.}
Now, we optimize the rate \eqref{DCGD+-complexity} of DCGD+ with respect to quantization steps $h_i = (h_{i;1},h_{i;2},\dots,h_{i;d})$ of compressor $\cQ_{h_i}$ controlled by $i^{th}$ node for all $i\in[n]$. The term in \eqref{DCGD+-complexity} affected by the compression is $\cL_{\max} = \max_{i\in[n]}\cL(\cC_i,\mL_i)$, which implies that each node $i\in[n]$ can minimize the impact of its own compression by minimizing $\cL(\cC_i,\mL_i)$ based on local smoothness matrix $\mL_i$.
Based on the upper bound \eqref{eq:002} and communication constraint given by $\|h^{-1}\|=\beta$ for some $\beta>0$, we get the following optimization problem to choose the optimal quantization parameters $h_{i}$ for node $i\in[n]$:
\begin{equation}\label{meta-opt}
\squeeze
\min\limits_{h\in\R^d} \|\diag(\mL_i)h\|, \quad s.t. \; \|h^{-1}\| = \beta,\; h_j > 0,\; j\in[d].
\end{equation}

This problem has the following closed form solution due to KKT conditions (see Appendix):
\begin{equation}\label{level-sols-dcgd}
\squeeze
h_{i;j} = \frac{1}{\beta}\sqrt{\frac{\sum_{t=1}^d \mL_{i;t}}{\mL_{i;j}}}, \quad i\in[n],\; j\in[d].
\end{equation}
With this choice of quantization steps we save $\cO(\min(n,d))$ times in communication.
\begin{theorem}\label{thm-prop:quant-dcgd+}
Assume both $\nu,\nu_1$ are $\cO(1)$ and $\beta = \cO(\nicefrac{d}{n})$. Then DCGD+ using quantization with varying steps \eqref{level-sols-dcgd} for all $i\in[n]$ reduces overall communication complexity by a factor of $\cO(\min(n,d))$ compared to the baseline of DCGD. Formally, the iteration complexity \eqref{DCGD+-complexity} can be upper bounded as
$
\squeeze
\frac{L}{\mu} + \frac{\cL_{\max}}{n\mu}
\le \frac{\nu}{n}\frac{L_{\max}}{\mu} + \frac{\nu_1}{\beta}\frac{L_{\max}}{n\mu}
=   \cO\( \frac{1}{n}\frac{L_{\max}}{\mu} + \frac{1}{d}\frac{L_{\max}}{\mu} \),
$
which is $\min(n,d)$ times smaller than the one for DCGD. As both methods communicate $\cO(\nicefrac{d}{n})$ bits per node per iteration, we get $\min(n,d)$ times savings in communication complexity.
\end{theorem}

{\bf 5.2. DIANA+ with varying quantization steps.}
Based on \eqref{eq:011}, each node $i\in[n]$ optimizes $\omega_{i} + \frac{\cL_{i}}{n\mu}$ with respect to its quantization parameters $h_{i}$, which is equivalent to the problem
\begin{equation}\label{meta-opt-diana+}
\squeeze
\min\limits_{h\in\R^d} \sum_{j=1}^d \(1+A_{ij}^2\)h_{j}^2, \quad s.t. \; \|h^{-1}\| = \beta,\; h_j > 0, j\in[d]
\end{equation}
where $A_{ij} \eqdef \nicefrac{\mL_{i;j}}{n\mu}$. Due to the KKT conditions (see Appendix), we get the following solution
\begin{equation}\label{level-sols-diana-main}
\squeeze
h_{i;j} = \frac{1}{\beta}\sqrt{\frac{\sum_{t=1}^d \sqrt{1+A^2_{it}}}{\sqrt{1+A^2_{ij}}}}.
\end{equation}
With this choice of quantization steps we save $\cO(\min(n,d))$ times in communication.
\begin{theorem}\label{thm-prop:quant-diana+}
Assume both $\nu,\nu_1$ are $\cO(1)$ and $\beta = \cO(\nicefrac{d}{n})$. Then DIANA+ using quantization with varying steps \eqref{level-sols-diana-main} for all $i\in[n]$ reduces overall communication complexity by a factor of $\cO(\min(n,d))$ compared to the baseline of DIANA. Formally, the iteration complexity \eqref{DIANA+-complexity} can be upper bounded as 
$
\squeeze
\omega_{\max} + \frac{L}{\mu} + \frac{\cL_{\max}}{n\mu}
\le \frac{\sqrt{2}d}{\beta} + \frac{\nu}{n}\frac{L_{\max}}{\mu} + \frac{\sqrt{2}\nu_1}{\beta n} \frac{L_{\max}}{\mu}
=   \cO \( n + \frac{1}{n}\frac{L_{\max}}{\mu} + \frac{1}{d}\frac{L_{\max}}{\mu} \),
$
which is $\min(n,d)$ times smaller than the one for DIANA (ignoring negligible term $n$).
\end{theorem}

\section{Experiments}\label{sec:exps}


{\bf 6.1. Setup.}
In this section we present two key experiments. Additional experiments can be found in the Appendix. We conduct a range of experiments with several datasets from the LibSVM repository \citep{chang2011libsvm} on the $\ell_2$-regularized logistic regression problem \eqref{main-opt-problem-dist}:
$$
\squeeze
\min\limits_{x\in\R^d} \frac{1}{n}\sum_{i=1}^n f_i(x),
\quad
f_i(x)= \frac{1}{m}\sum_{t=1}^m\log(1+\exp(- b_{i,t} \mathbf{A}_{i,t}^\top x)) + \frac{\lambda}{2}\|x\|^2,
$$
where $\mathbf{A}_{i,t}$ are data points sorted based on their norms before allocating to local workers for the heterogeneity. The experiments are performed on a workstation with Intel(R) Xeon(R) Gold 6246 CPU @ 3.30GHz cores. The \texttt{gather} and \texttt{broadcast} operations for the communications between master and workers are implemented based on the MPI4PY library \citep{dalcin2005mpi} and each CPU core is treated as a local worker. For each dataset, we run each algorithm multiples times with 5 random seeds for each worker. 
Due to space limitations, we present only two of our experiments here deferring the remaining experiments along with experimental details in the Appendix.

\begin{figure*}[htp] 
  \minipage{0.25\textwidth}
  \includegraphics[width=\linewidth]{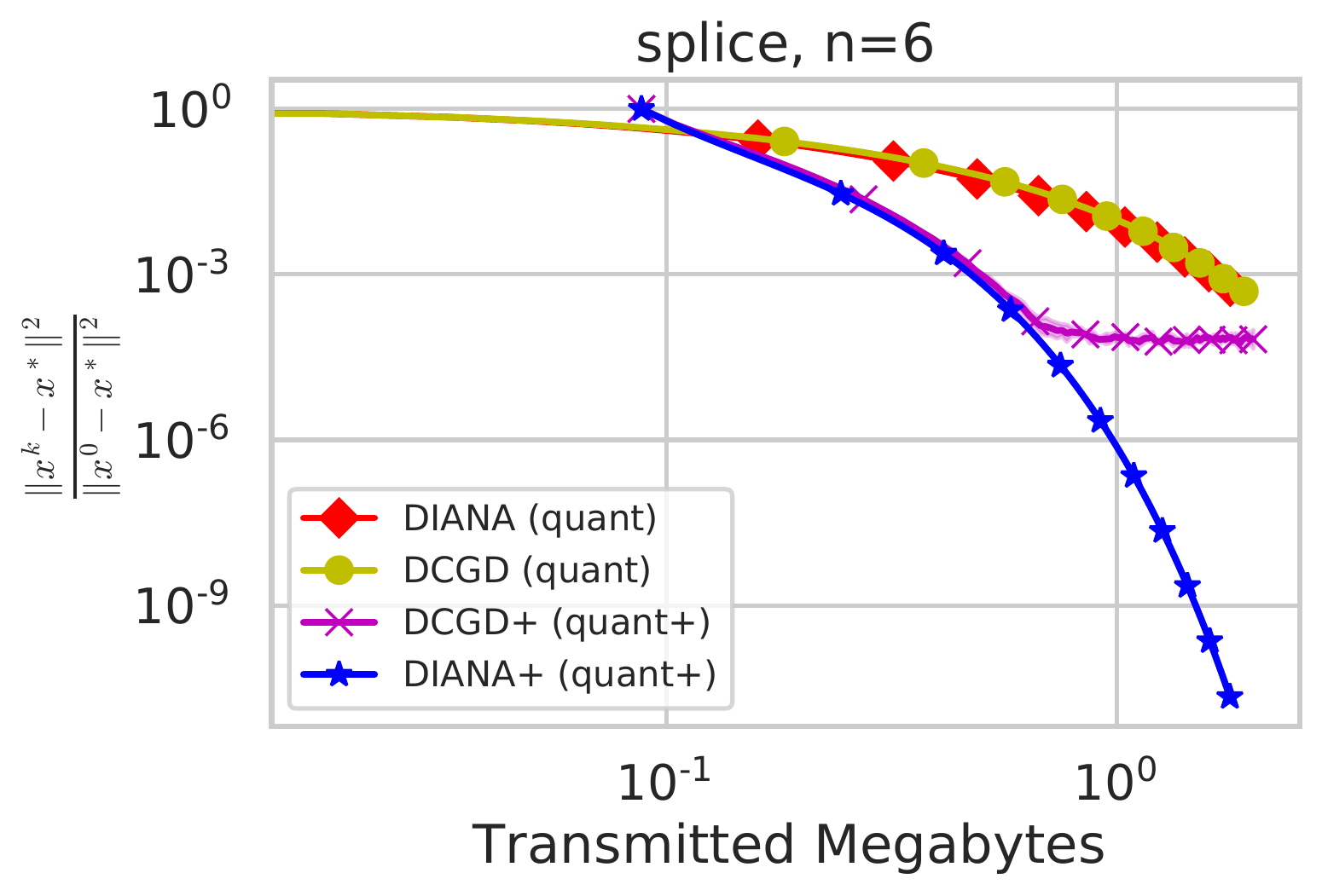}
  \endminipage\hfill
  \minipage{0.25\textwidth}
  \includegraphics[width=\linewidth]{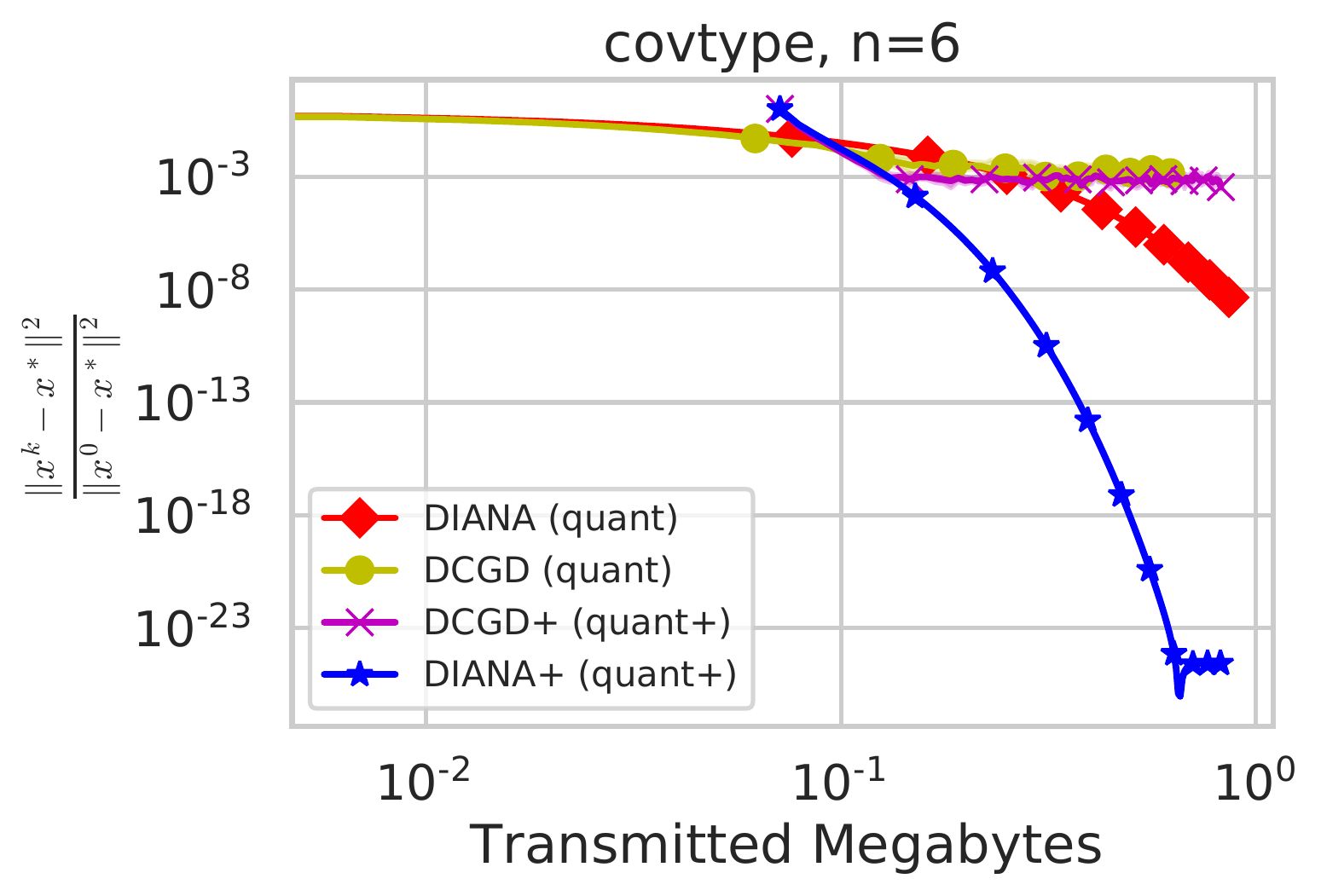}
  \endminipage\hfill
  \minipage{0.25\textwidth}
  \includegraphics[width=\linewidth]{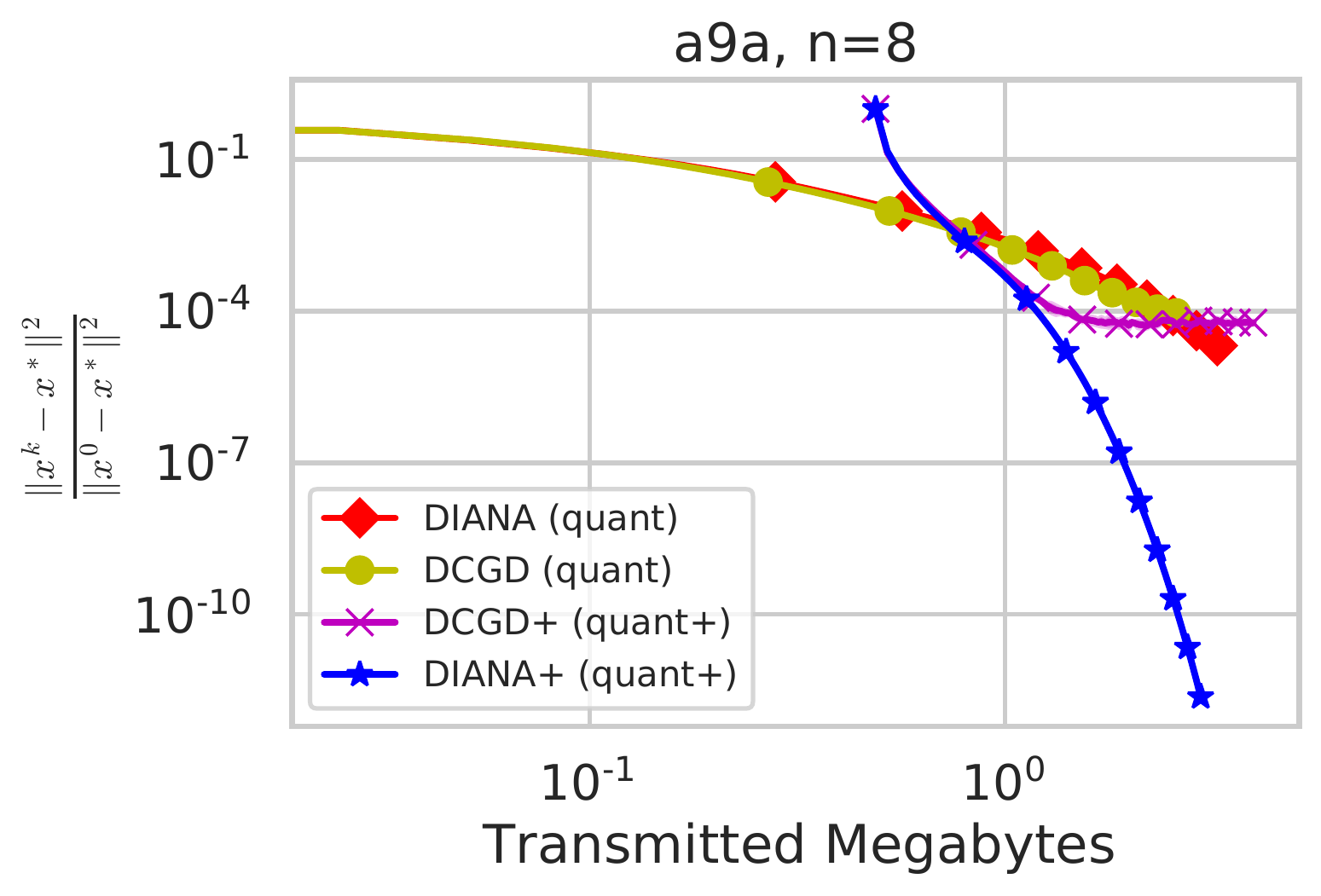}
  \endminipage\hfill
  \minipage{0.25\textwidth}
  \includegraphics[width=\linewidth]{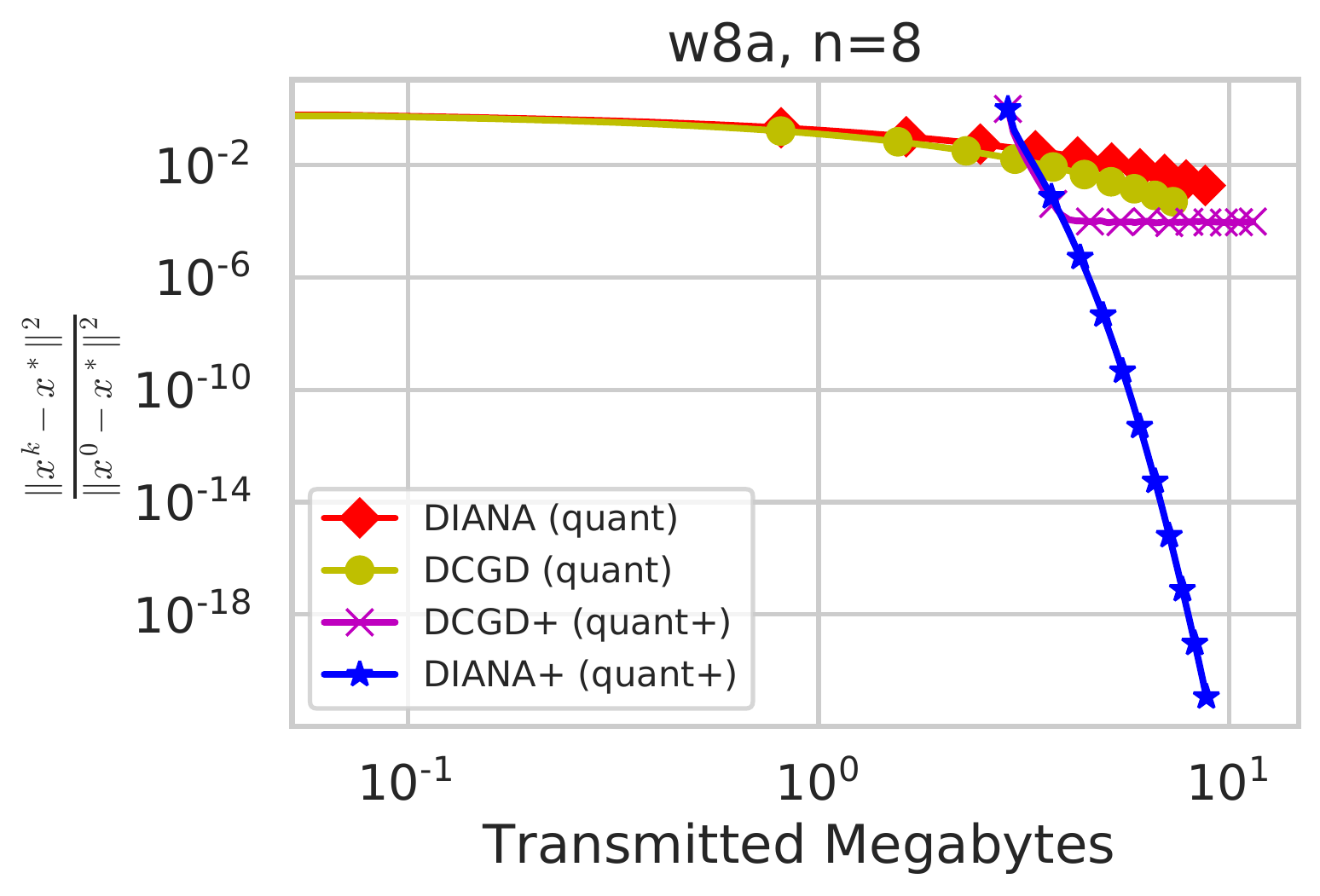}
  \endminipage\hfill
  \minipage{0.25\textwidth}
  \includegraphics[width=\linewidth]{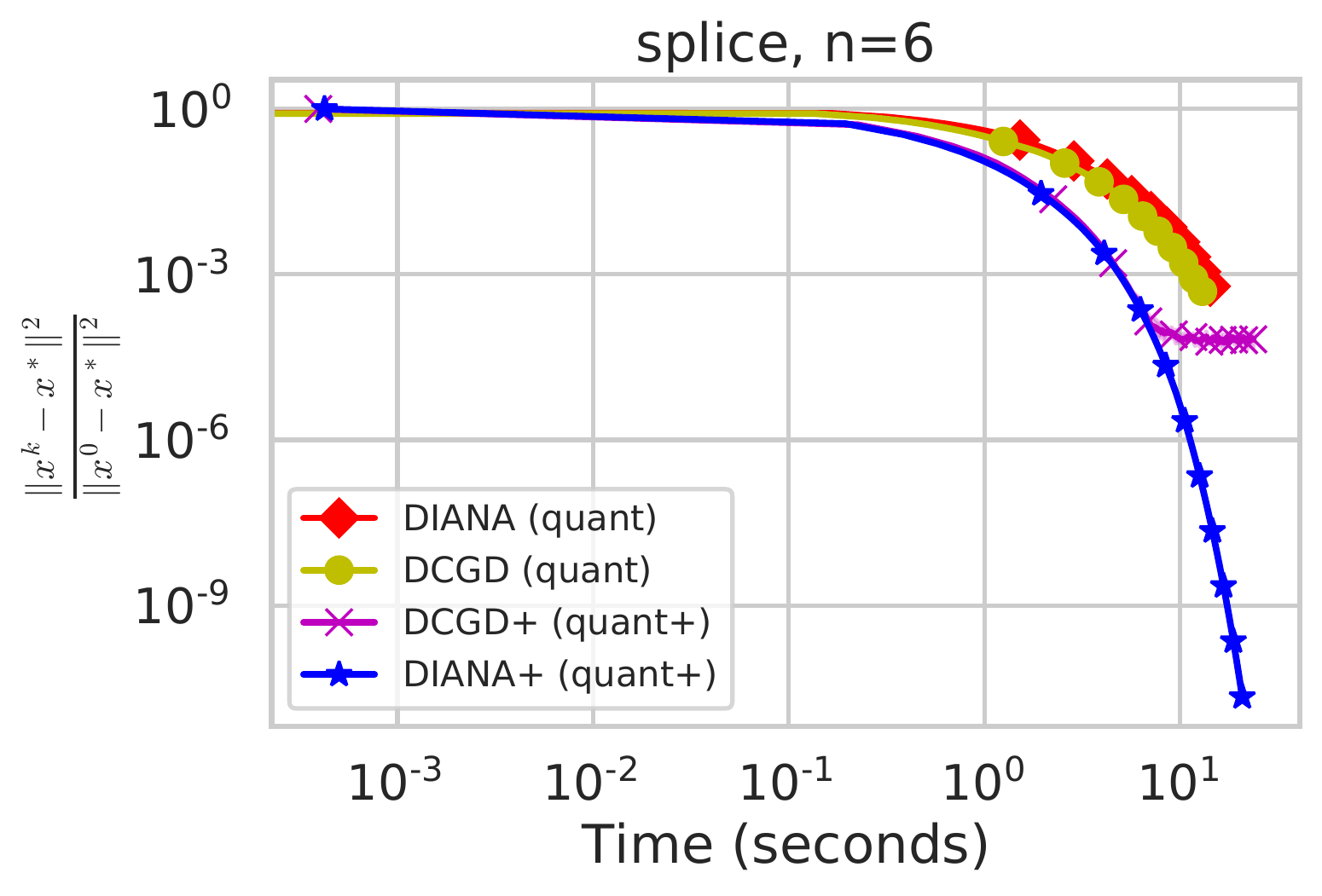}
  \endminipage\hfill  
  \minipage{0.25\textwidth}
  \includegraphics[width=\linewidth]{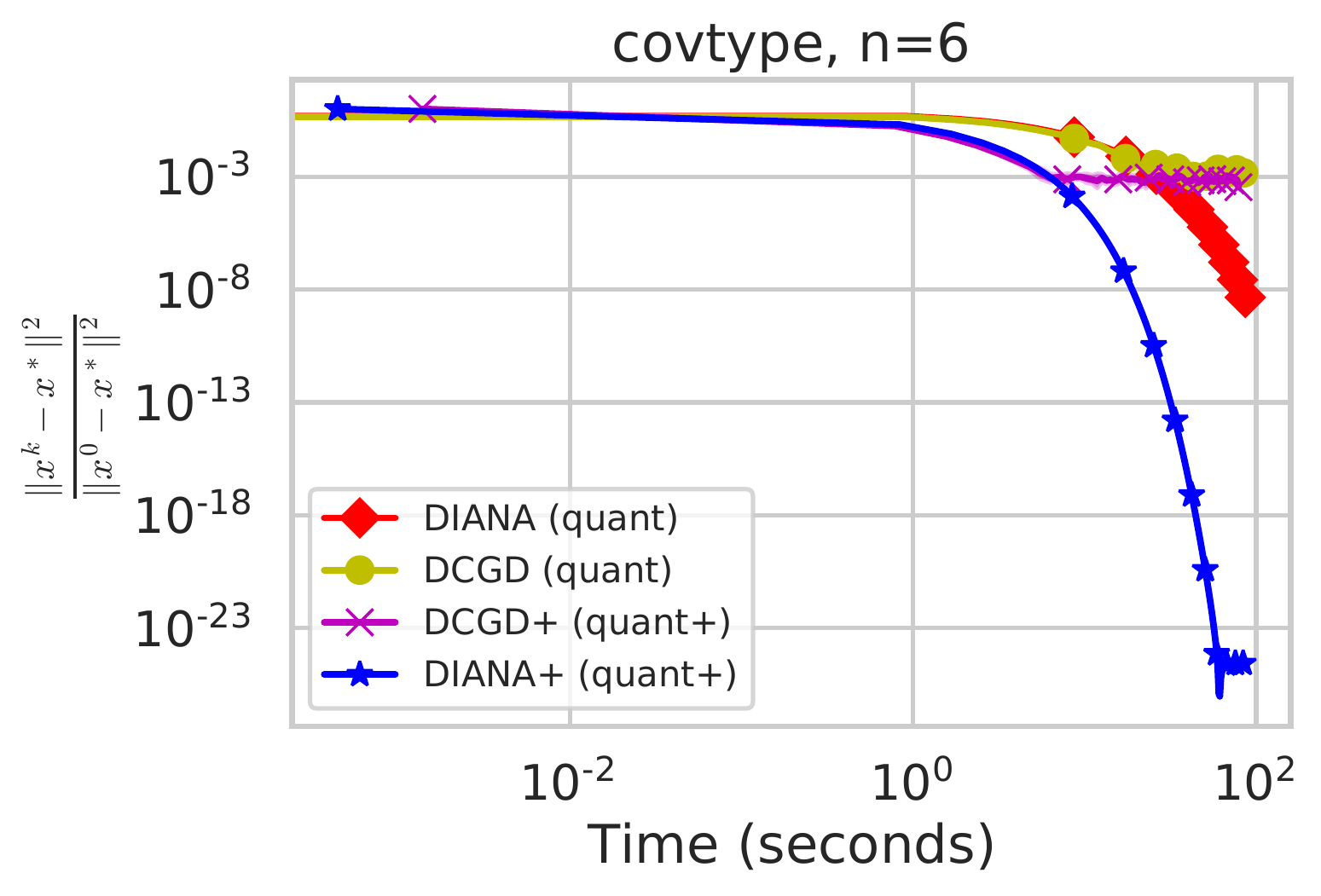}
  \endminipage\hfill 
  \minipage{0.25\textwidth}
  \includegraphics[width=\linewidth]{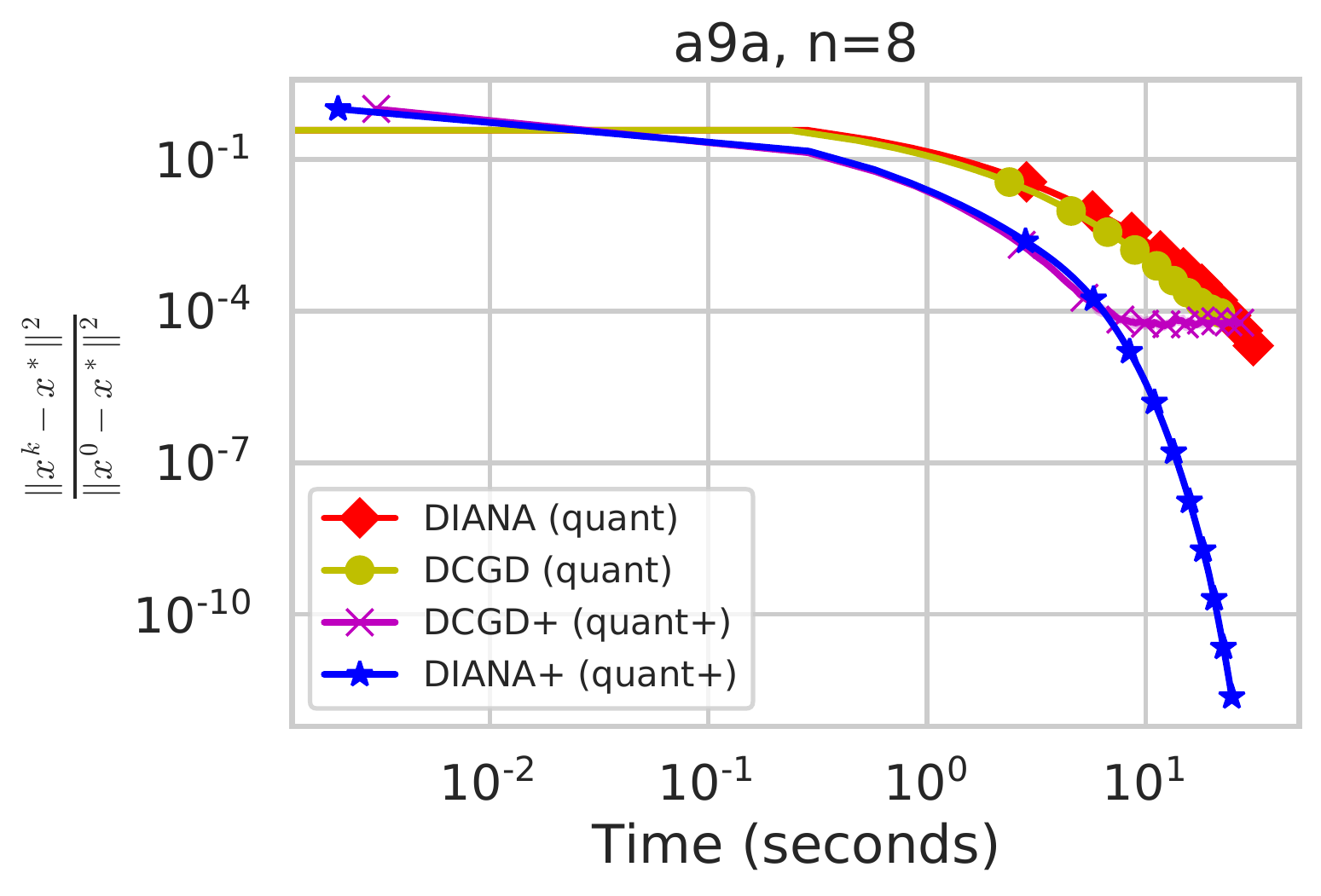}
  \endminipage\hfill  
  \minipage{0.25\textwidth}
  \includegraphics[width=\linewidth]{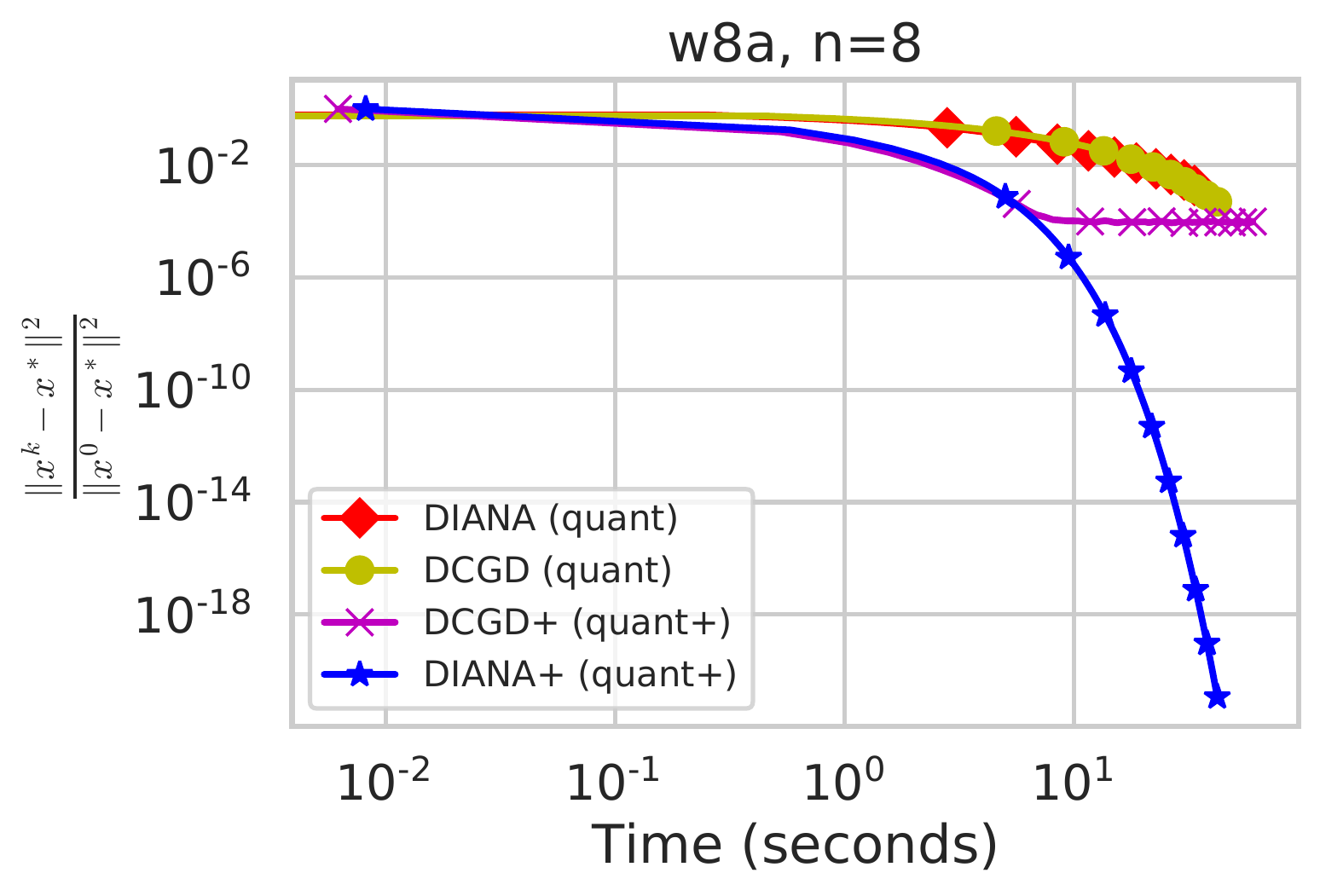}
  \endminipage\hfill  
  \caption{Comparison of smoothness-aware DCGD+/DIANA+ methods with varying-step quantization (\texttt{quant+}) to original DCGD/DIANA methods with standard quantization (\texttt{quant}). Note that in \texttt{quant+} workers need to send $\mL_i^{1/2}\in\R^{d\times d}$ and quantization steps $h_i\in\R^{d}$ to the master before the training. This leads to extra costs in communication bits and time, which are taken into consideration.}
  \label{fig-main:varying}
\end{figure*}

{\bf 6.2. Comparison to standard quantization techniques.}
In our first experiment, we compare smoothness-aware DCGD+ and DIANA+ methods with our varying-step quantization technique (\texttt{quant+}) to the original DCGD \citep{KFJ} and DIANA \citep{MGTR} methods with the standard quantization technique (\texttt{quant}) of \cite{alistarh2017qsgd}. Figure~\ref{fig-main:varying} demonstrates that DCGD+/DIANA+ with \texttt{quant+} lead to significant improvement in both transmitted megabytes and wall-clock time. An ablation study to disentangle the contributions of exploiting the smoothness matrix and utilizing varying number of levels can be found in Appendix \ref{sec:extra-exps}.

\begin{figure*}[htp]  
  \minipage{0.25\textwidth}
  \includegraphics[width=\linewidth]{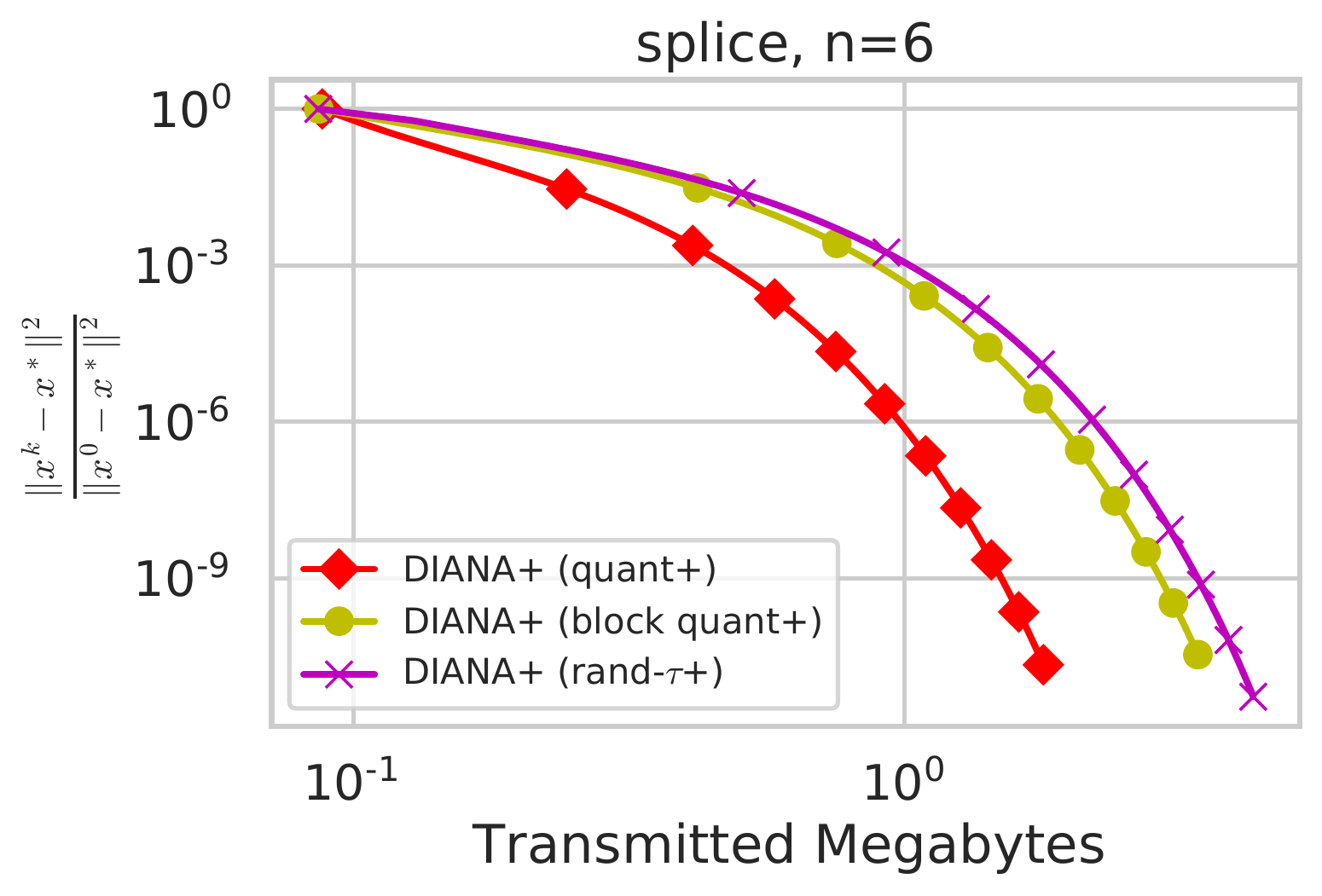}
  \endminipage\hfill  
  \minipage{0.25\textwidth}
  \includegraphics[width=\linewidth]{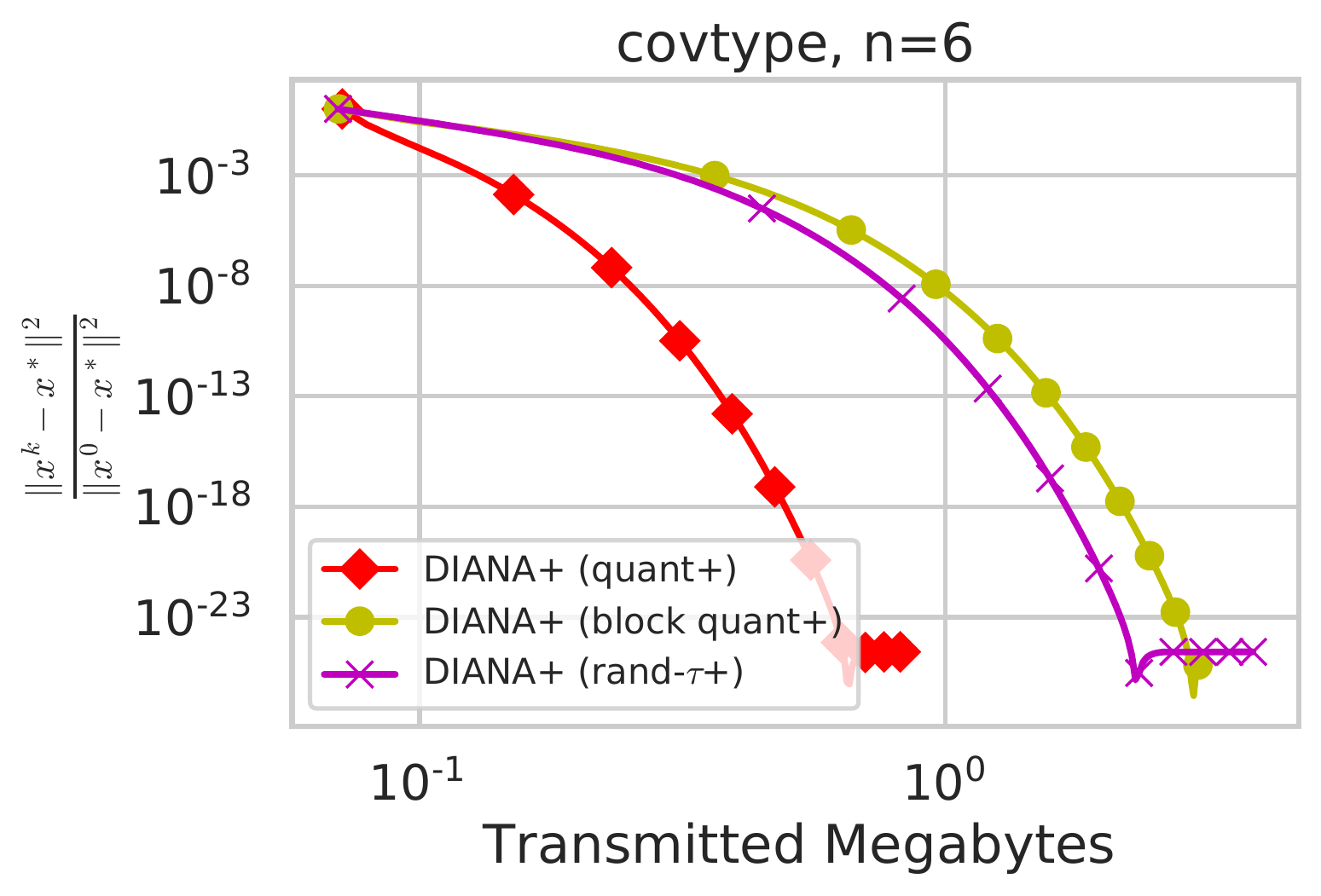}
  \endminipage\hfill  
  \minipage{0.25\textwidth}
  \includegraphics[width=\linewidth]{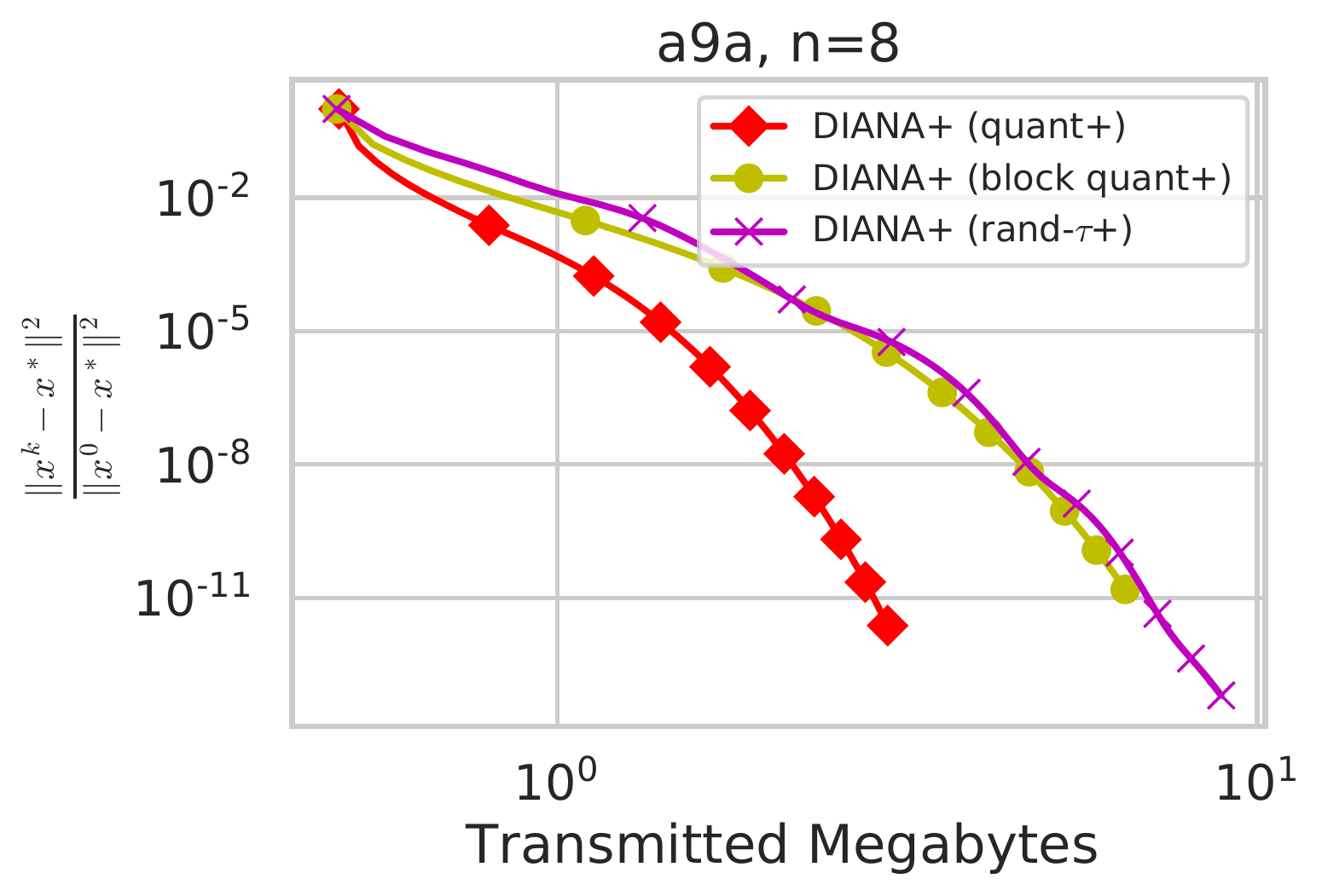}
  \endminipage\hfill  
  \minipage{0.25\textwidth}
  \includegraphics[width=\linewidth]{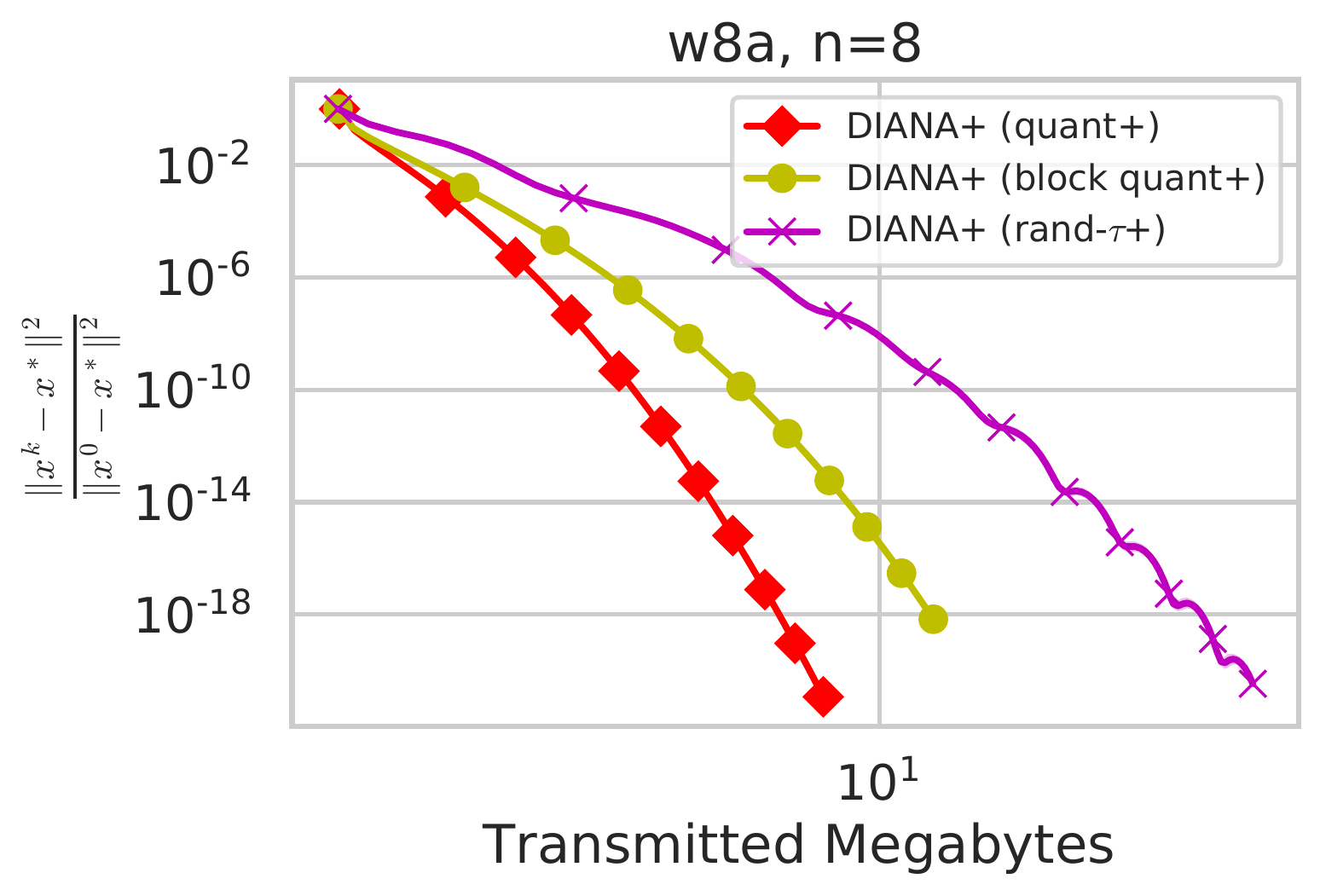}
  \endminipage\hfill  
  \minipage{0.25\textwidth}
  \includegraphics[width=\linewidth]{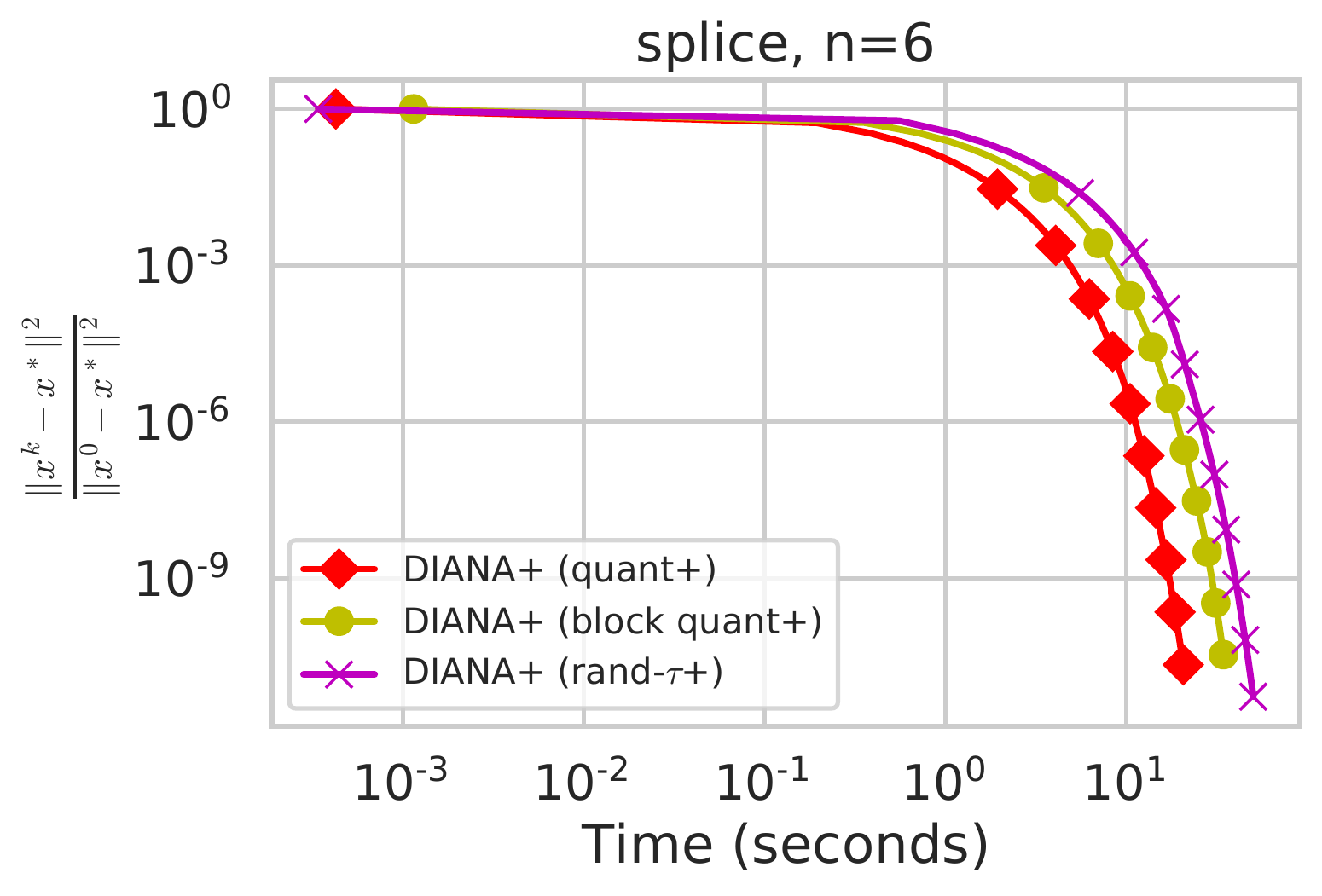}
  \endminipage\hfill  
  \minipage{0.25\textwidth}
  \includegraphics[width=\linewidth]{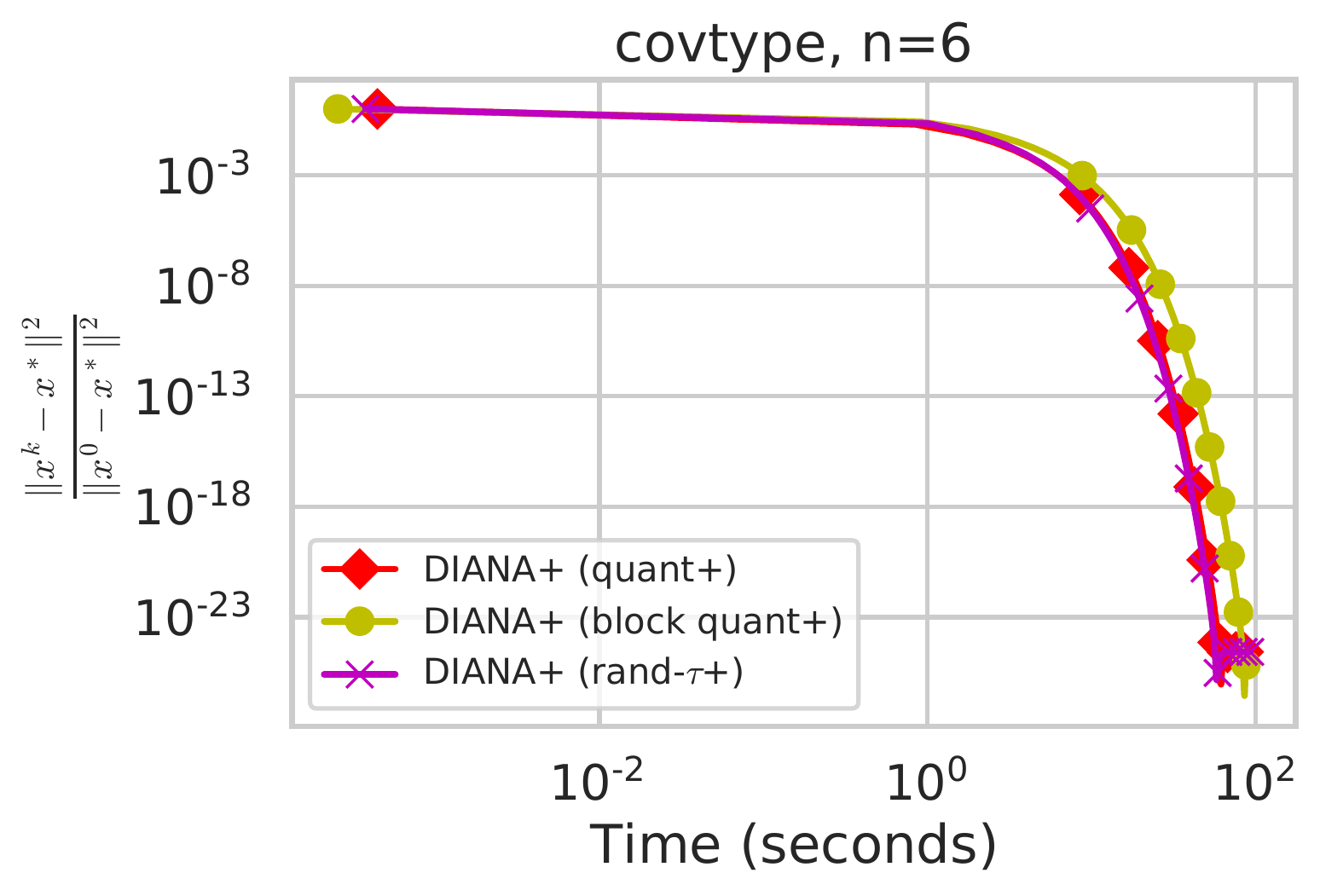}
  \endminipage\hfill  
  \minipage{0.25\textwidth}
  \includegraphics[width=\linewidth]{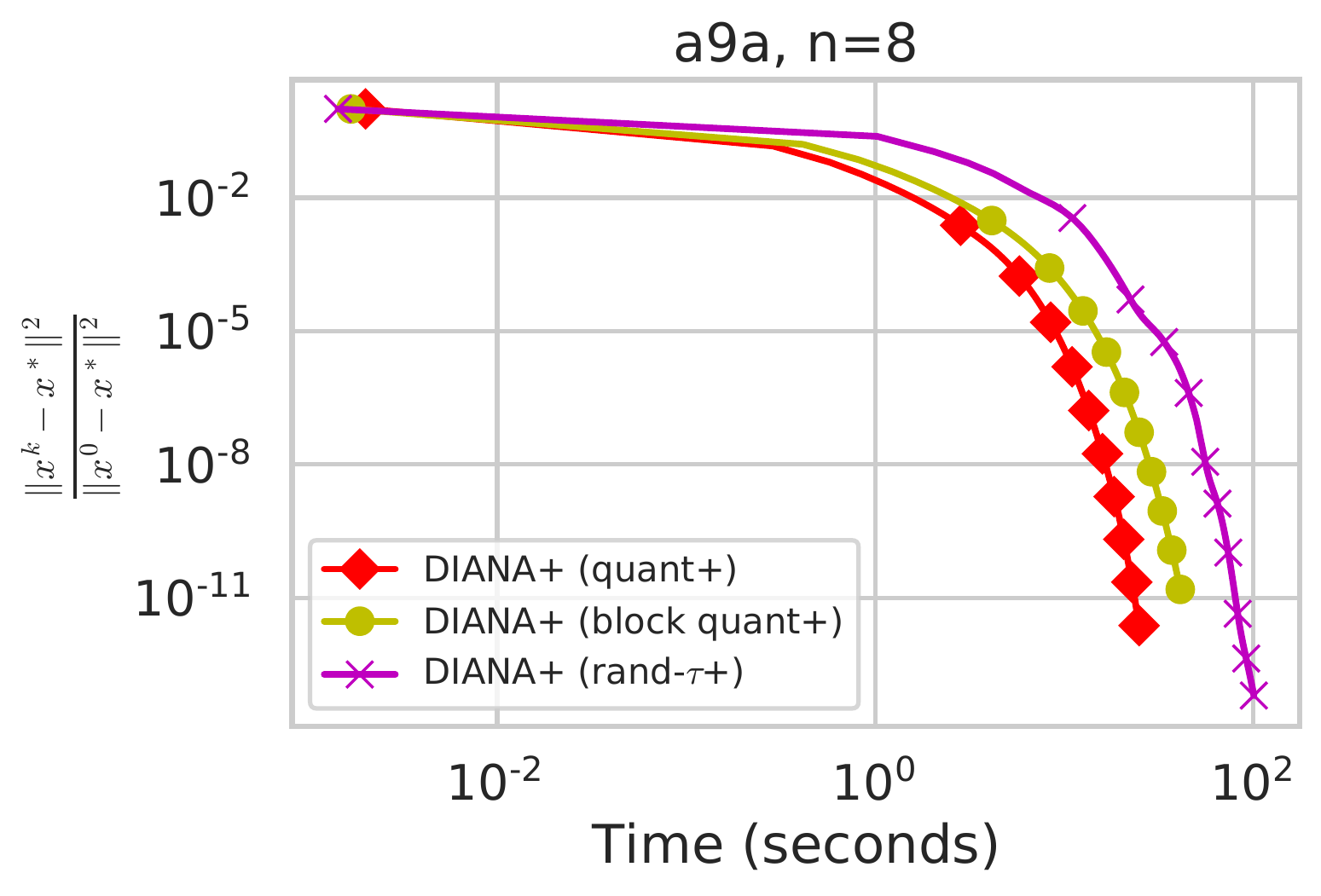}
  \endminipage\hfill  
  \minipage{0.25\textwidth}
  \includegraphics[width=\linewidth]{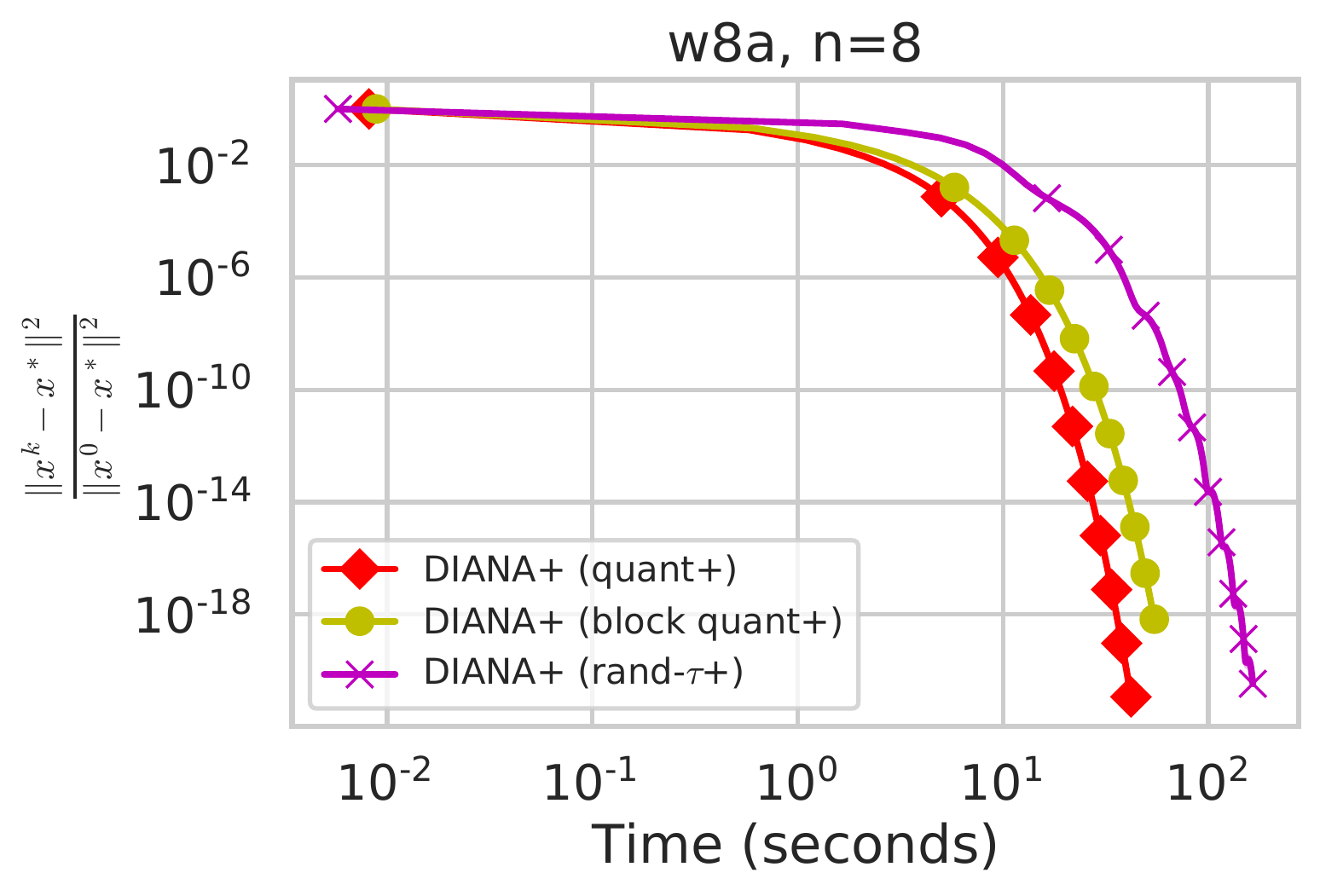}
  \endminipage\hfill  
  \caption{Comparison of three matrix-smoothness-aware compression techniques employed in DIANA+ method: varying-step quantization \texttt{quant+}, our variant of block quantization \texttt{block quant+}, and smoothness-aware sparsification \texttt{rand-$\tau$+} of \citet{safaryan2021smoothness}. }
  \label{fig-main:sp_vs_quant}
\end{figure*}

{\bf 6.3. Comparison to matrix-smoothness-aware sparsification.}
Second experiment is devoted to the performance of three smoothness-aware compression techniques ---block quantization (\texttt{block quant+}) of Section~\ref{sec:block-quant}, varying-step quantization (\texttt{quant+}) of Section~\ref{sec:quant} and smoothness-aware sparsification strategy (\texttt{rand-$\tau$+}) of \cite{safaryan2021smoothness}. All three compression techniques are shown to outperform the standard compression strategies by at most $\cO(n)$ times in theory. For the sparsification, we use the optimal probabilities and the sampling size $\tau = d/n$ as suggested in Section 5.3 of \citep{safaryan2021smoothness}. The empirical results in Figure~\ref{fig-main:sp_vs_quant} illustrate that the varying-step quantization technique (\texttt{quant+}) is always better than the smoothness-aware sparsification \citep{safaryan2021smoothness}, in terms of both communication cost and wall-clock time. Our block quantization technique also beats sparsification when the dimension of the model is relatively~high.

\bibliography{references}
\bibliographystyle{plainnat}
\clearpage

\clearpage

 \tableofcontents

\clearpage

\appendix
\part*{Appendix}

\section{Conclusions and Limitations}

In this work we extended the matrix-smoothness-aware sparsification strategy of \cite{safaryan2021smoothness} to arbitrary unbiased compression schemes. This significantly broadens the use of smoothness matrices in communication efficient distributed methods. 

\subsection{Generalization and quantization}
It is worth to mention that our results generalize those of \cite{safaryan2021smoothness} in a tight manner. That is, we recover the same convergence guarantees as a special case. Indeed, if compression operators $\cC_i$ are diagonal sketches $\MC_i$ generated independently from others and via arbitrary samplings, then
\begin{eqnarray*}
\cL_i
&=& \cL(\MC_i, \mL_i) \\
&=& \inf\left\{\cL\ge0 \colon \E\[\|\MC_i x - x\|^2_{\mL_i}\] \le \cL\|x\|^2 \; \forall x\in \R^d\right\} \\
&=& \inf\left\{\cL\ge0 \colon x^\top \E\[(\MC_i-\MI)\mL_i(\MC_i-\MI)\]x \le \cL\|x\|^2 \; \forall x\in \R^d\right\} \\
&=& \lambda_{\max}\( \E\[(\MC_i-\MI)\mL_i(\MC_i-\MI)\] \) \\
&=&   \lambda_{\max}\( \E\[\MC_i\mL_i\MC_i\] - \mL_i \) \\
&=& \lambda_{\max}(\overbar{\MP}_i \circ \ML_i - \ML_i) \\
& = &   \lambda_{\max}(\widetilde{\MP}_i \circ \ML_i),
\end{eqnarray*}
with the same probability matrices $\overbar{\MP}_i$ and $\widetilde{\MP}_i$ defined in \citep{safaryan2021smoothness}.

Further, we designed two novel quantization schemes (see Definitions \ref{def:block-quant} and \ref{def:quant}) capable of  properly utilizing matrix smoothness information of local loss functions in distributed optimization. We showed that the proposed quantization schemes can significantly outperform the key baselines both in theory and practice.

\subsection{Technical contributions}

We make  {\em two main technical contributions.}

First, we introduce the quantity $\cL(\fC,\mL)$ that properly captures {\em non-linear interaction} between the compressor $\fC$ and the smoothness matrix $\mL$. Due to the linearity of the sparsification (i.e., $\fC(x) = \mC x$), in previous work (Safaryan et al., 2021) it is easy to separate the sparsifier from the compressed gradient and combine it with the smoothness matrix:  $$\|\mL^{\half}\cC(\mL^{\dagger \half}\nabla f(x))\|^2 = \nabla f(x)^\top \mL^{\dagger \half} (\mC \mL\mC) \mL^{\dagger \half} \nabla f(x),$$ where $\mC \mL\mC$ shows a {\em linear interaction} between  $\mC$ and $\mL$. Once we came up with the proper notion of $\cL(\fC,\mL)$ (we had other approaches before we found the "right" one), the proofs of Theorems 1 and 2 followed standard steps. Note that $\cL(\fC,\mL)$ {\em recovers the previous quantity} when the compressor is specialized to a sparsifier (see Sec A.1). This contribution may seem simple from hindsight, but it is not.

Our second technical contribution is the introduction of {\em two non-linear compressors} that provably benefit from smoothness matrices. Specifically, we formulate and analytically solve 4 intermediate optimization problems (11), (14), (16), (18) to find out the best parameter setting for each quantization scheme based on the smoothness information. In fact, sections 4 and 5 outline the technical difficulties we managed to overcome in order to get $\min(n,d)$ speedup factors in each case. {\em Our key contribution is the proposal of these two modified quantization schemes.} We adapt methods DCGD+ and DIANA+ to showcase the potential of our quantization strategies in reducing communication complexity. We chose DCGD as it is the simplest gradient type method with communication compression, and DIANA as it is the variance-reduced version of DCGD. Of course, one can apply our quantization techniques to other distributed methods and gain similar improvements (see Sec A.2). However, we are not attempting to (and can't) be exhaustive in this direction as there are many methods in the literature employing communication compression.

\subsection{Limitations and possible workarounds}\label{sec:limits}

Next, we discuss main limitations of our work.

\begin{itemize}

\item Note while in this paper we redesigned only two methods, DCGD+ and DIANA+,  the modifications we suggest are not limited to these two methods and can be applied to other distributed methods. In particular, with a similar proof technique, ADIANA+ method of \cite{safaryan2021smoothness} introduced with sparsification can also be extended to arbitrary unbiased compression operator using the new notion of $\cL(\fC,\mL)$.

\item The computation or estimation of the smoothness matrix $\mL_i$ requires addiotional preprocessing. For generalized linear models (GLM) (e.g., linear/logistic regression, SVM with smooth hinge loss) the matrix $\mL_i$ can be {\em written in closed form using the local dataset} (see Lemma 1 of \citep{safaryan2021smoothness}). For example, $$\mL_i = \frac{1}{4m_i}\sum_{m=1}^{m_i}\mA_{im}^\top\mA_{im}$$ for logistic regression, where $\{\mA_{im} \colon m=1,\dots,m_i\}$ is the local data of device $i$. Beyond GLMs, $\mL_i$ can be difficult to compute. Note that we do not claim that the proposed method would be practical for high-dimensional deep learning problems - but perhaps this will be overcome in future research. One possibility is to treat $\mL_i$'s as hyper-parameters and learn some {\em rough approximations of the smoothness matrices from the first order information obtained by running a gradient type method}. This can be done initially as a preprocessing step, after which the matrices are considered ``learned'', and then our compression can be built and used.

\item The server is required to store $d\times d$ matrices $\ML_i^{\half}$ for all nodes $i\in[n]$ and multiply them by sparse updates $\fC_i^k(\mL_i^\phalf\nabla f_i(x^k))$ in each iteration. Moreover, each node $i$ is required to store only its smoothness matrix $\ML_i^{\phalf}$ and perform multiplication $\mL_i^\phalf\nabla f_i(x^k)$ in each iterate. Hence, our methods are practical when either dimension $d$ is not too big or smoothness matrices $\mL_i$ are of special structure (e.g., diagonal, low-rank).

\item We did not analyze the compression of the smoothness matrix before communication as it is transferred {\bf only once} before the training begins. Besides, we showed in our experiments that the overhead in communication cost is negligible when the number of iterations is large (the transmitted megabytes do not start from 0 in our plots).

However, in practice, compressing the matrix $\mL$ is a good idea. One option for that is to initially estimate a diagonal smoothness matrix that is as easy to communicate (still {\bf only once}) as one full precision gradient. Another option is to directly apply compression to the matrix $\mL$ so that the compressed matrix is an over-approximation. For example, let $\mL = \sum_{k=1}^d \lambda_k u_k u_k^\top$ be the eigendecomposition of $\mL$, where $\lambda_k$ is the $k^{th}$ largest eigenvalue corresponding to eigenvector $u_k$. Then
$$
\mL
\preceq \sum_{k=1}^r \lambda_k u_k u_k^\top + \sum_{k=r+1}^d \lambda_{r+1} u_k u_k^\top
= \sum_{k=1}^{r} \(\lambda_k - \lambda_{r+1}\) u_k u_k^{\top} + \lambda_{r+1} \mI.
$$

The latter over-approximation (which serves as a smoothness matrix for $f$) can be transferred with $rd+1$ floats where $r$ can be chosen small.

\item For the sake of presentation, we analyzed both DCGD+ and DIANA+ when exact local gradients, $\nabla f_i$, can be computed by all nodes in each iteration. However, we believe that it is possible to extend the analysis to stochastic local gradient oracles. Current tools handling stochastic gradients can be easily applied to our matrix-smoothness-aware compression techniques.

\item In our distributed methods we only compress uplink communication from nodes to the server, which is typically more bandwidth limited than downlink communication from the server to nodes. We believe that techniques that ensure compressed communication in both directions can be applied in our setting, too.

\item We developed all our theory for strongly convex objectives. Extending the theory to convex and non-convex problems in a tight manner seems to be  more challenging.
\end{itemize}

\section{Additional Experiments}\label{sec:extra-exps}

In this section we provide additional experiments to highlight effectiveness of our approach.

\subsection{Setup}
We run the experiments with several datasets listed in Table \ref{tab:exp_stats} from the LibSVM repository \citep{chang2011libsvm} on the $\ell_2$-regularized logistic regression problem described below:
$$
\min\limits_{x\in\R^d} \frac{1}{n}\sum_{i=1}^n f_i(x),
\quad \text{where} \quad
f_i(x)= \frac{1}{m}\sum_{t=1}^m\log(1+\exp(- b_{i,t} \mathbf{A}_{i,t}^\top x)) + \frac{\lambda}{2}\|x\|^2,
$$
where $x\in\R^d$, $\mathbf{A}_{i,l}\in\R^d$, $b_{i,l} \in \{-1,1\}$ are the feature and label of $l$-th data point on the $i$-th worker, where the features of each $\mathbf{A}_{i,l}$ are rescaled into $[-1,1]$. The data points are sorted based on their norms before allocating to local workers to ensure that the data split is heterogeneous. The experiments are performed on a workstation with Intel(R) Xeon(R) Gold 6246 CPU @ 3.30GHz cores. The \texttt{gather} and \texttt{broadcast} operations for the communications between master and workers are implemented based on the MPI4PY library \citep{dalcin2005mpi} and each CPU core is treated as a local worker. We set $\lambda = 10^{-3}$ for all datasets. For each dataset, we run each algorithm multiples times with 5 random seeds for each worker.

\begin{table}[htp]
  \caption{Information of the experiments on $\ell_2$-regularized logistic regression.}
  \renewcommand\arraystretch{1.8}
  \centering 
  \begin{tabular}{ccccc}\toprule[.1em]
    Dataset  & \#Instances $N$ & Dimension $d$  & \#Workers $n$ & \#Instances/worker $m$\\\midrule[.1em]
    german & 1,000 & 24 & 4 & 250\\
    svmguide3 & 1,243 & 21 & 4 & 310\\
    covtype & 581,012 & 54 & 6 &145,253\\
        splice & 1,000 & 60 & 6 & 166\\
     w8a & 49,749 & 300 & 8 & 6,218\\
     a9a & 22,696& 123 & 8 & 2,837\\
    \bottomrule[.1em]
  \end{tabular}
  \label{tab:exp_stats}
\end{table}

To implement Elias encoding and decoding, we utilize the EliasOmega library\footnote{\url{https://gist.github.com/robertofraile/483003}}. We compare the relative errors of different algorithms with respect to 3 measures: the number of iterations, the transmitted megabytes and wall-clock time. To be specific, the measured wall-clock time includes 1) the time of computation on each local worker in one iteration (e.g., local gradient computation, matrix multiplication, etc.); 2) the time of Elias coding and decoding; 3) the time of communication (\texttt{gather} and \texttt{broadcast}). It is worth noting that DCGD+ and DIANA+ require extra cost to transmit $\mL_i^{\half}$ beforehand. Moreover, when coupled with varying number of quantization levels, they also need to transmit $h_{i;j}$ before the start of training. These overheads are taken into consideration in our experimental results.

\subsection{Comparison to standard quantization techniques}

First, we compare DCGD+/DIANA+ with the block quantization technique (\texttt{block quant+}) described in Section~\ref{sec:block-quant} to DCGD \citep{KFJ}/DIANA \citep{MGTR} with the standard quantization technique (\texttt{quant}) in \citep{alistarh2017qsgd}. As shown in Figure~\ref{fig:varying}, DCGD+ (\texttt{block quant+}) and DIANA+ (\texttt{block quant+}) outperform DCGD (\texttt{quant}) and DIANA (\texttt{quant}) when $d$ is larger. This is understandable because the extra cost on communication $B$ norms becomes neglectable when the dimension is relatively high given the number of blocks, where splitting the whole parameters into blocks makes more sense.

\begin{figure*}[htp]
    \minipage{0.30\textwidth}
  \includegraphics[width=\linewidth]{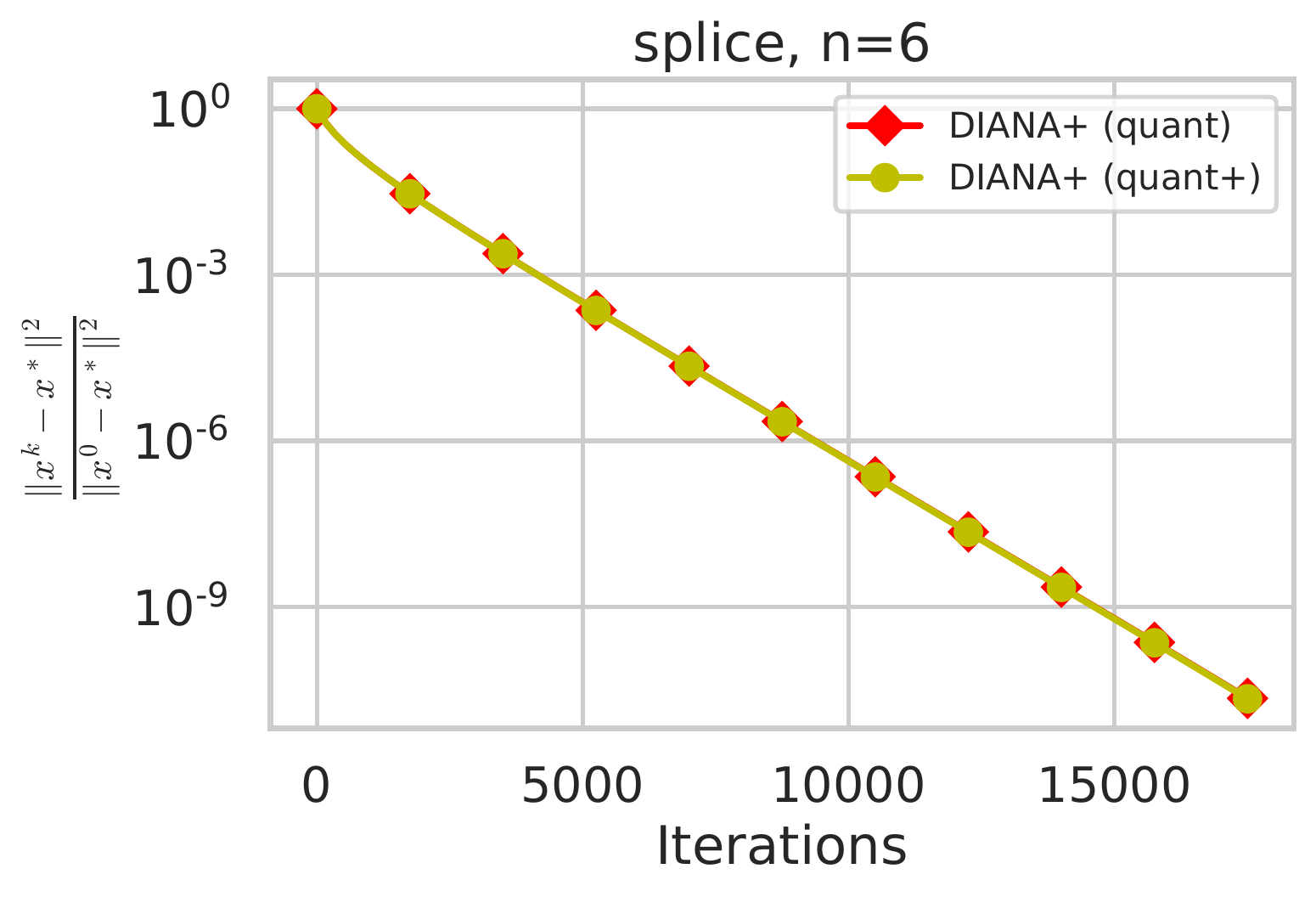}
  \endminipage\hfill  
   \minipage{0.30\textwidth}
  \includegraphics[width=\linewidth]{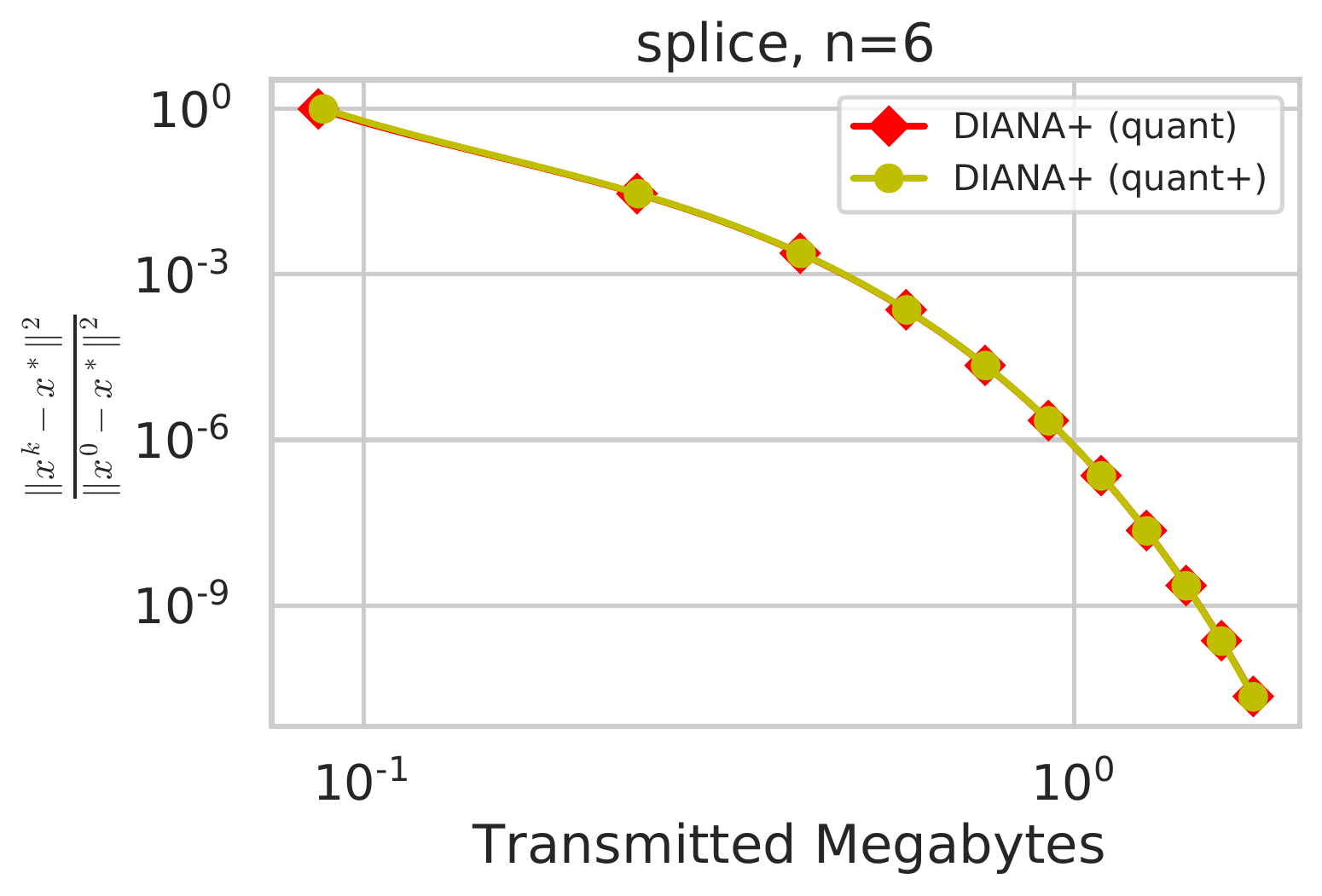}
  \endminipage\hfill  
  \minipage{0.30\textwidth}
  \includegraphics[width=\linewidth]{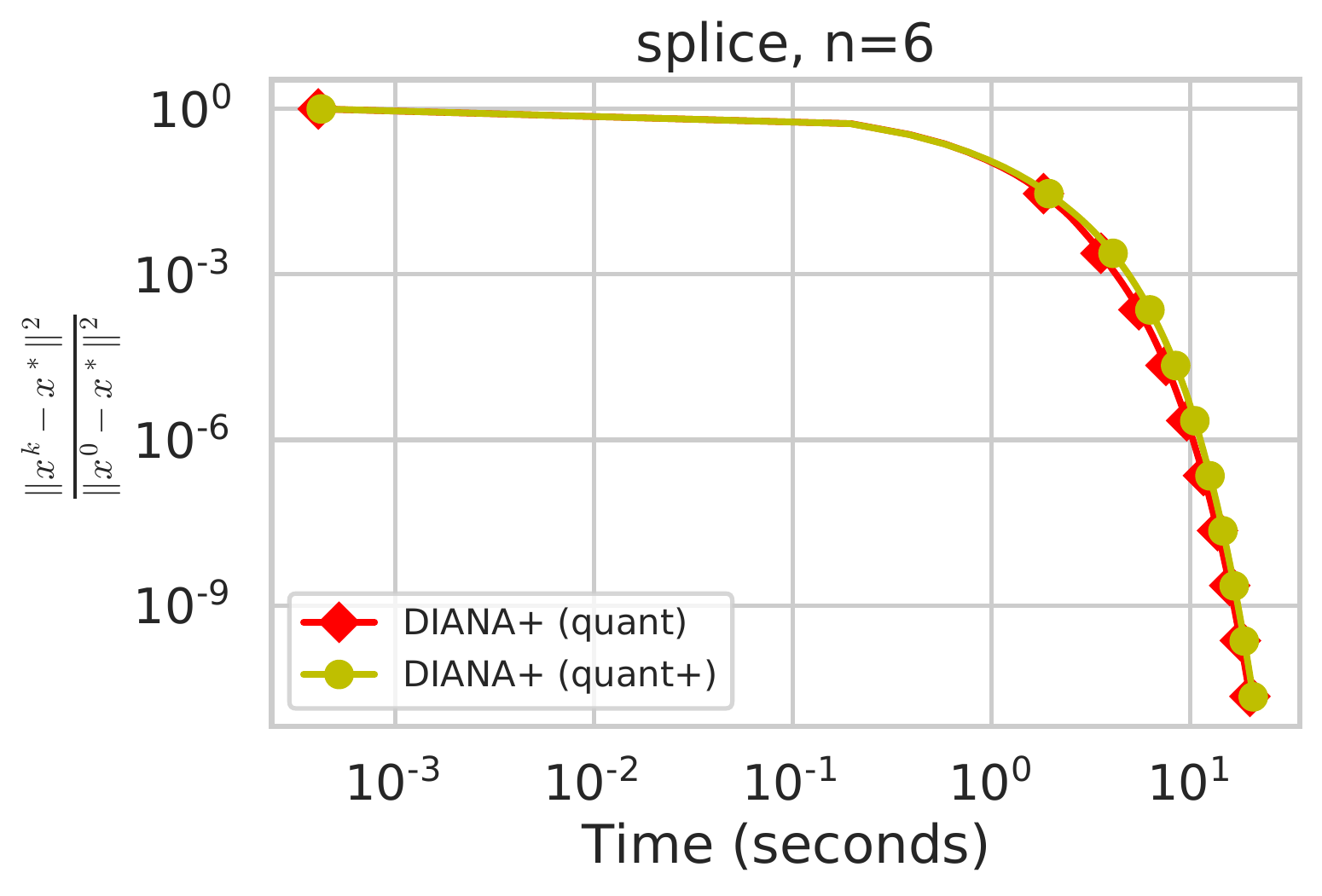}
  \endminipage\hfill  
  
  \minipage{0.30\textwidth}
  \includegraphics[width=\linewidth]{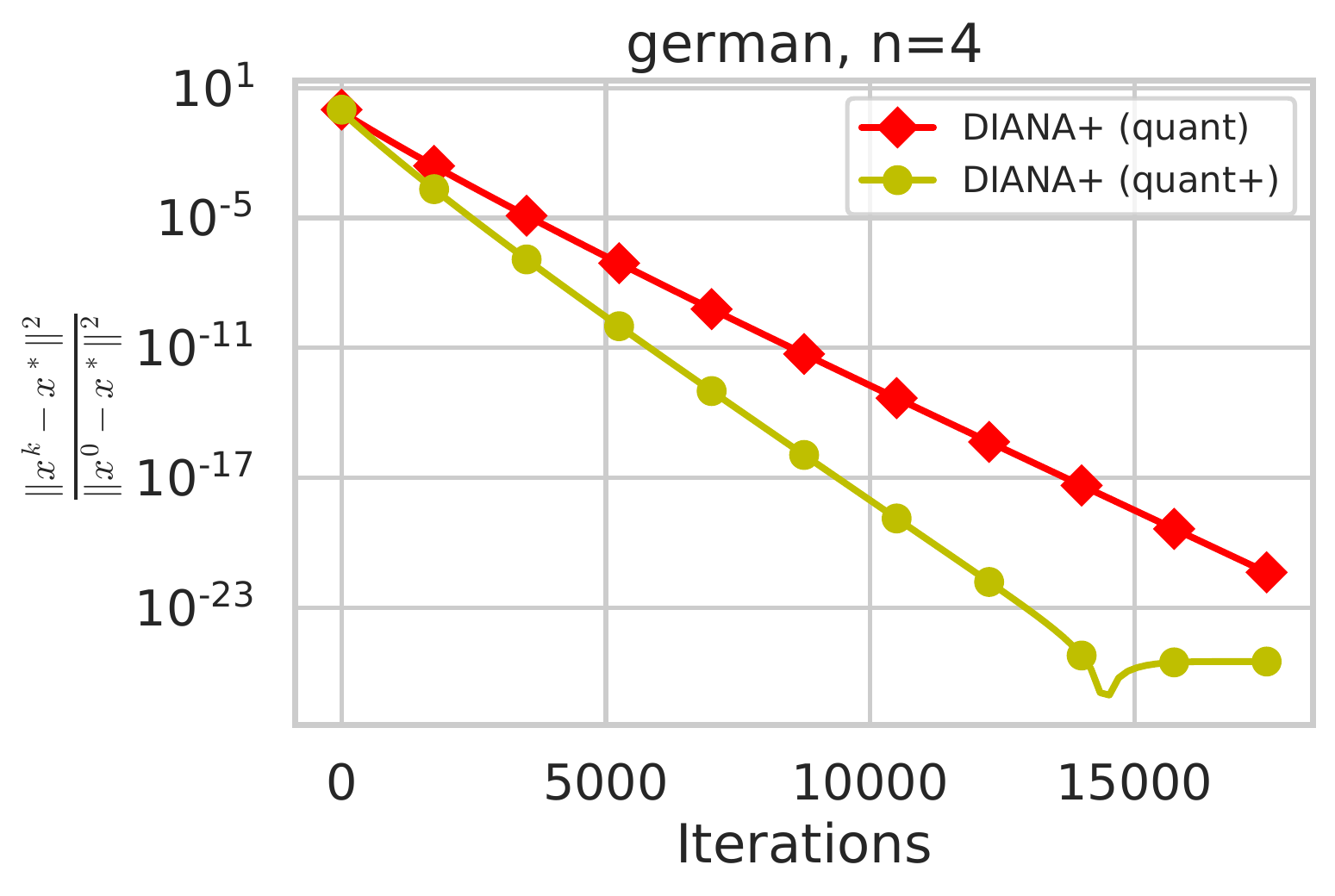}
  \endminipage\hfill  
  \minipage{0.30\textwidth}
  \includegraphics[width=\linewidth]{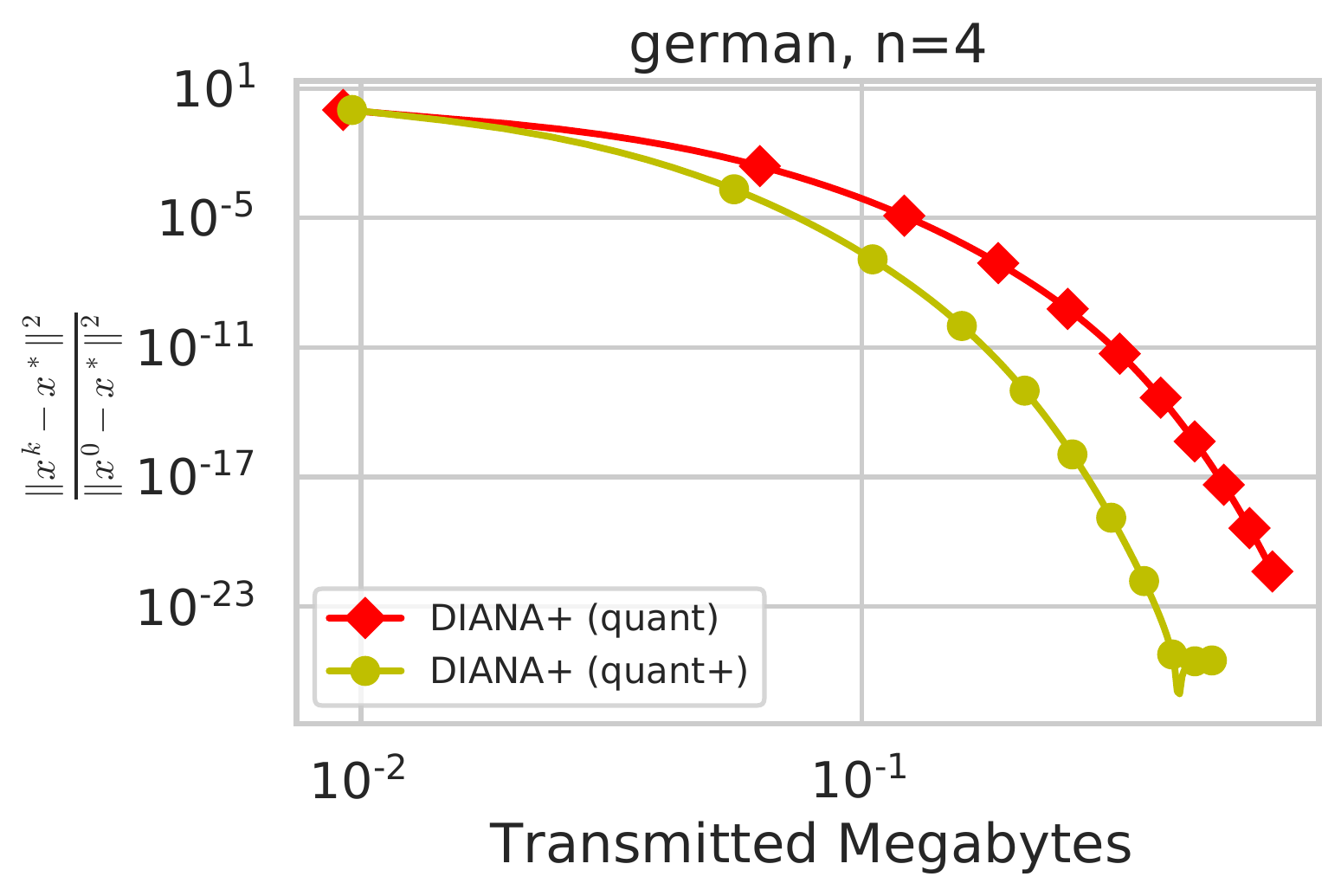}
  \endminipage\hfill
  \minipage{0.30\textwidth}
  \includegraphics[width=\linewidth]{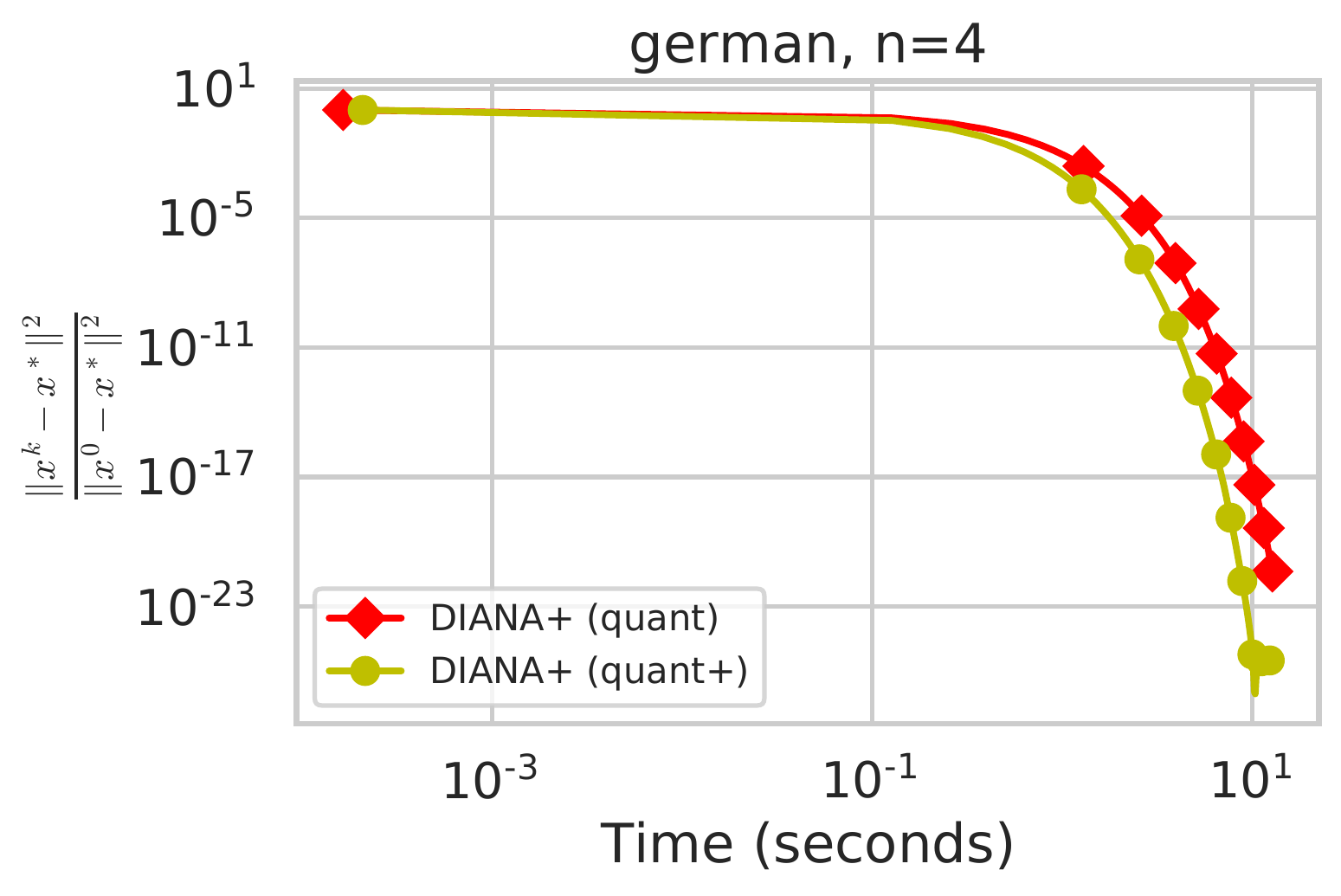}
  \endminipage\hfill    
  
  \minipage{0.30\textwidth}
  \includegraphics[width=\linewidth]{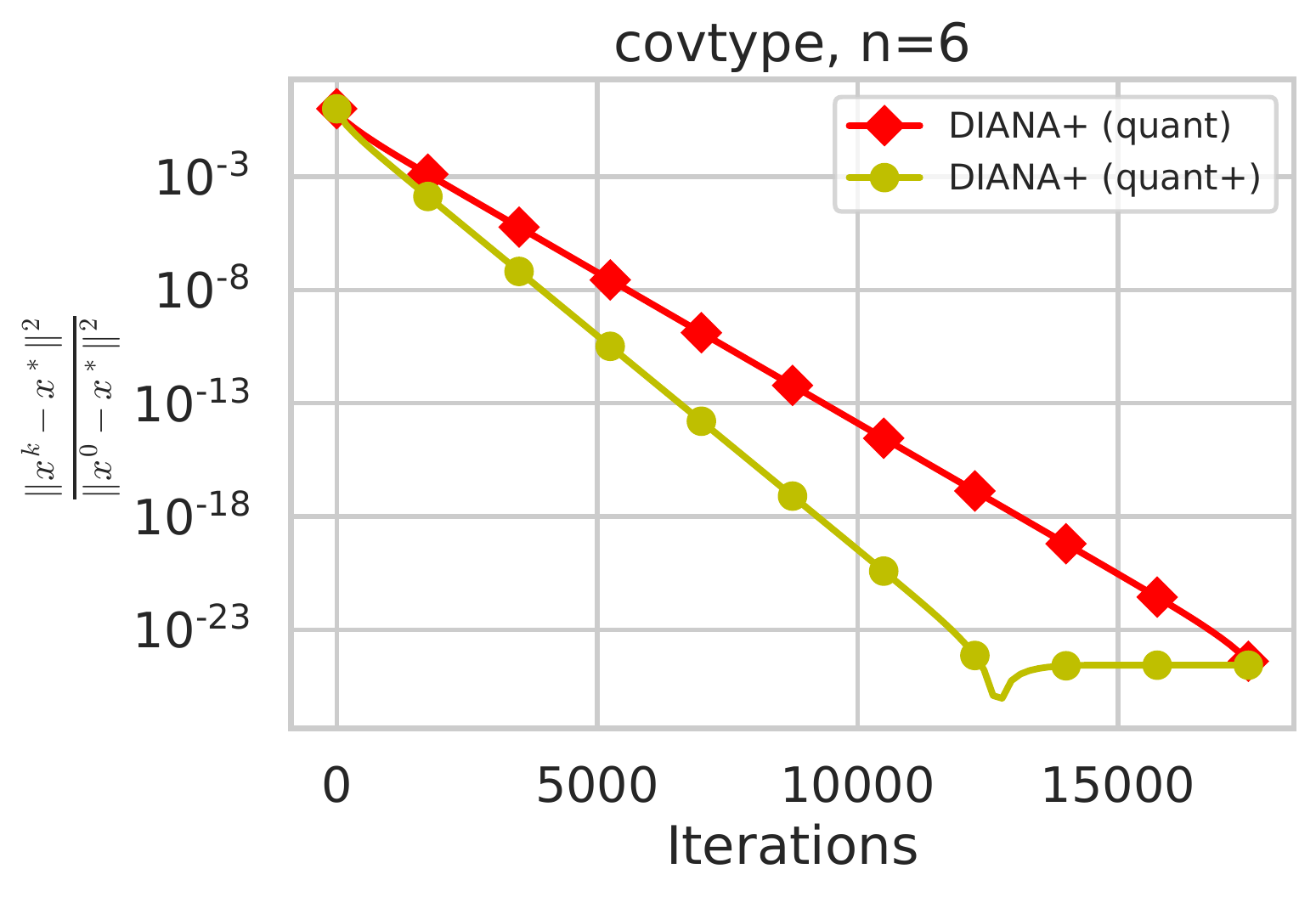}
  \endminipage\hfill  
  \minipage{0.30\textwidth}
  \includegraphics[width=\linewidth]{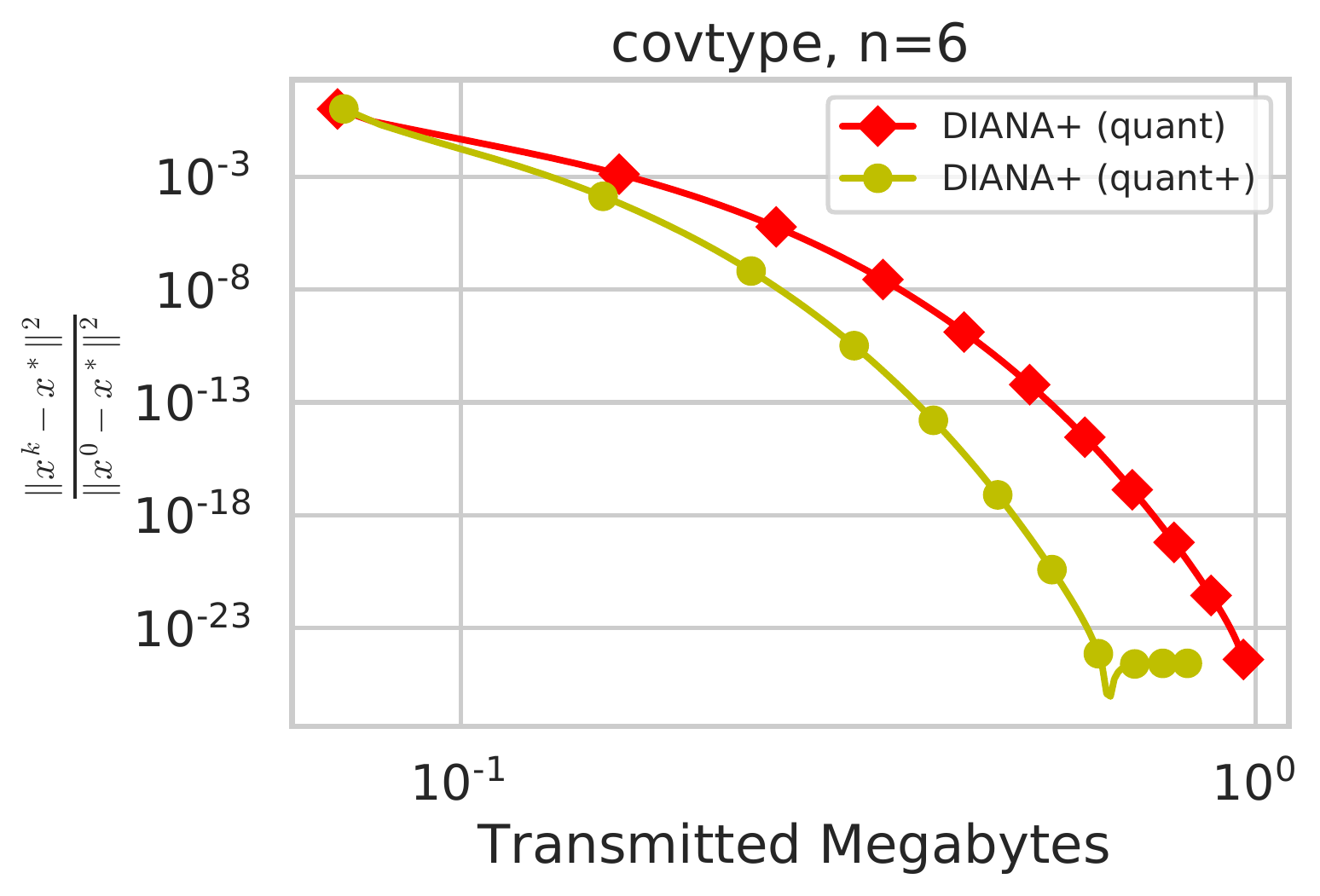}
  \endminipage\hfill  
    \minipage{0.30\textwidth}
  \includegraphics[width=\linewidth]{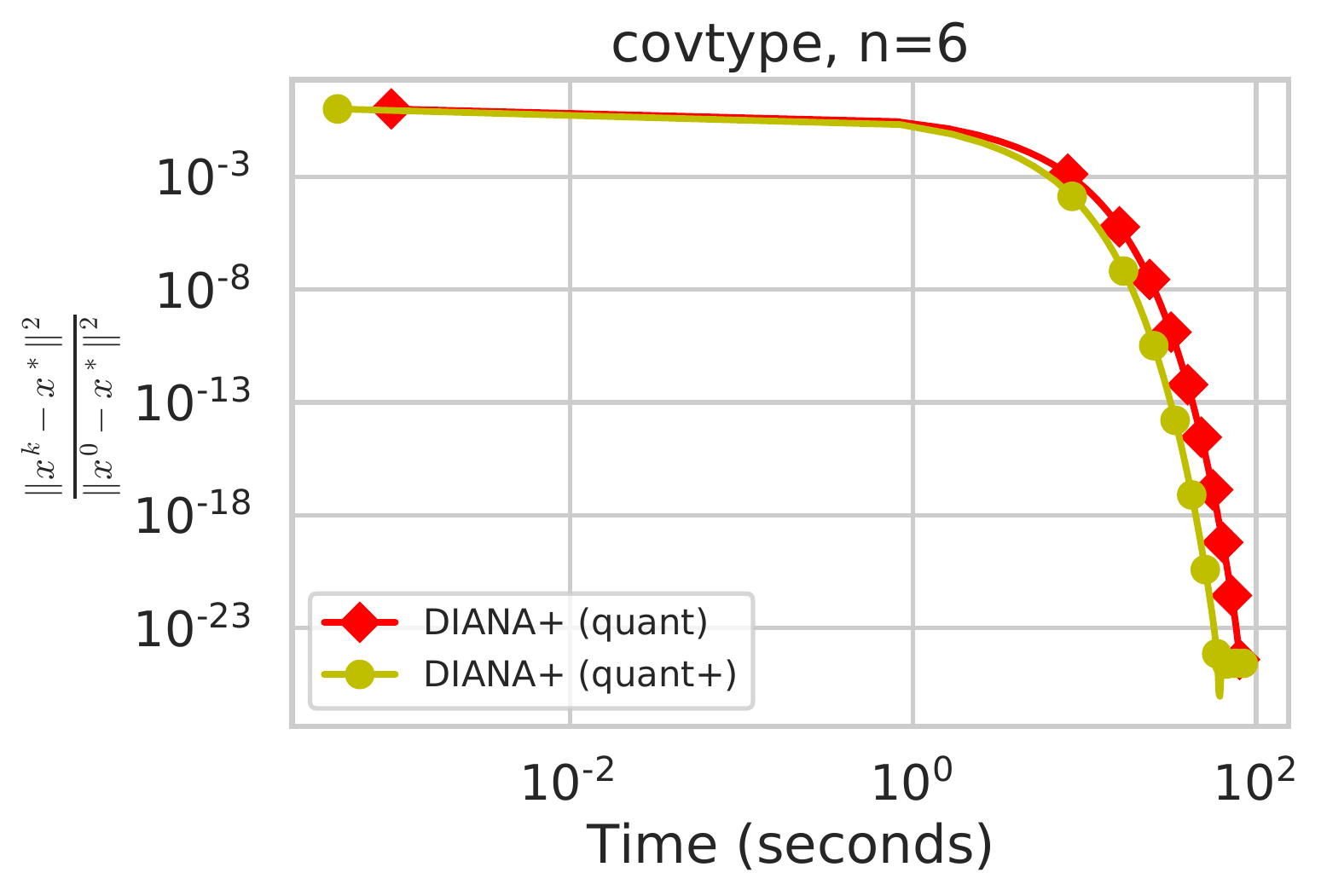}
  \endminipage\hfill  
  
  \minipage{0.30\textwidth}
  \includegraphics[width=\linewidth]{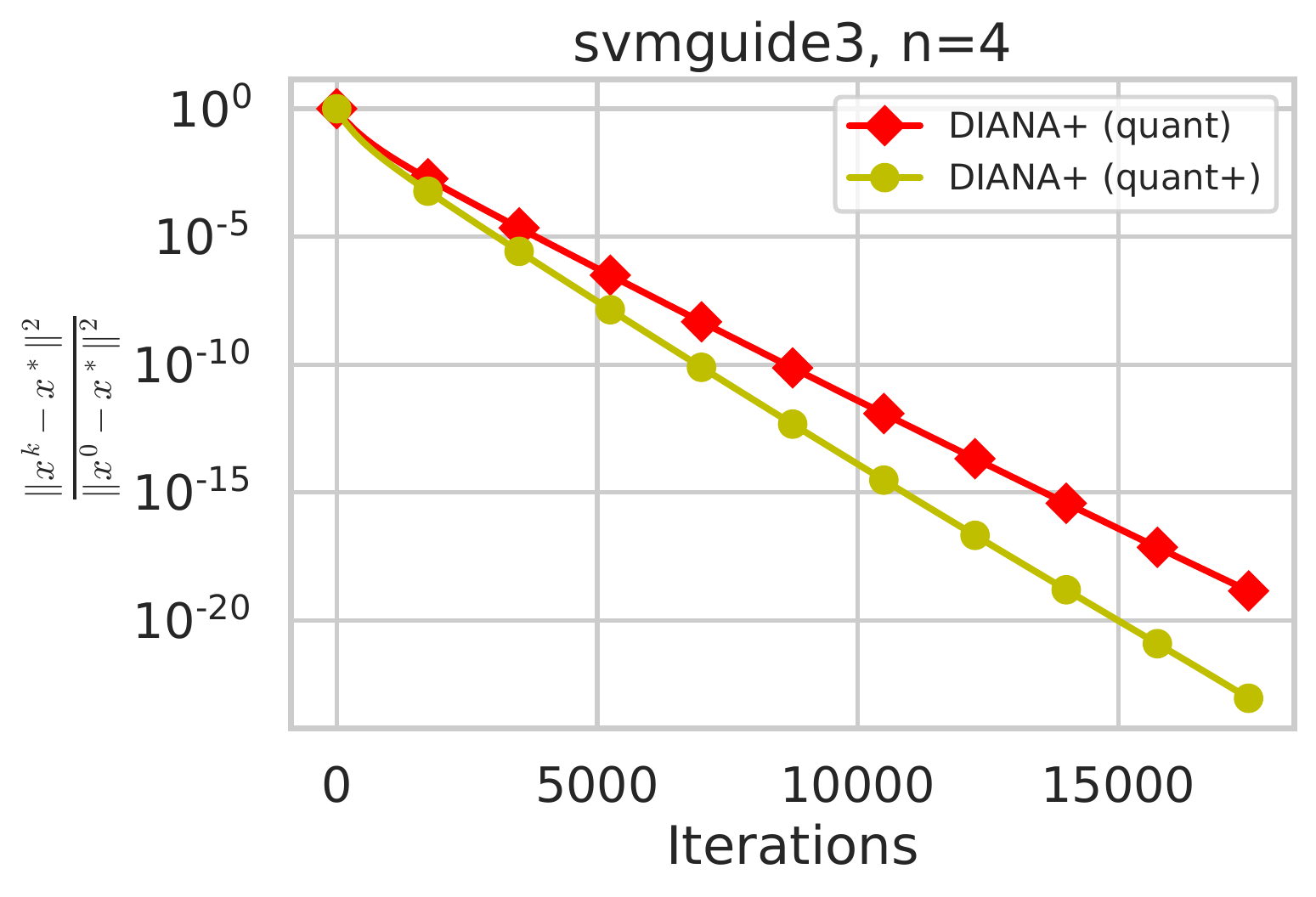}
  \endminipage\hfill  
   \minipage{0.30\textwidth}
  \includegraphics[width=\linewidth]{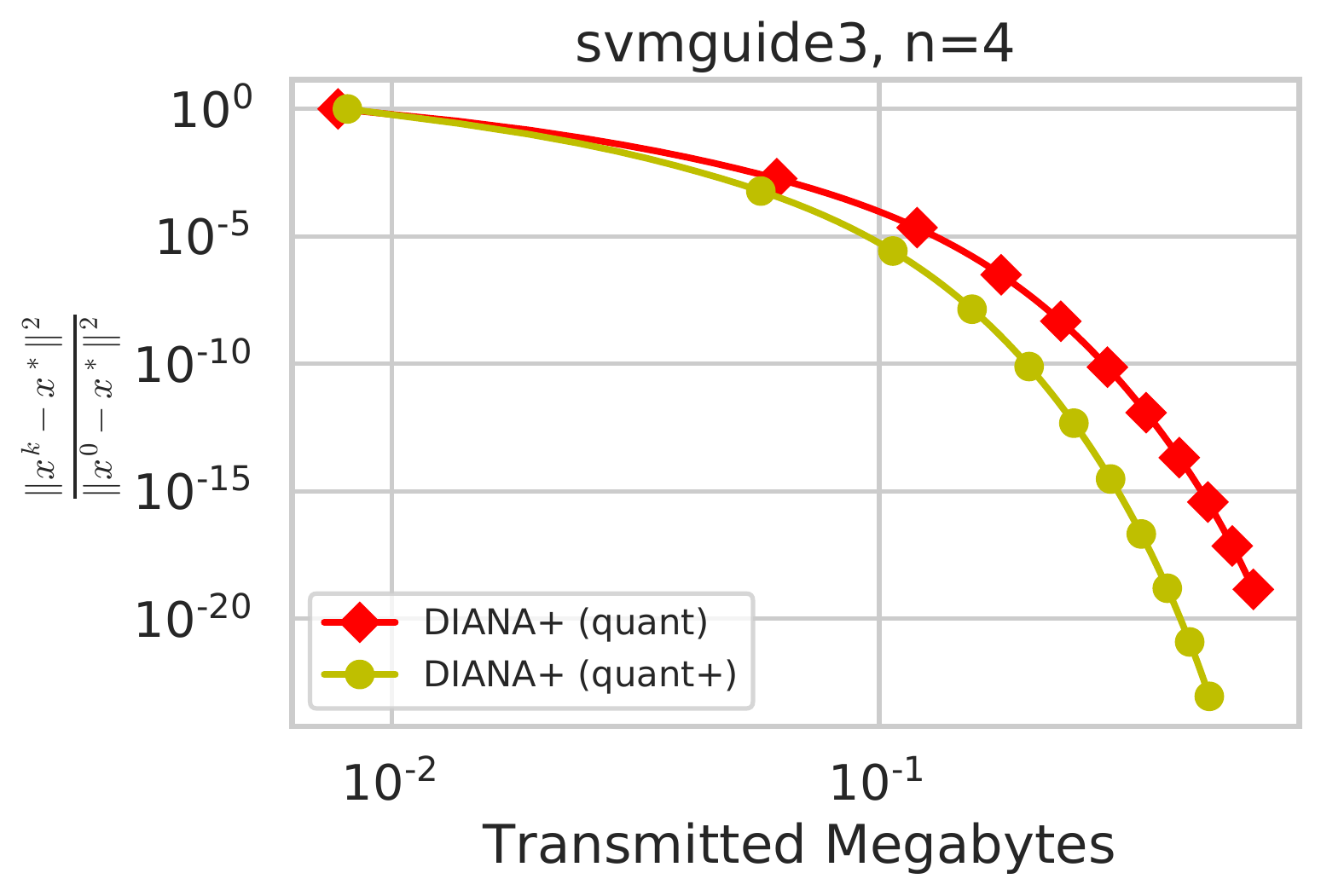}
  \endminipage\hfill  
    \minipage{0.30\textwidth}
  \includegraphics[width=\linewidth]{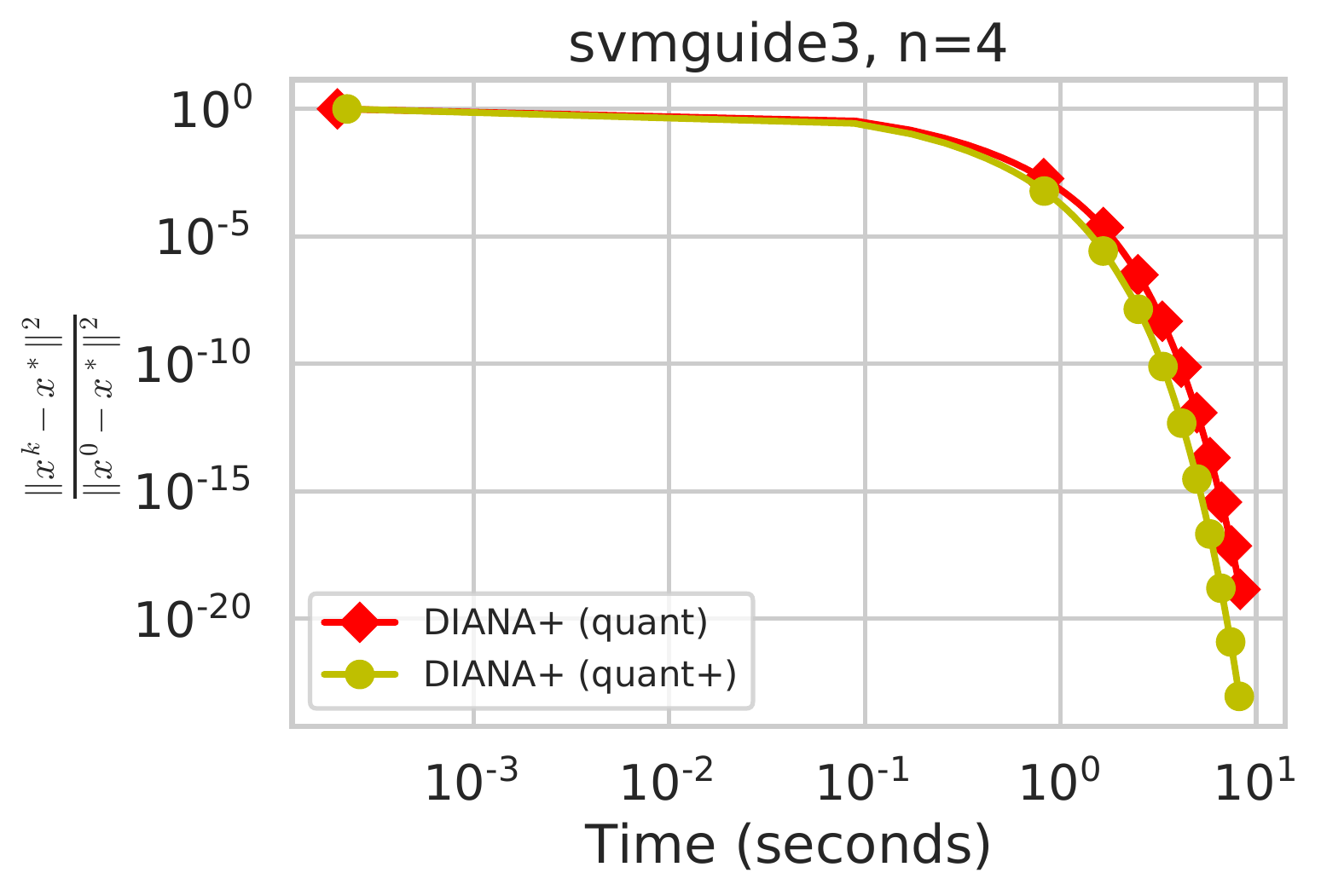}
  \endminipage\hfill  
  
    \minipage{0.30\textwidth}
  \includegraphics[width=\linewidth]{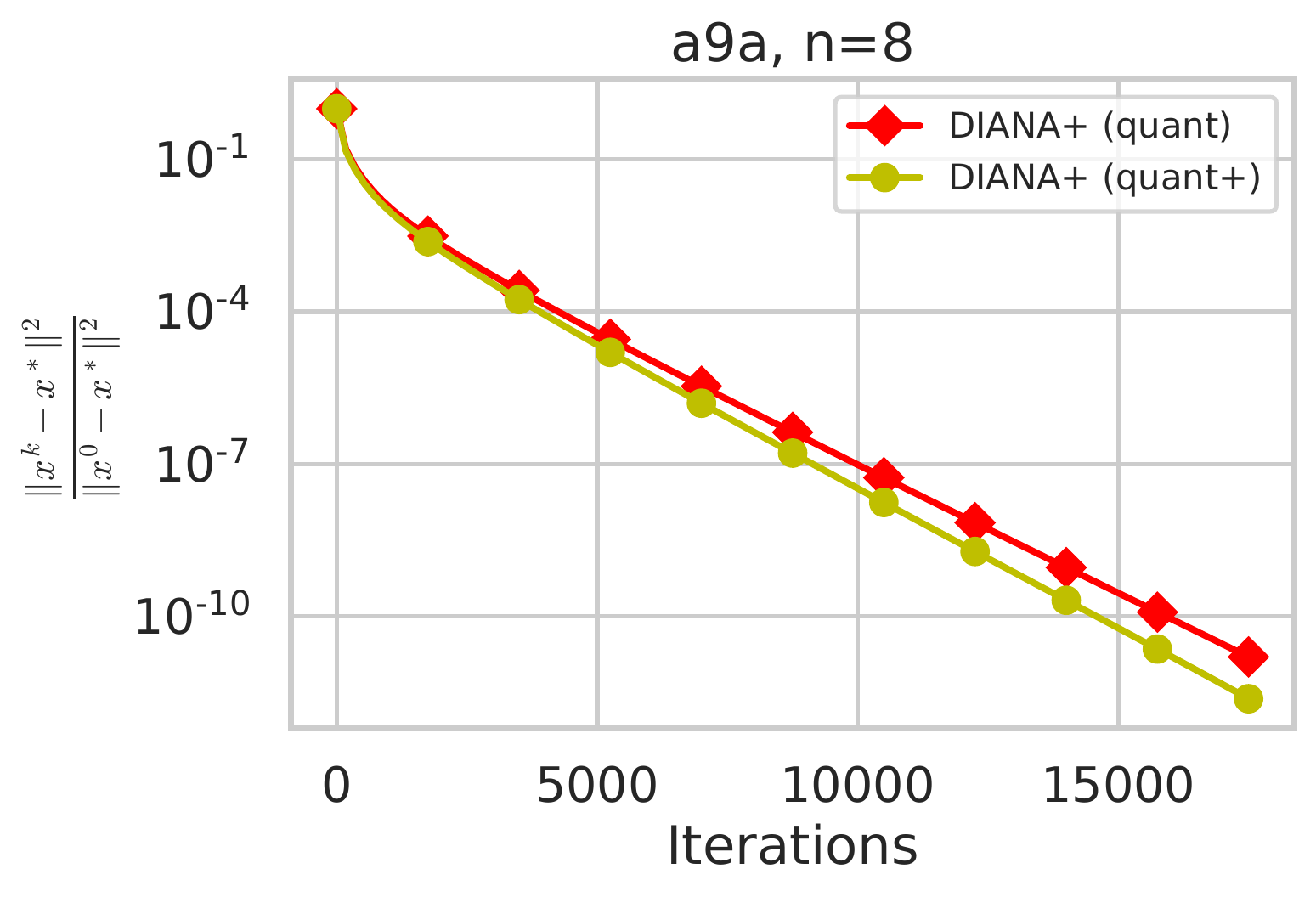}
  \endminipage\hfill  
    \minipage{0.30\textwidth}
  \includegraphics[width=\linewidth]{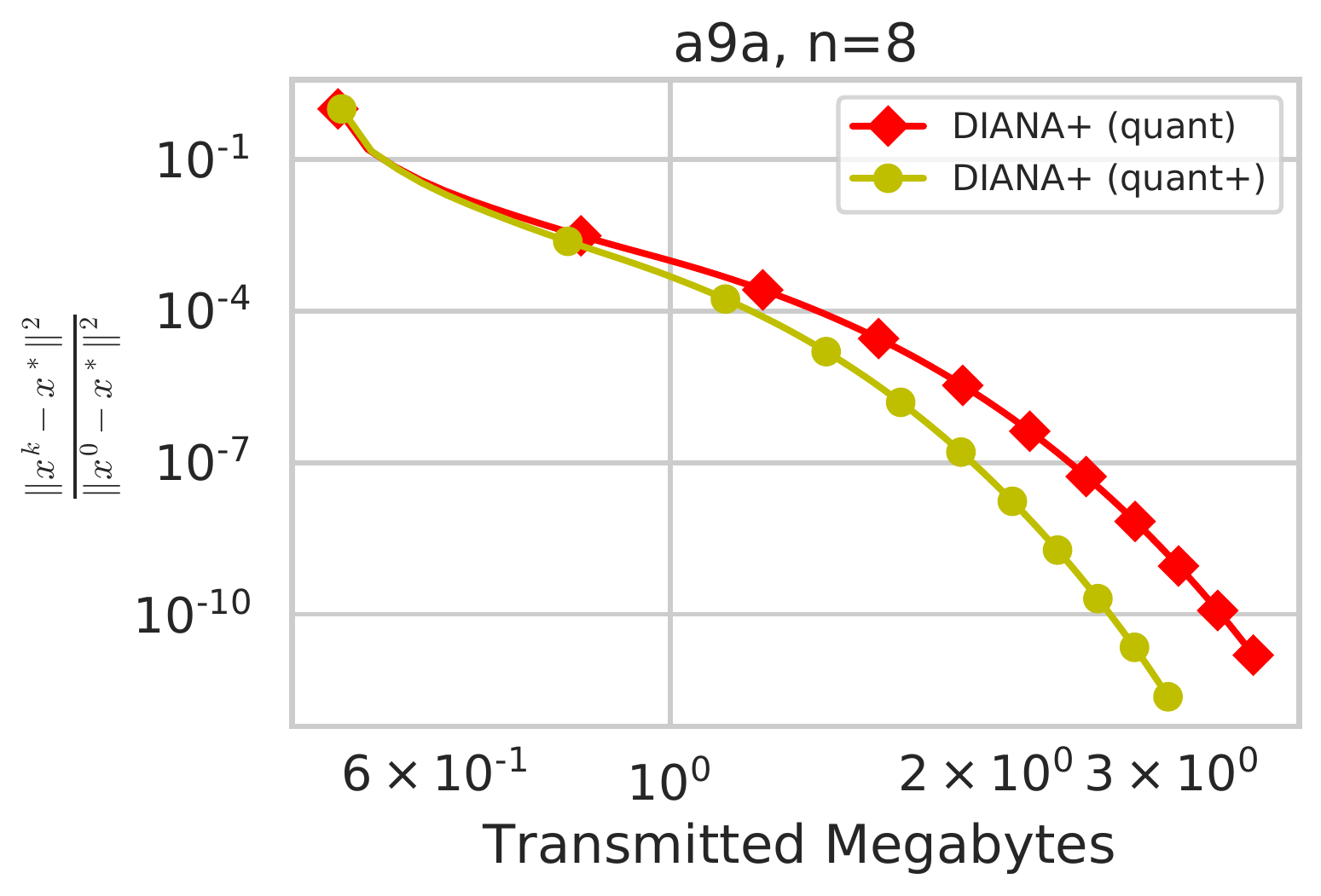}
  \endminipage\hfill  
   \minipage{0.30\textwidth}
 \includegraphics[width=\linewidth]{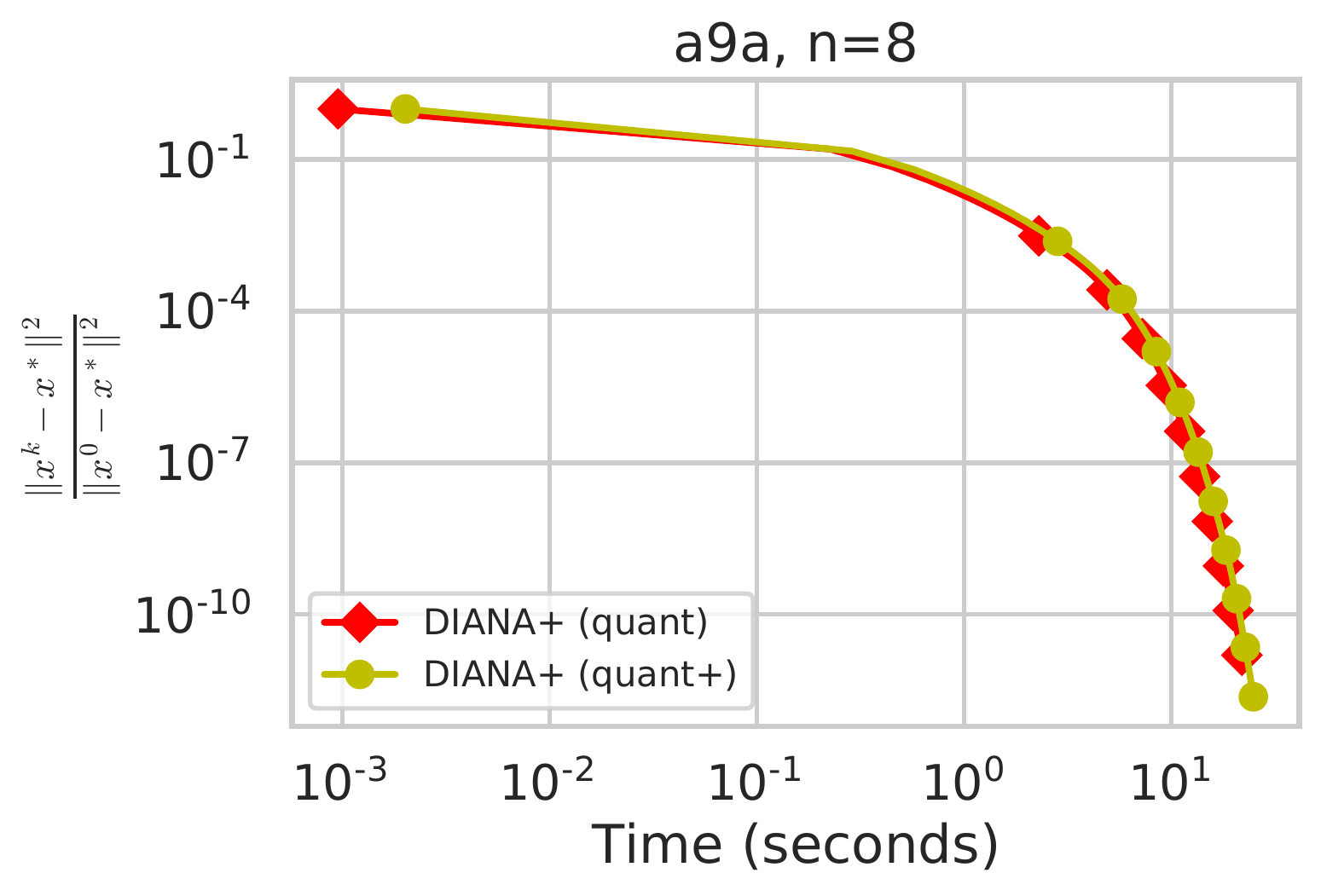}
 \endminipage\hfill  
  
    \minipage{0.30\textwidth}
  \includegraphics[width=\linewidth]{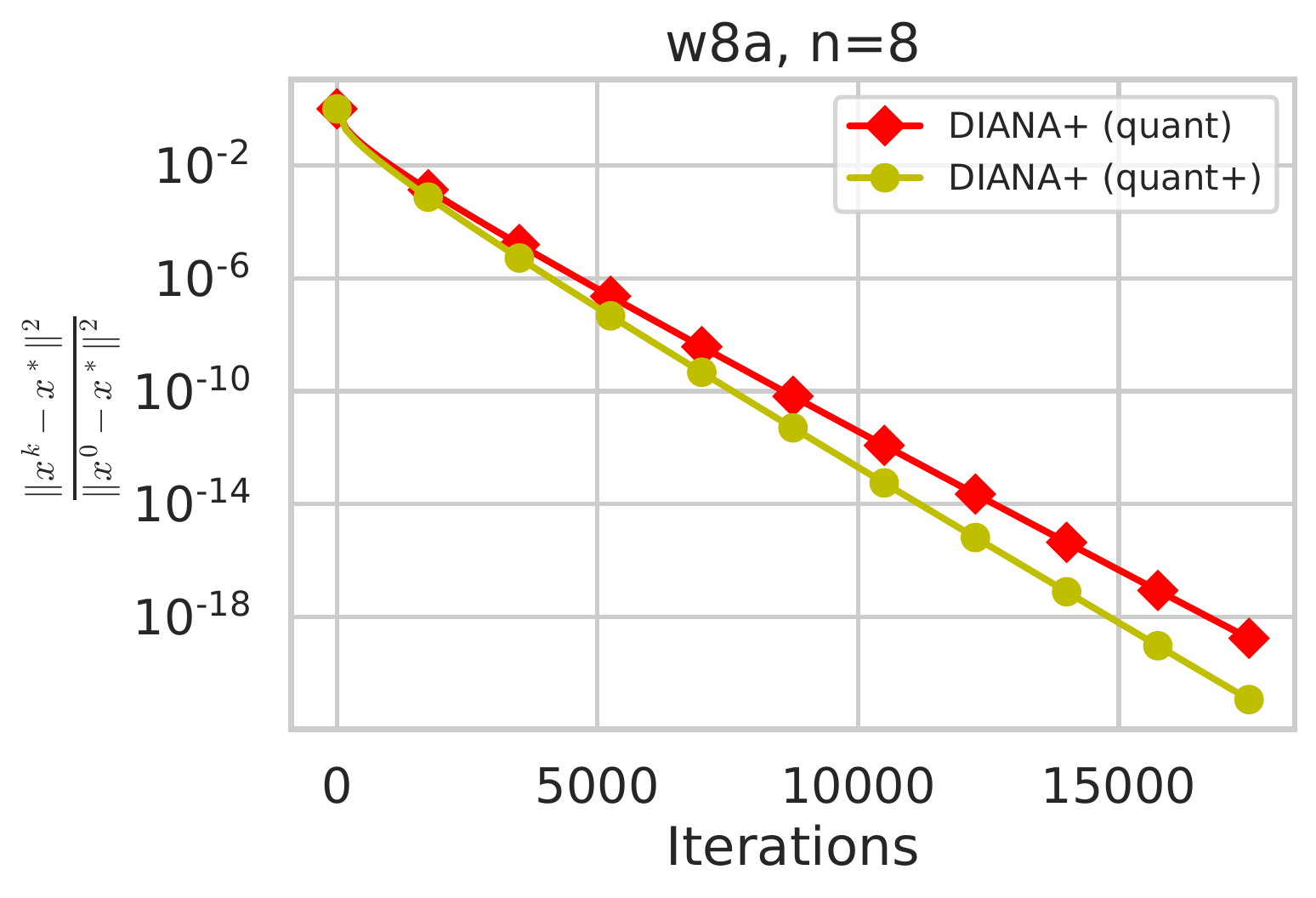}
  \endminipage\hfill  
  \minipage{0.30\textwidth}
  \includegraphics[width=\linewidth]{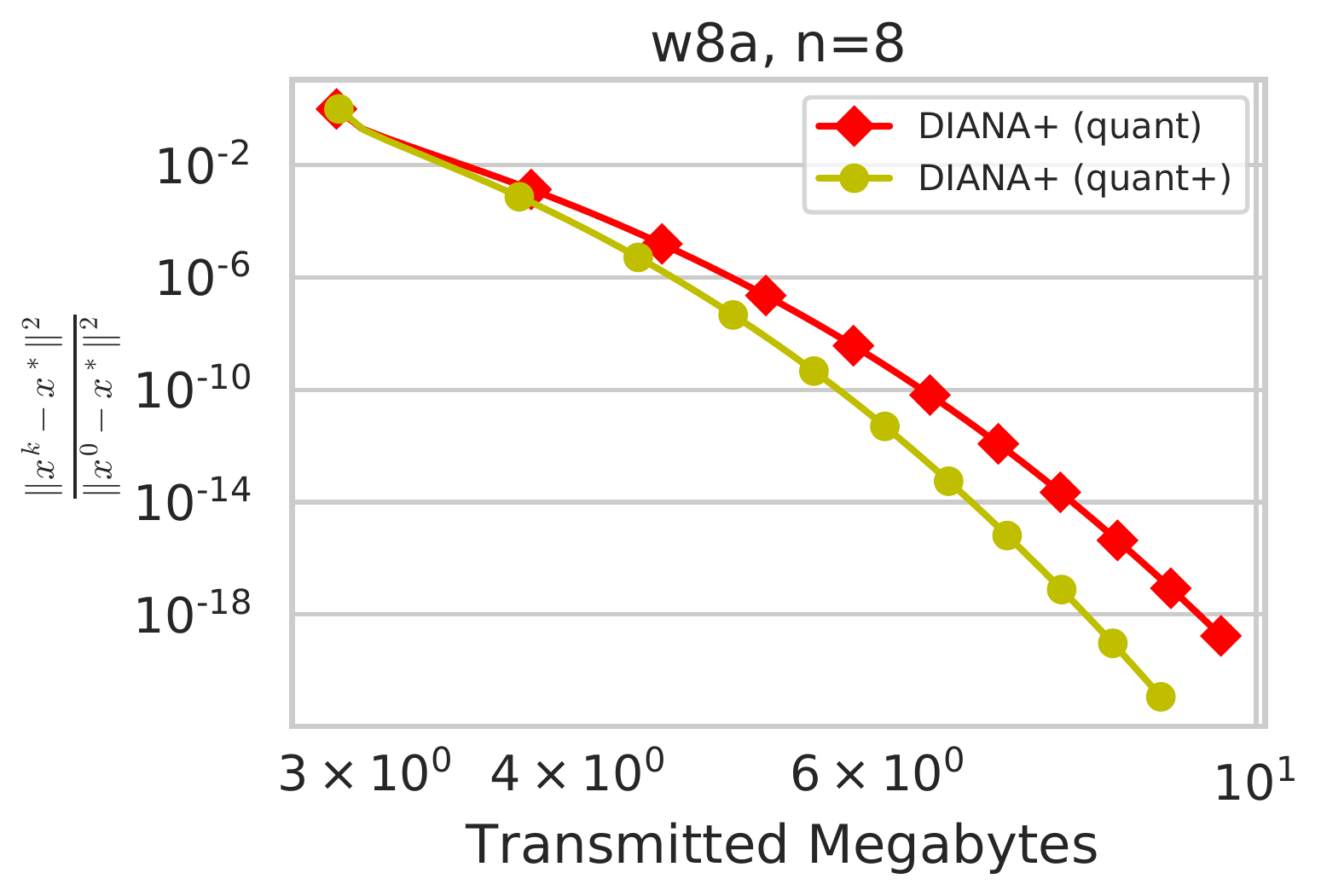}
  \endminipage\hfill  
  \minipage{0.30\textwidth}
  \includegraphics[width=\linewidth]{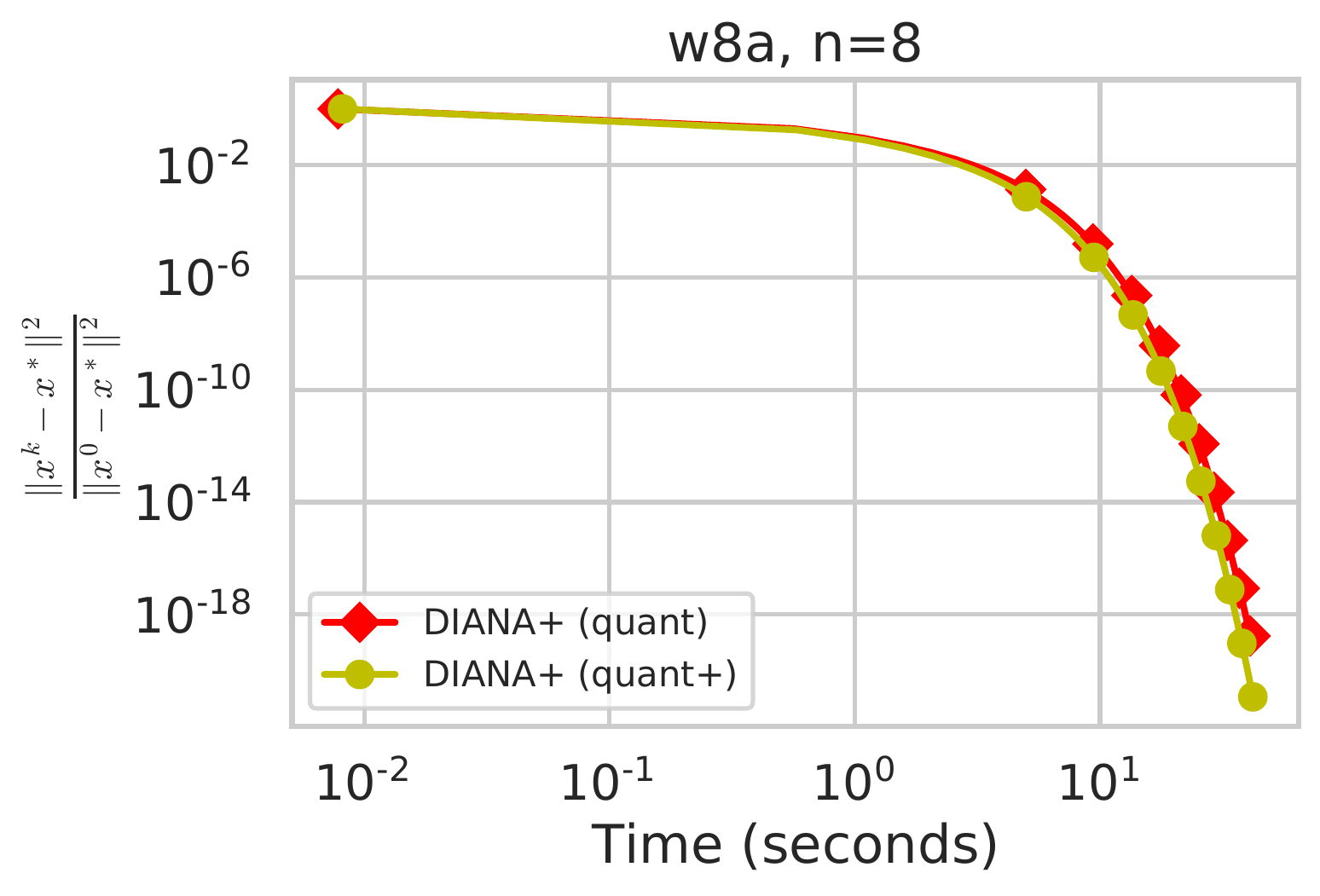}
  \endminipage\hfill  
  \caption{Comparison of DIANA+ with quantization that has varying or fixed number of levels.}
  \label{fig:nonuniform}
\end{figure*}

Next, we compare DCGD+/DIANA+ with our second quantization technique (\texttt{quant+}) that has varying number of quantization steps per coordinate to DCGD (\texttt{quant}) and DIANA (\texttt{quant}). Figure~\ref{fig:nonuniform} demonstrates that DCGD+ (\texttt{quant+}) and DIANA+ (\texttt{quant+}) lead to significant improvement. 

\subsection{Ablation study of DIANA+ (\texttt{block quant+}) and DIANA+ (\texttt{quant+})}

\begin{figure*}[htp]
    \minipage{0.30\textwidth}
    \includegraphics[width=\linewidth]{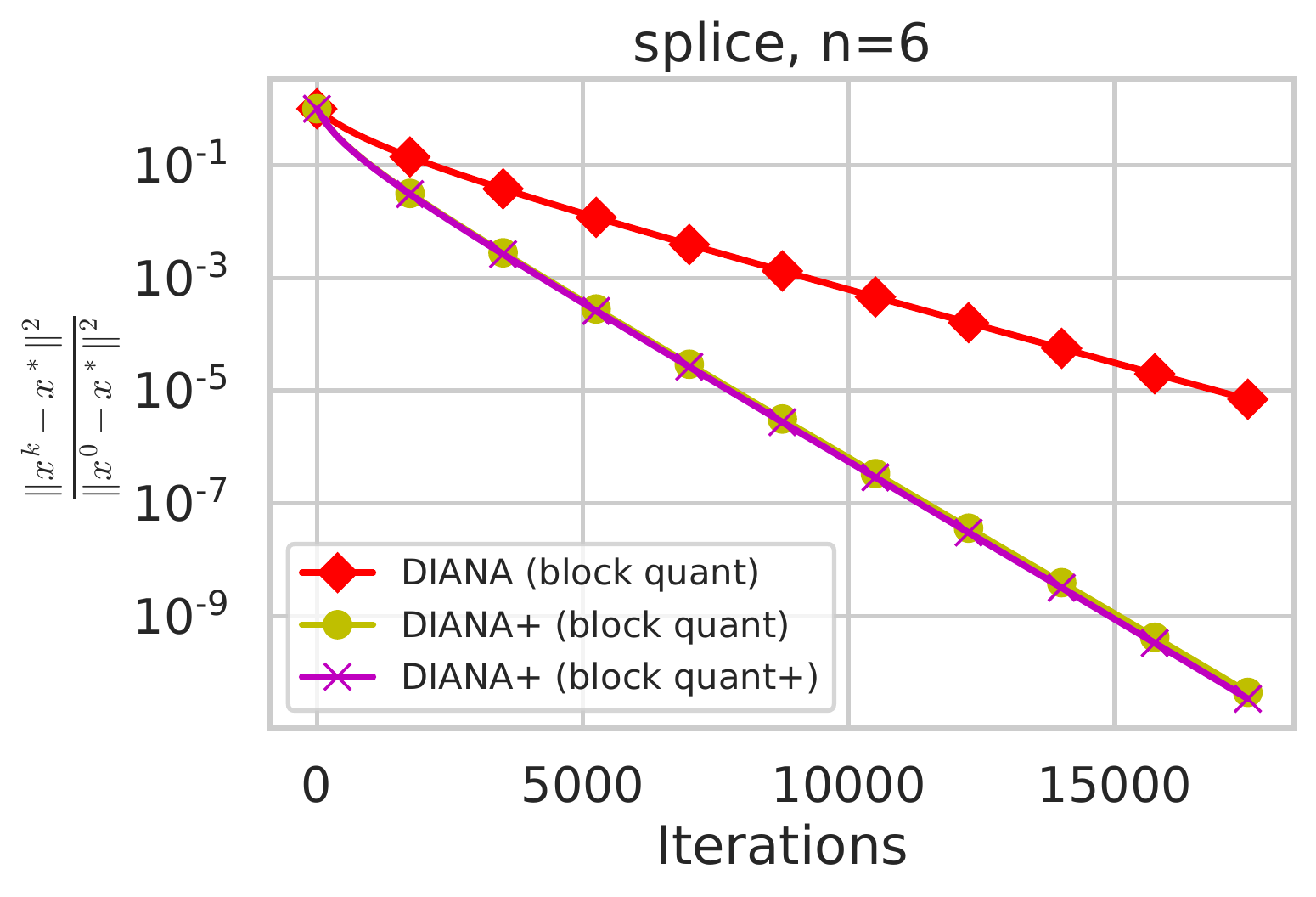}
    \endminipage\hfill  
        \minipage{0.30\textwidth}
    \includegraphics[width=\linewidth]{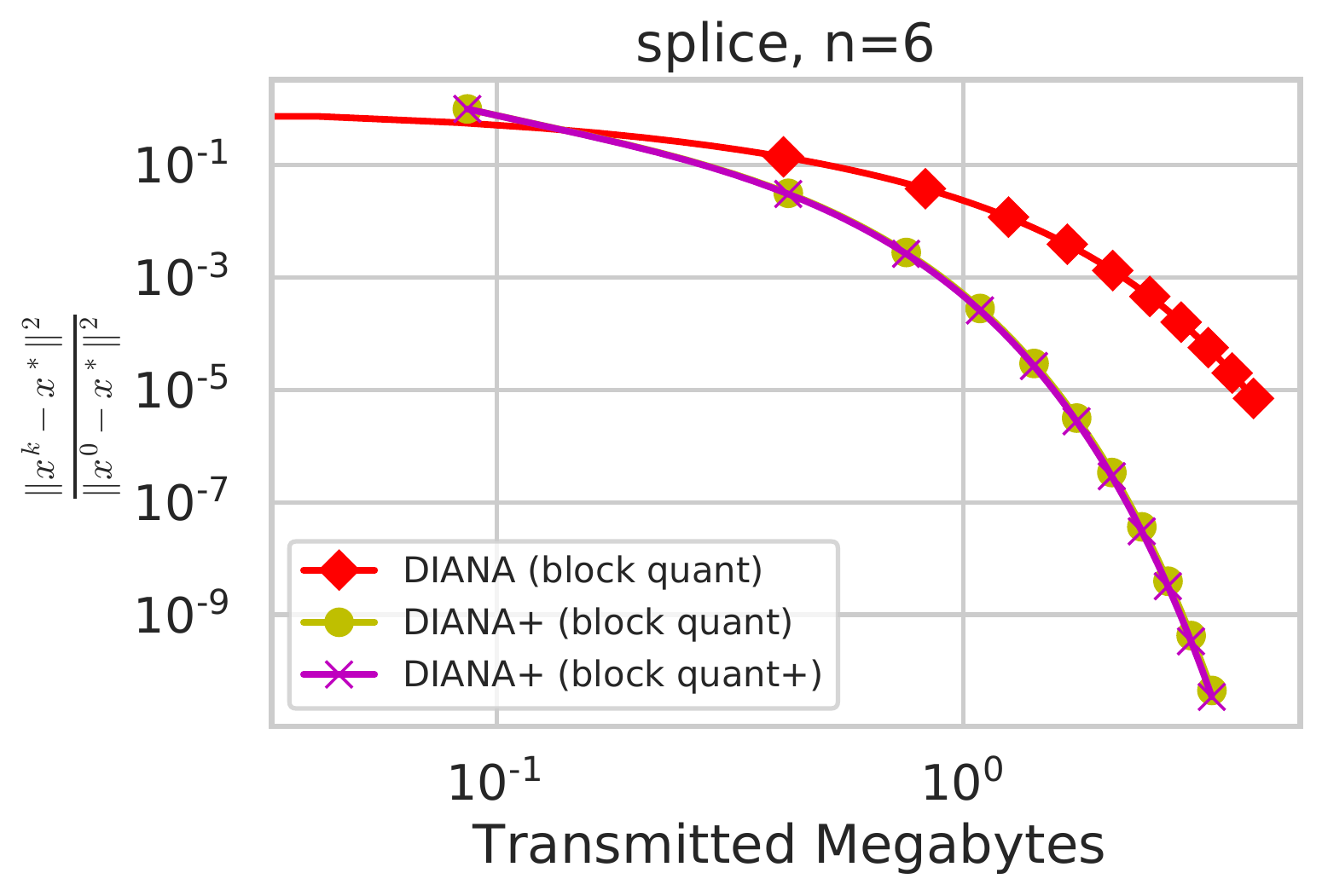}
    \endminipage\hfill  
    \minipage{0.30\textwidth}
    \includegraphics[width=\linewidth]{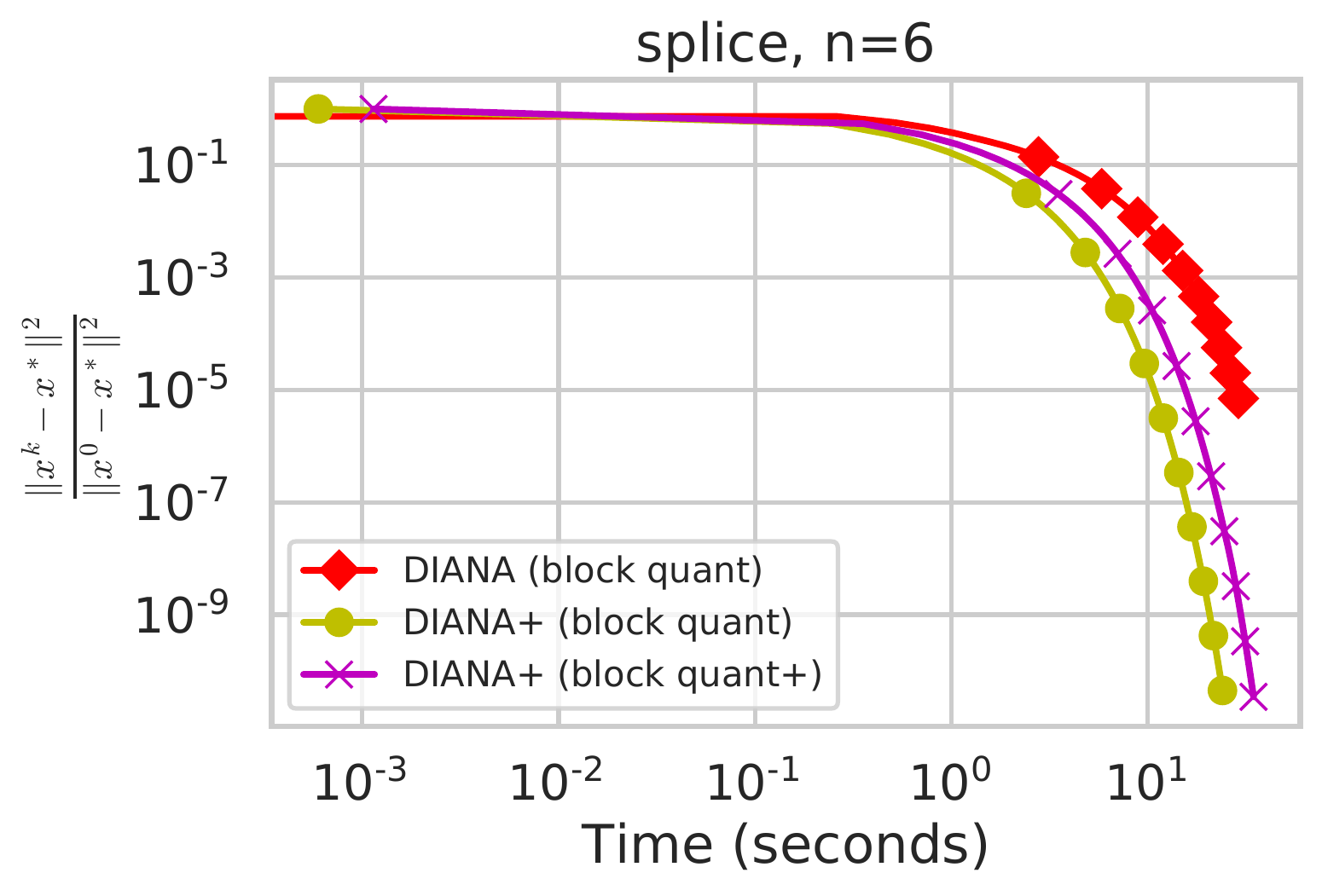}
    \endminipage\hfill

    \minipage{0.30\textwidth}
    \includegraphics[width=\linewidth]{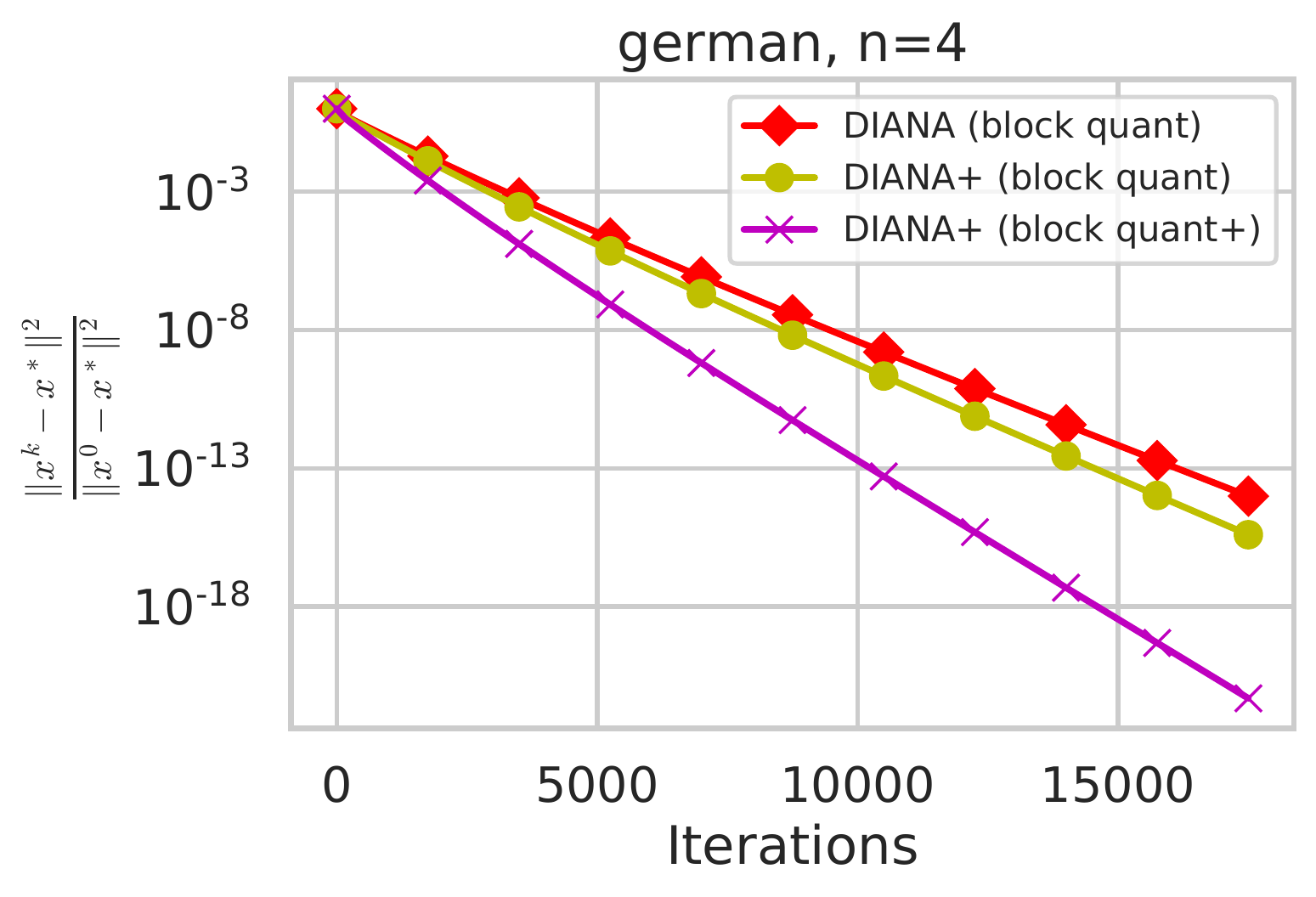}
    \endminipage\hfill  
        \minipage{0.30\textwidth}
    \includegraphics[width=\linewidth]{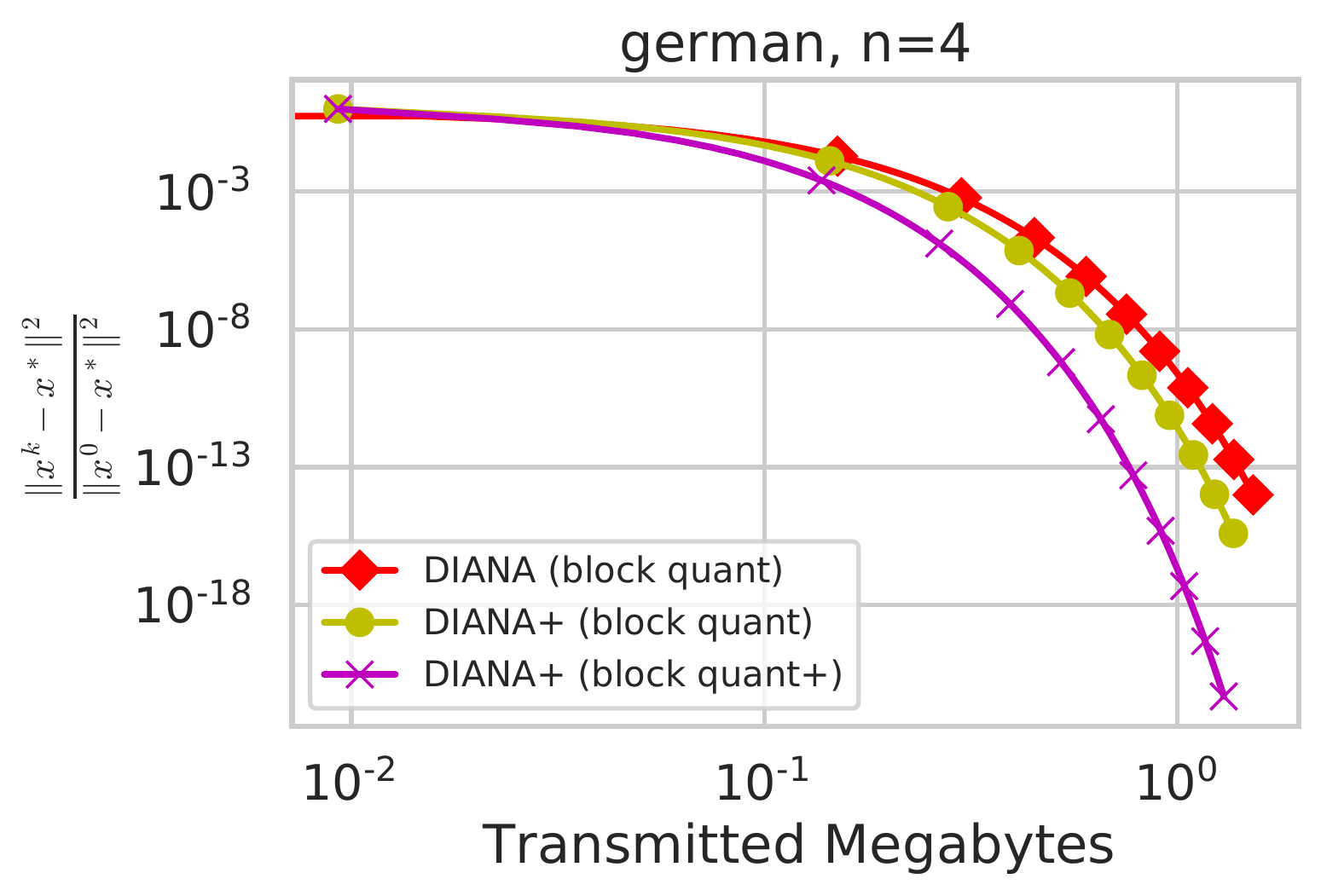}
    \endminipage\hfill
        \minipage{0.30\textwidth}
    \includegraphics[width=\linewidth]{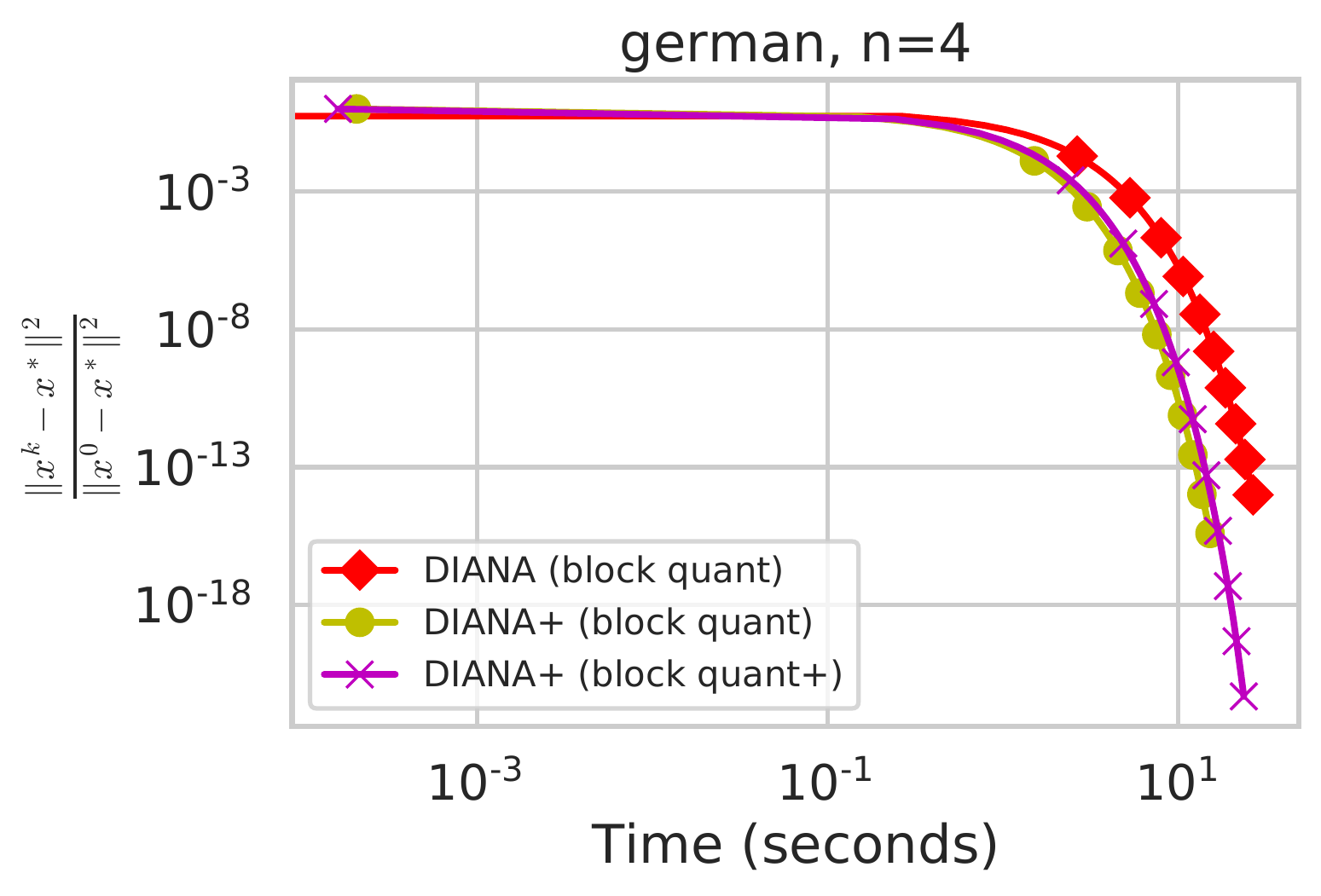}
    \endminipage\hfill

    \minipage{0.30\textwidth}
    \includegraphics[width=\linewidth]{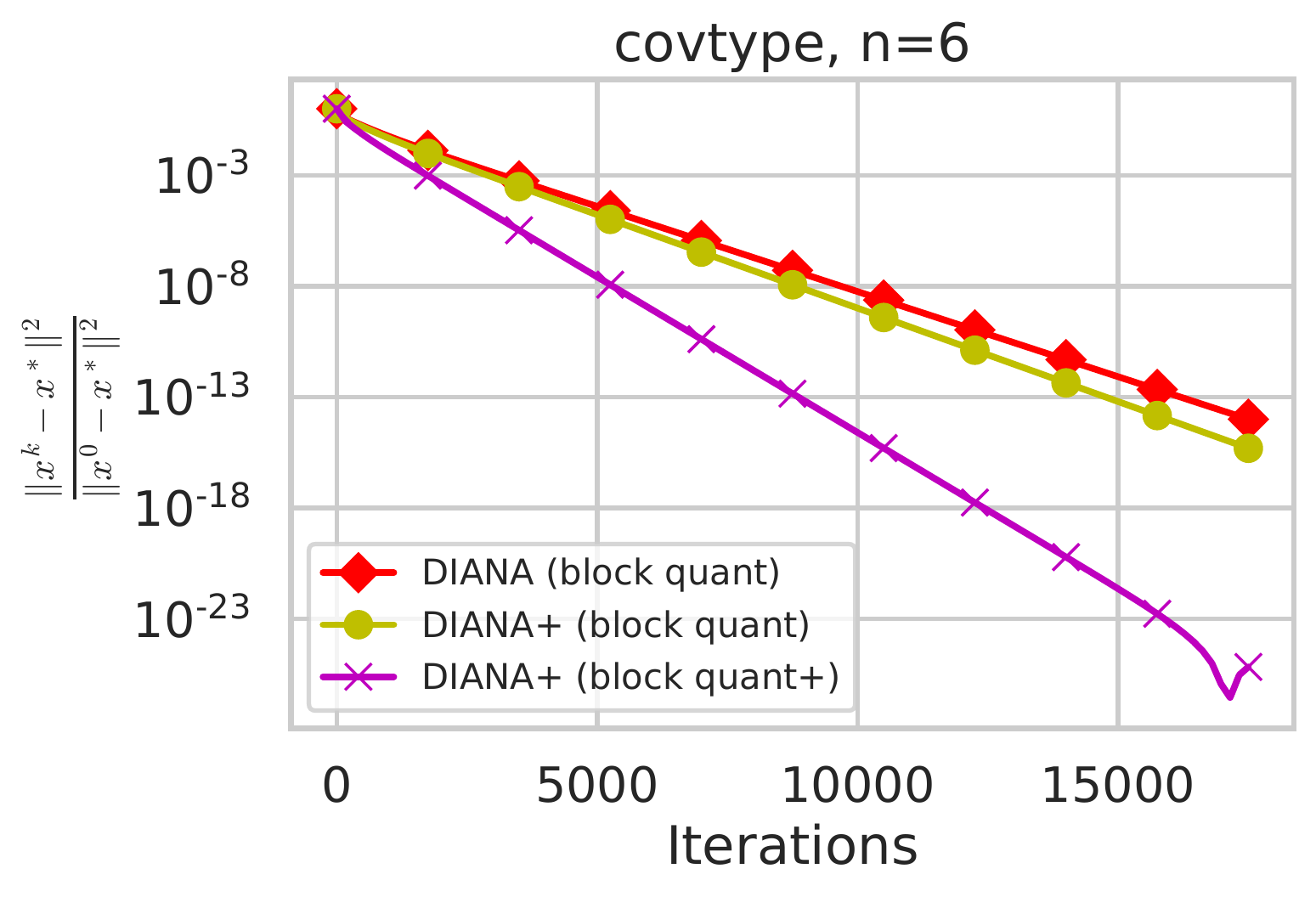}
    \endminipage\hfill  
        \minipage{0.30\textwidth}
    \includegraphics[width=\linewidth]{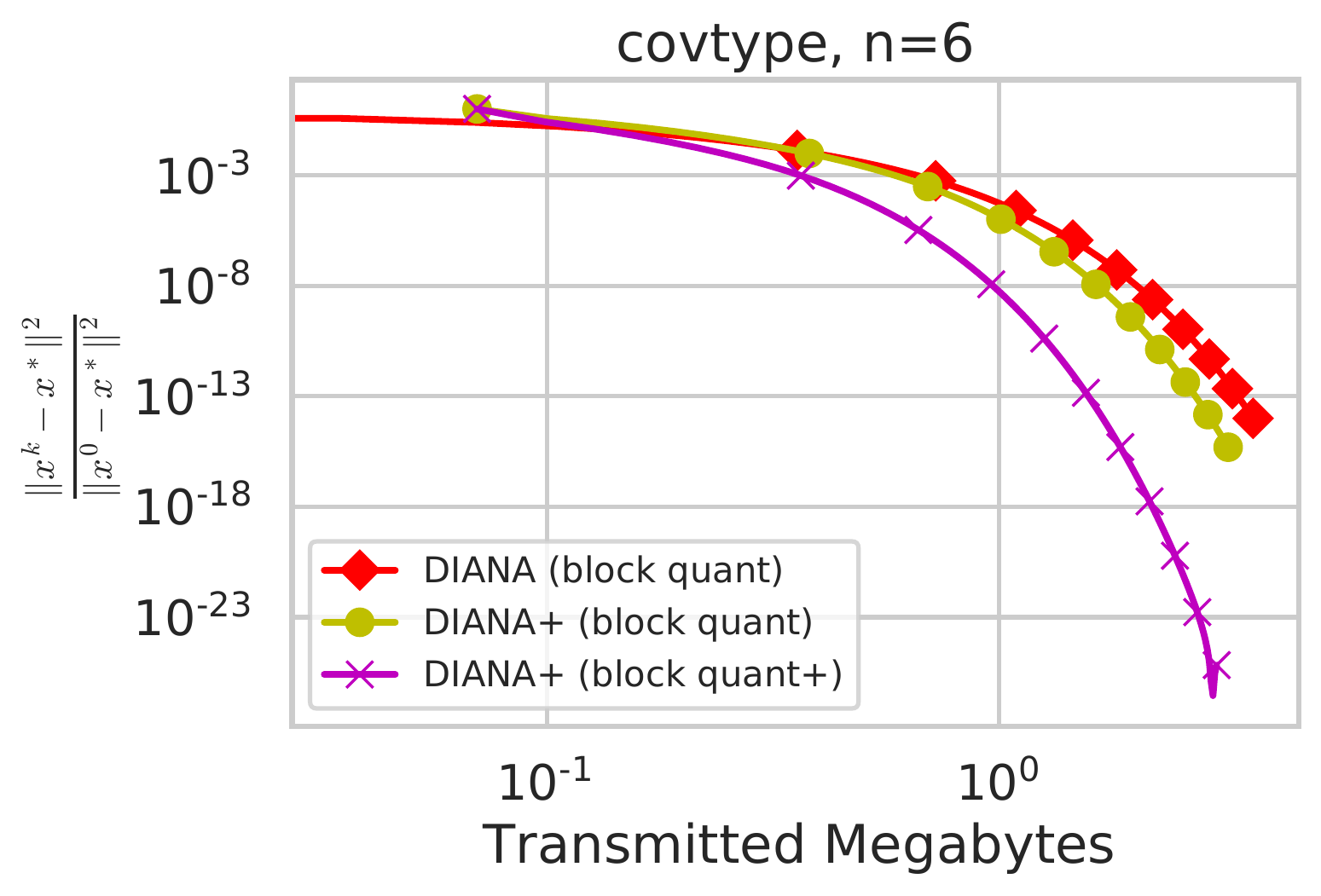}
    \endminipage\hfill
    \minipage{0.30\textwidth}
    \includegraphics[width=\linewidth]{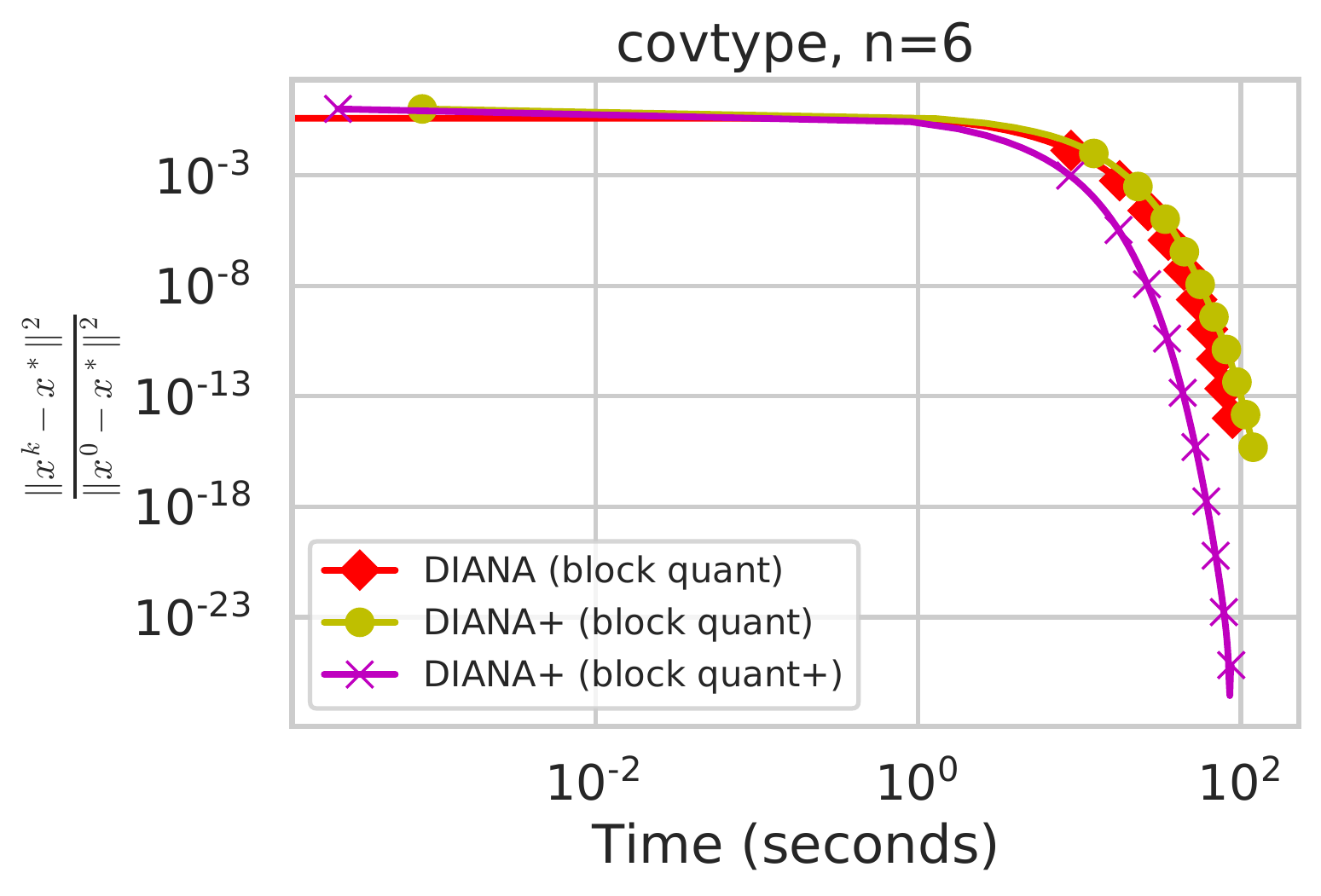}
    \endminipage\hfill  
    
    \minipage{0.30\textwidth}
    \includegraphics[width=\linewidth]{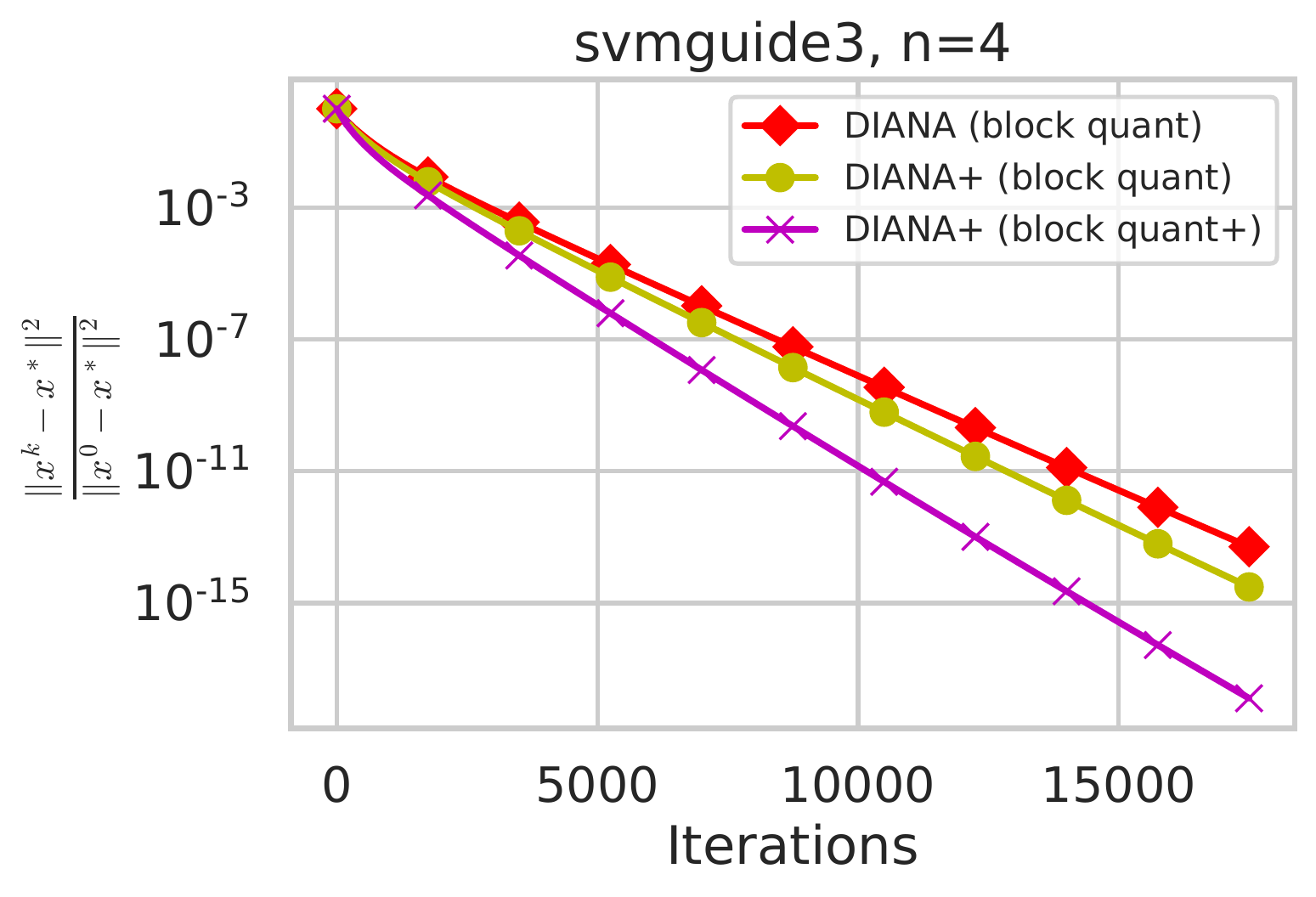}
    \endminipage\hfill  
        \minipage{0.30\textwidth}
    \includegraphics[width=\linewidth]{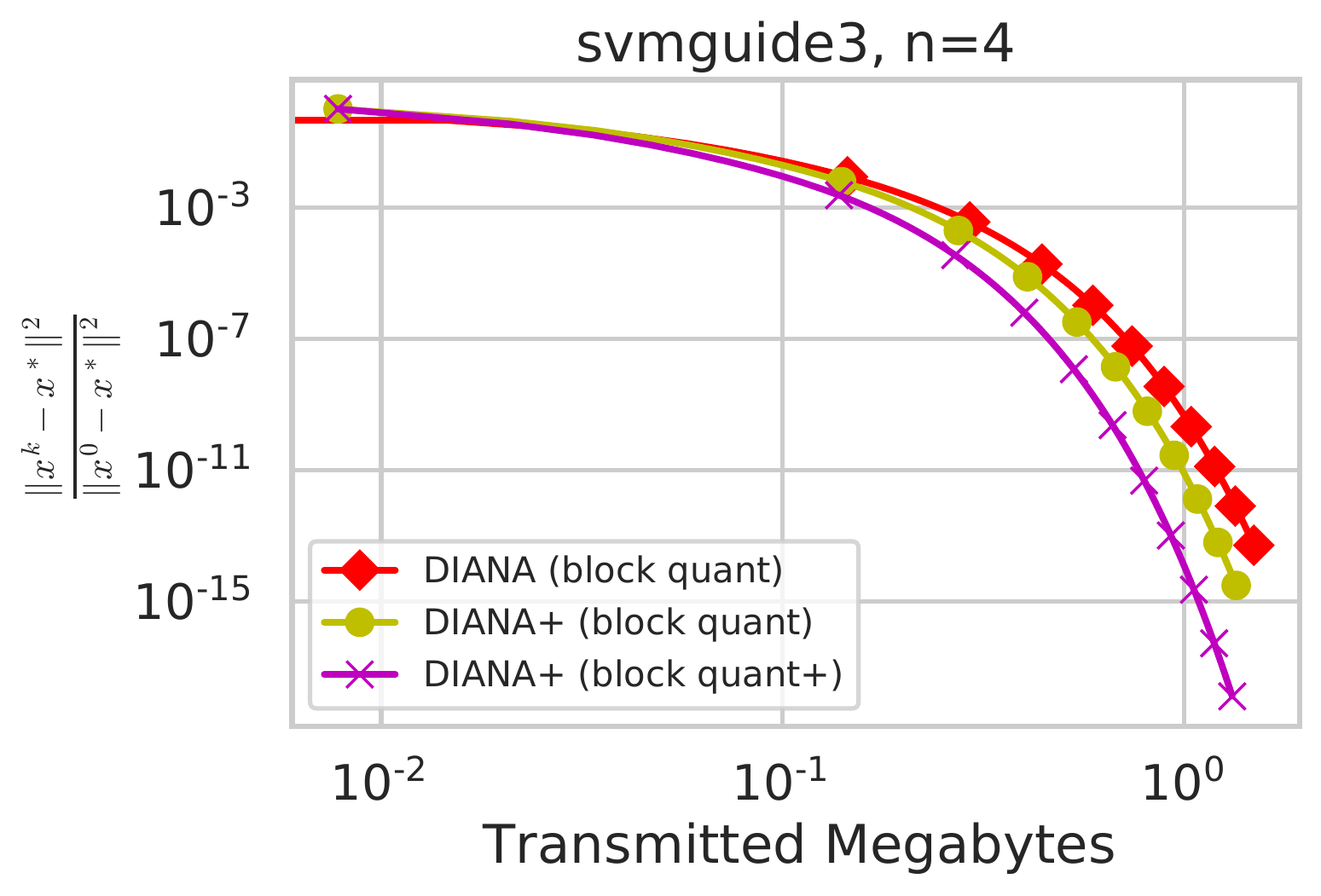}
    \endminipage\hfill
        \minipage{0.30\textwidth}
    \includegraphics[width=\linewidth]{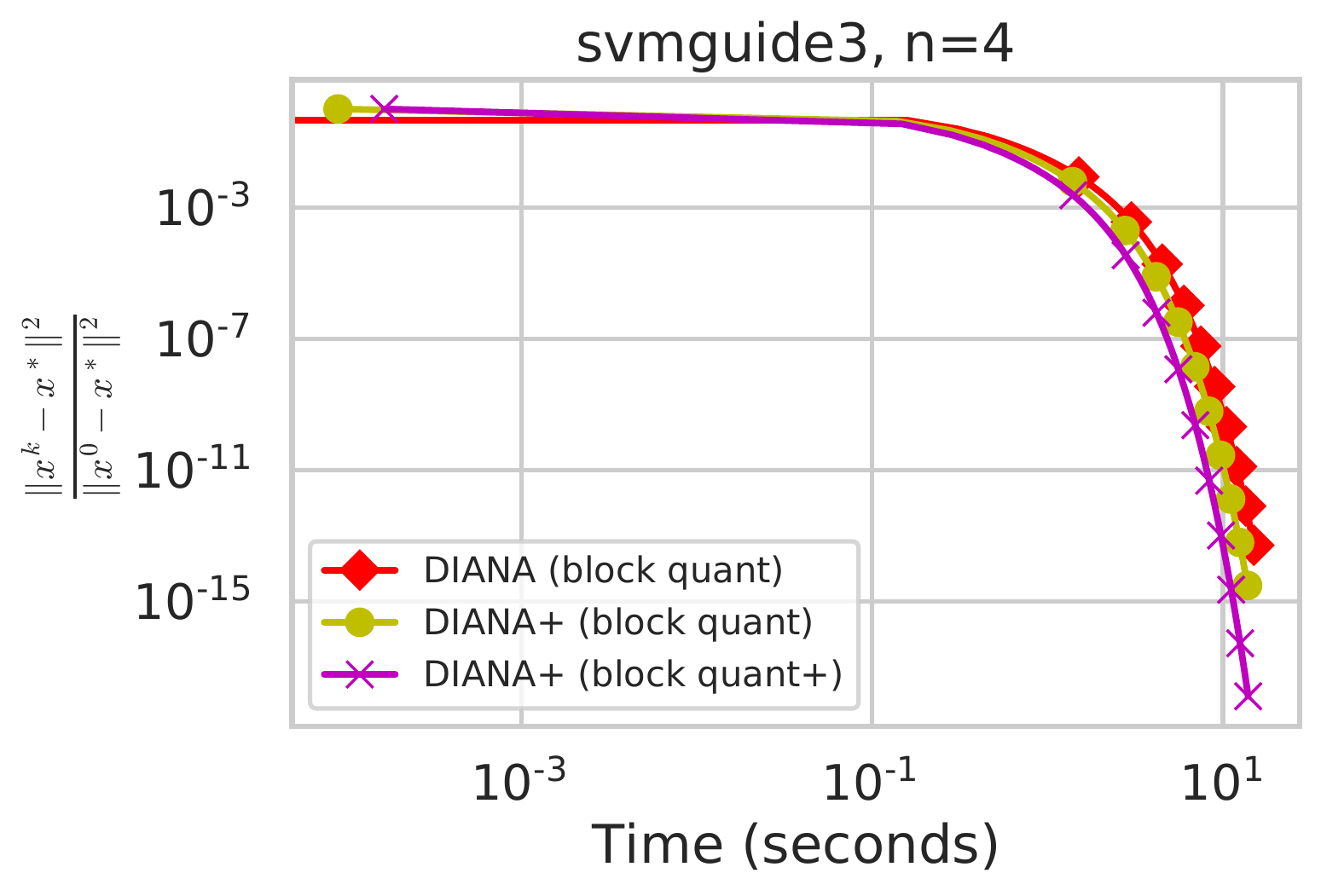}
    \endminipage\hfill

    \minipage{0.30\textwidth}
    \includegraphics[width=\linewidth]{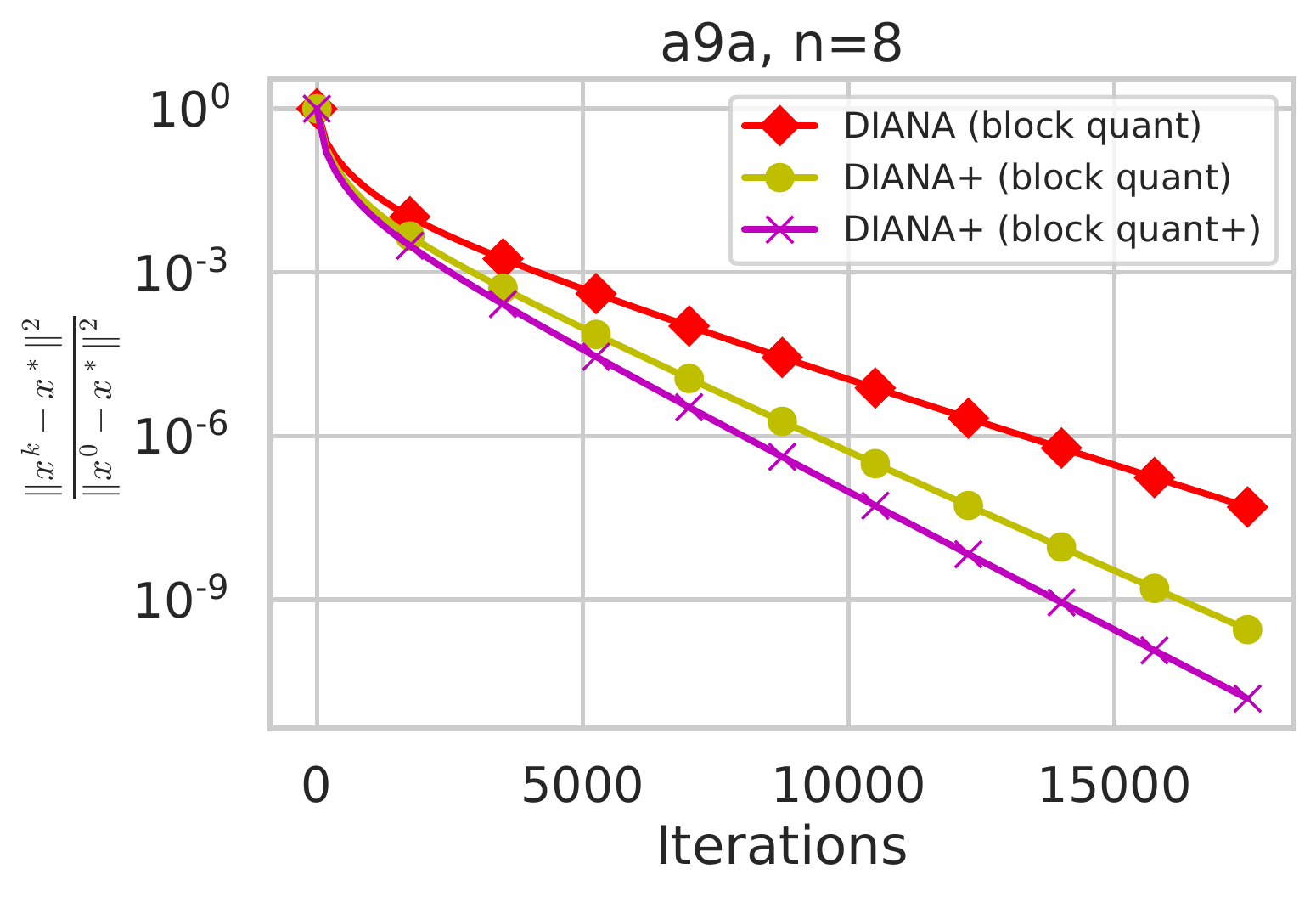}
    \endminipage\hfill  
        \minipage{0.30\textwidth}
    \includegraphics[width=\linewidth]{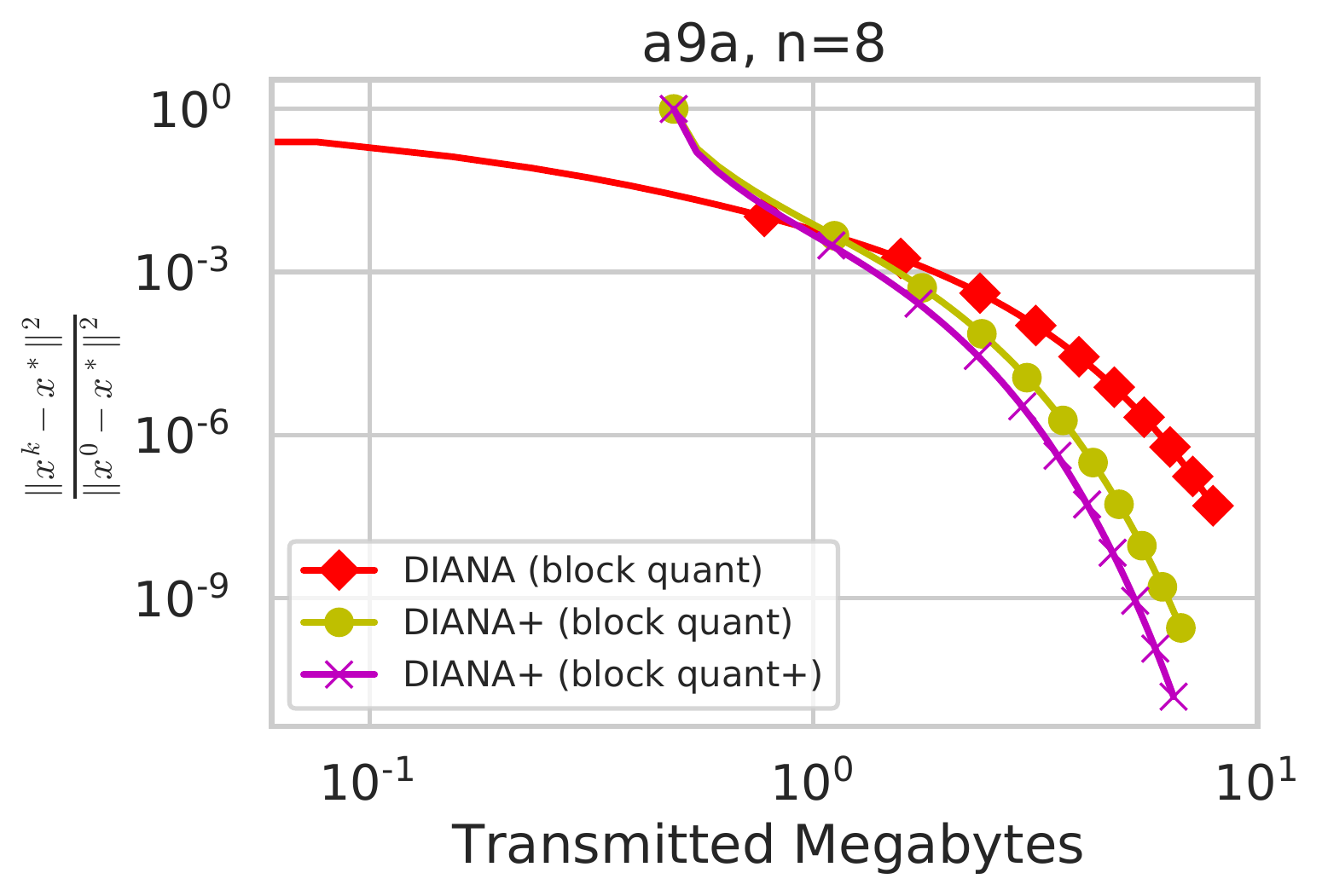}
    \endminipage\hfill
    \minipage{0.30\textwidth}
    \includegraphics[width=\linewidth]{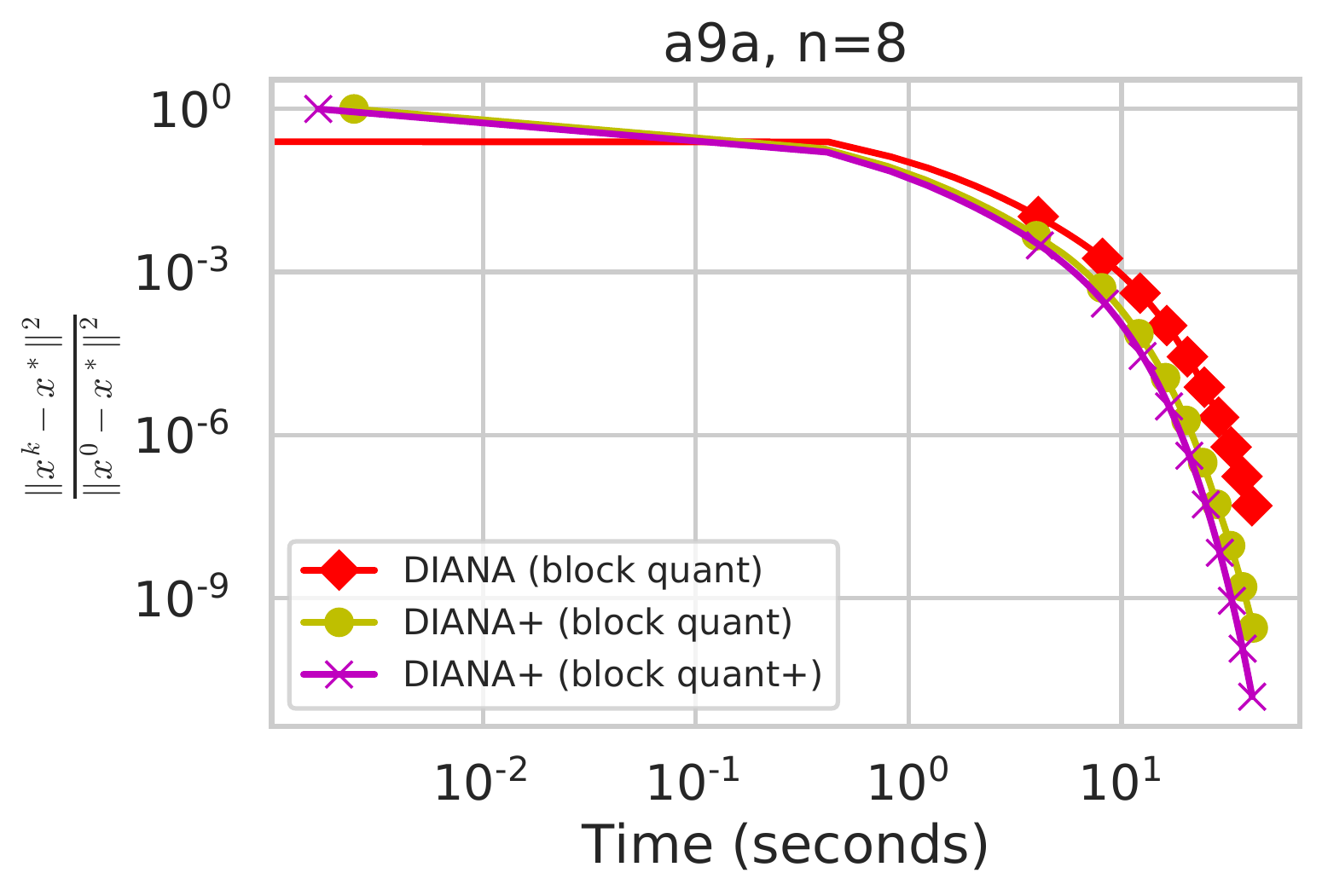}
    \endminipage\hfill  
    
    \minipage{0.30\textwidth}
    \includegraphics[width=\linewidth]{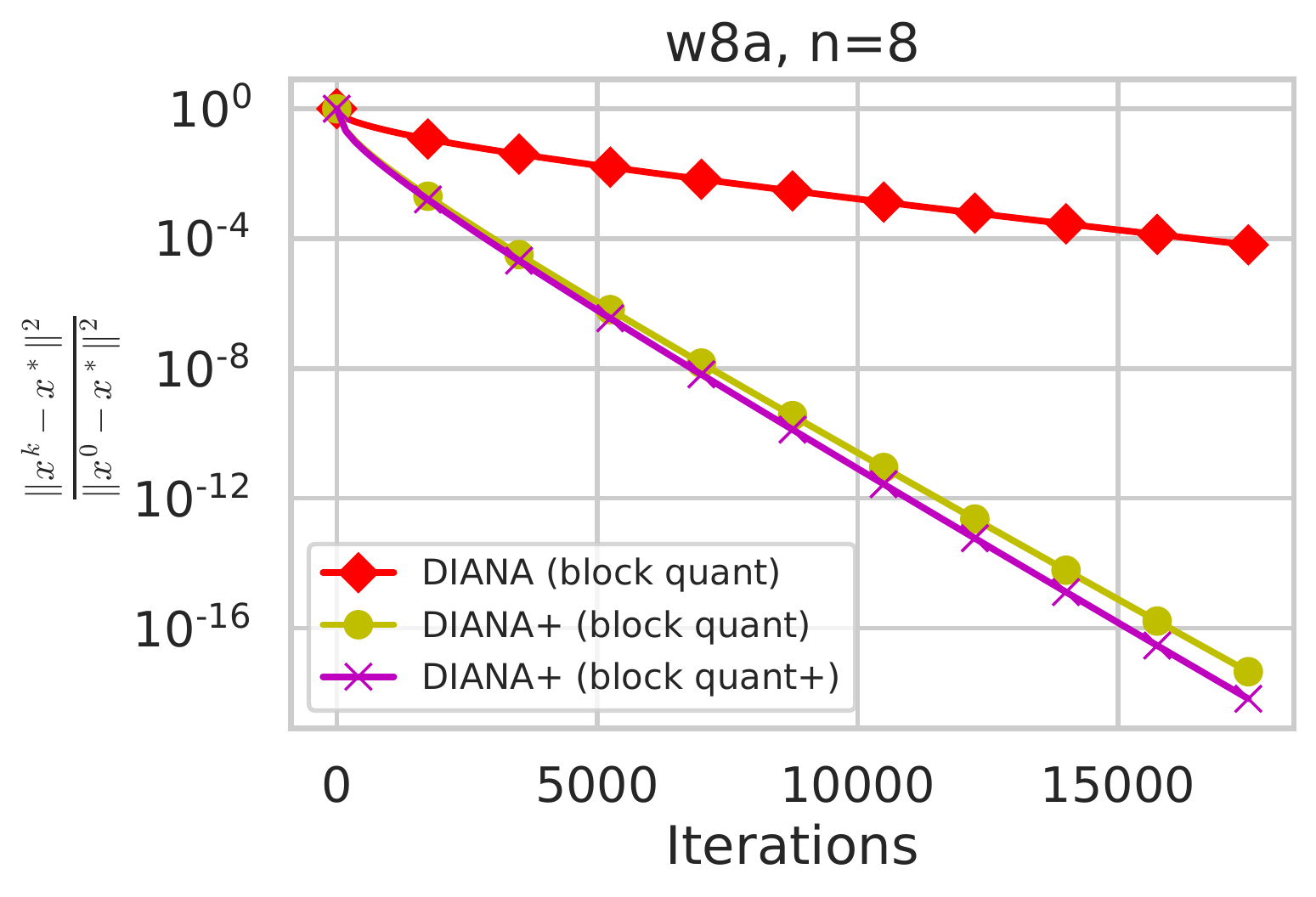}
    \endminipage\hfill  
    \minipage{0.30\textwidth}
    \includegraphics[width=\linewidth]{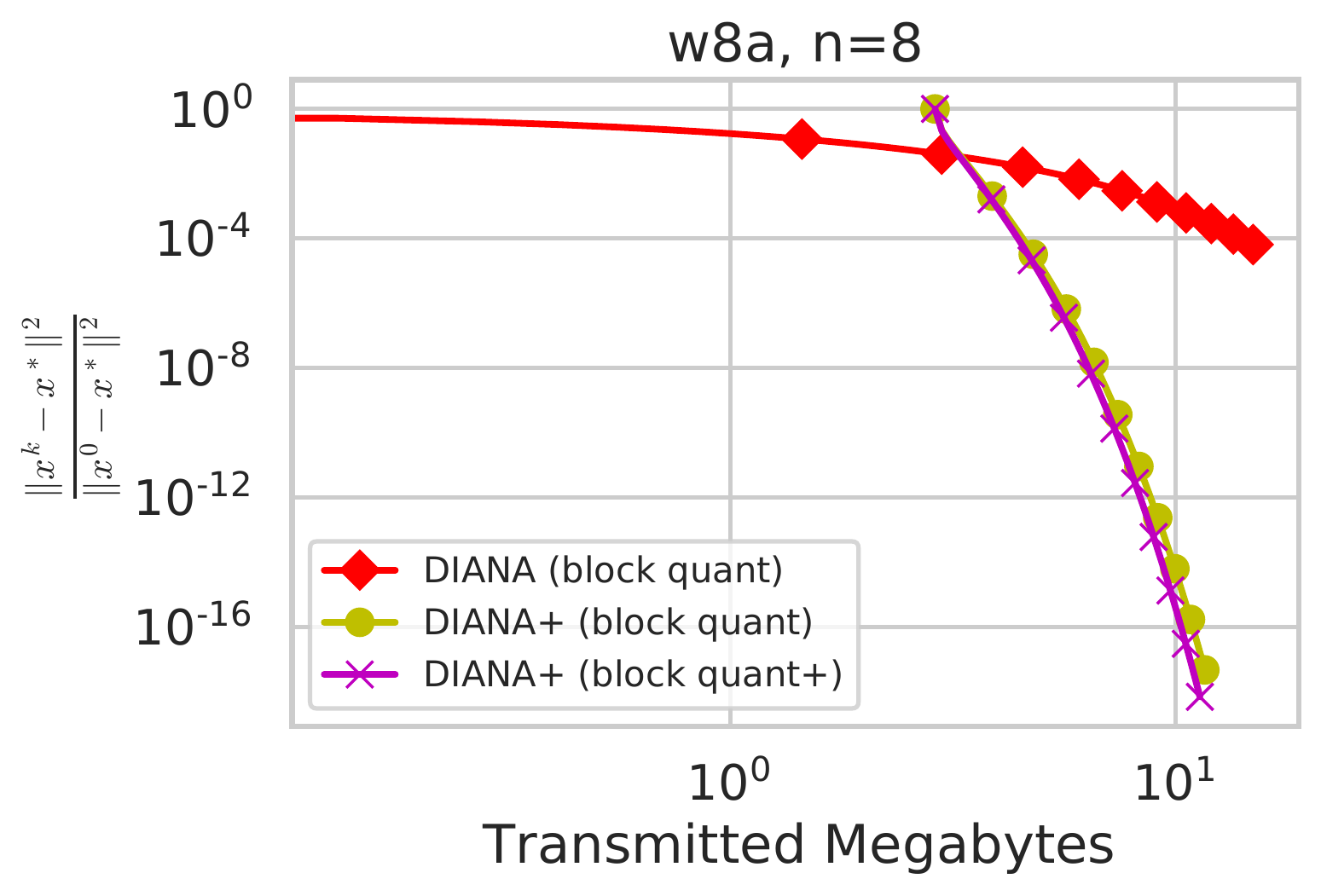}
    \endminipage\hfill
        \minipage{0.30\textwidth}
    \includegraphics[width=\linewidth]{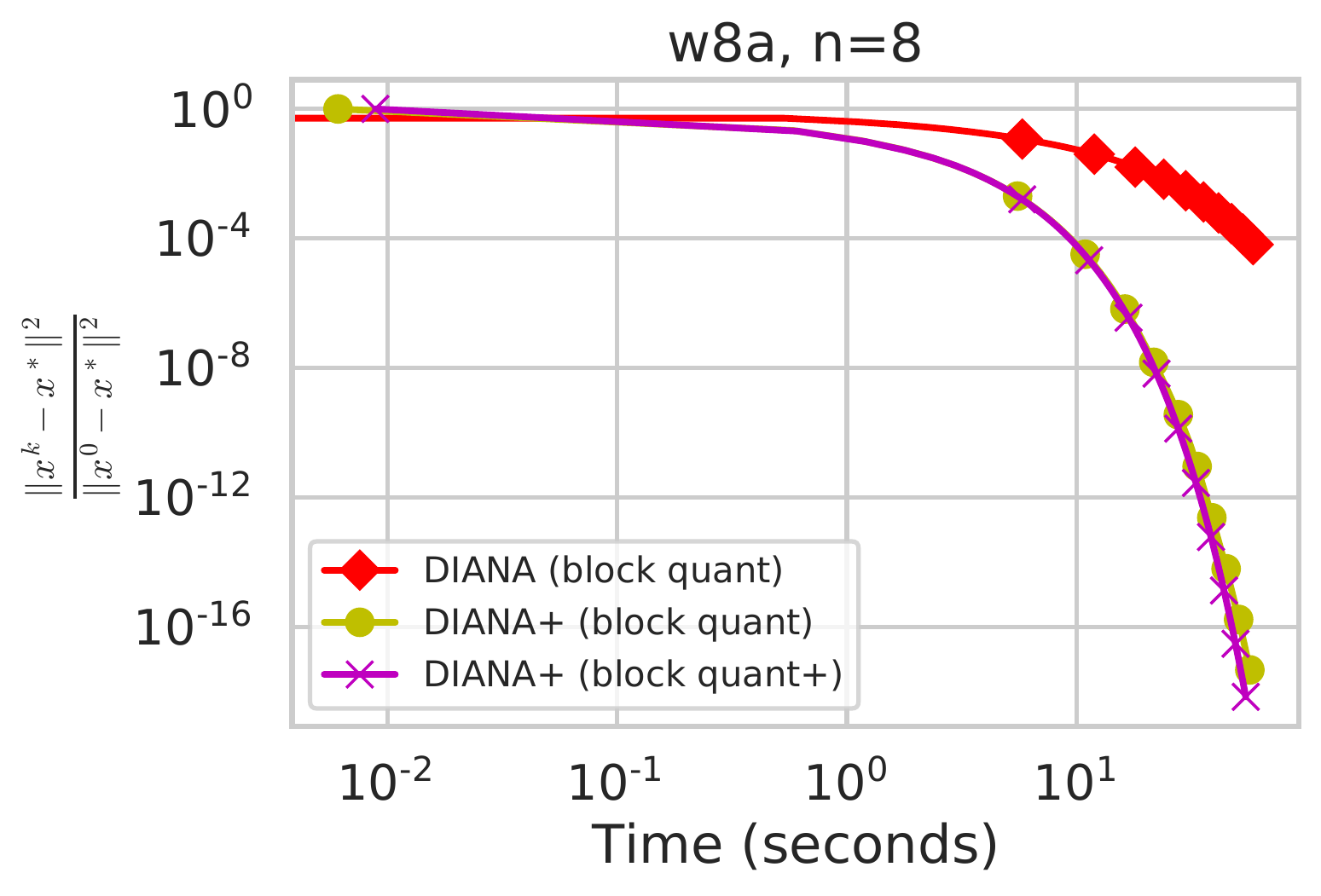}
    \endminipage\hfill  
    \caption{Comparison of DIANA+ (\texttt{block quant+}), DIANA+ (\texttt{block quant}), DIANA (\texttt{block quant}) and DIANA (\texttt{quant}).  }
    \label{fig:block_plus}
\end{figure*}

As mentioned by \cite{alistarh2017qsgd}, combining DCGD and block quantization can improve its iteration complexity at the cost of transmitting extra $32 B$ bits per iteration, which might also lead to better total communication complexity. Thus, the advantage of DIANA+ (\texttt{block quant+}) over DIANA (\texttt{quant}) may come from either splitting the features into blocks or exploiting the smoothness matrix. To further demistefy the improvement of DIANA+ (block quant+) , we compare the results of  DIANA+ (\texttt{block quant+}), DIANA+ (\texttt{block quant}), DIANA (\texttt{block quant}) and DIANA (\texttt{quant}) in Figure~\ref{fig:block_plus}. The difference between \texttt{block quant} and \texttt{block-quant+} is that the former one uses the same number of quantization levels for different blocks while the latter one uses varying numbers. It can be seen from Figure~\ref{fig:block_plus} that DIANA+ (\texttt{block-quant+}) consistently outperforms other methods because it optimally exploits the block structure and the smoothness matrix.

\begin{figure*}[htp]
  \minipage{0.30\textwidth}
\includegraphics[width=\linewidth]{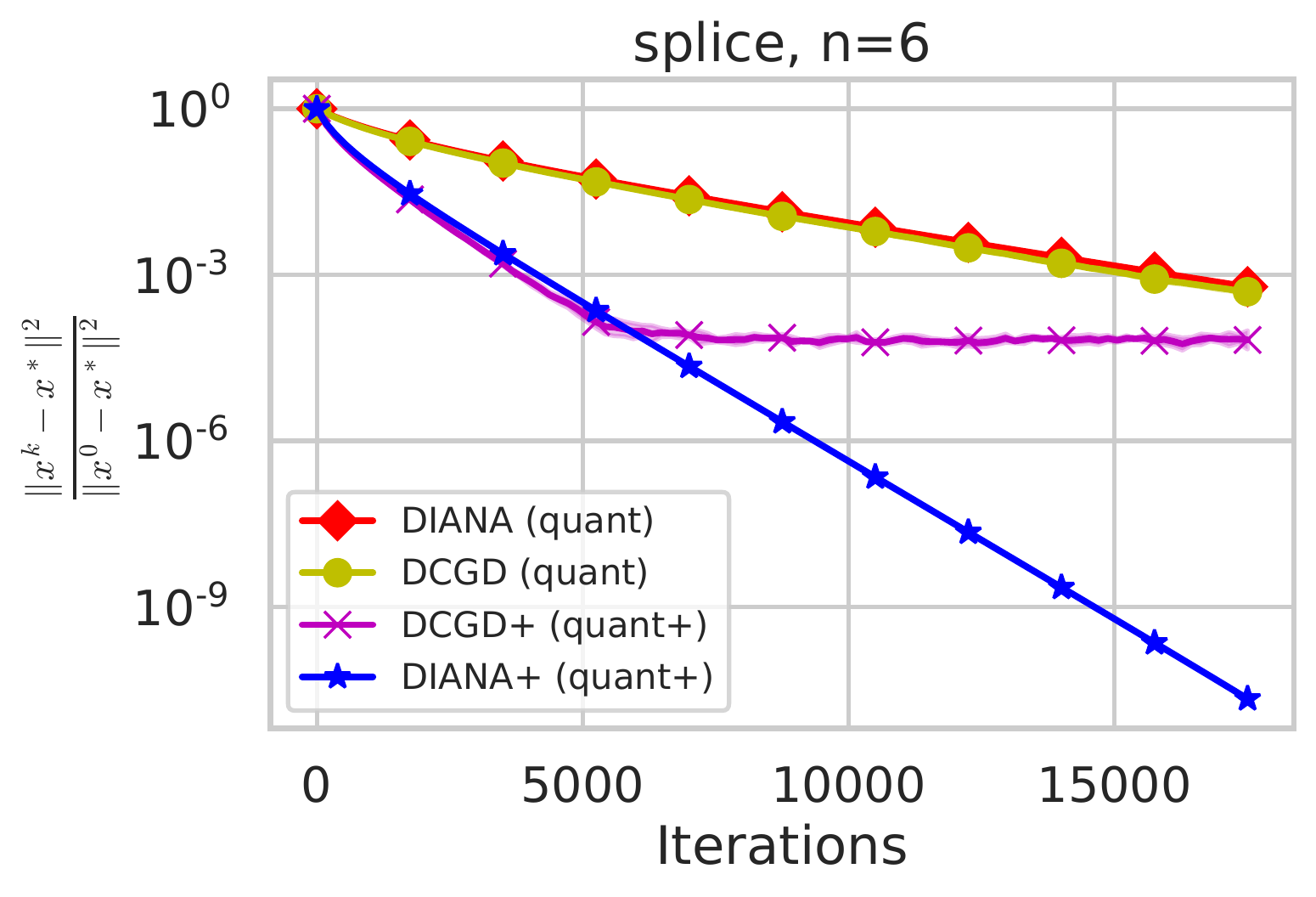}
\endminipage\hfill  
  \minipage{0.30\textwidth}
\includegraphics[width=\linewidth]{./figures/splice-scvx-6-inf/SD-DIANA-SD-DCGD-SD-DCGD-plus-SD-DIANA-plus_dist_trace_mbs.pdf}
\endminipage\hfill
 \minipage{0.30\textwidth}
\includegraphics[width=\linewidth]{./figures/splice-scvx-6-inf/SD-DIANA-SD-DCGD-SD-DCGD-plus-SD-DIANA-plus_dist_trace_time.pdf}
\endminipage\hfill  

  \minipage{0.30\textwidth}
  \includegraphics[width=\linewidth]{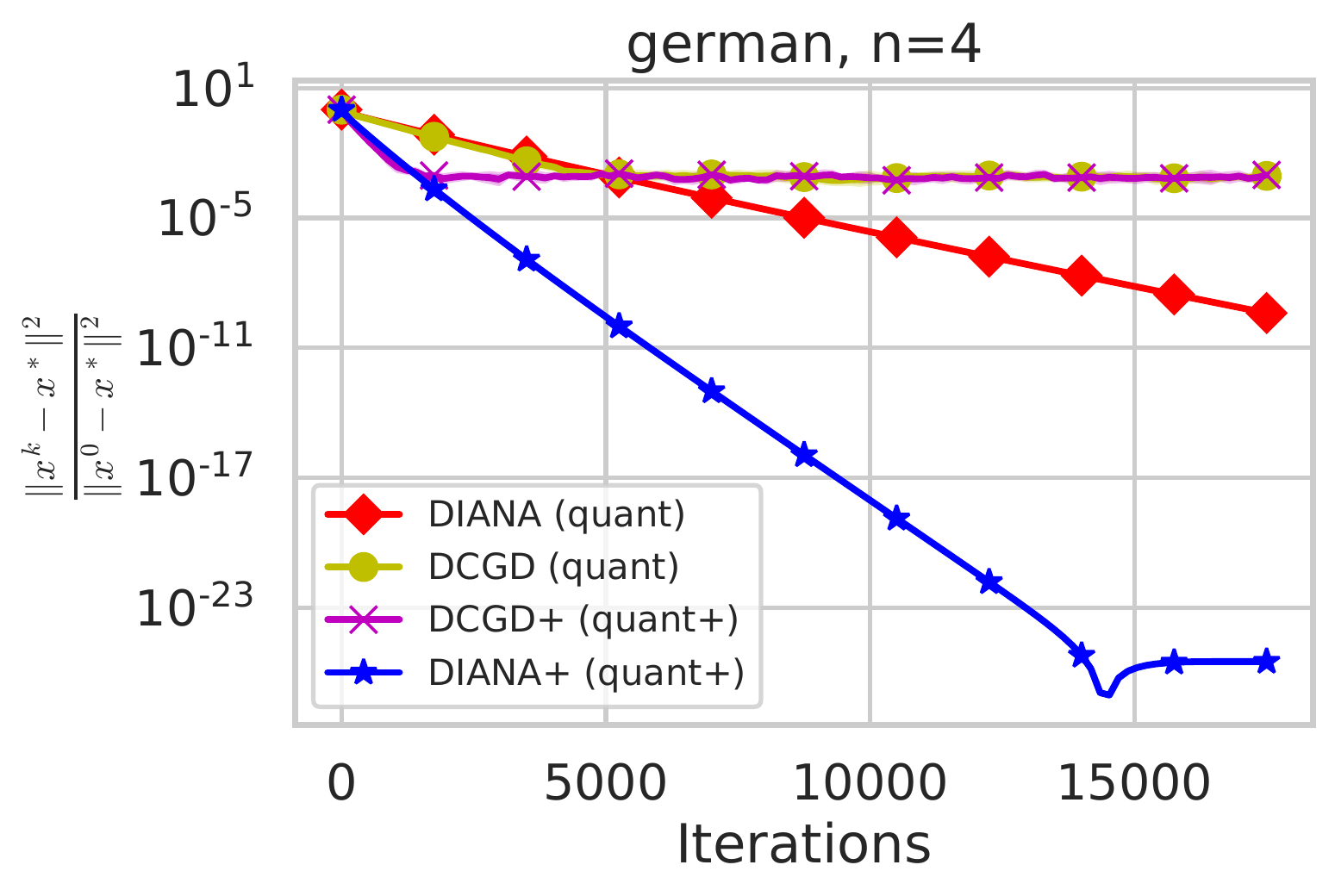}
  \endminipage\hfill  
   \minipage{0.30\textwidth}
  \includegraphics[width=\linewidth]{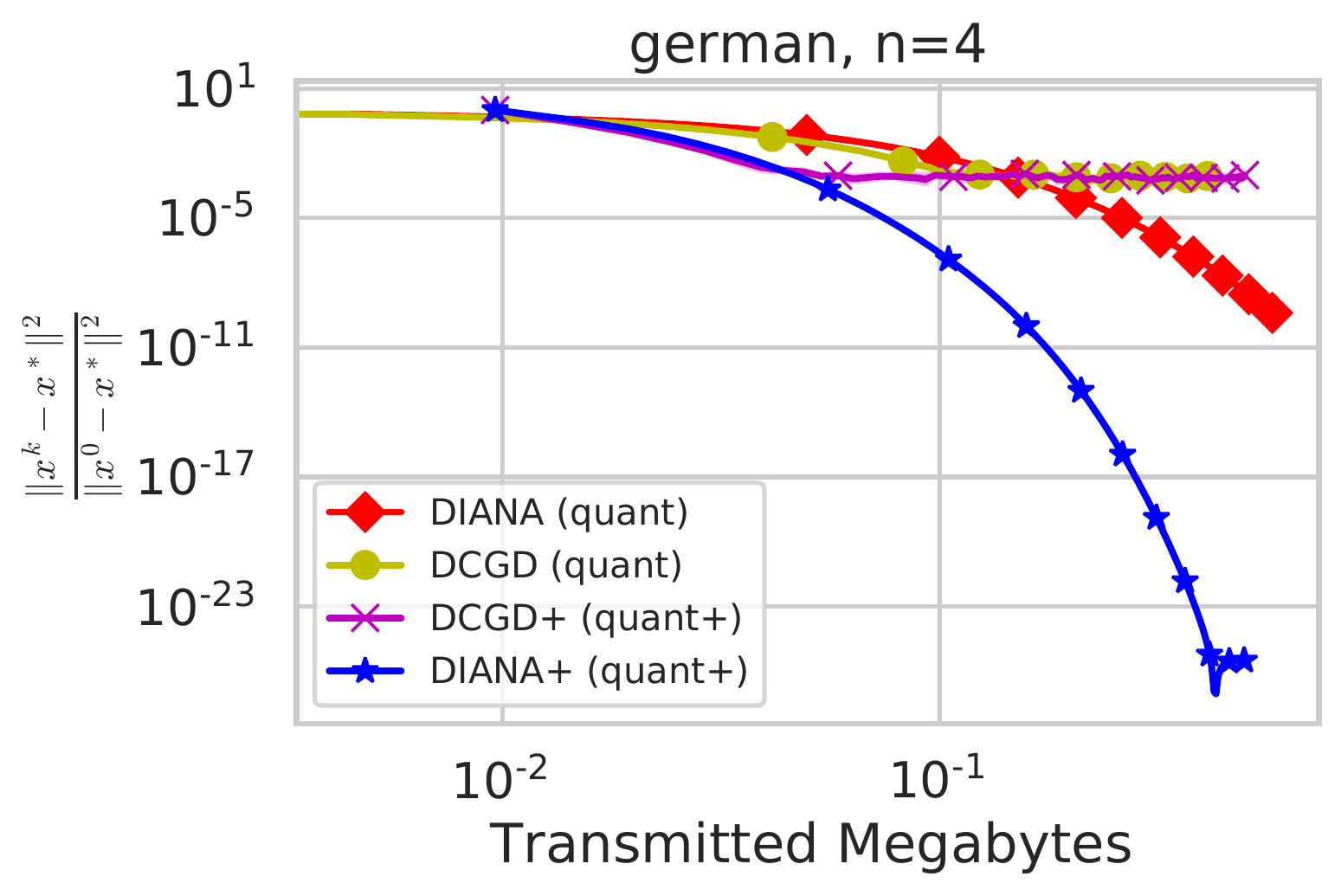}
  \endminipage\hfill
  \minipage{0.30\textwidth}
  \includegraphics[width=\linewidth]{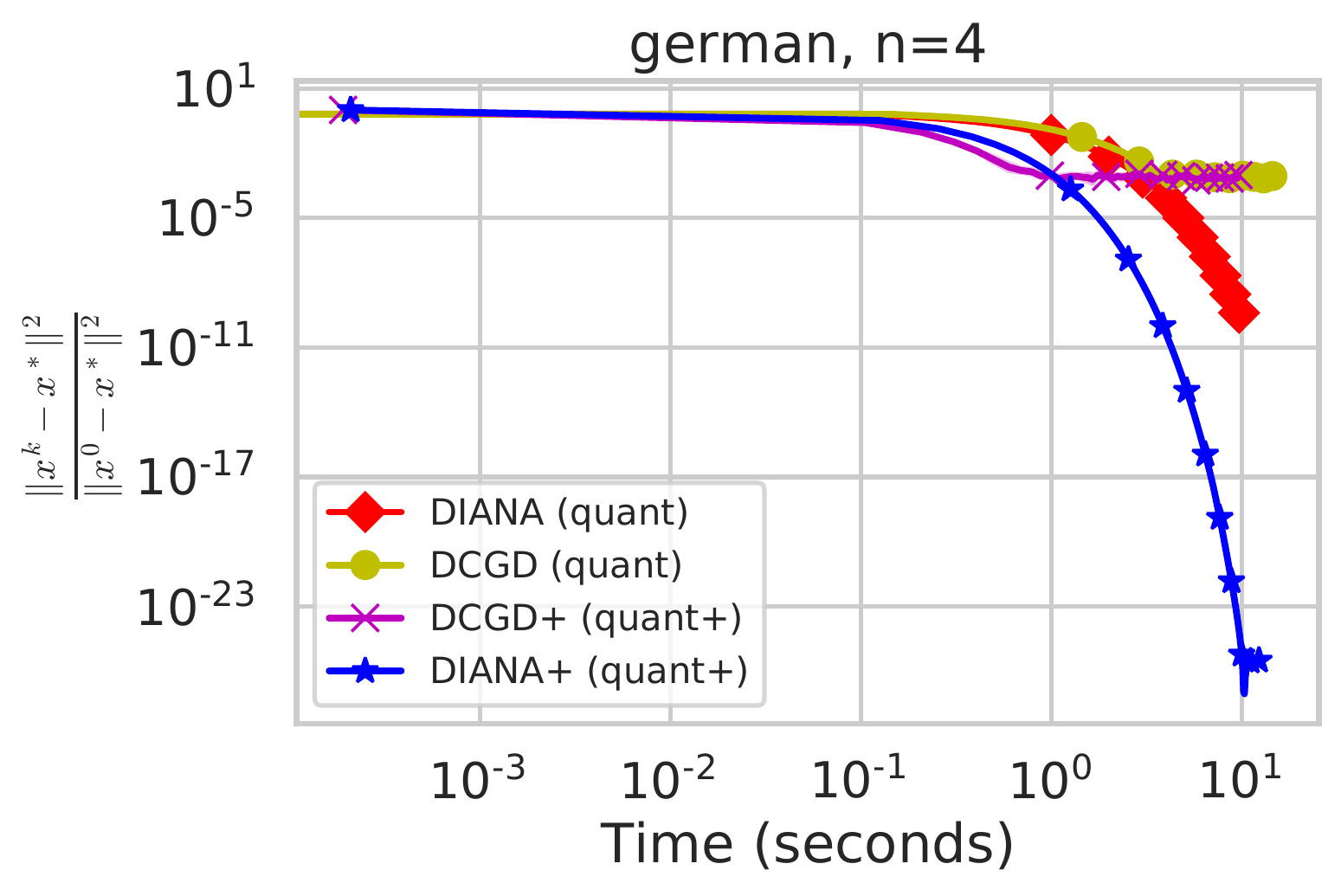}
  \endminipage\hfill  
  
  \minipage{0.30\textwidth}
  \includegraphics[width=\linewidth]{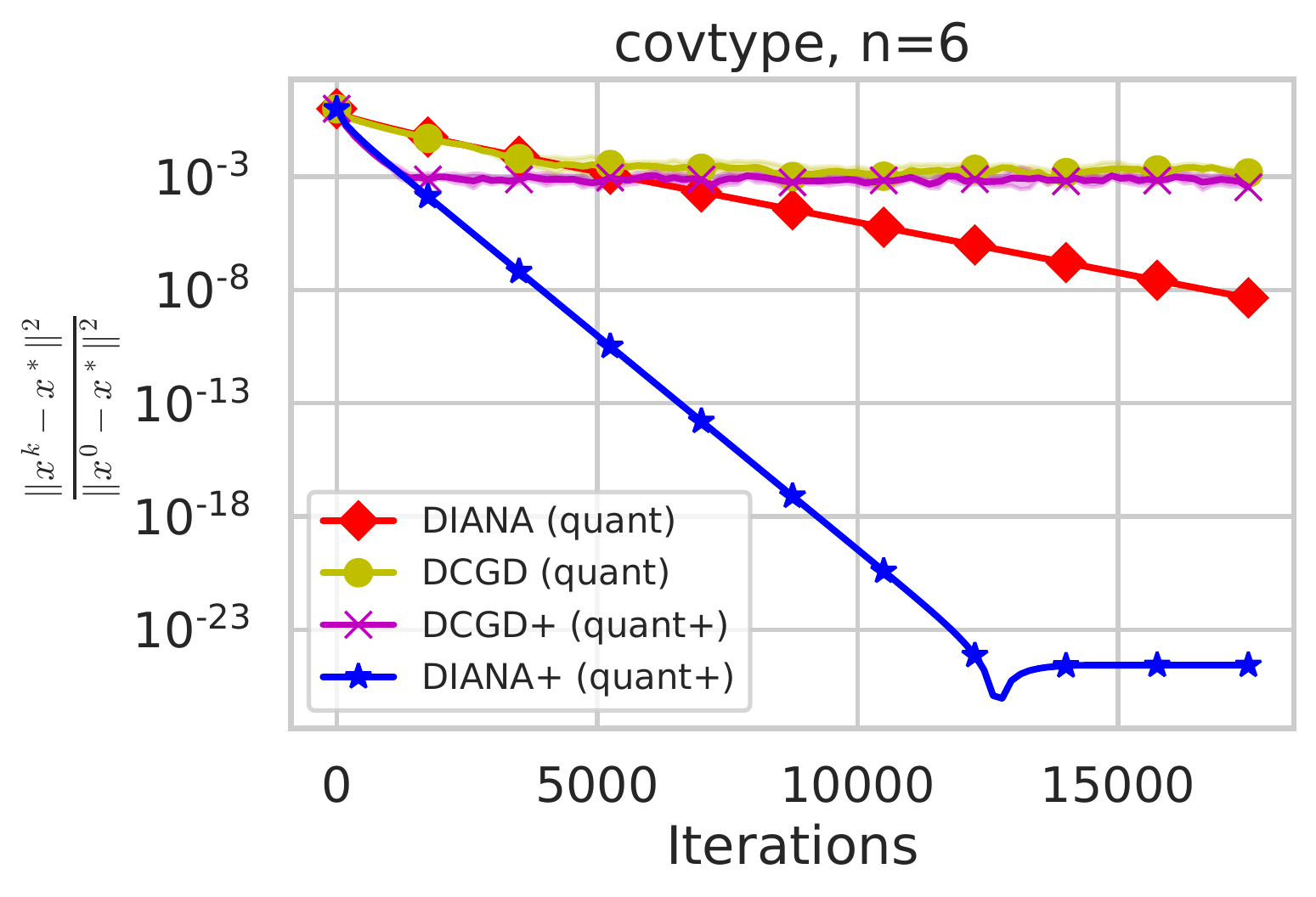}
  \endminipage\hfill  
    \minipage{0.30\textwidth}
  \includegraphics[width=\linewidth]{./figures/covtype-scvx-6-inf/SD-DIANA-SD-DCGD-SD-DCGD-plus-SD-DIANA-plus_dist_trace_mbs.pdf}
  \endminipage\hfill
   \minipage{0.30\textwidth}
  \includegraphics[width=\linewidth]{./figures/covtype-scvx-6-inf/SD-DIANA-SD-DCGD-SD-DCGD-plus-SD-DIANA-plus_dist_trace_time.pdf}
  \endminipage\hfill  
  
  \minipage{0.30\textwidth}
  \includegraphics[width=\linewidth]{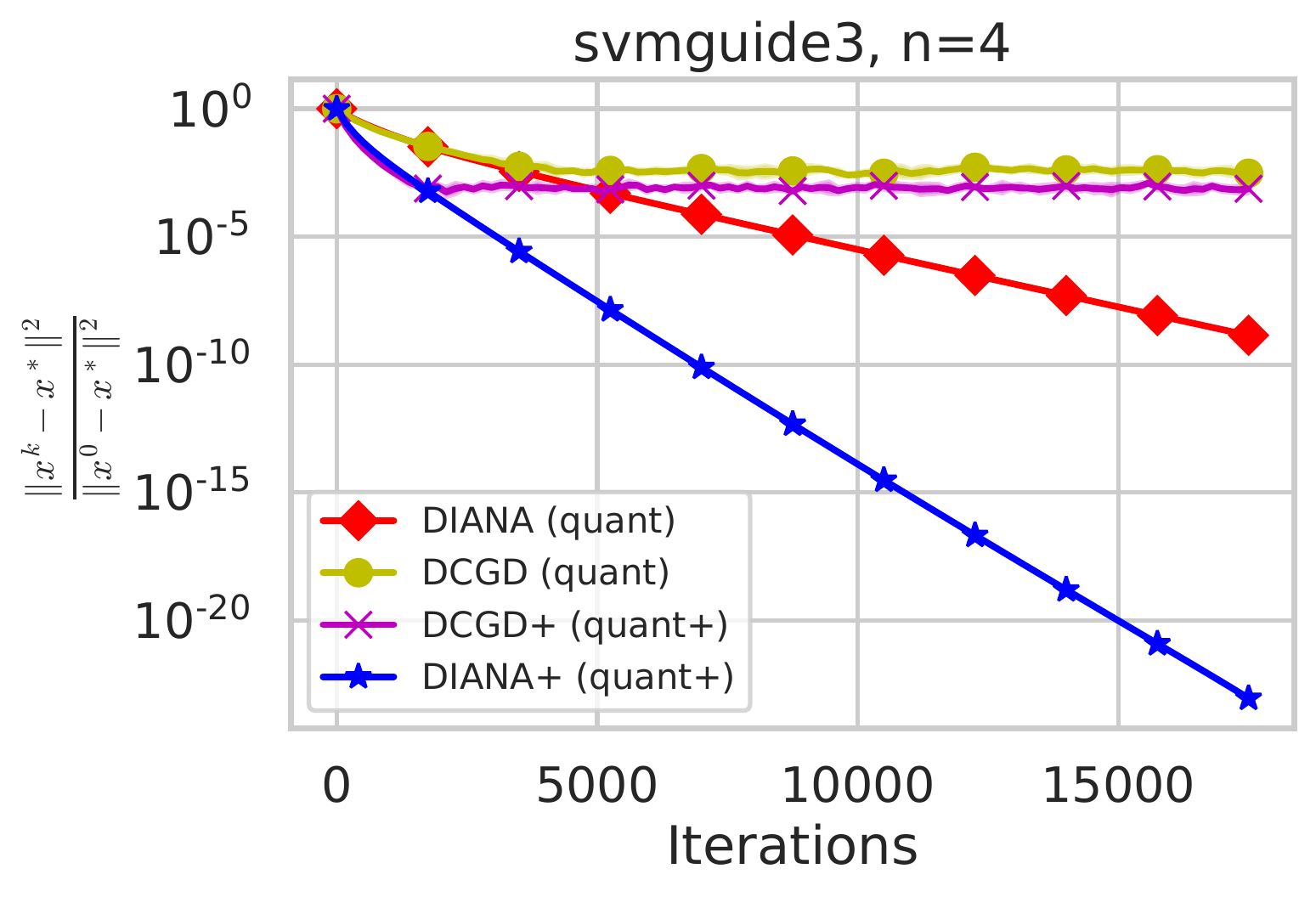}
  \endminipage\hfill  
      \minipage{0.30\textwidth}
  \includegraphics[width=\linewidth]{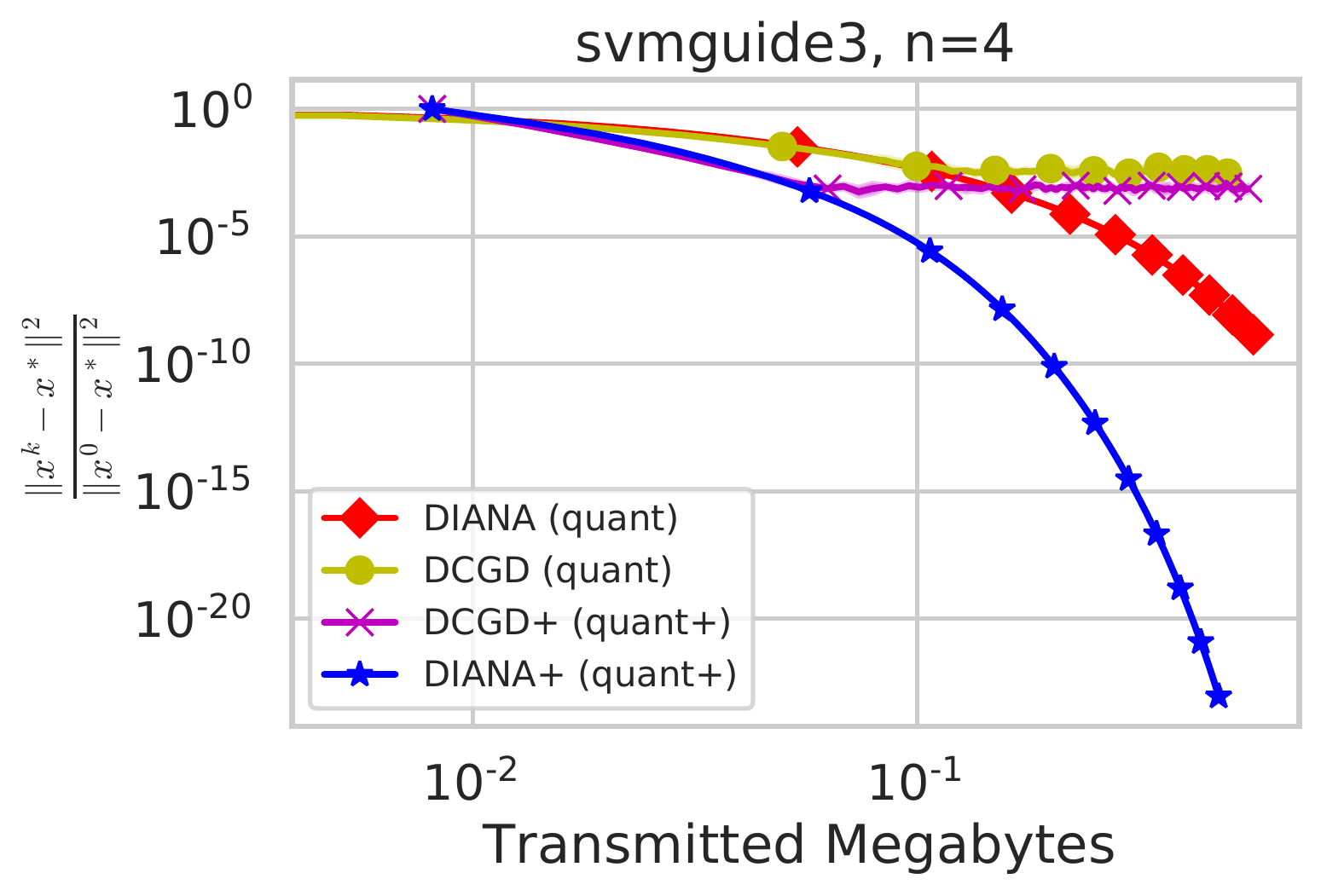}
  \endminipage\hfill
  \minipage{0.30\textwidth}
  \includegraphics[width=\linewidth]{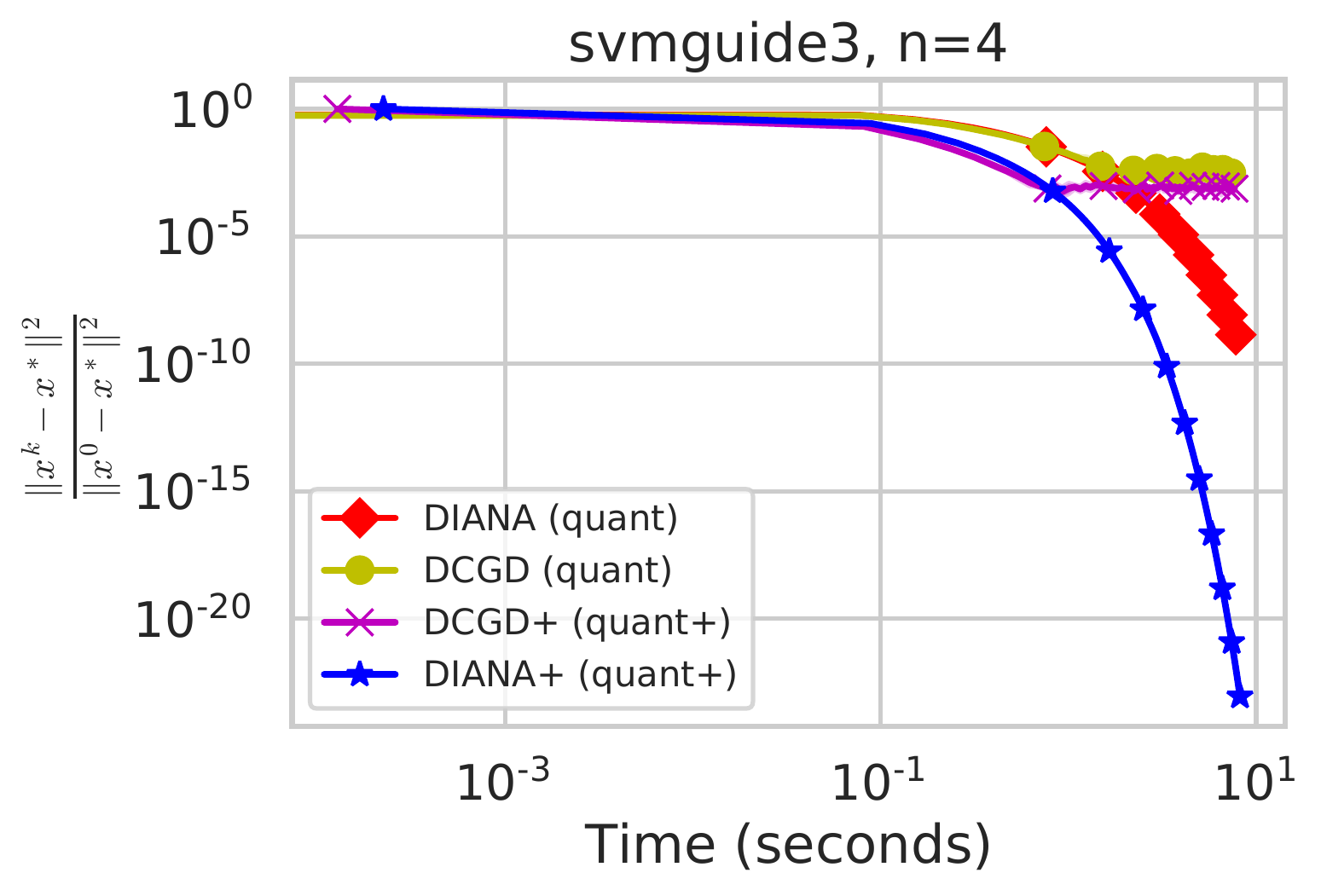}
  \endminipage\hfill  
  
  \minipage{0.30\textwidth}
  \includegraphics[width=\linewidth]{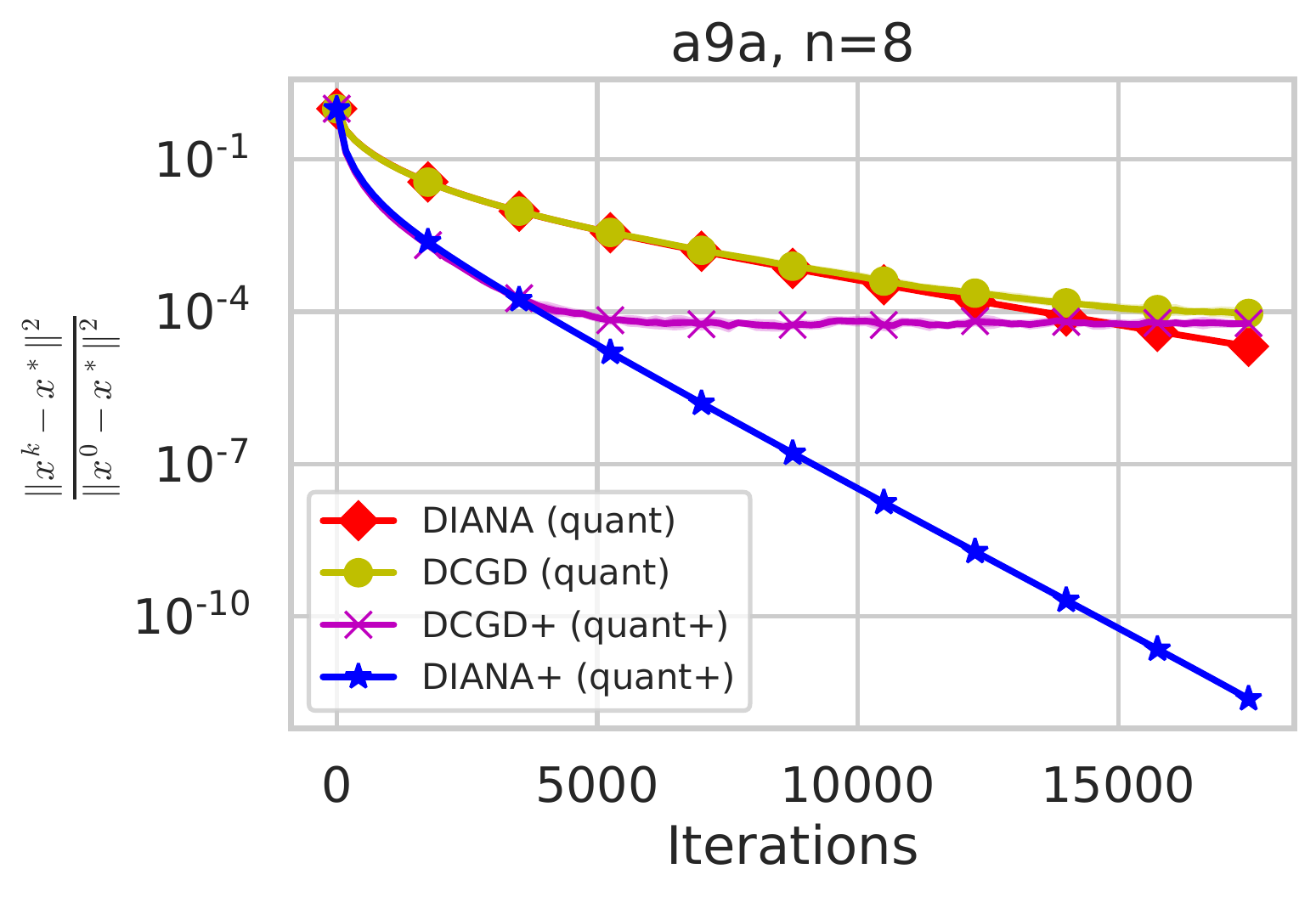}
  \endminipage\hfill  
    \minipage{0.30\textwidth}
  \includegraphics[width=\linewidth]{./figures/a9a-scvx-8-inf/SD-DIANA-SD-DCGD-SD-DCGD-plus-SD-DIANA-plus_dist_trace_mbs.pdf}
  \endminipage\hfill
  \minipage{0.30\textwidth}
\includegraphics[width=\linewidth]{./figures/a9a-scvx-8-inf/SD-DIANA-SD-DCGD-SD-DCGD-plus-SD-DIANA-plus_dist_trace_time.pdf}
\endminipage\hfill    
  
    \minipage{0.30\textwidth}
  \includegraphics[width=\linewidth]{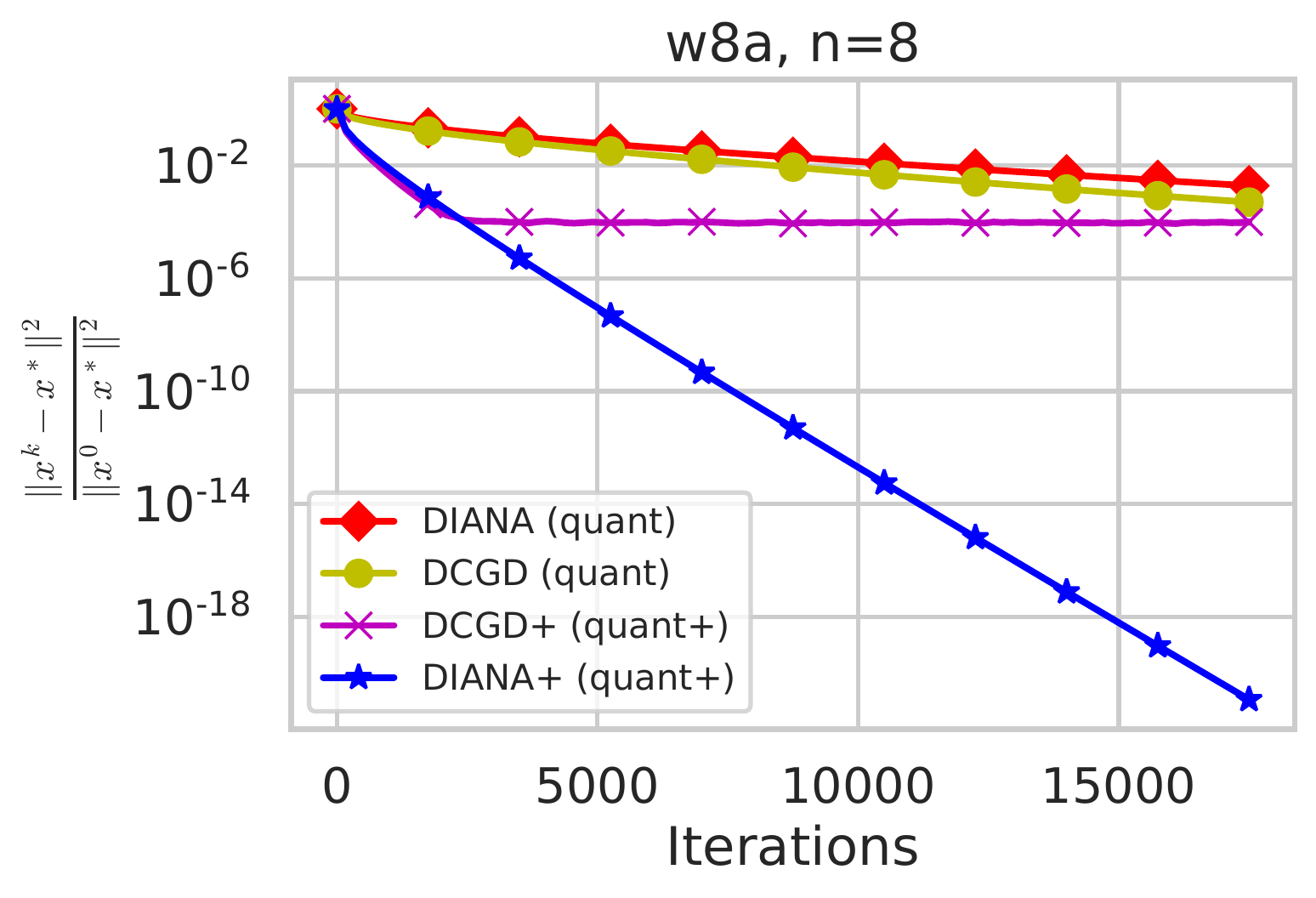}
  \endminipage\hfill  
     \minipage{0.30\textwidth}
\includegraphics[width=\linewidth]{./figures/w8a-scvx-8-inf/SD-DIANA-SD-DCGD-SD-DCGD-plus-SD-DIANA-plus_dist_trace_mbs.pdf}
\endminipage\hfill
  \minipage{0.30\textwidth}
\includegraphics[width=\linewidth]{./figures/w8a-scvx-8-inf/SD-DIANA-SD-DCGD-SD-DCGD-plus-SD-DIANA-plus_dist_trace_time.pdf}
\endminipage\hfill  
  \caption{Comparison of DCGD+ (\texttt{quant+}) and DIANA+ (\texttt{quant+}) with DCGD (\texttt{quant}) and DIANA (\texttt{quant}).  }
  \label{fig:varying}
\end{figure*}

We also demonstrate that how DIANA+ perform with varying or fixed number of levels. As seen in Figure~\ref{fig:varying}, the varying number of levels are beneficial on most of the datasets.

\subsection{Comparison to matrix-smoothness-aware sparsification}

\begin{figure*}[htp]    
        \minipage{0.30\textwidth}
    \includegraphics[width=\linewidth]{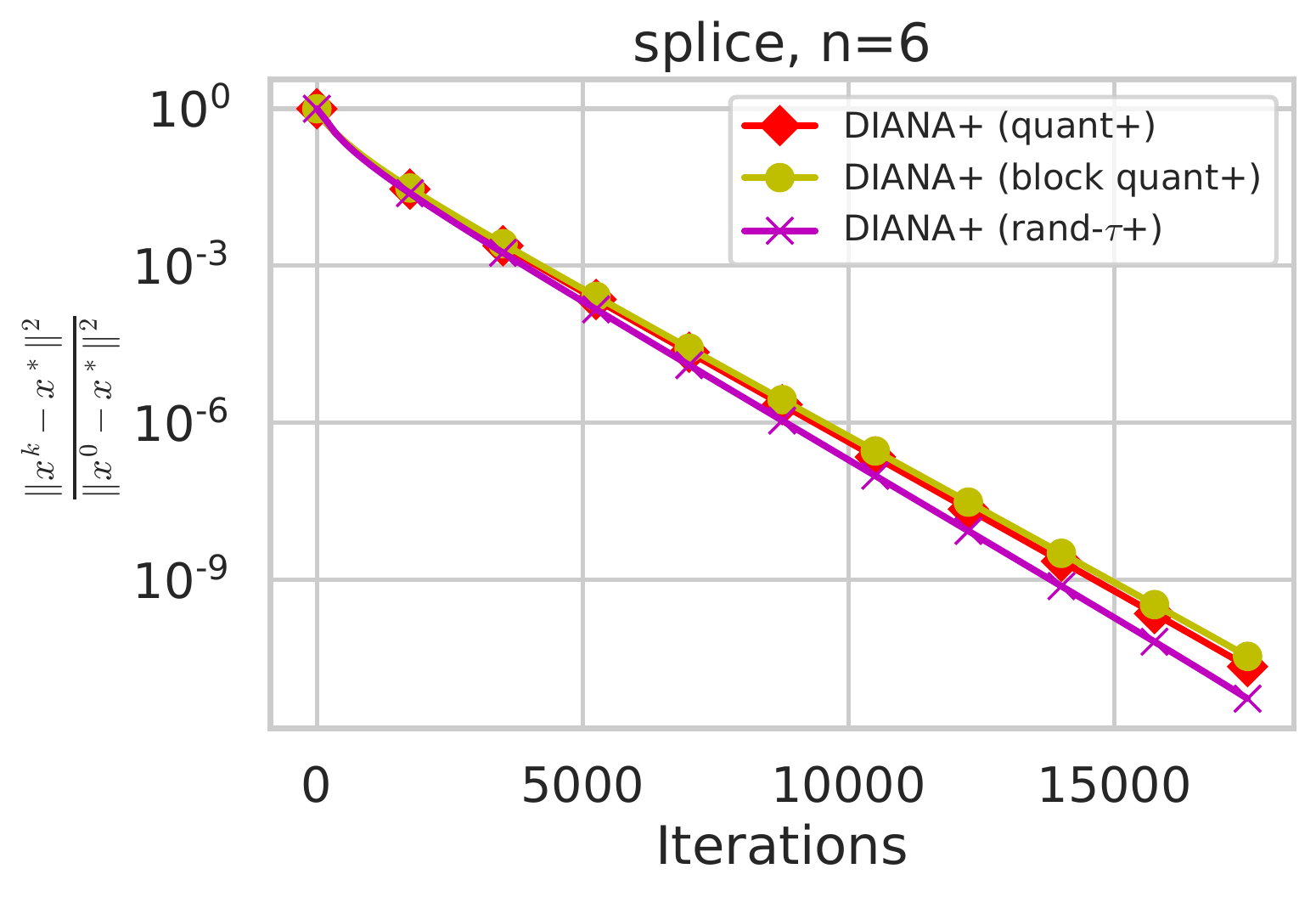}
    \endminipage\hfill  
    \minipage{0.30\textwidth}
    \includegraphics[width=\linewidth]{./figures/splice-scvx-6-inf/SD-DIANA-plus-BL-DIANA-plus-Sparse-DIANA-plus_dist_trace_mbs.pdf}
    \endminipage\hfill  
    \minipage{0.30\textwidth}
    \includegraphics[width=\linewidth]{./figures/splice-scvx-6-inf/SD-DIANA-plus-BL-DIANA-plus-Sparse-DIANA-plus_dist_trace_time.pdf}
    \endminipage\hfill  

    \minipage{0.30\textwidth}
\includegraphics[width=\linewidth]{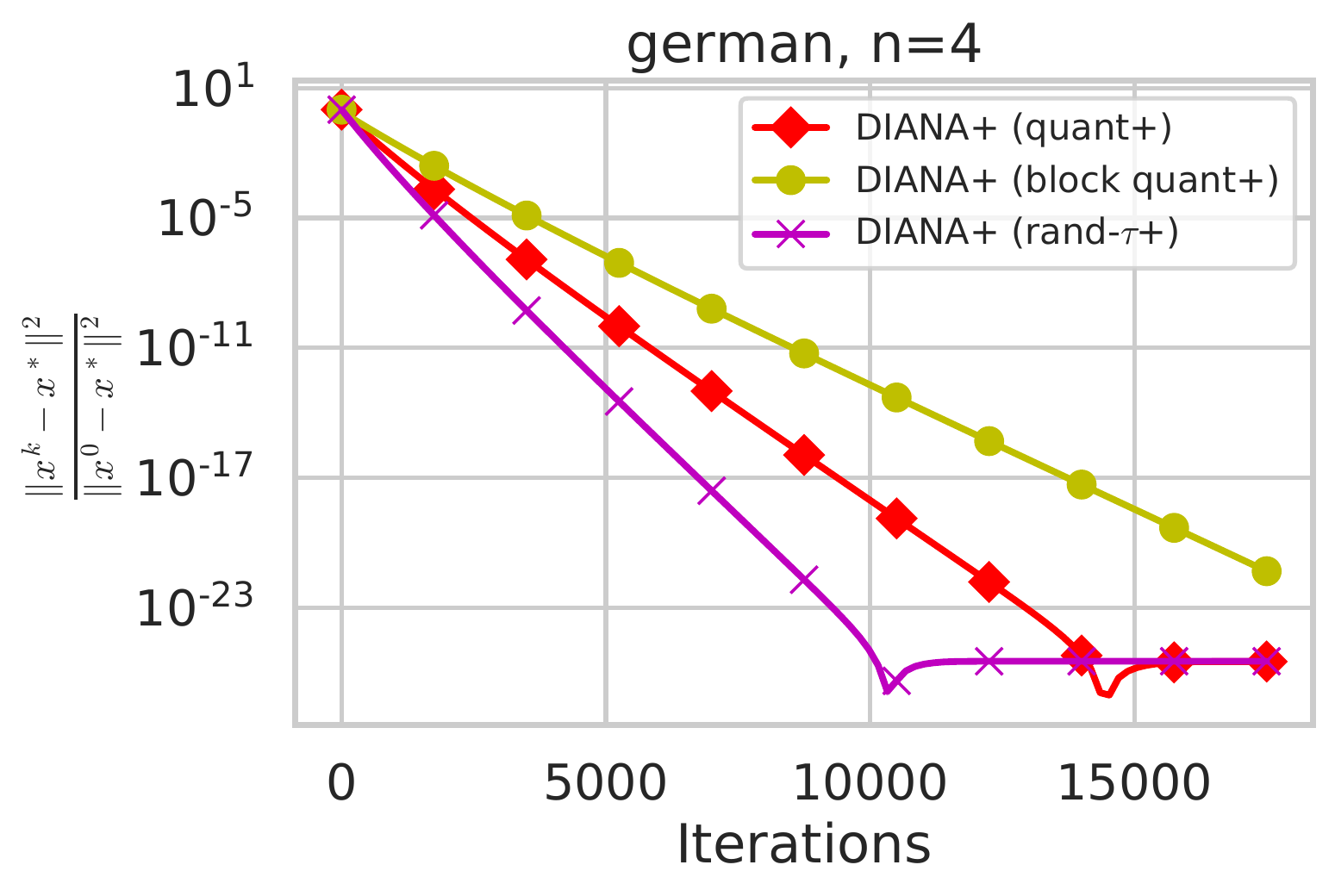}
\endminipage\hfill      
    \minipage{0.30\textwidth}
    \includegraphics[width=\linewidth]{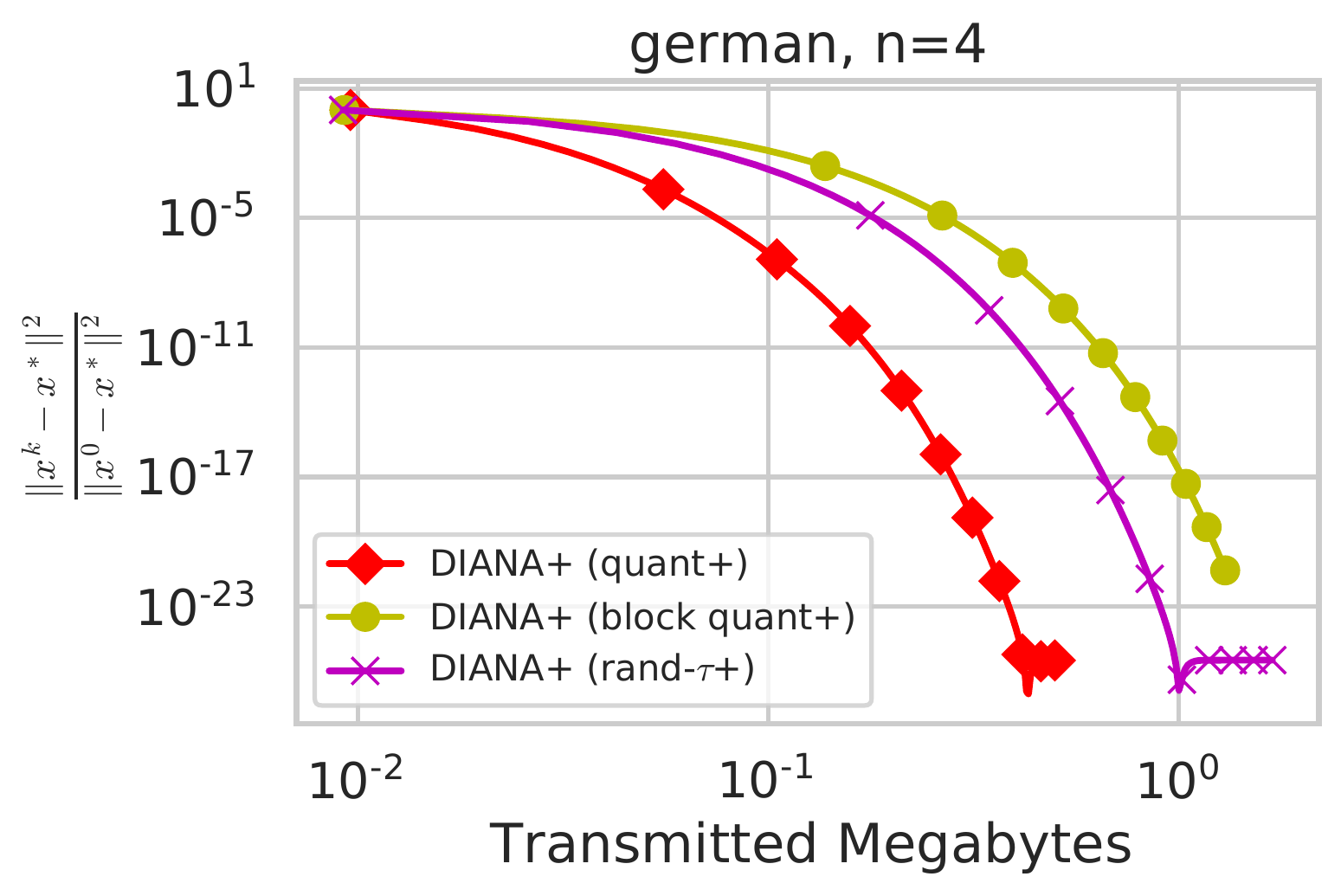}
    \endminipage\hfill  
    \minipage{0.30\textwidth}
    \includegraphics[width=\linewidth]{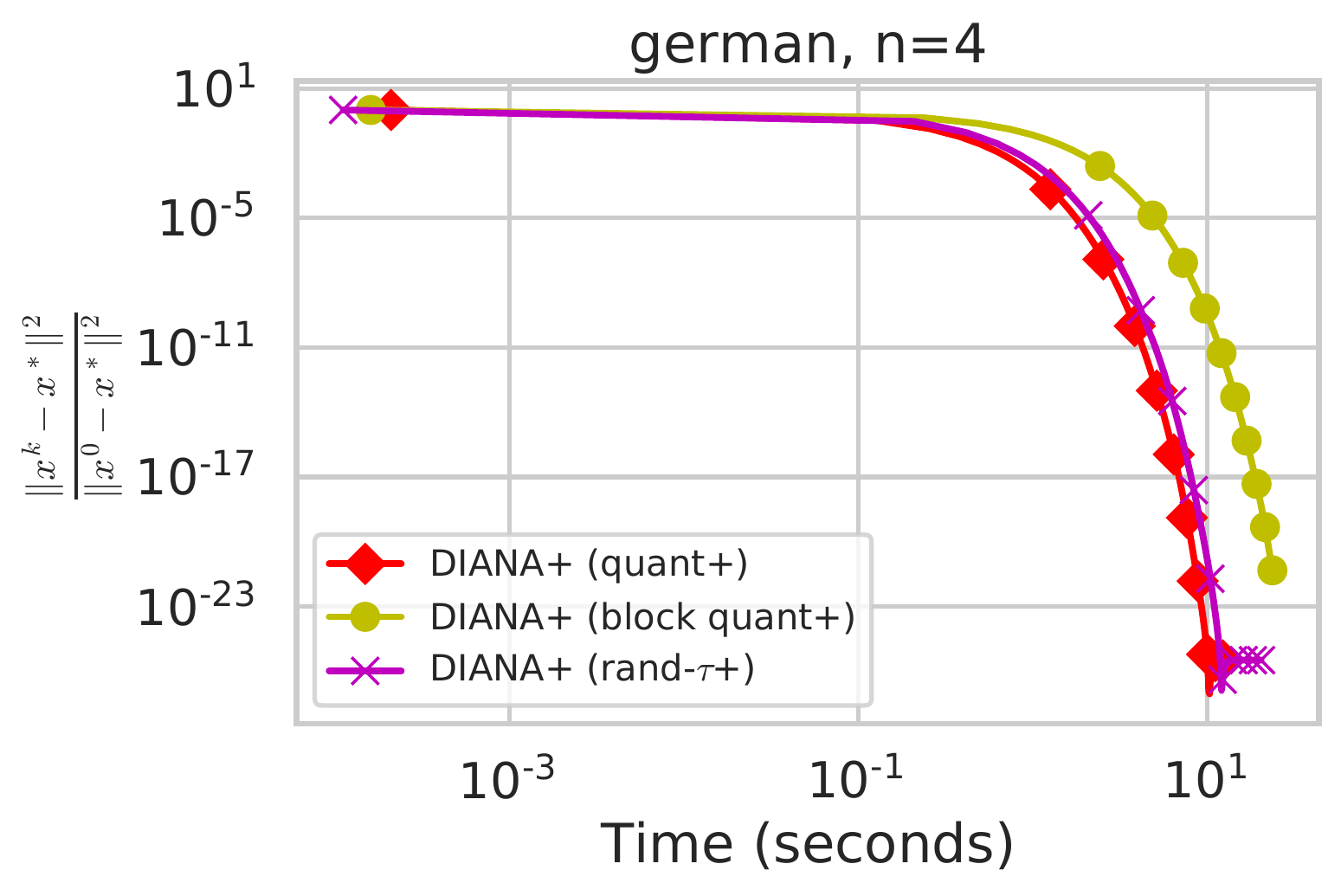}
    \endminipage\hfill  
    
        \minipage{0.30\textwidth}
    \includegraphics[width=\linewidth]{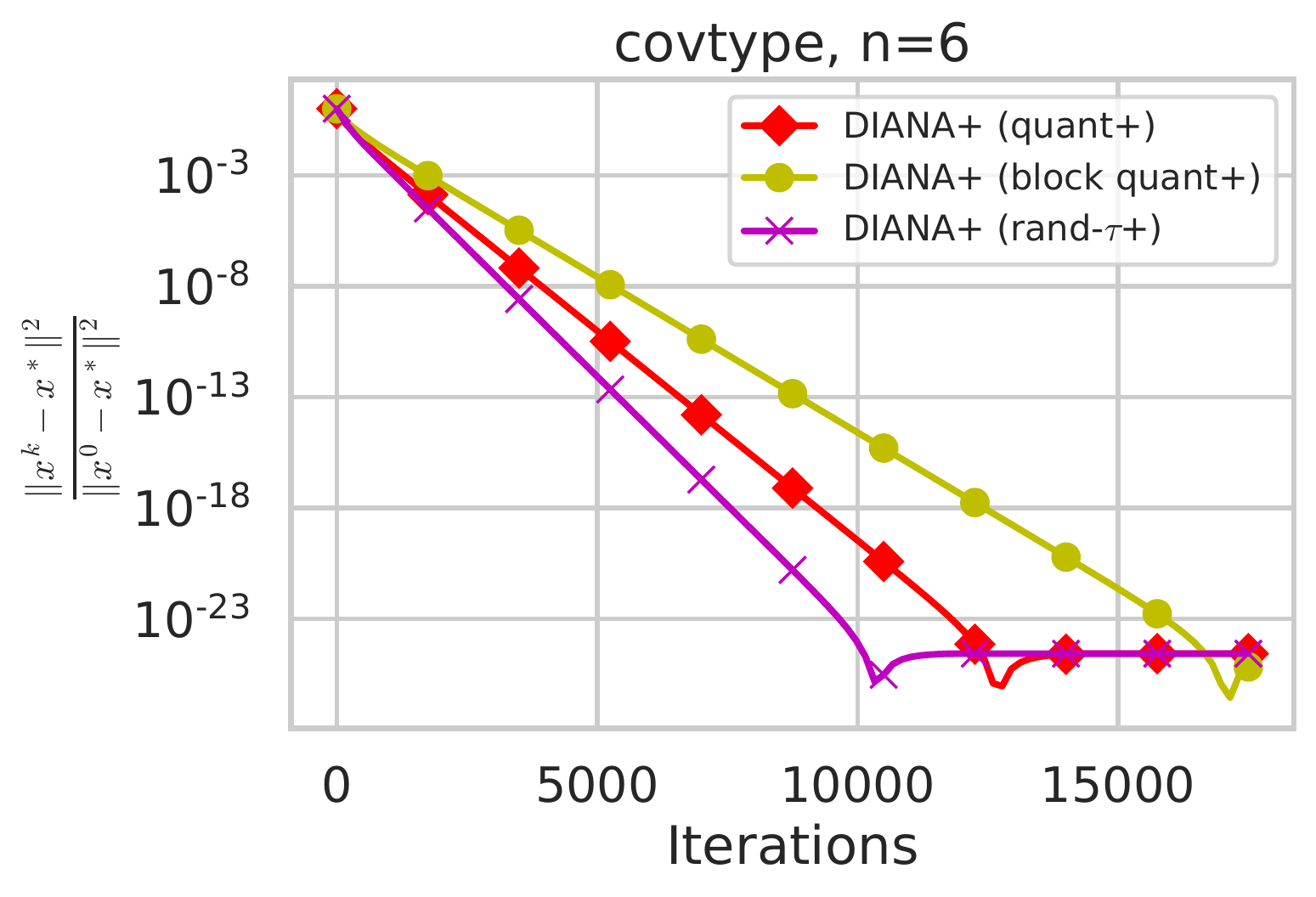}
    \endminipage\hfill  
    \minipage{0.30\textwidth}
    \includegraphics[width=\linewidth]{./figures/covtype-scvx-6-inf/SD-DIANA-plus-BL-DIANA-plus-Sparse-DIANA-plus_dist_trace_mbs.pdf}
    \endminipage\hfill  
    \minipage{0.30\textwidth}
    \includegraphics[width=\linewidth]{./figures/covtype-scvx-6-inf/SD-DIANA-plus-BL-DIANA-plus-Sparse-DIANA-plus_dist_trace_time.pdf}
    \endminipage\hfill  
    
    \minipage{0.30\textwidth}
    \includegraphics[width=\linewidth]{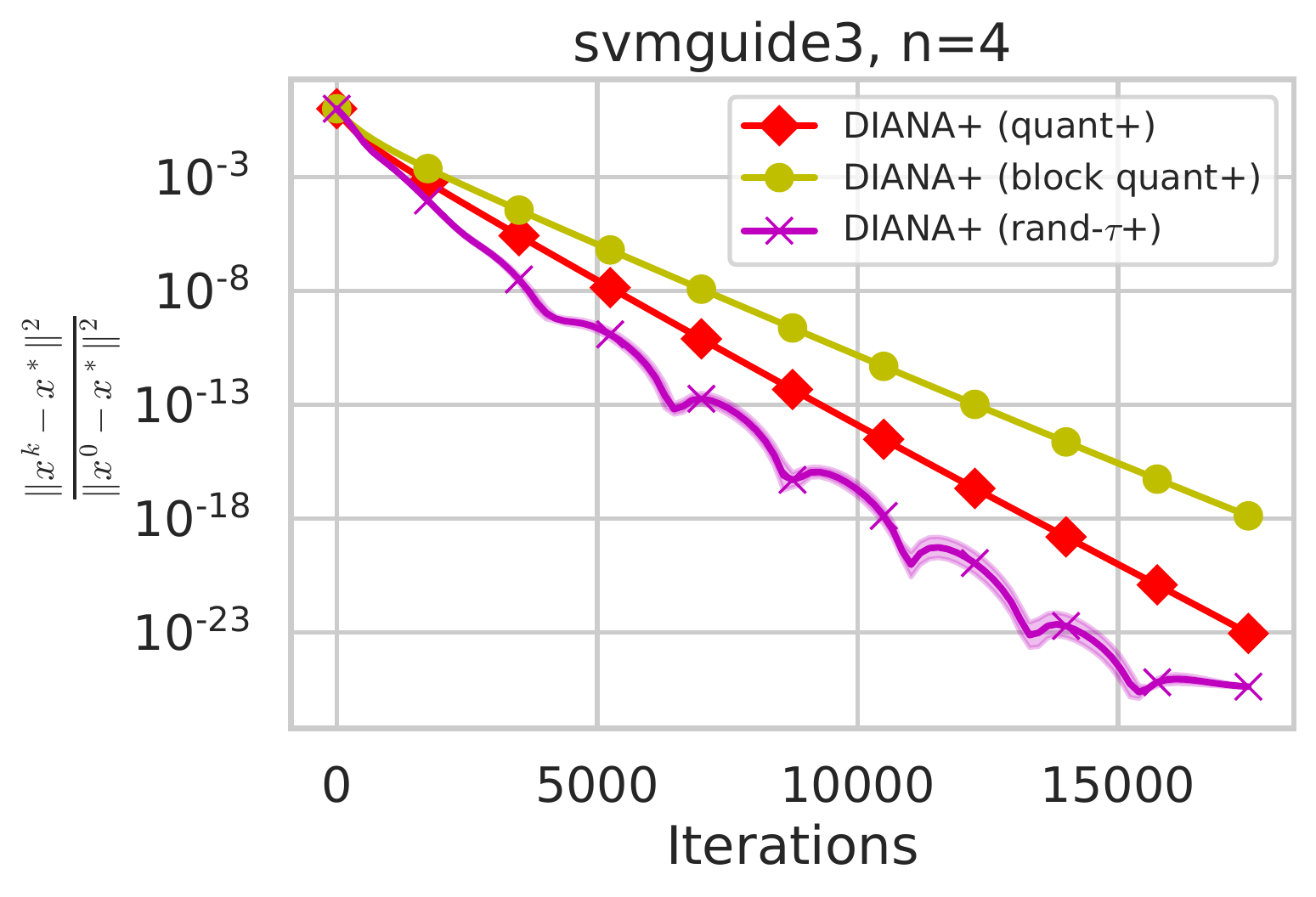}
    \endminipage\hfill  
    \minipage{0.30\textwidth}
    \includegraphics[width=\linewidth]{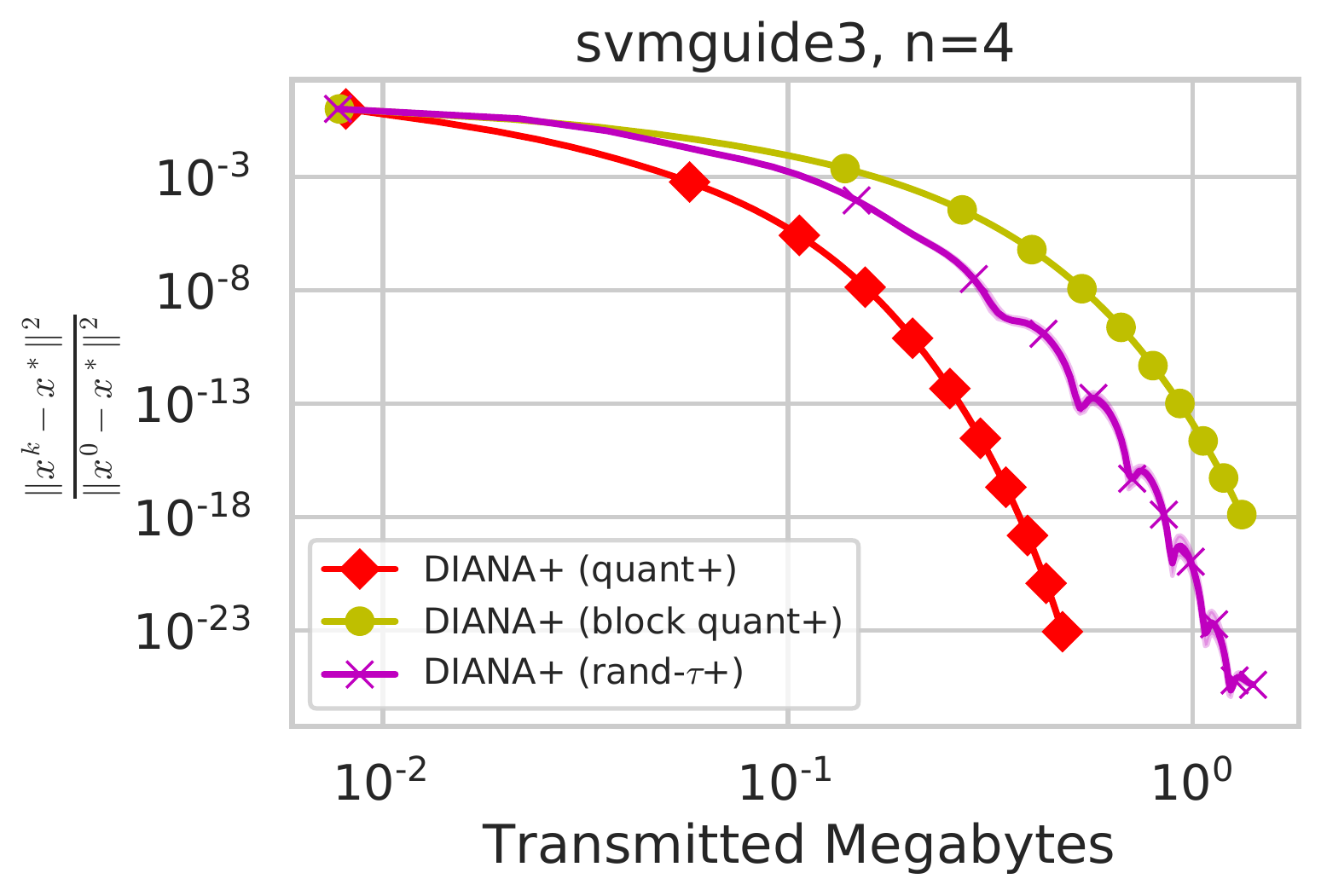}
    \endminipage\hfill  
    \minipage{0.30\textwidth}
    \includegraphics[width=\linewidth]{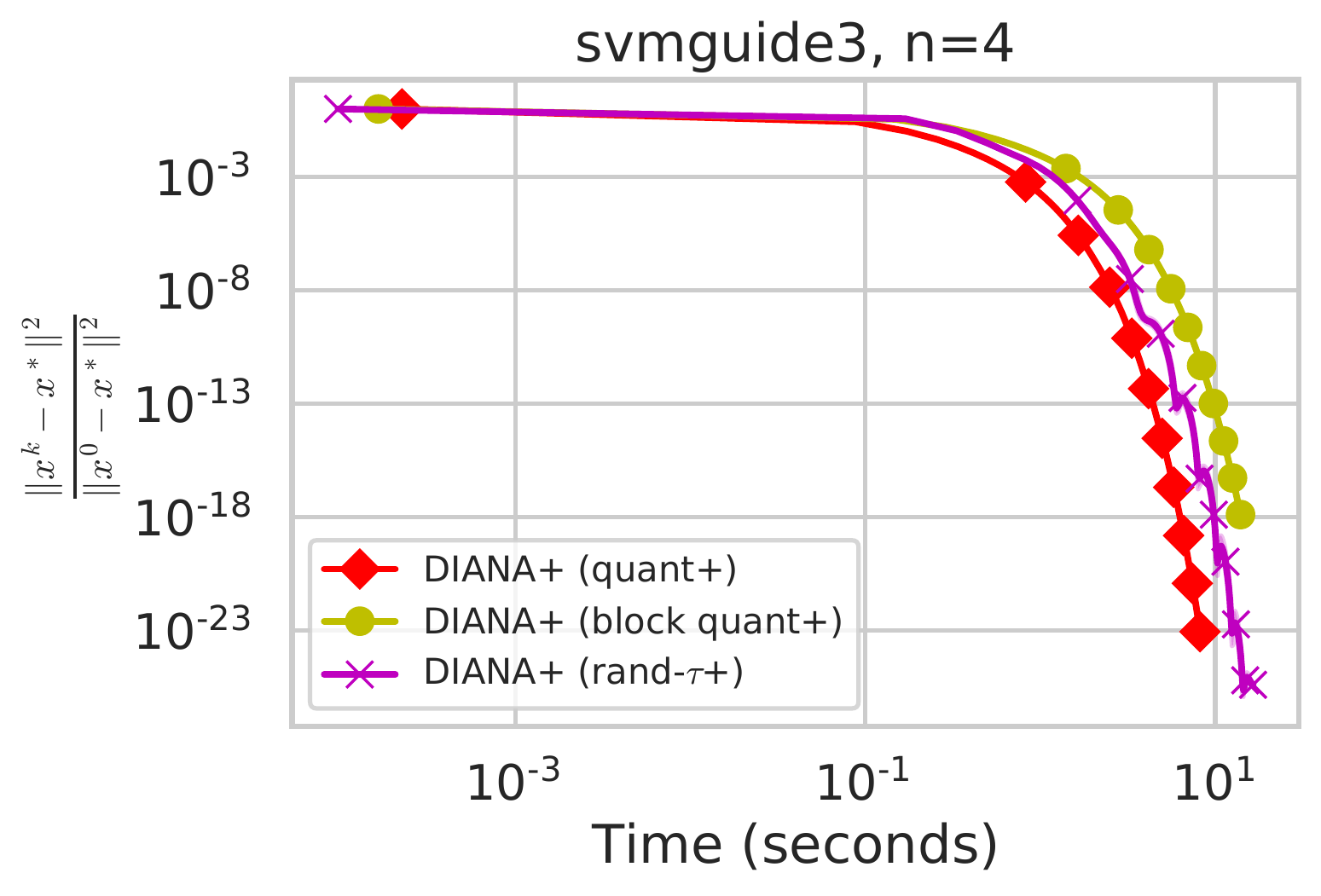}
    \endminipage\hfill  
    
    \minipage{0.30\textwidth}
    \includegraphics[width=\linewidth]{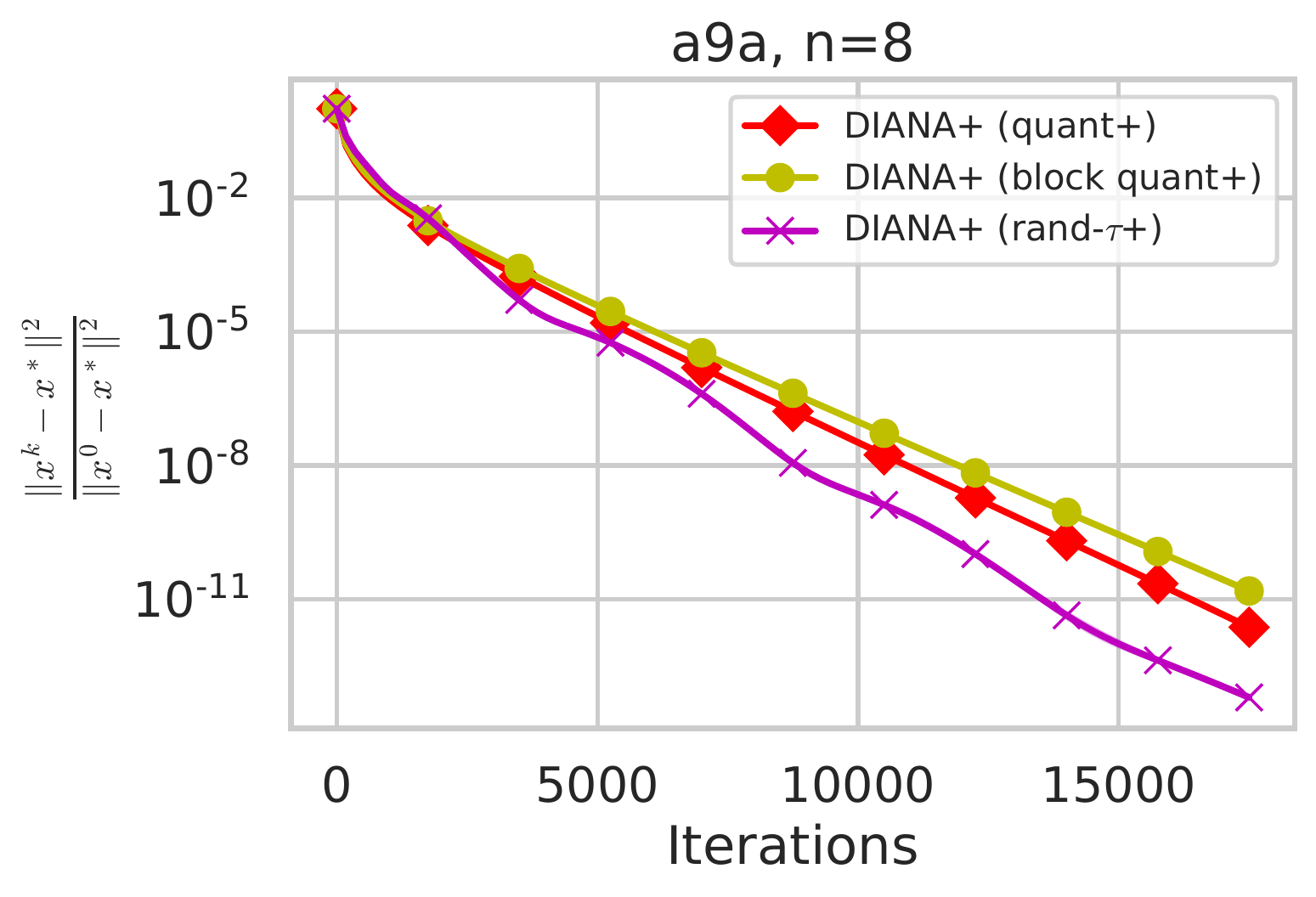}
    \endminipage\hfill  
    \minipage{0.30\textwidth}
    \includegraphics[width=\linewidth]{./figures/a9a-scvx-8-inf/SD-DIANA-plus-BL-DIANA-plus-Sparse-DIANA-plus_dist_trace_mbs.pdf}
    \endminipage\hfill  
    \minipage{0.30\textwidth}
\includegraphics[width=\linewidth]{./figures/a9a-scvx-8-inf/SD-DIANA-plus-BL-DIANA-plus-Sparse-DIANA-plus_dist_trace_time.pdf}
\endminipage\hfill 
    
    \minipage{0.30\textwidth}
    \includegraphics[width=\linewidth]{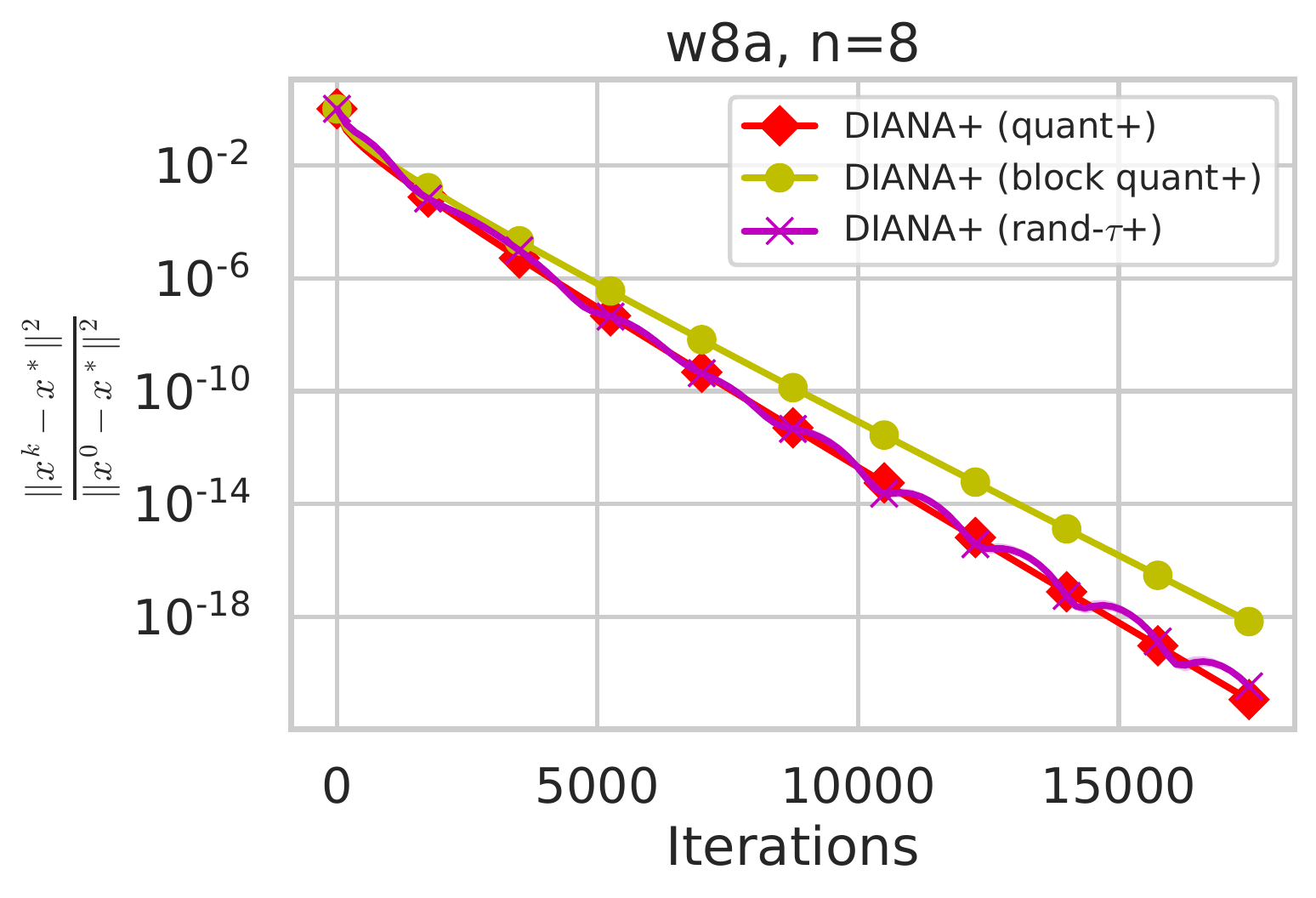}
    \endminipage\hfill  
    \minipage{0.30\textwidth}
    \includegraphics[width=\linewidth]{./figures/w8a-scvx-8-inf/SD-DIANA-plus-BL-DIANA-plus-Sparse-DIANA-plus_dist_trace_mbs.pdf}
    \endminipage\hfill  
        \minipage{0.30\textwidth}
    \includegraphics[width=\linewidth]{./figures/w8a-scvx-8-inf/SD-DIANA-plus-BL-DIANA-plus-Sparse-DIANA-plus_dist_trace_time.pdf}
    \endminipage\hfill  
    \caption{Comparison of smoothness-aware DCGD+/DIANA+ methods with varying-step quantization (\texttt{quant+}) to original DCGD/DIANA methods with standard quantization (\texttt{quant}). Note that in \texttt{quant+} workers need to send $\mL_i^{1/2}\in\R^{d\times d}$ and quantization steps $h_i\in\R^{d}$ to the master before the training. This leads to extra costs in communication bits and time, which are taken into consideration. }
    \label{fig:sp_vs_quant}
\end{figure*}

Moreover, we also compare the performance of three smoothness-aware compression techniques ---block quantization (\texttt{block quant+}) of Section~\ref{sec:block-quant}, varying-step quantization (\texttt{quant+}) of Section~\ref{sec:quant} and smoothness-aware sparsification strategy (\texttt{rand-$\tau$+}) of \cite{safaryan2021smoothness}. All three compression techniques are shown to outperform the standard compression strategies by at most $\cO(n)$ times in theory. For the sparsification, we use the optimal probabilities and the sampling size $\tau = d/n$ as suggested in Section 5.3 of \citep{safaryan2021smoothness}. The empirical results in Figure~\ref{fig:sp_vs_quant} illustrate that the varying-step quantization technique (\texttt{quant+}) is always better than the smoothness-aware sparsification \citep{safaryan2021smoothness}, in terms of both communication cost and wall-clock time. Our block quantization technique also beats sparsification when the dimension of the model is relatively~high.

\subsection{Numerical verification of Assumption~\ref{asm:quant-bits}}

\begin{figure}    
    \centering
    \includegraphics[width=0.5\linewidth]{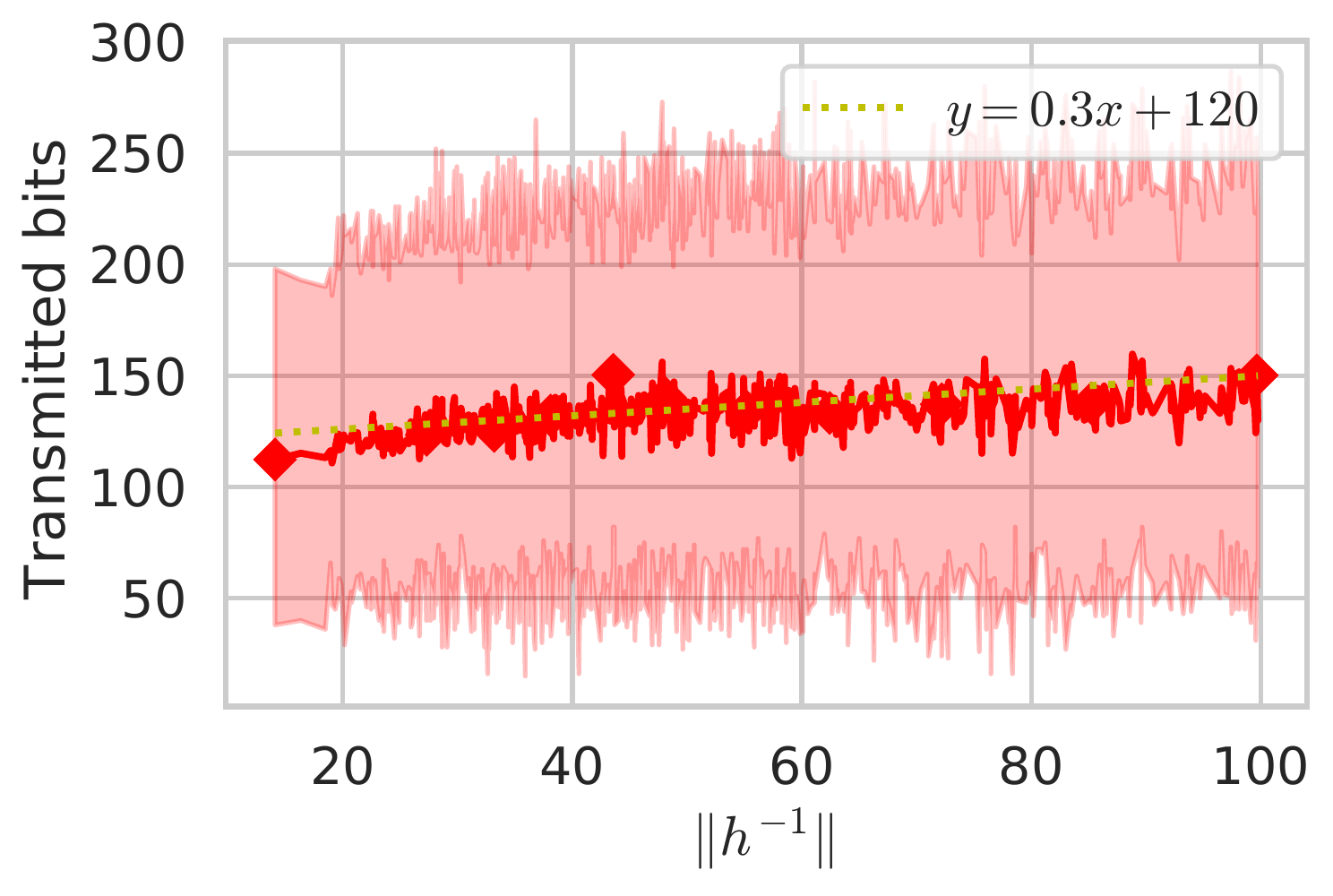}
\caption{Experiment to verify the Assumption~\ref{asm:quant-bits}.  We randomly generate 1000 quantization step vectors $h\in\mathbb{R}^{50}$, each component of $h$ is $h_j = |\tilde{h}_j|$ and $\tilde{h}_j$ is independently sampled from $\mathcal{N}(0,1)$. For each $h$, we randomly generate multiple sparse vectors to quantize $x$, which is sampled from Poisson distribution with $\lambda = \{1,10,100\}$ and density $\{0.25, 0.5, 0.75, 1.0\}$. }
\label{fig:assum_verification}
\end{figure}

We provide a numerical experiment to verify Assumption~\ref{asm:quant-bits} that $\Norm{h^{-1}}$ and the communicated bits are positively correlated. Figure~\ref{fig:assum_verification} shows that the communicated bits and $\Norm{h^{-1}}$ are indeed positively correlated.

\begin{figure*}[htp]
    \minipage{0.3\textwidth}
    \includegraphics[width=\linewidth]{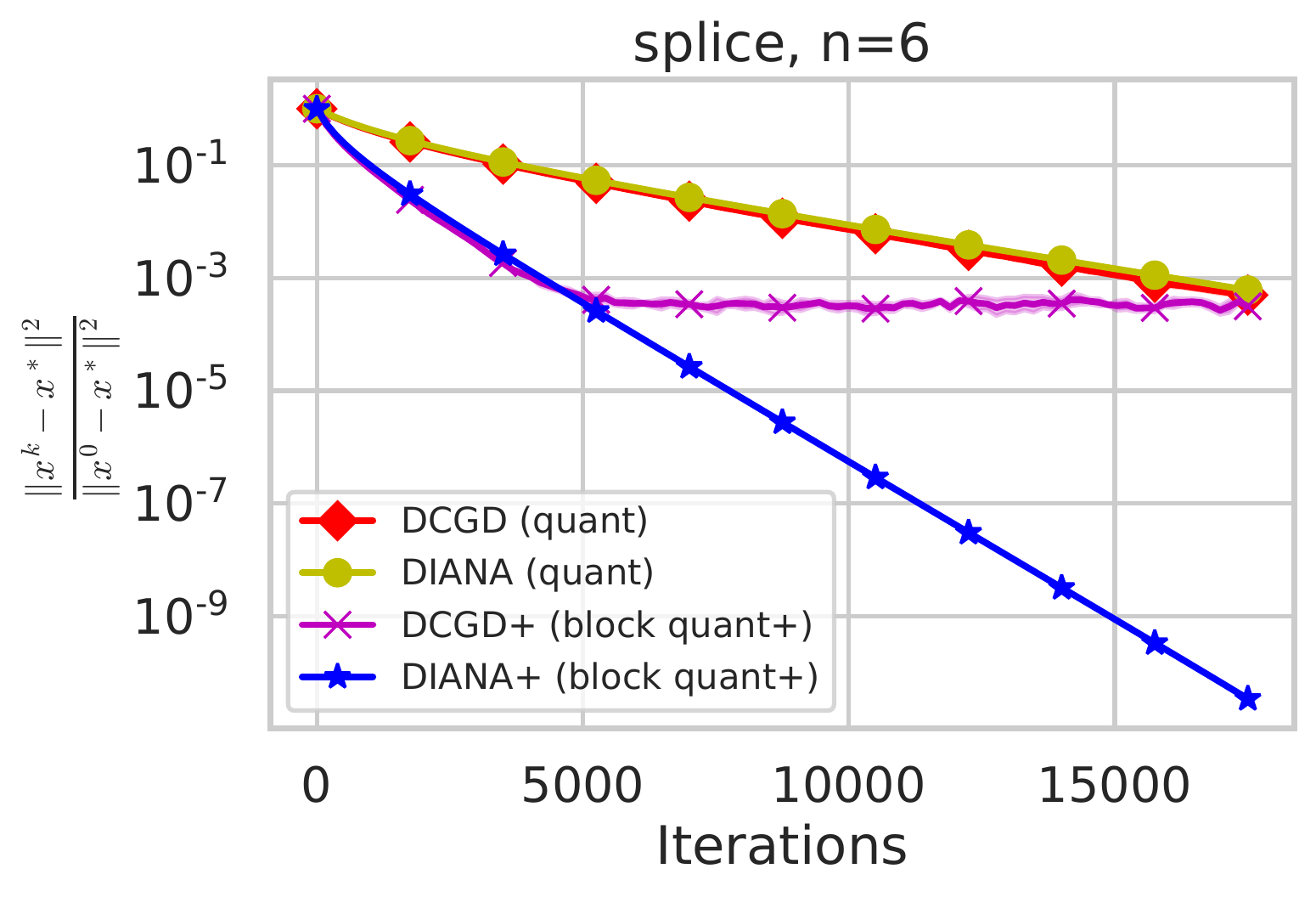}
        \endminipage\hfill  
    \minipage{0.3\textwidth}
    \includegraphics[width=\linewidth]{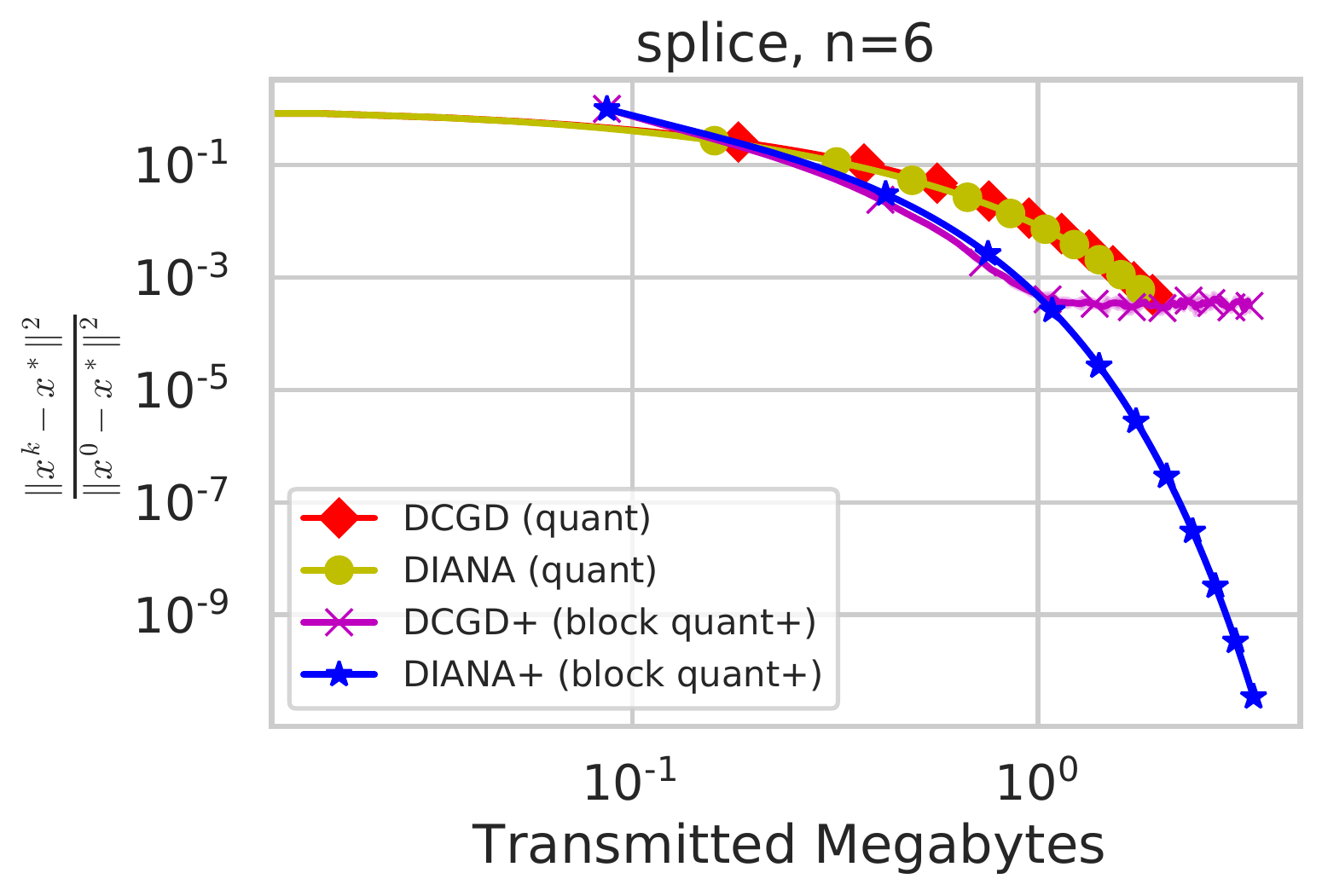}
    \endminipage\hfill
    \minipage{0.3\textwidth}
    \includegraphics[width=\linewidth]{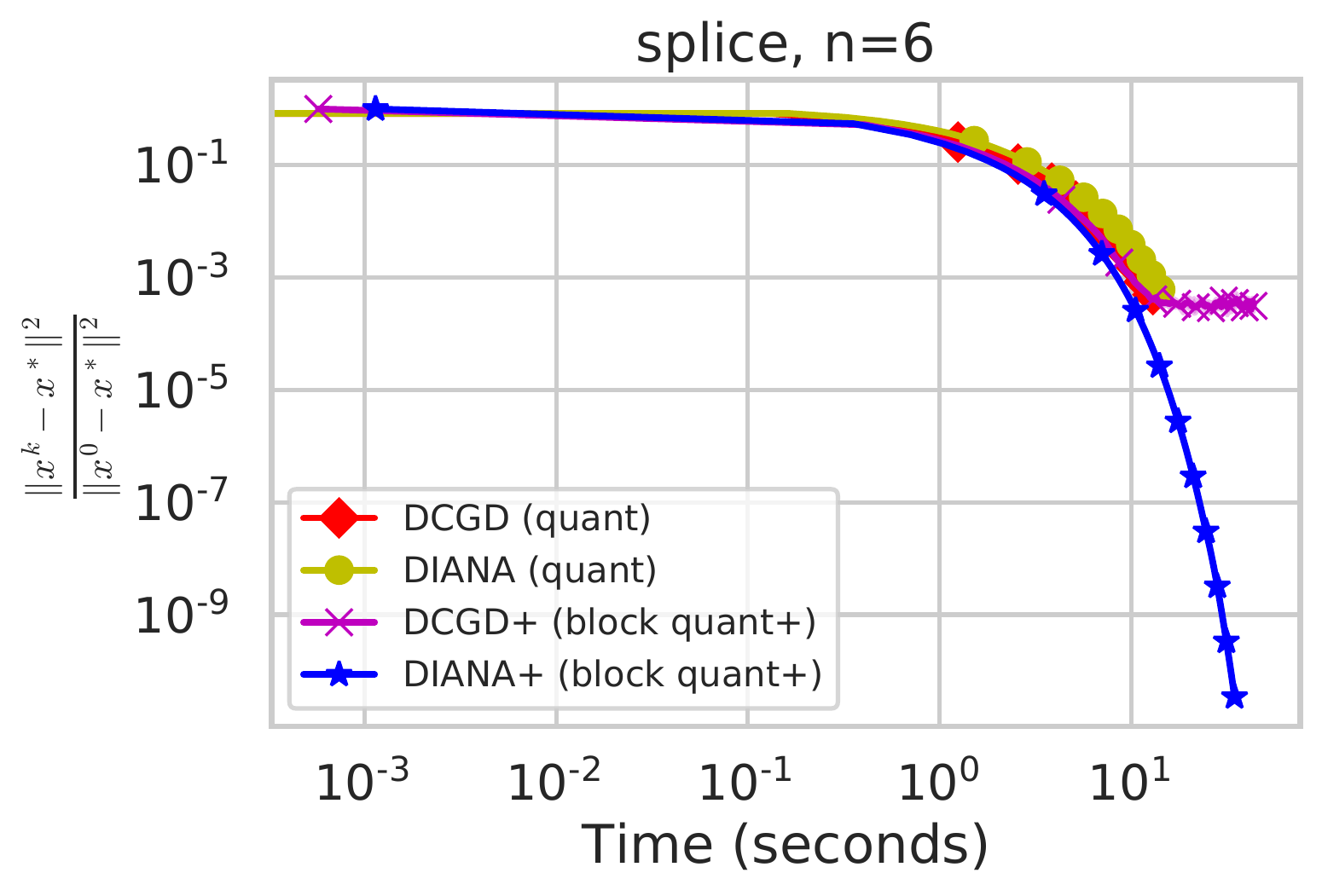}
    \endminipage\hfill

    \minipage{0.3\textwidth}
    \includegraphics[width=\linewidth]{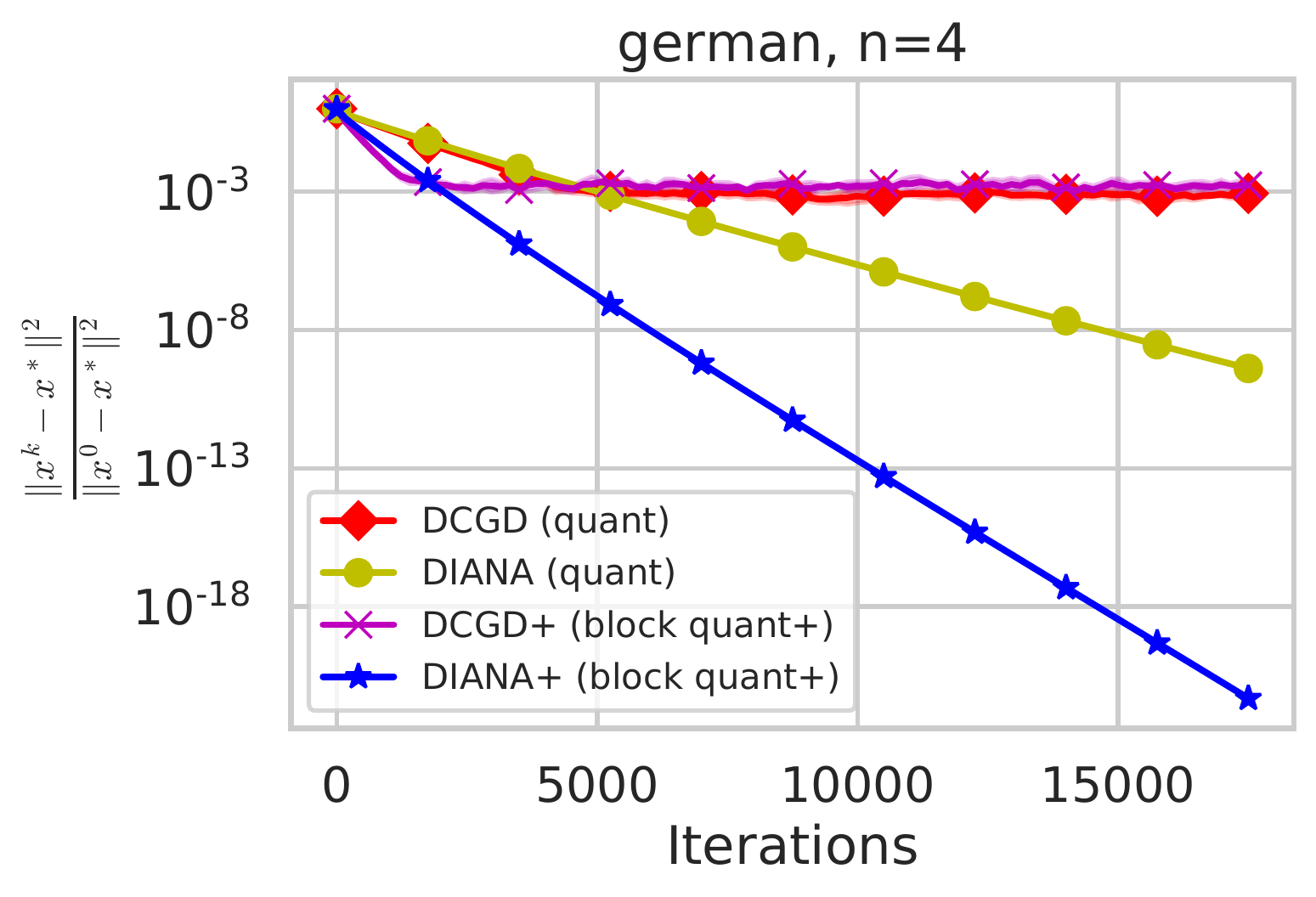}
        \endminipage\hfill  
    \minipage{0.3\textwidth}
    \includegraphics[width=\linewidth]{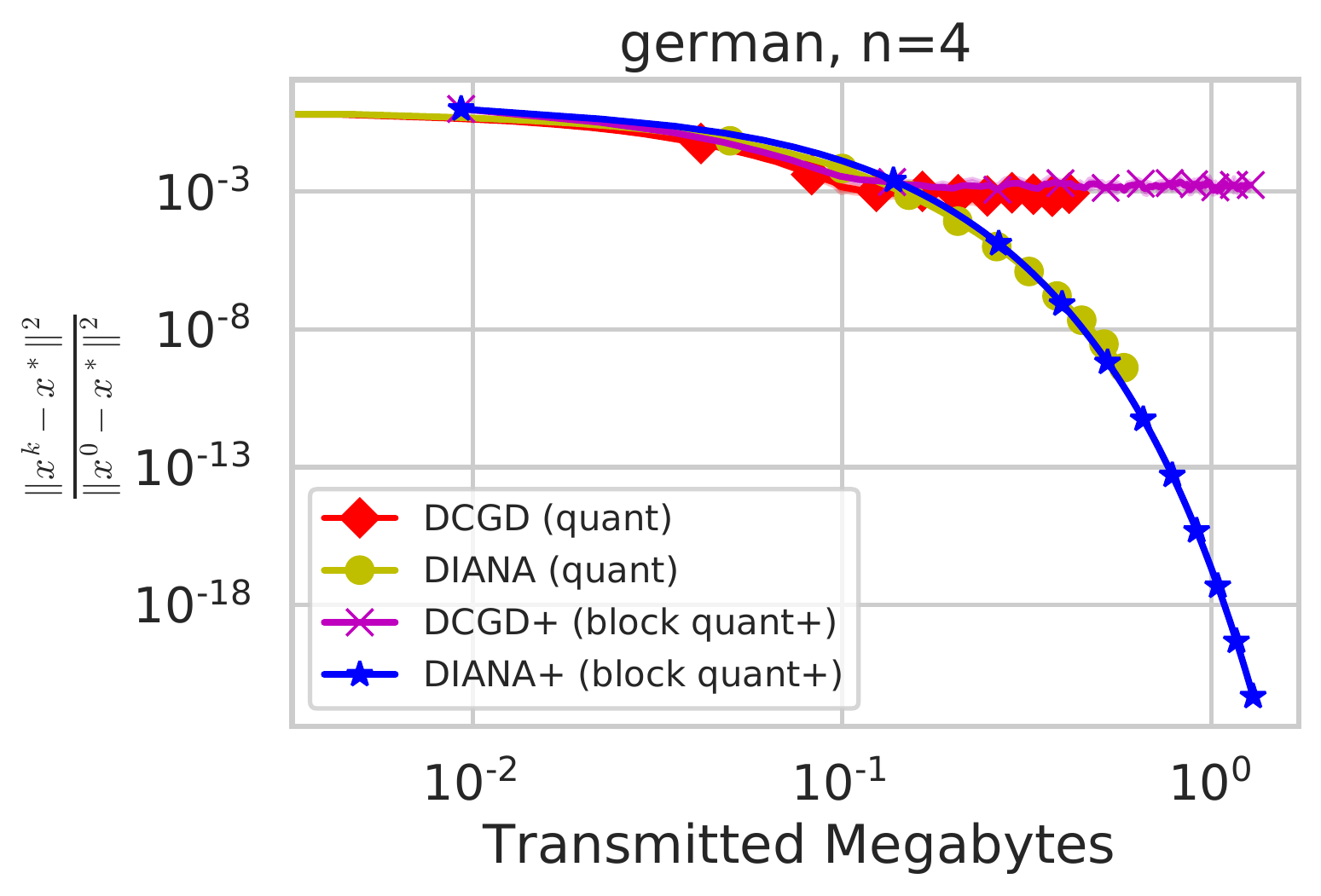}
    \endminipage\hfill
        \minipage{0.3\textwidth}
    \includegraphics[width=\linewidth]{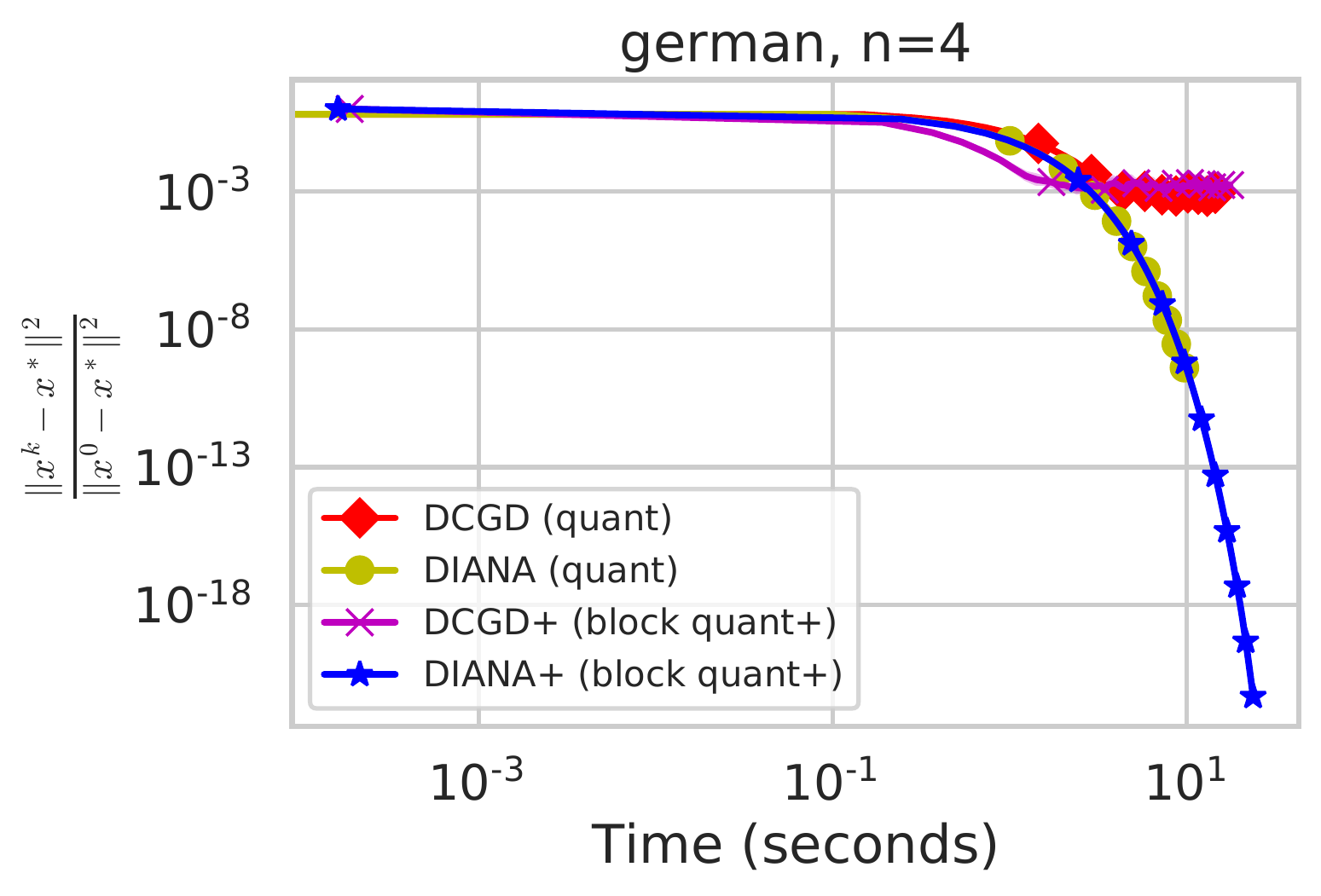}
    \endminipage\hfill

    \minipage{0.3\textwidth}
    \includegraphics[width=\linewidth]{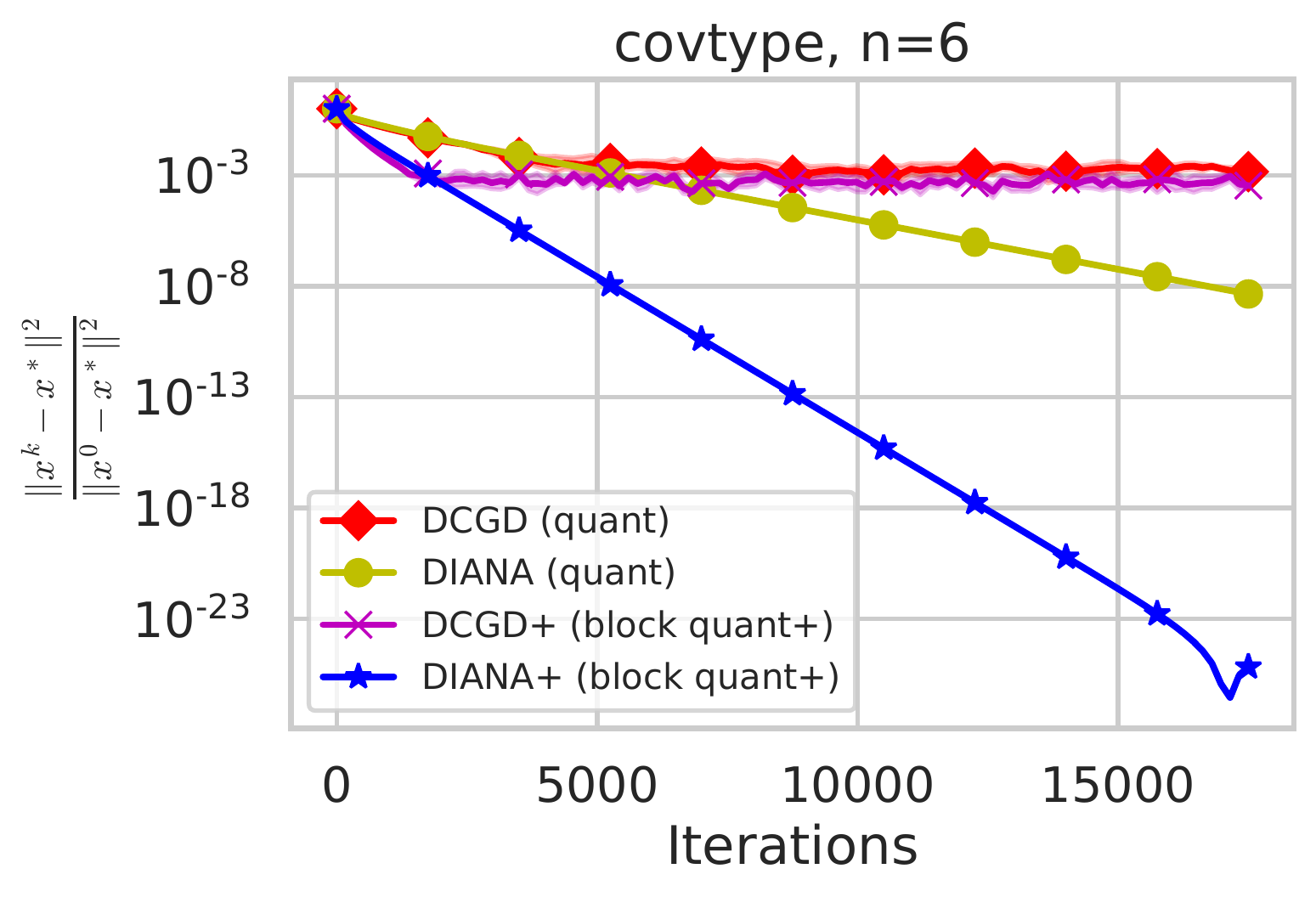}
    \endminipage\hfill  
    \minipage{0.3\textwidth}
    \includegraphics[width=\linewidth]{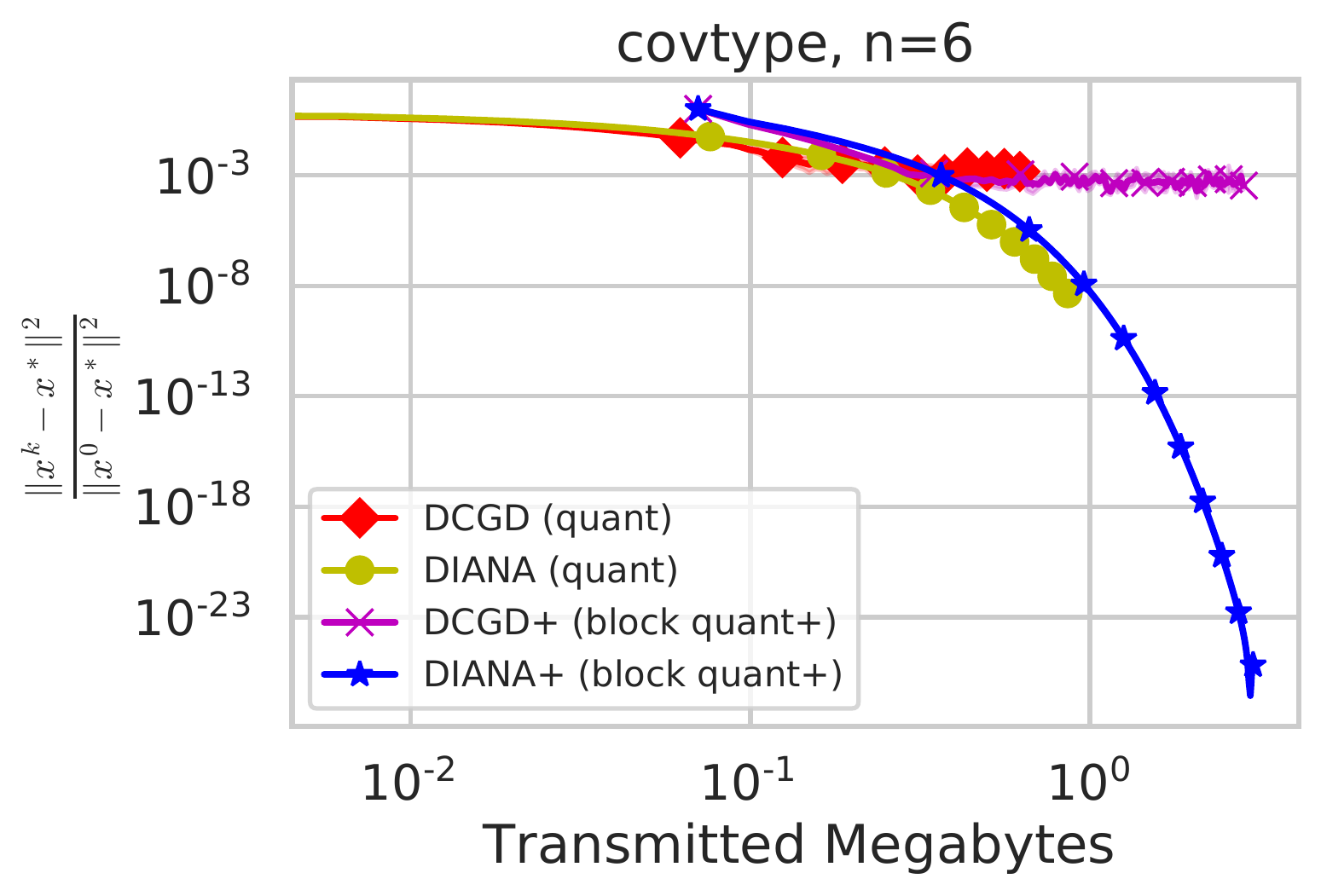}
    \endminipage\hfill  
    \minipage{0.3\textwidth}
    \includegraphics[width=\linewidth]{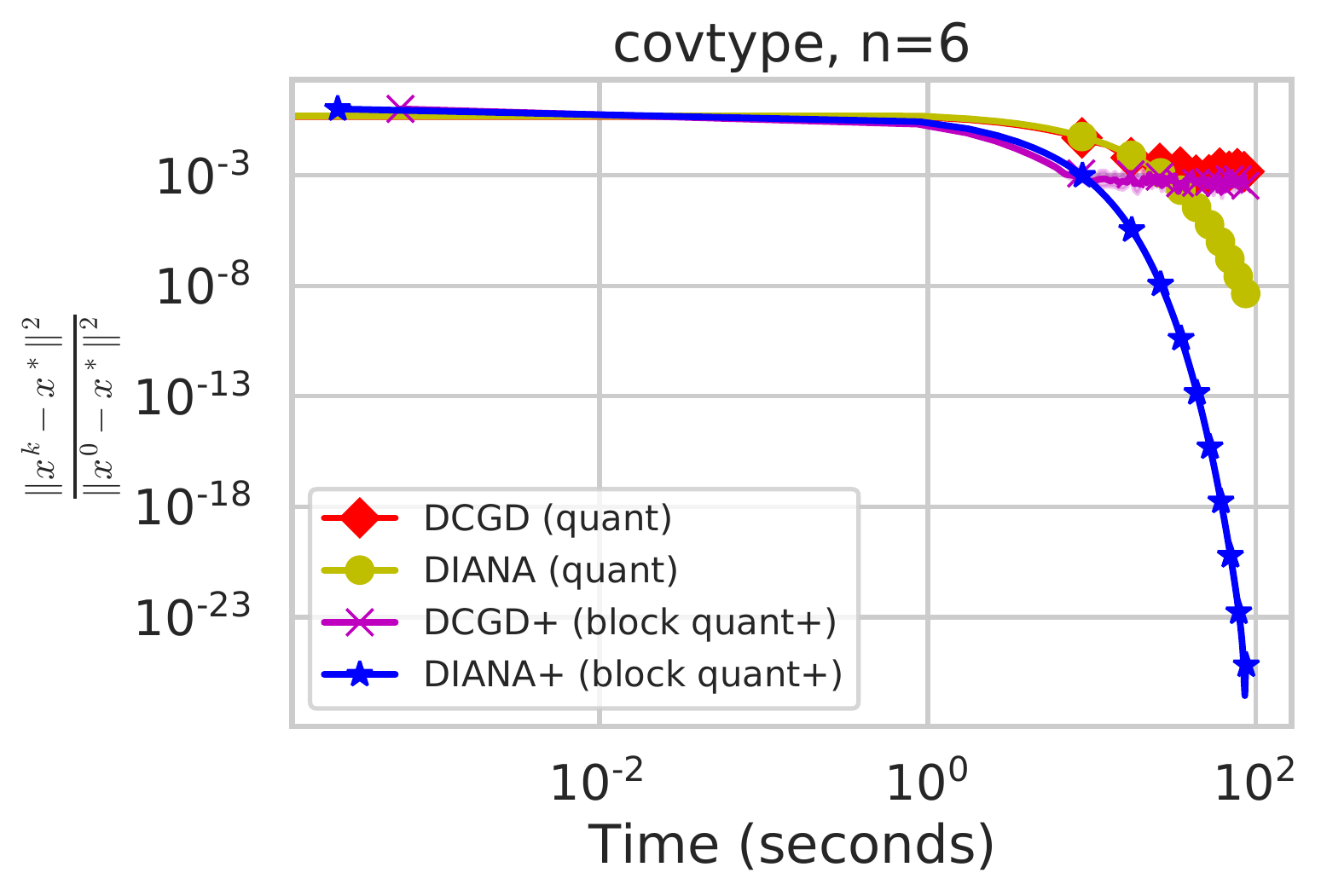}
    \endminipage\hfill  
    
    \minipage{0.3\textwidth}
    \includegraphics[width=\linewidth]{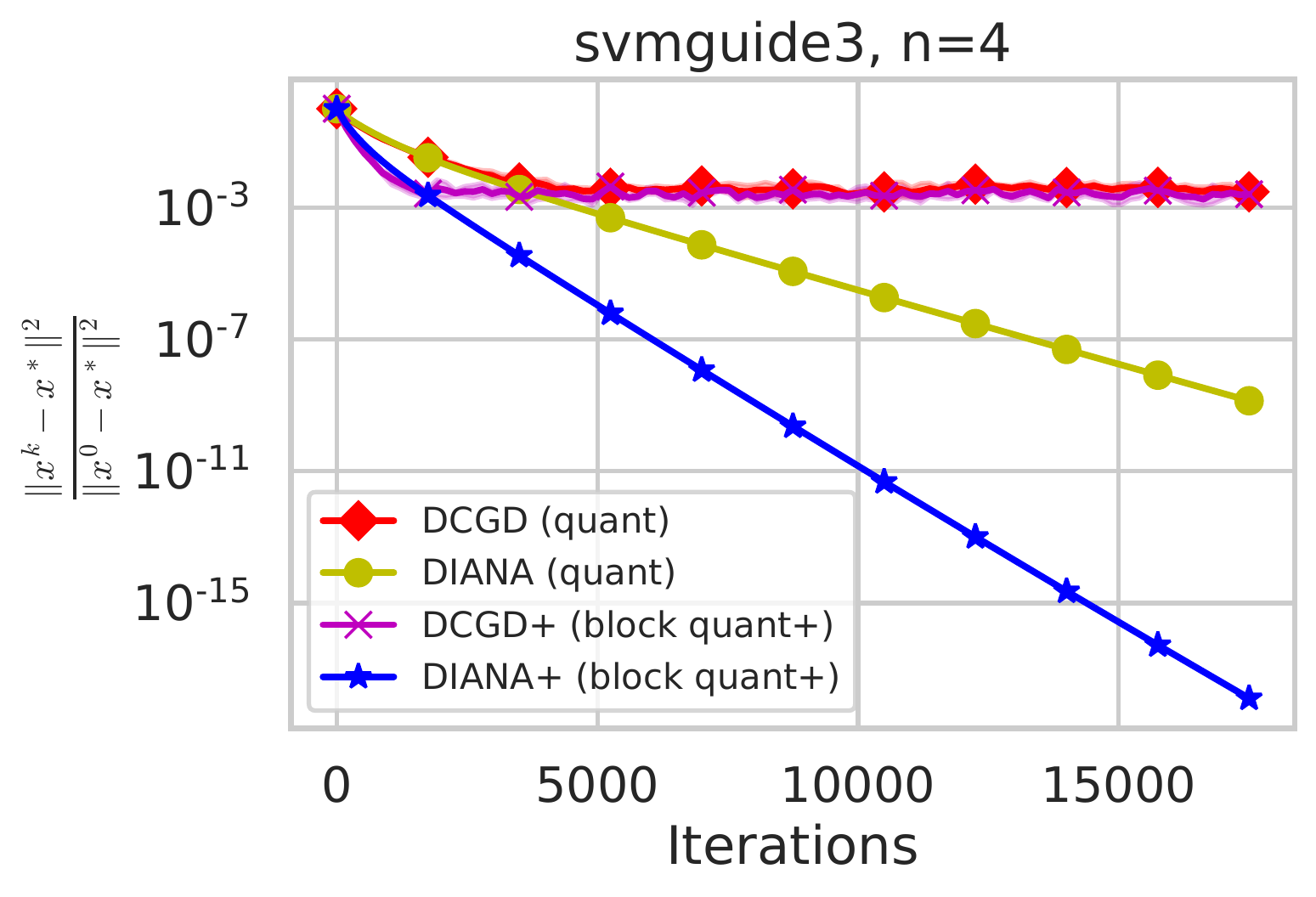}
    \endminipage\hfill  
    \minipage{0.3\textwidth}
    \includegraphics[width=\linewidth]{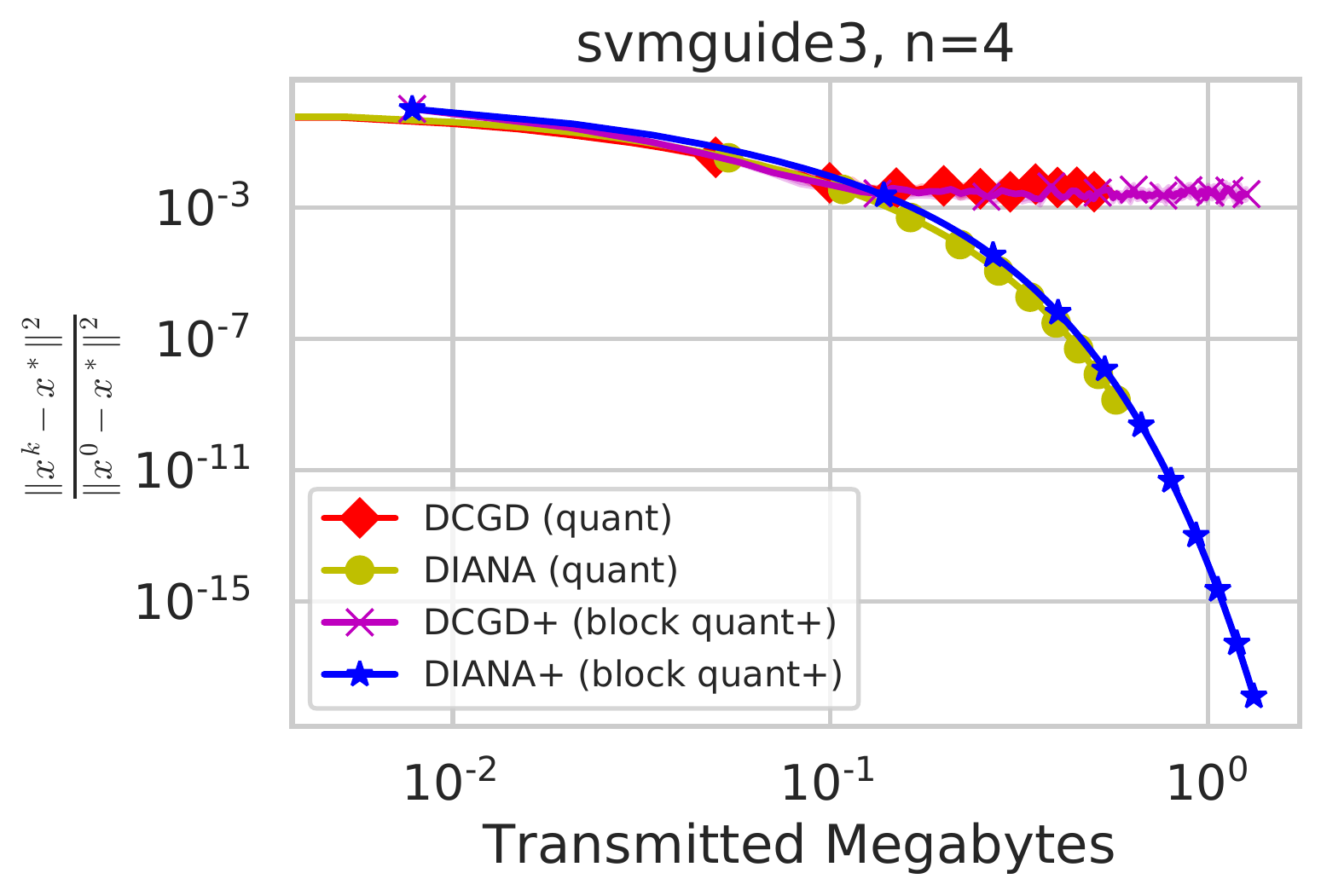}
    \endminipage\hfill
    \minipage{0.3\textwidth}
    \includegraphics[width=\linewidth]{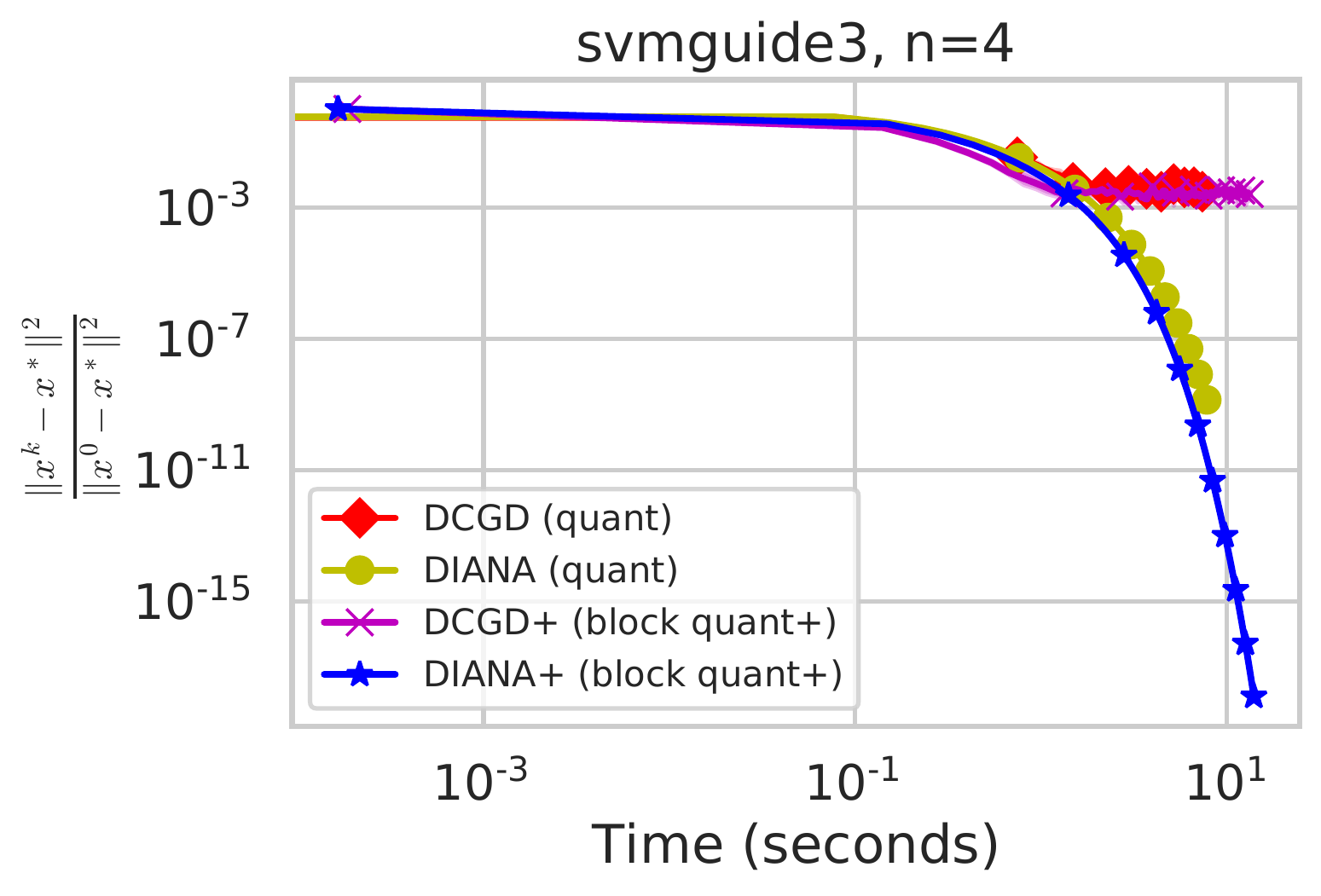}
    \endminipage\hfill

    \minipage{0.3\textwidth}
    \includegraphics[width=\linewidth]{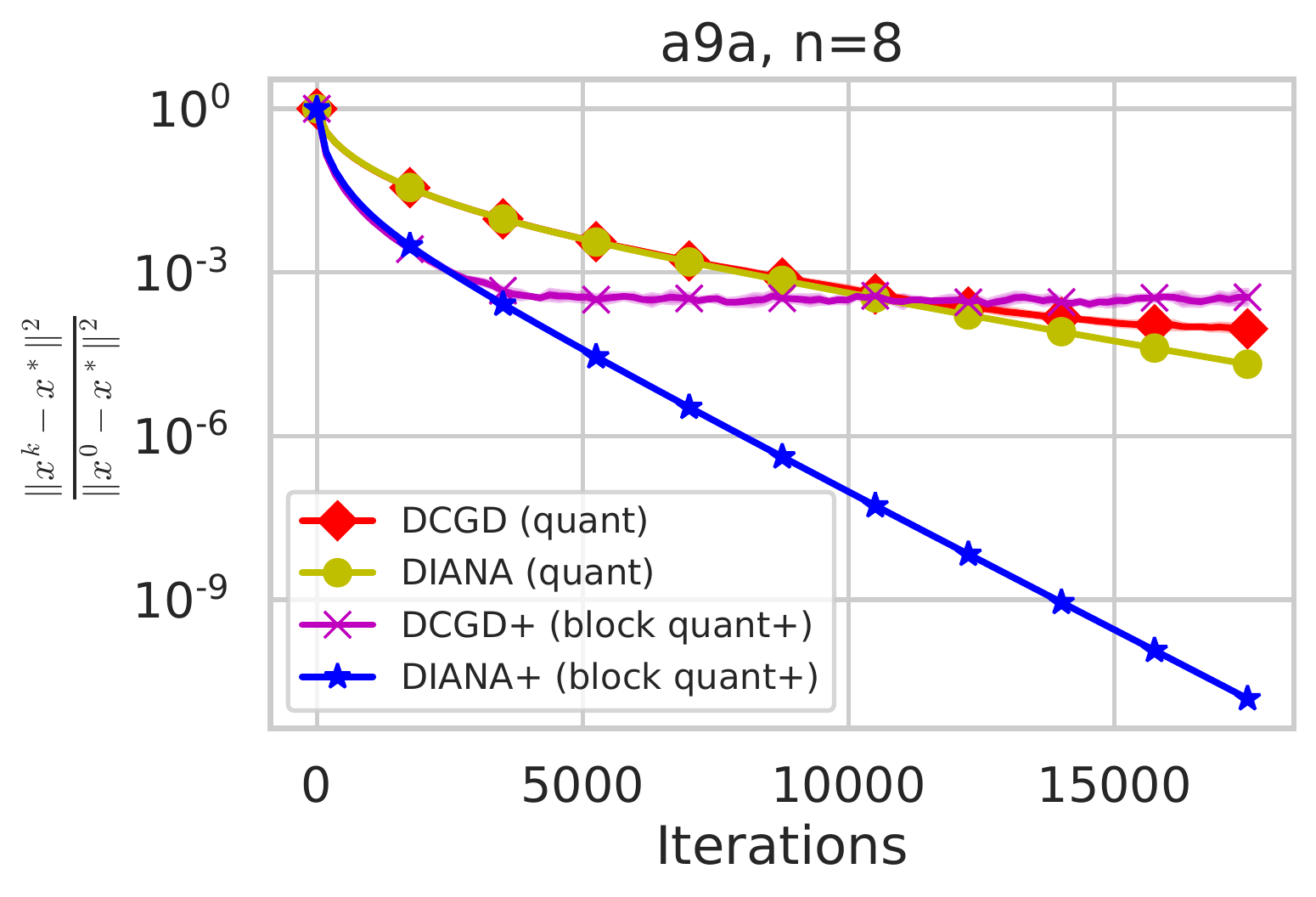}
    \endminipage\hfill  
    \minipage{0.3\textwidth}
    \includegraphics[width=\linewidth]{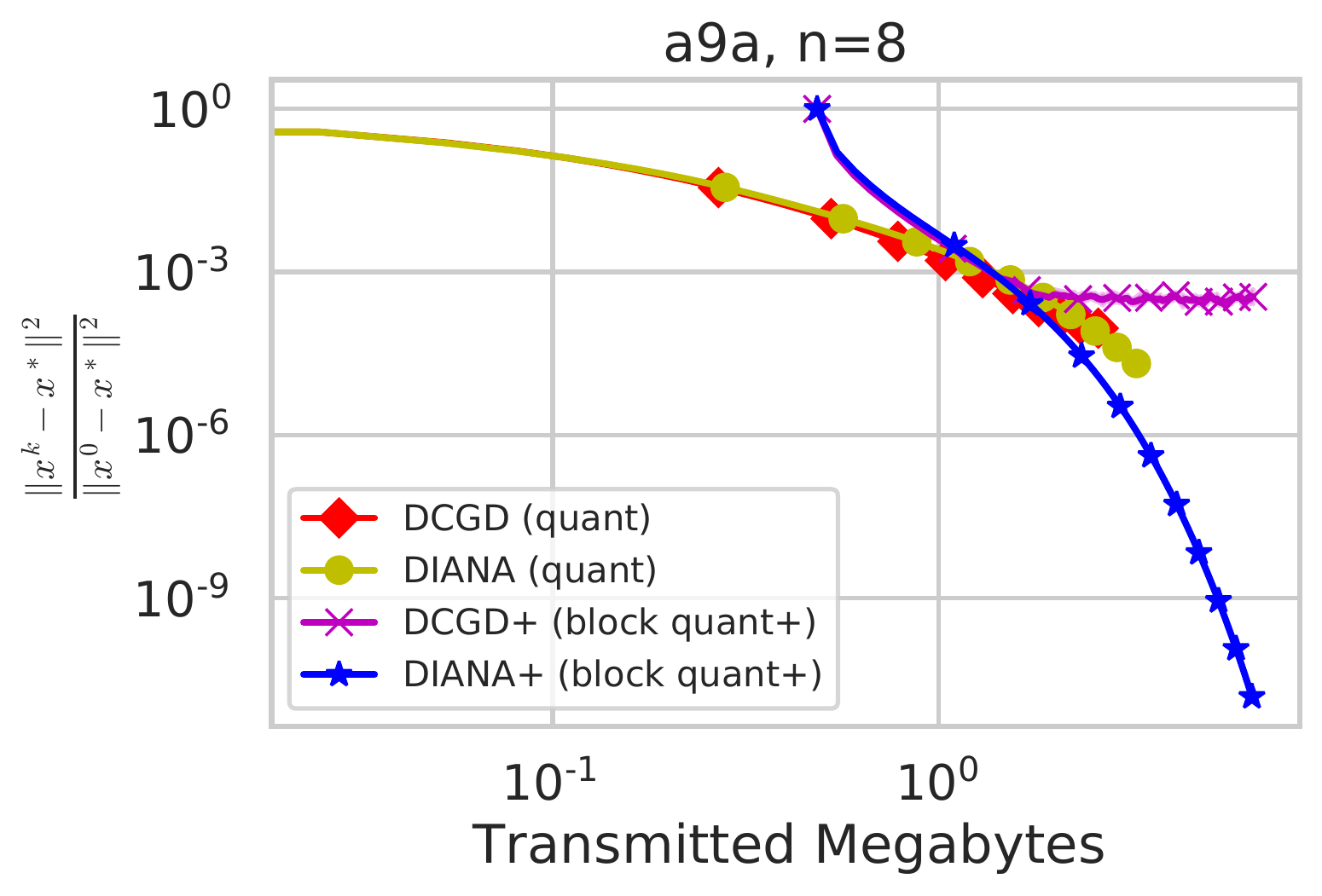}
    \endminipage\hfill
    \minipage{0.3\textwidth}
    \includegraphics[width=\linewidth]{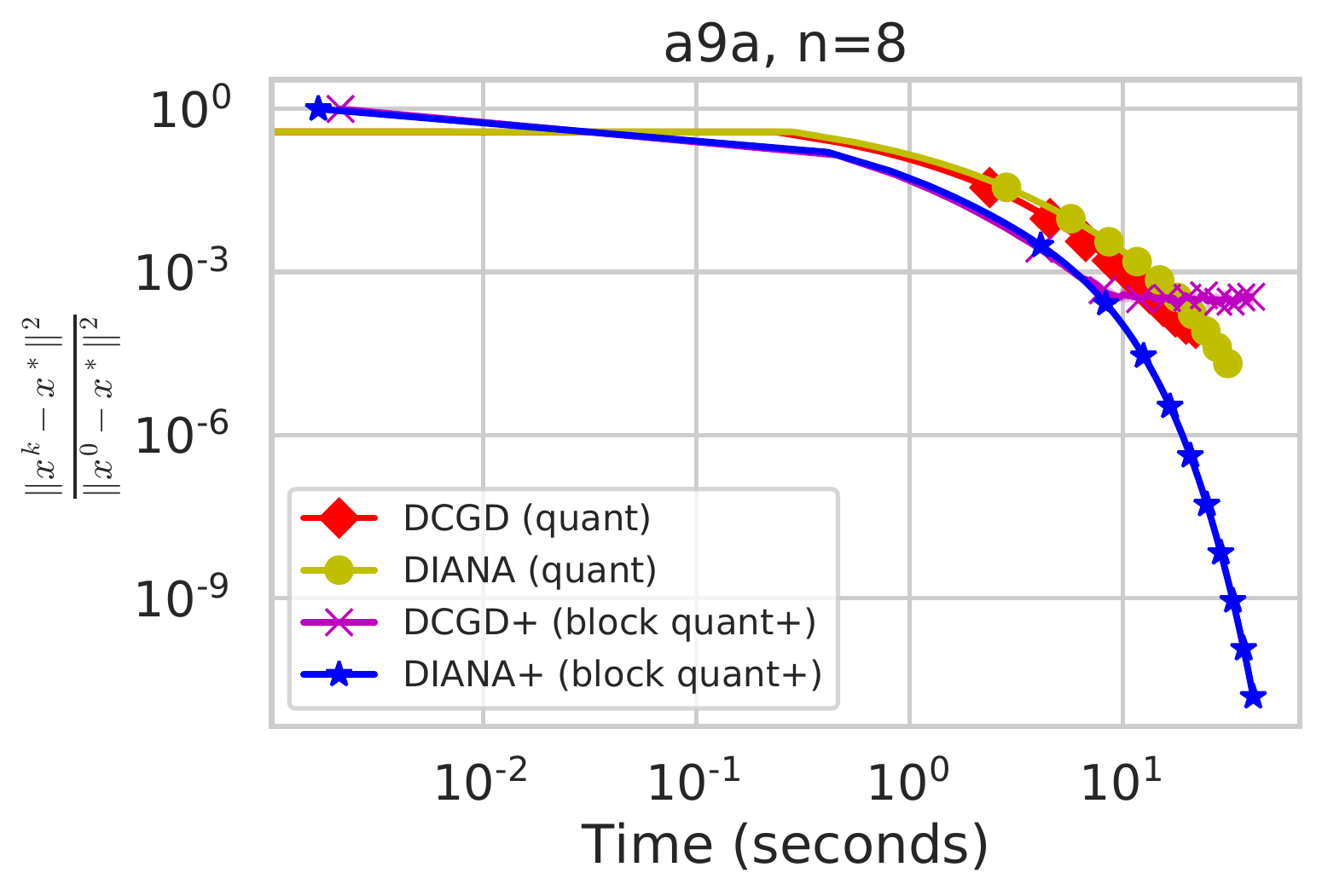}
    \endminipage\hfill  
    
    \minipage{0.3\textwidth}
    \includegraphics[width=\linewidth]{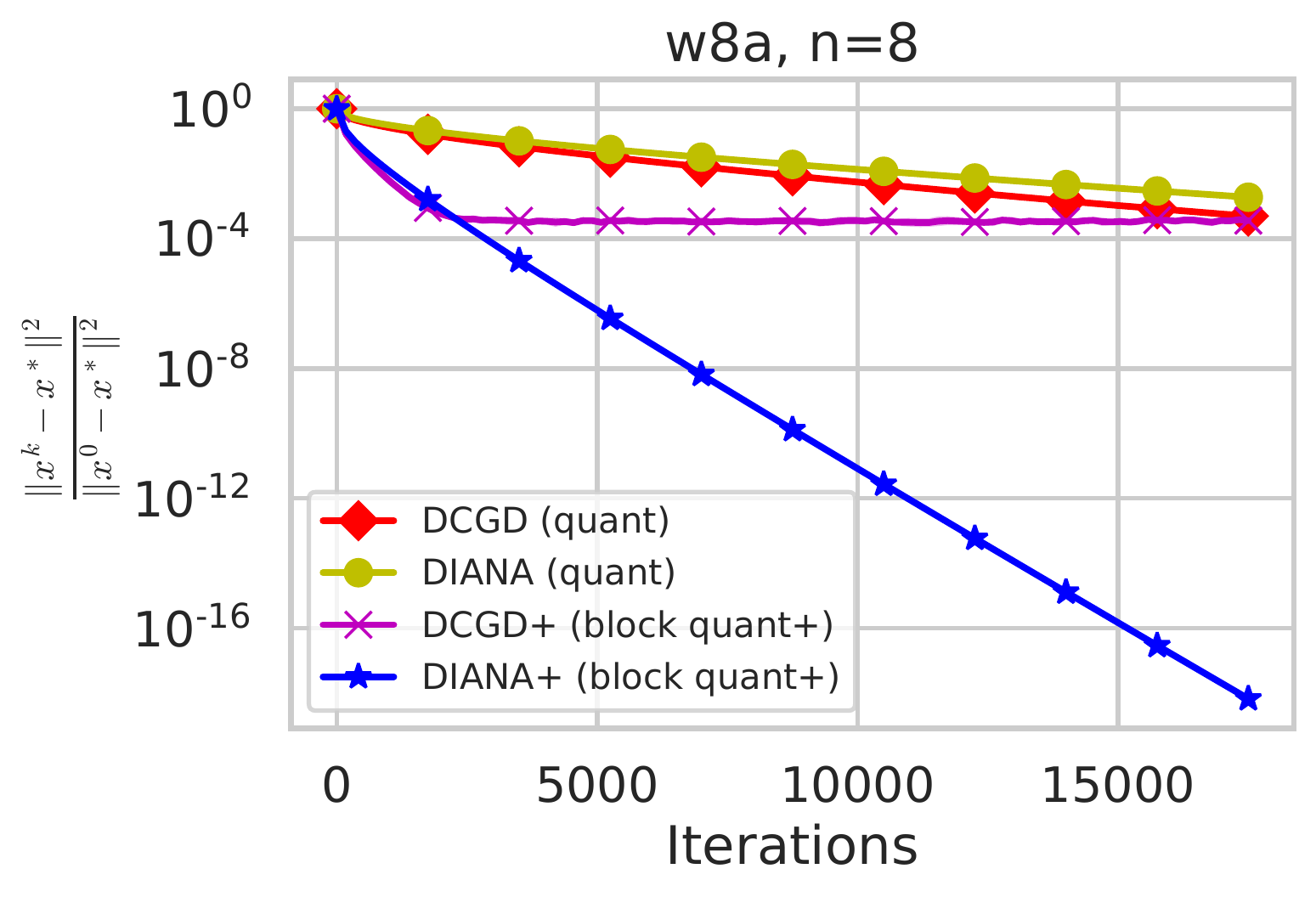}
    \endminipage\hfill  
    \minipage{0.3\textwidth}
    \includegraphics[width=\linewidth]{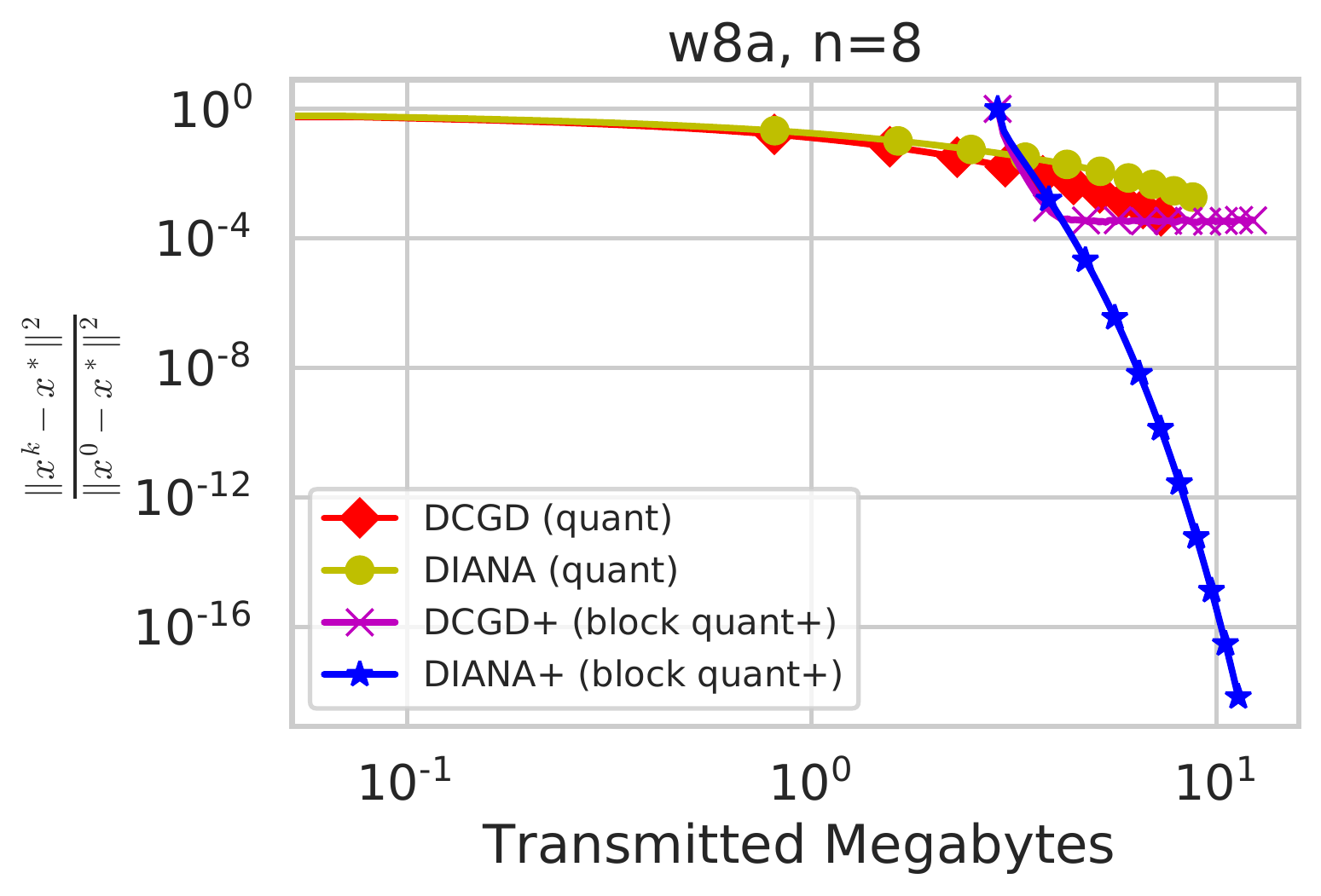}
    \endminipage\hfill
    \minipage{0.3\textwidth}
    \includegraphics[width=\linewidth]{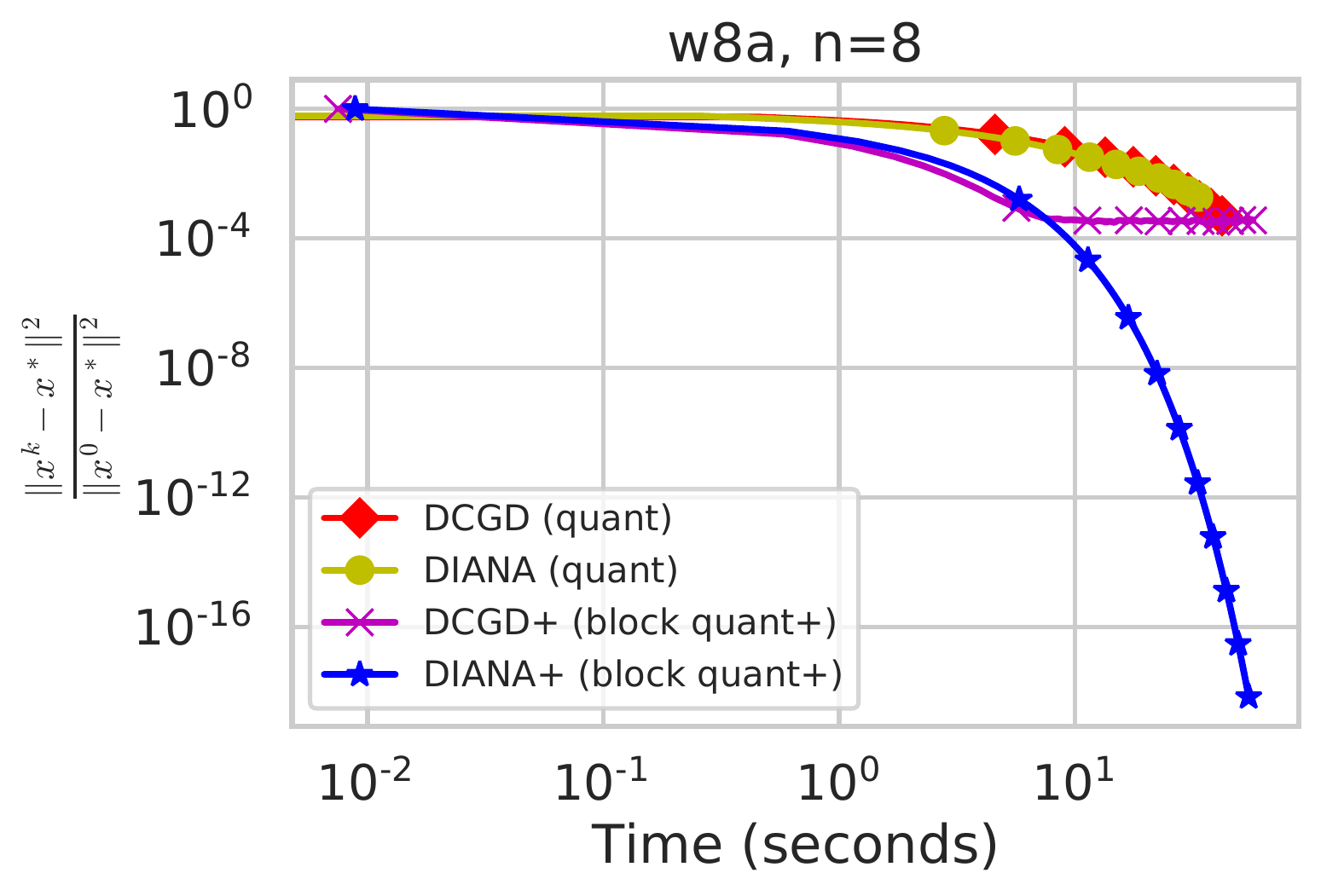}
    \endminipage\hfill  
    \caption{Comparison of DCGD+ (\texttt{block quant+}) and DIANA+ (\texttt{block quant+}) with DCGD (\texttt{quant}) and DIANA (\texttt{quant}). }
    \label{fig:block}
\end{figure*}

\clearpage
\section{Proofs for Section \ref{sec:ext-arb-comp}: Smoothness-Aware Distributed Methods with General Compressors}

Here we provide the proofs of Theorem \ref{thm:dcgd+_arbitrary} and Theorem \ref{thm:diana+_arbitrary}. Both proofs follow similar steps done for sparsification in \citep{safaryan2021smoothness}.

\subsection{Proof of Theorem \ref{thm:dcgd+_arbitrary}: DCGD+ with arbitrary unbiased compression}\label{apx:dcgd+_arbitrary}

To simplify the notation, let us skip the iteration count $k$ in the derivations. We are going to estimate the quantity $\E\[\|g(x)-\nabla f(x^*)\|^2\]$ and establish the following bound for the gradient estimator $g(x) = \frac{1}{n}\sum_{i=1}^n \ML_i^\half\fC_i(\ML_i^\phalf \nabla f_i(x))$:
\begin{equation*}
\E\[\|g(x) - \nabla f(x^*)\|^2\] \le 2\(L+\frac{2\cL_{\max}}{n}\) D_{f}(x,x^*) + \frac{2\sigma_+^*}{n},
\end{equation*}
where $D_{f}(x,x^*)$ is the Bregman divergence between $x$ and $x^*$ with respect to $f$. Due to Lemma E.3 \citep{GJS-HR}, we have $\nabla f_i(x) = \ML_i^{\nicefrac{1}{2}}r_i$ for some $r_i$. Therefore, 
\begin{equation}\label{unbiased-est}
\E\[ \ML_i^{\nicefrac{1}{2}}\fC_i(\ML_i^{\dagger\nicefrac{1}{2}} \ML_i^{\nicefrac{1}{2}}r_i ) \]
= \ML_i^{\nicefrac{1}{2}} \E\[ \fC_i(\ML_i^{\dagger\nicefrac{1}{2}} \ML_i^{\nicefrac{1}{2}}r_i ) \]
= \ML_i^{\nicefrac{1}{2}}\ML_i^{\dagger\nicefrac{1}{2}} \ML_i^{\nicefrac{1}{2}}r_i
= \ML_i^{\nicefrac{1}{2}}r_i
= \nabla f_i(x),
\end{equation}
which implies unbiasedness of the estimator $g(x)$, namely $\E\[g(x)\] = \nabla f(x)$.  Note that:
\begin{eqnarray}
  \E\[\|g(x)-\nabla f(x)\|^2\]
  &=&
  \E\[ \left\|\frac{1}{n}\sum_{i=1}^n \ML_i^\half\fC_i(\ML_i^\phalf \nabla f_i(x)) - \nabla f_i(x) \right\|^2 \] \notag \\
  &\stackrel{\clubsuit}{=}&
  \frac{1}{n^2}\sum_{i=1}^n  \E\[ \left\|\ML_i^\half\fC_i(\ML_i^\phalf \nabla f_i(x)) - \ML_i^\half \ML_i^\phalf \nabla f_i(x) \right\|^2 \] \notag \\
  &=&
  \frac{1}{n^2}\sum_{i=1}^n  \E\[ \left\|\fC_i(\ML_i^\phalf \nabla f_i(x)) - \ML_i^\phalf \nabla f_i(x) \right\|^2_{\ML_i} \] \notag \\
  &\stackrel{\spadesuit}{\le}&
  \frac{1}{n^2}\sum_{i=1}^n  \cL(\fC_i,\ML_i) \|\nabla f_i(x)\|^2_{\ML_i^\dagger} \notag \\
  &\stackrel{\diamondsuit}{\le}&
  \frac{2}{n^2}\sum_{i=1}^n \cL_i \|\nabla f_i(x) - \nabla f_i(x^*)\|^2_{\ML_i^\dagger}
  + \frac{2}{n^2}\sum_{i=1}^n \cL_i\|\nabla f_i(x^*)\|^2_{\ML_i^\dagger} \notag \\
  &\stackrel{\bigstar}{\le}&
  \frac{4}{n^2}\sum_{i=1}^n \cL_i D_{f_i}(x,x^*) + \frac{2\sigma_+^*}{n} \notag \\
  &\le&
  \frac{4\cL_{\max}}{n} D_{f}(x,x^*) + \frac{2\sigma_+^*}{n}, \notag
\end{eqnarray}
where $\clubsuit$ is due to $\E\left[\ML_i^\half\fC_i(\ML_i^\phalf \nabla f_i(x))\right] = \nabla f_i(x)$ and $\nabla f_i(x) = \ML_i^{\nicefrac{1}{2}}\ML_i^{\dagger\nicefrac{1}{2}} \ML_i^{\nicefrac{1}{2}}r_i = \ML_i^{\nicefrac{1}{2}}\ML_i^{\dagger\nicefrac{1}{2}} \nabla f_i(x)$ based on (\ref{unbiased-est}) and $\nabla f_i(x) = \ML_i^{\nicefrac{1}{2}}r_i$ for some $r_i$. $\spadesuit$ can be directly obtained by noticing the definition of $\cL(\fC, \mL)$ in Table~\ref{tbl:notation}. $\diamondsuit$ is based on the definition of $\cL_i$ and the fact $\Norm{x+y}^2 \leq 2\Norm{x}^2 + 2\Norm{y}^2$. $\bigstar$ is due to Lemma E.3 of \cite{GJS-HR} and the definition of $\sigma_+^*$ defined in Theorem~\ref{thm:dcgd+_arbitrary}. The inequality above together with convexity and $L$-smoothness of $f$ implies
\begin{eqnarray*}
\E\[\|g(x) - \nabla f(x^*)\|^2\]
&=& \|\nabla f(x) - \nabla f(x^*)\|^2 + \E\[\|g(x) - \nabla f(x)\|^2\] \\
&\le& 2L D_{f}(x,x^*) + \frac{4\cL_{\max}}{n} D_{f}(x,x^*) + \frac{2\sigma_+^*}{n} \\
&\le& 2\(L + \frac{2\cL_{\max}}{n}\) D_{f}(x,x^*) + \frac{2\sigma_+^*}{n}.
\end{eqnarray*}

Applying the result of \cite{sigma_k} we conclude the proof.

\subsection{Proof of Theorem \ref{thm:diana+_arbitrary}: DIANA+ with arbitrary unbiased compression}\label{apx:diana+_arbitrary}

We start with the unbiasedness of the estimator
$$
g^k = \frac{1}{n}\sum_{i=1}^n \ML_i^\half\fC_i\(\ML_i^\phalf (\nabla f_i(x) - u_i^k)\) + u_i^k.
$$
In (\ref{unbiased-est}), we showed unbiasedness using inclusion $\nabla f_i(x^k)\in\range(\ML_i)$. Assuming $u_i^k\in\range(\ML_i)$ for all $k\ge0$, we get $\nabla f_i(x^k) - u_i^k\in\range(\ML_i)$ for all $k\ge0$. Hence, in the same way we can show unbiasedness of $g^k$ as
\begin{eqnarray*}
\E_k\[g^k\]
&=& \frac{1}{n}\sum_{i=1}^n \ML_i^\half \E_k\[\fC_i\(\ML_i^\phalf (\nabla f_i(x) - u_i^k)\)\] + u_i^k \\
&=& \frac{1}{n}\sum_{i=1}^n \ML_i^\half \ML_i^\phalf (\nabla f_i(x) - u_i^k) + u_i^k \\
&=& \frac{1}{n}\sum_{i=1}^n \nabla f_i(x^k)
= \nabla f(x^k).
\end{eqnarray*}
The inclusion $u_i^k\in\range(\ML_i)$ directly follows from the initialization $u_i^0\in\range(\ML_i)$ (see line 1 of Algorithm \ref{alg:DIANA+}) and linear update rule of $u_i^{k+1} = u_i^k + \alpha\ML_i^{\nicefrac{1}{2}}\Delta_i^k$ (see line 5 of Algorithm \ref{alg:DIANA+}). As both $\nabla f_i(x^k)$ and $u_i^k$ belong to $\range(\ML_i)$, denote $\nabla f_i(x^k)-u_i^k = \ML_i^{\nicefrac{1}{2}}r_i^k$. Next we bound

\begin{eqnarray}
  \E\[\|g(x)-\nabla f(x)\|^2\]
  &=&
  \E\[ \left\|\frac{1}{n}\sum_{i=1}^n \ML_i^\half\fC_i\(\ML_i^\phalf (\nabla f_i(x) - u_i^k)\) + u_i^k - \nabla f_i(x) \right\|^2 \] \notag \\
  &=&
  \frac{1}{n^2}\sum_{i=1}^n  \E\[ \left\|\ML_i^\half\fC_i^k\(\ML_i^\phalf (\nabla f_i(x) - u_i^k)\) - \ML_i^\half \ML_i^\phalf (\nabla f_i(x) - u_i^k) \right\|^2 \] \notag \\
  &=&
  \frac{1}{n^2}\sum_{i=1}^n  \E\[ \left\|\fC_i^k\(\ML_i^\phalf (\nabla f_i(x) - u_i^k)\) - \ML_i^\phalf (\nabla f_i(x) - u_i^k) \right\|^2_{\ML_i} \] \notag \\
  &\le&
  \frac{1}{n^2}\sum_{i=1}^n  \cL(\fC_i,\ML_i) \|\nabla f_i(x) - u_i^k\|^2_{\ML_i^\dagger} \notag \\
  &\le&
  \frac{2\cL_{\max}}{n^2}\sum_{i=1}^n \|\nabla f_i(x) - \nabla f_i(x^*)\|^2_{\ML_i^\dagger}
  + \frac{2\cL_{\max}}{n^2}\sum_{i=1}^n \|u_i^k - \nabla f_i(x^*)\|^2_{\ML_i^\dagger} \notag \\
  &\le&
  \frac{4\cL_{\max}}{n^2}\sum_{i=1}^n D_{f_i}(x,x^*) + \frac{2\cL_{\max}}{n} \sigma_+^k \notag \\
  &=&
  \frac{4\cL_{\max}}{n} D_{f}(x,x^*) + \frac{2\cL_{\max}}{n}\sigma_+^k, \notag
\end{eqnarray}
where $\sigma_+^k \eqdef \frac{1}{n}\sum_{i=1}^n \|u_i^k - \nabla f_i(x^*)\|^2_{\ML_i^\dagger}$ is the error in the gradient learning process. To proceed, we need to establish contractive recurrence relation for $\sigma_+^k$. For each summand, we have

\begin{eqnarray}
  && \E_k\[\left\| u_i^{k+1} - \nabla f_i(x^*) \right\|^2_{\ML_i^{\dagger}}\] \notag \\
  &=&
  \E_k\[\left\| u_i^k - \nabla f_i(x^*) + \alpha\overbar{\Delta}_i^k \right\|^2_{\ML_i^{\dagger}}\] \notag \\
  &=&
  \left\| u_i^k - \nabla f_i(x^*) \right\|^2_{\ML_i^{\dagger}}
  + 2\alpha\<u_i^k - \nabla f_i(x^*), \nabla f_i(x^k)-u_i^k\>_{\ML_i^{\dagger}}
  + \alpha^2 \E\[\left\| \ML_i^\half\fC_i\(\ML_i^\phalf (\nabla f_i(x) - u_i^k)\) \right\|^2_{\ML_i^{\dagger}}\] \notag \\
  &\le&
  \left\| u_i^k - \nabla f_i(x^*) \right\|^2_{\ML_i^{\dagger}}
  + 2\alpha\<u_i^k - \nabla f_i(x^*), \nabla f_i(x^k)-u_i^k\>_{\ML_i^{\dagger}}
  + \alpha^2 \E\[\left\| \fC_i\(\ML_i^\phalf (\nabla f_i(x) - u_i^k)\) \right\|^2\] \notag \\
  &\le&
  \left\| u_i^k - \nabla f_i(x^*) \right\|^2_{\ML_i^{\dagger}}
  + 2\alpha\<u_i^k - \nabla f_i(x^*), \nabla f_i(x^k)-u_i^k\>_{\ML_i^{\dagger}}
  + \alpha^2(1+\omega_i) \left\| \nabla f_i(x^k)-u_i^k \right\|^2_{\ML_i^{\dagger}} \notag \\
  &\le&
  \left\| u_i^k - \nabla f_i(x^*) \right\|^2_{\ML_i^{\dagger}}
  + 2\alpha\<u_i^k - \nabla f_i(x^*), \nabla f_i(x^k)-u_i^k\>_{\ML_i^{\dagger}}
  + \alpha \left\| \nabla f_i(x^k)-u_i^k \right\|^2_{\ML_i^{\dagger}} \notag \\
  &=&
  (1-\alpha)\left\| u_i^k - \nabla f_i(x^*) \right\|^2_{\ML_i^{\dagger}}
  + \alpha\left\| \nabla f_i(x^k)- \nabla f_i(x^*) \right\|^2_{\ML_i^{\dagger}}, \notag \\
  &\le&
  (1-\alpha)\left\| u_i^k - \nabla f_i(x^*) \right\|^2_{\ML_i^{\dagger}}
  + 2\alpha D_{f_i}(x^k,x^*), \notag 
\end{eqnarray}
where we used bounds $\alpha \le \frac{1}{1+\omega_i}$ and
$
0\preceq \ML_i^{\nicefrac{1}{2}}\ML_i^{\dagger}\ML_i^{\nicefrac{1}{2}} \preceq \MI.
$
Thus, with $\alpha \le \frac{1}{1+\omega_{\max}}$, the estimator $g^k$ of DIANA+ satisfies
\begin{align*}
  \E_k\[g^k\] &= \nabla f(x^k) \\
  \E_k\[\|g^k - \nabla f(x^*)\|^2\] &\le 2\(L + \frac{2\cL_{\max}}{n}\)D_f(x^k,x^*) + \frac{2\cL_{\max}}{n}\sigma_+^k \\
  \E_k\[\sigma_+^{k+1}\] &\le (1-\alpha)\sigma_+^k + 2\alpha D_f(x^k,x^*).
\end{align*}

Again, we apply the generic result of \cite{sigma_k} to complete the proof.

\clearpage
\section{Proofs for Section \ref{sec:block-quant}: Block Quantization}\label{apx:block-quant}

Here we provide the missing proofs of Section \ref{sec:block-quant}.

\subsection{Proof of the variance bound \eqref{var-bound:block-q}}
Using Definition \ref{def:block-quant} of compression operator $\cQ^B_h$, we have
\begin{eqnarray*}
\E\[\|\cQ^B_h(x) - x\|_{\mL}^2\]
&=& \sum_{l=1}^B \|x^l\|^2 \E\[\left\|\xi_l\(\frac{|x^l|}{\|x^l\|}\) - \frac{|x^l|}{\|x^l\|}\right\|_{\mL^{ll}}^2\] \\
&\le& \sum_{l=1}^B \|x^l\|^2 \min\( h^2_l\sum_{j=1}^{d_l}\mL^{ll}_{jj}, h_l\sqrt{\sum_{j=1}^{d_l}\[\mL^{ll}_{jj}\]^2} \) \\
&\le& \max_{1\le l\le B} \min\( h^2_l\sum_{j=1}^{d_l}\mL^{ll}_{jj}, h_l\sqrt{\sum_{j=1}^{d_l}\[\mL^{ll}_{jj}\]^2} \) \|x\|^2 \\
&=& \max_{1\le l\le B} \min\( h^2_l\|\diag(\mL^{ll})\|_1, h_l\|\diag(\mL^{ll})\| \) \|x\|^2.
\end{eqnarray*}
From the definition of $\cL(\cQ^B_h,\mL)$ we get
$$
\cL(\cQ^B_h,\mL) \le \max_{1\le l\le B} \min\( h^2_l\|\diag(\mL^{ll})\|_1, h_l\|\diag(\mL^{ll})\| \),
$$
which implies \eqref{var-bound:block-q} if we ignore the first term.

\subsection{Proof of Theorem \ref{thm-prop:block-dcgd+}: DCGD+ with block quantization}
First, recall that quantization steps $h_i$ are given by
$$
h_{i,l} = \frac{\delta_{i,B}}{\|\diag(\mL_i^{ll})\|},\; l\in[B],
\quad\text{where}\;
\delta_{i,B} \le \frac{d}{\beta-B}T_{i,B} + \sqrt{\frac{d}{\beta-B}}T_{i,1}.
$$
Then, we have
\begin{eqnarray}
\frac{\cL_{\max}}{n}
&=&   \frac{1}{n}\max_{i\in[n]}\cL(\cQ^B_{h_i},\mL_i) \notag \\
&\le& \frac{1}{n} \max_{i\in[n]}\delta_{i,B} \notag \\
&\le& \frac{1}{n}\max_{i\in[n]} \[\frac{d}{\beta-B}T_{i,B} + \sqrt{\frac{d}{\beta-B}}T_{i,1}\] \notag \\
&\le& \[\frac{\nicefrac{d}{n}}{\beta-B}\] \max_{i\in[n]}T_{i,B} + \sqrt{\frac{\nicefrac{d}{n}}{\beta-B}} \max_{i\in[n]}\frac{T_{i,1}}{\sqrt{n}} \notag.
\end{eqnarray}
Set $\beta = \nicefrac{d}{n} + n$ and $B=n$. Since $n=\cO(\sqrt{d})$, we have $\beta=\cO(\nicefrac{d}{n})$ and hence $\frac{\nicefrac{d}{n}}{\beta-B} = 1$. For the sake of simplicity, assume $d_l=\nicefrac{d}{n}$. Next
\begin{eqnarray*}
\frac{T_{i,1}}{\sqrt{n}}
&\le& \frac{1}{\sqrt{nd}}\sum_{j=1}^d \mL_{i;jj}
\le \frac{\nu_1 L_{\max}}{\sqrt{nd}} \\
T_{i,n}
&\le& \frac{1}{d}\sum_{l=1}^n \sqrt{d_l}\sum_{j=1}^{d_l}\mL^{ll}_{jj} \\
&=& \frac{\max_{l\in[n]}\sqrt{d_l}}{d}\sum_{j=1}^d\mL_{jj} \\
&=& \frac{\max_{l\in[n]}\sqrt{d_l}}{d} \nu_1 L_{\max}
\le \frac{\nu_1 L_{\max}}{\sqrt{nd}}.
\end{eqnarray*}

Regardless of the choice $h_{i}$, using the following inequalities with respect to matrix order
\begin{equation}\label{L-bound-06}
\ML \preceq \frac{1}{n}\sum_{i=1}^n\ML_i, \quad \ML_i \preceq n\ML,
\end{equation}
we bound $L$ as follows
\begin{equation}\label{L-bound}
L = \lambda_{\max}\(\ML\)
\overset{(\ref{L-bound-06})}{\le}
    \lambda_{\max}\(\frac{1}{n}\sum_{i=1}^n \ML_i\)
\le
    \frac{1}{n}\sum_{i=1}^n \lambda_{\max}\(\ML_i\)
=
    \frac{1}{n} \sum_{i=1}^n L_i
\overset{(\ref{def:nu})}{\le}
    \frac{\nu}{n} L_{\max}.
\end{equation}

Hence
$$
\frac{L}{\mu} + \frac{\cL_{\max}}{\mu n}
\le \frac{\nu}{n}\frac{L_{\max}}{\mu} + \frac{2\nu_1}{\sqrt{nd}}\frac{L_{\max}}{\mu}
= \cO\(\frac{1}{n}\frac{L_{\max}}{\mu}\),
$$
which guarantees $n$ times fewer communication rounds with the same number of bits per round. In other words, each node communicates $\cO(\nicefrac{d}{n})$ bits to the master in each iteration, which gives us $\cO(d)$ communication per communication round. Thus, overall communication complexity to achieve $\varepsilon>0$ accuracy is
$$
\cO\( \frac{d}{n}\frac{L_{\max}}{\mu}\log\frac{1}{\varepsilon} \).
$$

\subsection{Proof of Theorem \ref{thm-prop:block-diana+}: DIANA+ with block quantization}

As already mentioned, for DIANA+ each node aims to minimize $\omega_i + \frac{1}{n\mu}\cL(\cQ^B_{h_i},\mL_i)$ with respect to its quantization steps $h_i$. Notice that
\begin{eqnarray*}
\omega_{i} + \frac{1}{n\mu}\cL(\cQ^B_{h_i},\mL_i)
&\le& \max_{l\in[B]} h_{i,l}\sqrt{d_l} + \max_{l\in[B]} \frac{h_{i,l}}{\mu n}\|\diag(\mL_i^{ll})\| \\
&\le& 2\max_{l\in[B]} h_{i,l}\(\sqrt{d_l} + \frac{1}{\mu n}\|\diag(\mL_i^{ll})\|\).
\end{eqnarray*}

This leads to the following optimization problem with respect to $h$:
\begin{equation}\label{meta-opt-4_}
\begin{aligned}
\min_{h\in\R^B}          \quad & \max_{1\le l\le B} h_l\(\sqrt{d_l} + \frac{1}{\mu n}\|\diag(\mL_i^{ll})\|\) \\
\textrm{s.t.} \quad & \sum_{l=1}^B\(\frac{1}{h_l^2} + \frac{\sqrt{d_l}}{h_l}\) + B = \beta,\; h_l>0. \\
\end{aligned}
\end{equation}

which is solved similar to \eqref{meta-opt-3}.
Denote
$$
A_{il} \eqdef \sqrt{d_l} + \frac{1}{\mu n}\|\diag(\mL_i^{ll})\|,
\quad
\widetilde{T}_{iB} \eqdef \frac{1}{d}\sum_{l=1}^B \sqrt{d_l}A_{il}.
$$

Analogous to \eqref{meta-opt-3}, the solution of \eqref{meta-opt-4_} has the following form
$$
h_{il} = \frac{\widetilde{\delta}_{iB}}{A_{il}},\; l\in[B],
$$
where $\widetilde{\delta}_{iB}$ is determined by the constraint equality of \eqref{meta-opt-4_} as
$$
\widetilde{\delta}_{iB} = \frac{d \widetilde{T}_{i,B}}{2(\beta-B)} + \sqrt{\frac{d^2 \widetilde{T}_{i,B}^2}{4(\beta-B)^2} + \frac{d \widetilde{T}_{i,1}^2}{\beta-B}}
\le \frac{d}{\beta-B}\widetilde{T}_{i,B} + \sqrt{\frac{d}{\beta-B}}\widetilde{T}_{i,1}.
$$

Let us estimate $\widetilde{T}_{i,1}$ and $\widetilde{T}_{i,n}$ using the assumptions $B=n$ and (for the sake of simplicity) $d_l=\nicefrac{d}{n}$.
\begin{eqnarray*}
\widetilde{T}_{i1}
&=& \frac{1}{\sqrt{d}}\( \sqrt{d} + \frac{1}{\mu n}\|\diag(\mL_i)\| \)
= 1 + \frac{1}{\mu n \sqrt{d}}\sum_{j=1}^d \mL_{i;jj}
\le 1 + \frac{\nu_1 L_{\max}}{\mu n \sqrt{d}} \\
\widetilde{T}_{in}
&=& \frac{1}{d}\sum_{l=1}^n \sqrt{\frac{d}{n}}\( \sqrt{\frac{d}{n}} + \frac{1}{\mu n}\|\diag(\mL_i^{ll})\| \)
= 1 + \frac{1}{\mu n \sqrt{nd}}\sum_{l=1}^n \|\diag(\mL_i^{ll})\| \\
&\le& 1 + \frac{1}{\mu n \sqrt{nd}}\sum_{j=1}^d \mL_{i;jj}
= 1 + \frac{\nu_1 L_{\max}}{\mu n \sqrt{nd}}.
\end{eqnarray*}

Next, using $\beta=\nicefrac{d}{n}+n$ and $\nu_1 = \cO(1)$, we get
\begin{eqnarray*}
\omega_i + \frac{1}{n\mu}\cL(\cQ^n_{h_i},\mL_i)
&\le& 2 \widetilde{\delta}_{in} \\
&\le& \frac{2d}{\beta-n}\widetilde{T}_{in} + 2\sqrt{\frac{d}{\beta-n}}\widetilde{T}_{i1} \\
&=&   2n \widetilde{T}_{in} + 2\sqrt{n}\widetilde{T}_{i1} \\
&\le& 2n \( 1 + \frac{\nu_1 L_{\max}}{\mu n \sqrt{nd}} \) + 2\sqrt{n} \( 1 + \frac{\nu_1 L_{\max}}{\mu n \sqrt{d}} \)
= \cO\( n + \frac{1}{\sqrt{nd}}\frac{L_{\max}}{\mu} \).
\end{eqnarray*}

Together with \eqref{L-bound}, we complete the proof with the following iteration complexity:
$$
\cO\( n + \frac{1}{n}\frac{L_{\max}}{\mu} + \frac{1}{\sqrt{nd}}\frac{L_{\max}}{\mu} \).
$$

\clearpage
\section{Proofs for Section \ref{sec:quant}: Quantization with varying steps}

In this part of the appendix we provide missing proofs and detailed arguments of Section \ref{sec:quant}.

\subsection{An encoding scheme for $\cQ_h$ operator}

To communicate a vector of the form $\cQ_h(x)$, we adapt the encoding scheme of \cite{albasyoni2020optimal}. From the definition, we have $$\[\cQ_h(x)\]_j = \|x\| \cdot \sign(x_j \hat{k}_j) \cdot \hat{k}_jh_j$$ for all $j\in[d]$, where $\hat{k}_j\ge0$ are non-negative random variables coming from \eqref{quant-xi}. Thus, we need to encode the magnitude $\|x\|$, signs $\sign(x_j\hat{k}_j)$ and non-negative integers $\hat{k}_j$.

For the magnitude $\|x\|$ we need just $31$ bits. Let $n_0 \eqdef |\{j\in[d] \colon \hat{k}_j=0\}|$ be the number of coordinates $x_j$ that are compressed to 0. To communicate signs $\{\sign(x_j\hat{k}_j) \colon j\in[d]\}$, we first send the locations of those $n_0$ coordinates and then $d-n_0$ bits for the values $\pm 1$. Sending $n_0$ positions can be done by sending $\log d$ bits representing the number $n_0$, followed by $\log\binom{d}{n_0}$ bits for the positions. For the signs, we need $\log d + \log\binom{d}{\hat{n}_0} + d-\hat{n}_0 \le \log d + d\log 3$ bits at most. Finally, it remains to encode $\hat{k}_j$'s for which we only need to send nonzero entries since the positions of $\hat{k}_j = 0$ are already encoded. We encode $\hat{k}_j \ge 1$ with $\hat{k}_j$ bits: $\hat{k}_j-1$ ones followed by 0. Hence, the expected number of bits to encode $\hat{k}_j$'s is
$$
\E\[\sum_{j=1}^d \hat{k}_j\]
\overset{\eqref{quant-xi}}{=} \sum_{j=1}^d \frac{v_j}{h_j} \le \sqrt{\sum_{j=1}^d v_j^2}\sqrt{\sum_{j=1}^d \frac{1}{h_j^2}} = \sqrt{\sum_{j=1}^d \frac{1}{h_j^2}} = \|h^{-1}\|,
$$
where $v_j=\frac{|x_j|}{\|x\|}$. 

In total, $\cQ_h(x)$ can be encoded by $$31 + \log d + \log\binom{d}{\hat{n}_0} + d-\hat{n}_0 + \|h^{-1}\|$$ bits.  Lastly, the $\log\binom{d}{\hat{n}_0}$ term can be upper bounded by the binary entropy function $H_2(t) \eqdef -t\log t - (1-t)\log(1-t)$ (see \citep{albasyoni2020optimal} for more details), and the expected number of encoding bits for $\cQ_h(x)$ can be upper bounded by $$31 + \log d + d H_2\left(\frac{\|\hat{x}\|_0}{d}\right) + \|\hat{x}\|_0 + \|h^{-1}\|,$$ where $\hat{x} = \cQ_h(x)$.

\clearpage

\subsection{Proof of the variance bound \eqref{eq:002}}

Let $v\in\R^d$ be the unit vector with non-negative entries $v_j = \nicefrac{|x_j|}{\|x\|}$ for $j\in[d]$. Then
\begin{eqnarray}
\E\[\|\cQ_h(x) - x\|^2_{\mL}\]
&=& \E\[\left\|\|x\| \cdot \sign(x) \cdot \xi\(\frac{|x|}{\|x\|}\) - \|x\| \cdot \sign(x) \cdot \frac{|x|}{\|x\|}\right\|^2_{\mL}\] \notag \\
&=& \|x\|^2 \E\[\left\| \xi\(v\) - v\right\|^2_{\mL}\] \notag \\
&=& \|x\|^2 \E\[ \sum_{j,l=1}^d \mL_{jl}\(\xi_j(v_j)-v_j\)\(\xi_l(v_l)-v_l\) \] \notag \\
&=& \|x\|^2 \sum_{j=1}^d \mL_{jj} \E\[ \(\xi_j(v_j)-v_j\)^2 \] \label{eq:001} \\
&=& \|x\|^2 \sum_{j=1}^d \mL_{jj} \(v_j-k_jh_j\)\((k_j+1)h_j-v_j\) \notag \\
&=& \|x\|^2 \sum_{j=1}^d \mL_{jj}h_j^2 \(\frac{v_j}{h_j}-k_j\)\[1- \(\frac{v_j}{h_j}-k_j\)\] \notag \\
&\le& \|x\|^2 \sum_{j=1}^d \mL_{jj}h_j^2 \min\(1, \frac{v_j}{h_j}\) \notag \\
&\le& \min\( \sum_{j=1}^d \mL_{jj}h_j^2, \sum_{j=1}^d \mL_{jj}h_jv_j \) \|x\|^2  \notag\\
&\le& \min\( \sum_{j=1}^d \mL_{jj}h_j^2, \sqrt{\sum_{j=1}^d \mL_{jj}^2h_j^2} \) \|x\|^2. \notag \\
&=&   \min\( \|\diag(\mL)h^2\|_1, \|\diag(\mL)h\| \) \|x\|^2, \notag
\end{eqnarray}
which implies \eqref{eq:002}.

\subsection{Proof of Theorem \ref{thm-prop:quant-dcgd+}: DCGD+ with varying quantization steps}

Based on the upper bound \eqref{eq:002} and the communication constraint given by $\|h_i^{-1}\|=\beta$ for some $\beta>0$, we get the  optimization problem 
\begin{equation}\label{meta-opt}
\begin{aligned}
\min_{h_i}        \quad & \Norm{\diag(\mL_i) h_i}\quad \textrm{subject to} \quad  \Norm{h_i^{-1}} = \beta ,
\end{aligned}
\end{equation}
for choosing the optimal quantization parameters $h_{i;j}$. This problem has a closed form solution. Indeed, due to the KKT conditions, we have
\begin{align*}
        \frac{\mL_{i;j}^2 h_{ij}^4}{\sqrt{\sum_{t=1}^d\mL_{i;t}^2h_{it}^2}} = 2 \zeta,\quad \zeta\left(\sum_{t=1}^d h_{ij}^2 - \beta^2\right)=0,
\end{align*}
where $\zeta$ is the multiplier. Solving this leads to the solution:
\begin{equation}\label{level-sols-dcgd}
h_{i;j} = \frac{1}{\beta}\sqrt{\frac{\sum_{t=1}^d \mL_{i;t}}{\mL_{i;j}}}.
\end{equation}

For the solution \eqref{level-sols-dcgd} we have
\begin{eqnarray}\label{eq:005}
\widetilde{\cL}(\cQ_{h_i},\mL_i)
&\le& \sqrt{\sum_{j=1}^d \mL^2_{i;jj}h_{i;j}^2}
= \frac{1}{\beta} \sqrt{\sum_{j=1}^d \mL^2_{i;jj} \frac{ \sum_{l=1}^d \mL_{i;ll} }{\mL_{i;jj}}}
= \frac{1}{\beta} \sum_{j=1}^d \mL_{i;jj} \notag \\
&\le& \frac{\nu_1}{\beta}\max_{j\in[d]}\mL_{i;jj}
\le \frac{\nu_1}{\beta}L_i = \frac{\nu_1}{\beta}L_{\max}.
\end{eqnarray}

Therefore,
if both parameters $\nu$ and $\nu_2$ are $\cO(1)$, then the rate \eqref{DCGD+-complexity} of DCGD+ becomes $\cO(\frac{L_{\max}}{n\mu} + \frac{L_{\max}}{\beta n\mu})$. To make a fair comparison against DCGD, we need to fix $\cO(\frac{d}{n})$ number of bits each node communicates to the master server.
Now, to make DCGD+ communicate the same number of bits, we set $\beta=\cO(\frac{d}{n})$. Hence we have the following iteration complexity for DCGD+ based on solution \eqref{level-sols-dcgd}:
$$
\cO\( \frac{1}{n}\frac{L_{\max}}{\mu} + \frac{1}{d}\frac{L_{\max}}{\mu} \)
$$
which is $\min(n,d)$ times better than the one of DCGD.

\subsection{Proof of Theorem \ref{thm-prop:quant-diana+}: DIANA+ with varying quantization steps}

Denote $A_{ij} \eqdef \frac{\mL_{i;jj}}{n\mu}$. Note that
\begin{eqnarray}
 \omega_i + \frac{\cL_i}{n\mu} &\le& \min\( \sum_{j=1}^d h_{i;j}^2, \sqrt{\sum_{j=1}^d h_{i;j}^2} \) + \frac{1}{n\mu}\min\( \sum_{j=1}^d \mL_{i;jj}h_{i;j}^2, \sqrt{\sum_{j=1}^d \mL_{i;jj}^2h_{i;j}^2} \) \notag \\
&=& \min\( \sum_{j=1}^d h_{i;j}^2, \sqrt{\sum_{j=1}^d h_{i;j}^2} \)
+ \min\( \sum_{j=1}^d \frac{\mL_{i;jj}}{n\mu}h_{i;j}^2, \sqrt{\sum_{j=1}^d \(\frac{\mL_{i;jj}}{n\mu}\)^2h_{i;j}^2} \) \notag \\
&\le& \min\( \sum_{j=1}^d h_{i;j}^2 + \sum_{j=1}^d \frac{\mL_{i;jj}}{n\mu}h_{i;j}^2, \sqrt{\sum_{j=1}^d h_{i;j}^2} + \sqrt{\sum_{j=1}^d \(\frac{\mL_{i;jj}}{n\mu}\)^2h_{i;j}^2}\) \notag \\
&\le& \min\( \sum_{j=1}^d \(1+A_{ij}\)h_{i;j}^2, \sqrt{2\sum_{j=1}^d \(1+A_{ij}^2\)h_{i;j}^2}\) \notag \\
&\le& \sum_{j=1}^d \(1+A_{ij}\)h_{i;j}^2 . \notag
\end{eqnarray}
We solve the optimization problem
\begin{equation}\label{meta-opt-split-diana}
\begin{aligned}
\min_{h_i}        \quad & \sum_{j=1}^d \(1+A_{ij}\)h_{i;j}^2 \quad \textrm{subject to} \quad \Norm{h^{-1}} = \beta,
\end{aligned}
\end{equation}
which has a closed form solution. Indeed, due to the KKT conditions, we have:
\begin{equation}\label{level-sols-diana}
h_{i;j} = \frac{1}{\beta}\sqrt{\frac{\sum_{l=1}^d \sqrt{1+A^2_{il}}}{\sqrt{1+A^2_{ij}}}}.
\end{equation}

For the solution \eqref{level-sols-diana} we have
\begin{eqnarray*}
\omega_i + \frac{\widetilde{\cL}_i}{n\mu}
&\le& \sqrt{2\sum_{j=1}^d \(1+A_{ij}^2\)h_{i;j}^2}
= \frac{\sqrt{2}}{\beta} \sum_{j=1}^d \sqrt{1+A_{ij}^2}
= \frac{\sqrt{2}}{\beta} \sum_{j=1}^d \(1+A_{ij}\) \notag \\
&=& \frac{\sqrt{2}d}{\beta} + \frac{\sqrt{2}}{\beta n\mu} \sum_{j=1}^d \mL_{i;jj}
\le \frac{\sqrt{2}d}{\beta} + \frac{\sqrt{2}\nu_1}{\beta n} \frac{L_{\max}}{\mu},
\end{eqnarray*}
which further leads to $\cO(n + \frac{1}{n}\frac{L_{\max}}{\mu} + \frac{1}{d}\frac{L_{\max}}{\mu})$ iteration complexity if $\nu_1=\cO(1)$ and $\beta=\cO(\frac{d}{n})$.


\clearpage
\section{Notation Table}

\begin{table}[!h]
\caption{Notation we use throughout the paper.}
\label{tbl:notation}
\begin{center} \footnotesize
\begin{tabular}{|c|l|c|}
\hline 
\multicolumn{3}{|c|}{{\bf Basic} } \\
\hline 
$d$ & number of the model parameters to be trained & \\
\hline
$n$ & number of the nodes/workers in distributed system & \\
\hline
$[n]$ & $\eqdef \{1, 2, \dots, n\}$ & \\
\hline
$f: \R^d \to \R$ & overall empirical loss/risk & \eqref{main-opt-problem-dist}\\
\hline
$f_i: \R^d \to \R$ & local loss function associated with data owned by the node $i\in[n]$ & \eqref{main-opt-problem-dist}\\
\hline
$R: \R^d \to \R$ & (possibly non-smooth) regularization & \eqref{main-opt-problem-dist}\\
\hline
$x^*$ & trained model, i.e. optimal solution to \eqref{main-opt-problem-dist} & \\
\hline
$\varepsilon$ & target accuracy & \\
\hline
$\|x\|_0$ & $\eqdef \#\{j\in[d] \colon x_j \ne 0\}$, number of nonzero entries of $x\in\R^d$ & \\
\hline
$\|x\|$ & $\eqdef \sqrt{\sum_{j=1}^d x_j^2}$, the standard Euclidean norm of $x\in\R^d$ & \\
\hline
$D_{f}(x,y)$ & Bregman divergence between $x$ and $y$ with respect to $f$ for $x,y\in\R^d$ & \\
\hline
\hline 
\multicolumn{3}{|c|}{{\bf Standard} }\\
\hline 
$\mu$ & strong convexity parameter of $f$ & Asm.~\ref{asm:mu-convex} \\
\hline
$L$ & smoothness constant of $f$, namely $L = \lambda_{\max}(\mL)$ & \eqref{eq:scalar-smooth} \\
\hline
$L_i$ & smoothness constant of $f_i$, namely $L_i = \lambda_{\max}(\mL_i)$ & \\
\hline
$L_{\max}$ & $\eqdef \max_{i\in[n]} L_i$ & \\
\hline
$\fC$ & (possibly randomized) compression operator $\fC\colon\R^d\to\R^d$ &  \\
\hline
$\bB^d(\omega)$ & class of compressors with $\E\[\cC(x)\] = x, \; \E\[\|\fC(x) - x\|^2\] \le \omega\|x\|^2, \; \forall x\in\R^d$ & \\
\hline
$\fC_i$ & compression operator controlled by node $i$ & \\
\hline
$\omega_i$ & variance of compression operator $\fC_i$ & \\
\hline
$\omega_{\max}$ & $\eqdef \max_{i\in[n]} \omega_i$ & \\
\hline
$\gamma$ & step-size parameter in DCGD+ and DIANA+ methods & \\
\hline
$\alpha$ & learning rate for the local optimal gradients in DIANA+ & \\
\hline
\hline 
\multicolumn{3}{|c|}{{\bf Matrix Smoothness} }\\
\hline 
$\mL$ & smoothness matrix of $f$ & \eqref{eq:matrix-smooth} \\
\hline
$\mL^{\half}$ & square root of symmetric and positive semidefinite matrix $\mL$ & \\
\hline
$\mL^{\dagger}$ & Moore–Penrose inverse of matrix $\mL$ & \\
\hline
$\mL_i$ & smoothness matrix of $f_i$ & \\
\hline
$\mL_{i;j},\mL_{i;jj}$ & $j^{th}$ diagonal element of $\mL_i$ & \\
\hline
$\cL(\fC, \mL)$ & $\eqdef \inf\left\{\cL\ge0 \colon \E{\|\fC(x) - x\|^2_{\mL}} \le \cL\|x\|^2 \; \forall x\in \R^d\right\} \le \omega \lambda_{\max}( \mL)$ & \\
\hline
$\cL_i$ & $\eqdef \cL(\fC_i, \mL_i)$ & \eqref{def:exp-smooth-L} \\
\hline
$\cL_{\max}$ & $\eqdef \max_{i\in[n]}\cL(\fC_i, \mL_i) = \max_{i\in[n]}\cL_i$ & \eqref{def:exp-smooth-L} \\
\hline
$\nu,\; \nu_1$ & $\nu \eqdef \frac{\sum_{i=1}^n L_i}{\max_{i\in[n]} L_i}$ and $\nu_1 \eqdef \max_{i\in[n]}\frac{\sum_{j=1}^d \ML_{i;j}}{\max_{j\in[d]}\ML_{i;j}}$ & Def.~\ref{def:nu} \\
\hline
\hline 
\multicolumn{3}{|c|}{{\bf Quantization} }\\
\hline 
$s$ & number of quantization levels & \\
\hline
$B$ & number of blocks to divide the space $\R^d$ & \\
\hline
$l$ & index for blocks, i.e. $l\in[B]$ & \\
\hline
$d_l$ & dimension of the $l^{th}$ subspace in $\R^d$, in particular $\sum_{l=1}^B d_l = d$ & \\
\hline
$x^l$ & $l^{th}$ block of coordinates of $x\in\R^d$ & \\
\hline
$\mL^{ll}$ &  $l^{th}$ diagonal block matrix of $\mL$ with sizes $d_l\times d_l$ & \\
\hline
$h_{i;l}$ & quantization step of $l^{th}$ block for node $i$ & \\
\hline
$\beta$ & parameter controlling the number of encoding bits & \\
\hline
$j$ & index for coordinates, i.e. $j\in[d]$ & \\
\hline
$h_{i;j}$ & quantization step of $j^{th}$ coordinate for node $i$ & \eqref{level-sols-dcgd} \\
\hline
\end{tabular}
\end{center}
\end{table}

\end{document}